\documentclass[sigconf]{acmart}

\usepackage{amsmath,amsfonts}
\usepackage{algorithmic}
\usepackage{array}
\usepackage{textcomp}
\usepackage{stfloats}
\usepackage{url}
\usepackage{verbatim}
\usepackage{graphicx}
\usepackage{balance}
\usepackage{booktabs}
\usepackage{subcaption}
\usepackage{tabularx}
\usepackage{soul}
\usepackage{microtype}
\usepackage{xspace}
\usepackage{caption}
\usepackage{multirow}
\usepackage{mathtools}
\usepackage{pifont}
\usepackage{appendix}
\usepackage{bm}

\newcommand{\W}{$\mathcal{W}$\xspace}
\newcommand{\Wp}{$\mathcal{W}+$\xspace}

\newcolumntype{P}[1]{>{\centering\arraybackslash}p{#1}}
\newcolumntype{L}[1]{>{\raggedright\arraybackslash}p{#1}}
\def\etal{{\textit{et~al.~}}}

\AtBeginDocument{%
  \providecommand\BibTeX{{%
    \normalfont B\kern-0.5em{\scshape i\kern-0.25em b}\kern-0.8em\TeX}}}

\setcopyright{acmlicensed}
\acmConference[MM '22]{Proceedings of the 30th ACM International Conference on Multimedia}{October 10--14, 2022}{Lisboa, Portugal}
\copyrightyear{2022}
\acmYear{2022}
\acmBooktitle{Proceedings of the 30th ACM International Conference on Multimedia (MM '22), October 10--14, 2022, Lisboa, Portugal}
\acmPrice{15.00}
\acmDOI{}
\acmISBN{}

\begin{document}

\title{Cycle Encoding of a StyleGAN Encoder for Improved Reconstruction and Editability}


\author{Xudong Mao$^{1}$, Liujuan Cao$^{2*}$, Aurele T. Gnanha$^{3}$, Zhenguo Yang$^{4}$, Qing Li$^{5}$, Rongrong Ji$^{2}$}
\affiliation{%
  \institution{$^{1}$Guangdong Key Laboratory of Big Data Analysis and Processing, Sun Yat-sen University, China\\
  $^{2}$School of Informatics, Xiamen University, China\\
  $^{3}$Department of Computer Science, City University of Hong Kong, China\\
  $^{4}$Department of Computer Science, Guangdong University of Technology, China\\
  $^{5}$Department of Computing, Hong Kong Polytechnic University, China\\
  }
}
\email{maoxd3@mail.sysu.edu.cn, caoliujuan@xmu.edu.cn}
\thanks{*Corresponding author}



\begin{abstract}
GAN inversion aims to invert an input image into the latent space of a pre-trained GAN. Despite the recent advances in GAN inversion, there remain challenges to mitigate the tradeoff between distortion and editability, i.e. reconstructing the input image accurately and editing the inverted image with a small visual quality drop. The recently proposed pivotal tuning model makes significant progress towards reconstruction and editability, by using a two-step approach that first inverts the input image into a latent code, called pivot code, and then alters the generator so that the input image can be accurately mapped into the pivot code. Here, we show that both reconstruction and editability can be improved by a proper design of the pivot code. We present a simple yet effective method, named cycle encoding, for a high-quality pivot code. The key idea of our method is to progressively train an encoder in varying spaces according to a cycle scheme: $\mathcal{W}$$\rightarrow$$\mathcal{W}+$$\rightarrow$$\mathcal{W}$. This training methodology preserves the properties of both \W and \Wp spaces, i.e. high editability of \W and low distortion of \Wp. To further decrease the distortion, we also propose to refine the pivot code with an optimization-based method, where a regularization term is introduced to reduce the degradation in editability. Qualitative and quantitative comparisons to several state-of-the-art methods demonstrate the superiority of our approach.

\end{abstract}



\begin{teaserfigure}
    \setlength{\tabcolsep}{0.5pt}
	\renewcommand{\arraystretch}{0.75}
    \centering
        {\small
        \begin{tabular}{c c c c @{\hspace{2.5pt}} c c c c}
                \includegraphics[width=0.12\textwidth]{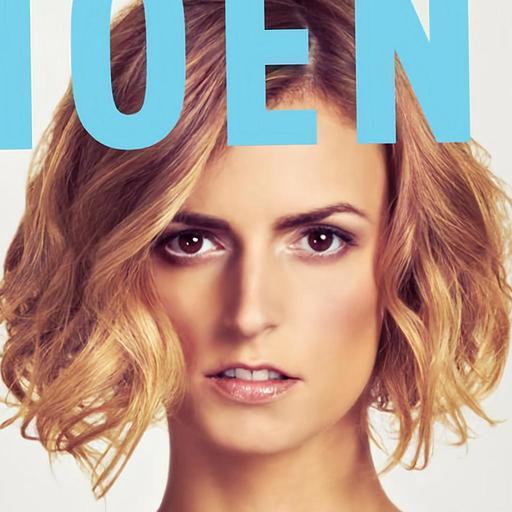}&
                \includegraphics[width=0.12\textwidth]{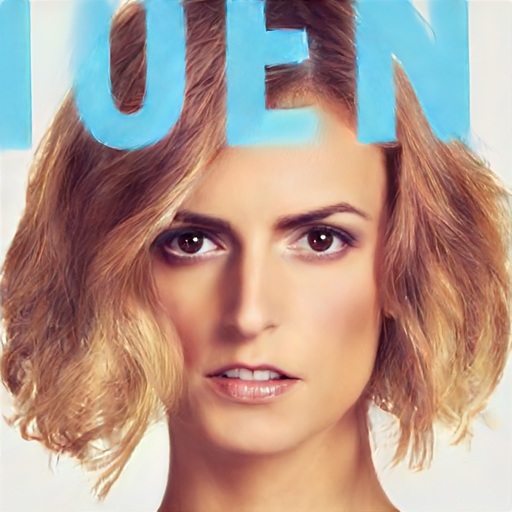}&
                \includegraphics[width=0.12\textwidth]{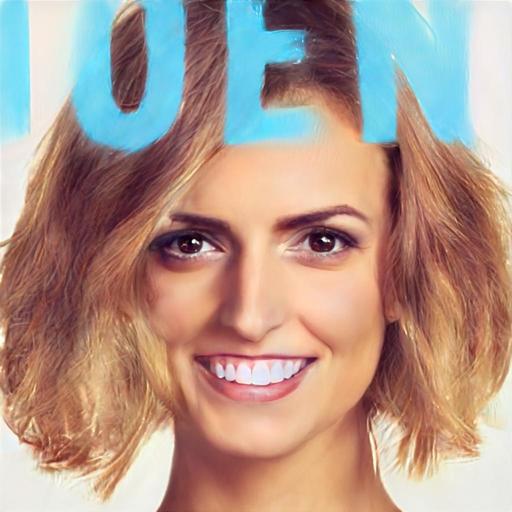}&
                \includegraphics[width=0.12\textwidth]{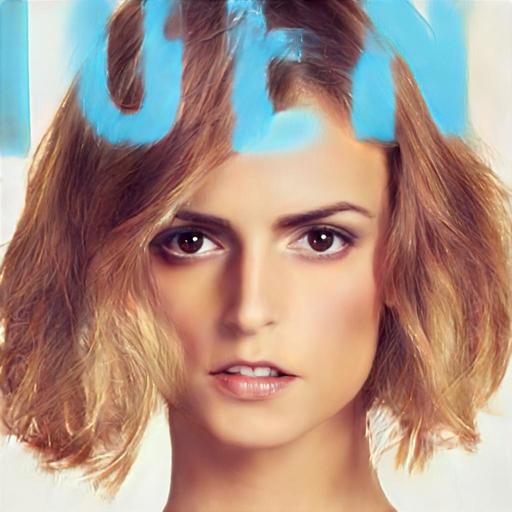}&        
                \includegraphics[width=0.12\textwidth]{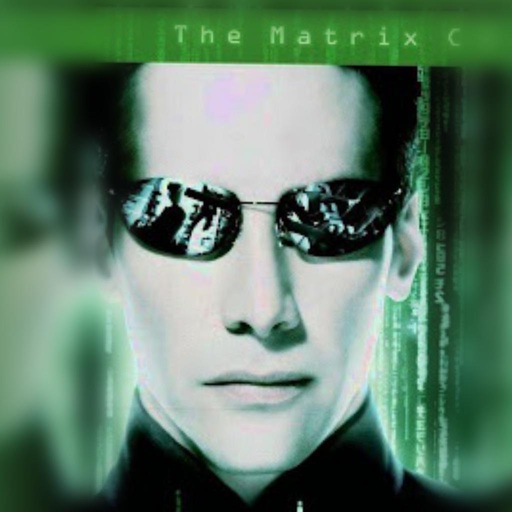}&
                \includegraphics[width=0.12\textwidth]{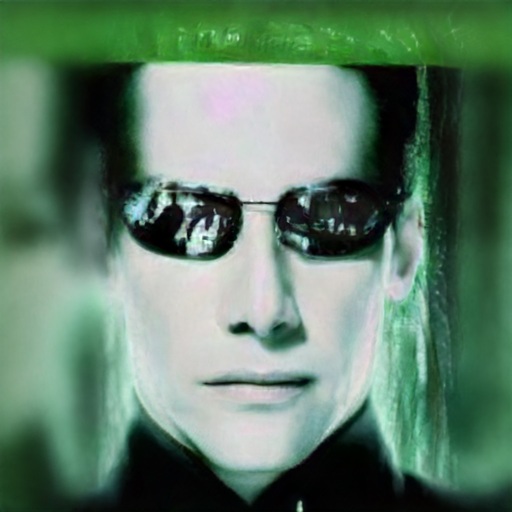}&
                \includegraphics[width=0.12\textwidth]{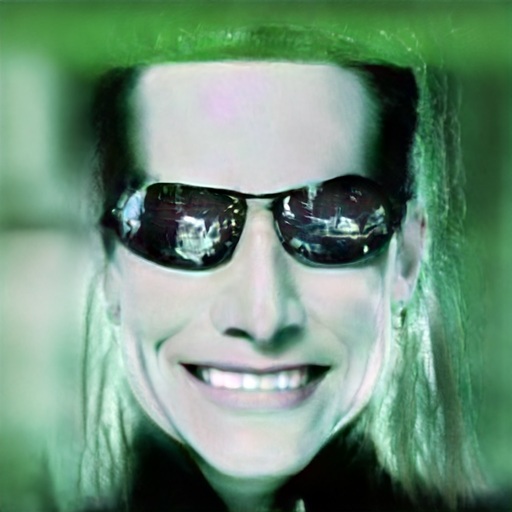}&
                \includegraphics[width=0.12\textwidth]{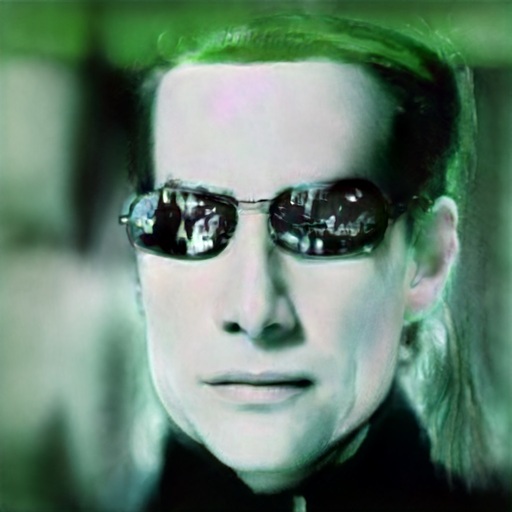}        
                \tabularnewline
                \includegraphics[width=0.12\textwidth]{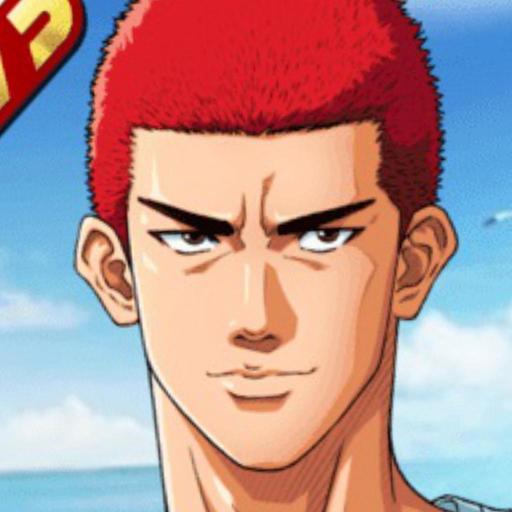}&
                \includegraphics[width=0.12\textwidth]{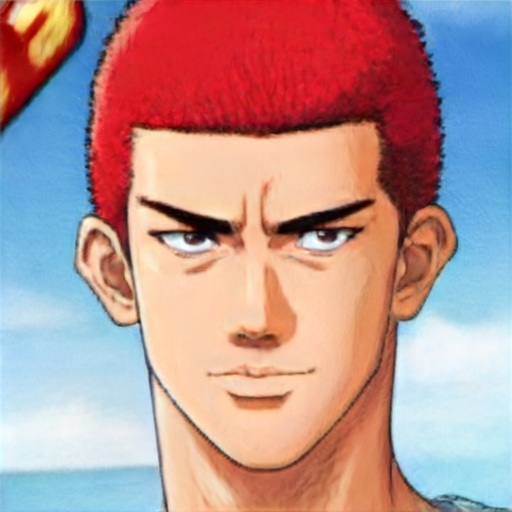}&
                \includegraphics[width=0.12\textwidth]{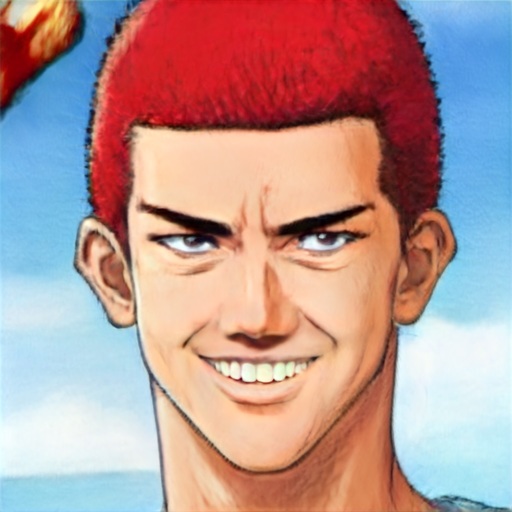}&
                \includegraphics[width=0.12\textwidth]{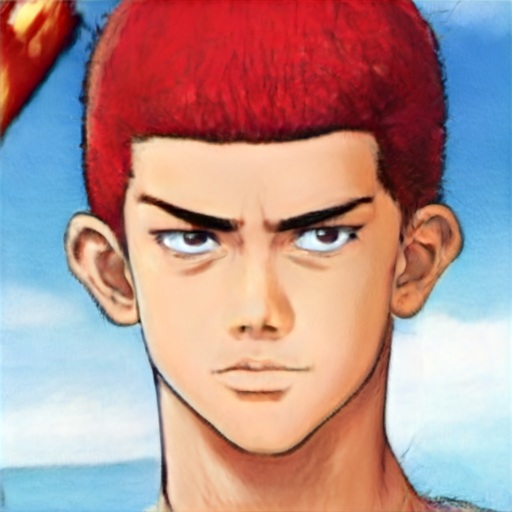}&  				
                \includegraphics[width=0.12\textwidth]{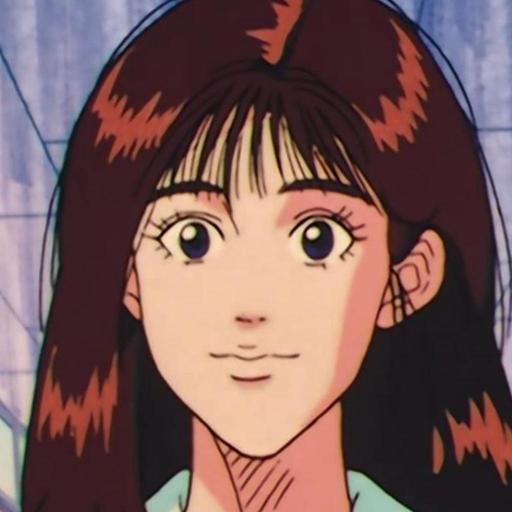}&
                \includegraphics[width=0.12\textwidth]{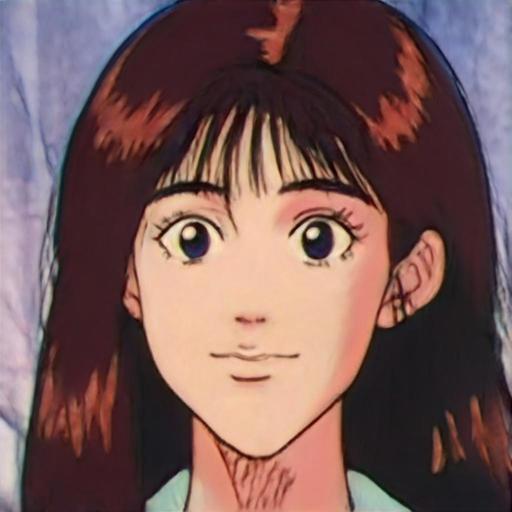}&
                \includegraphics[width=0.12\textwidth]{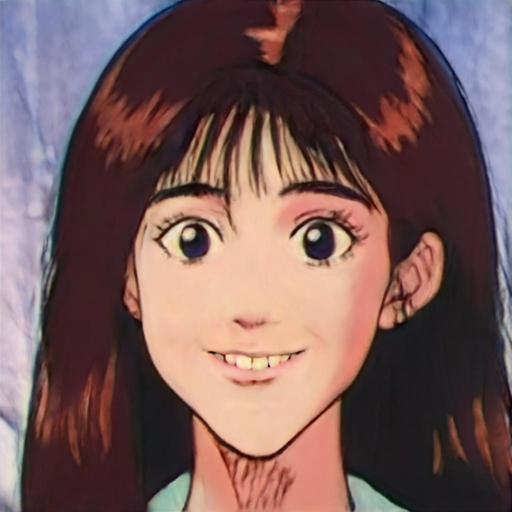}&
                \includegraphics[width=0.12\textwidth]{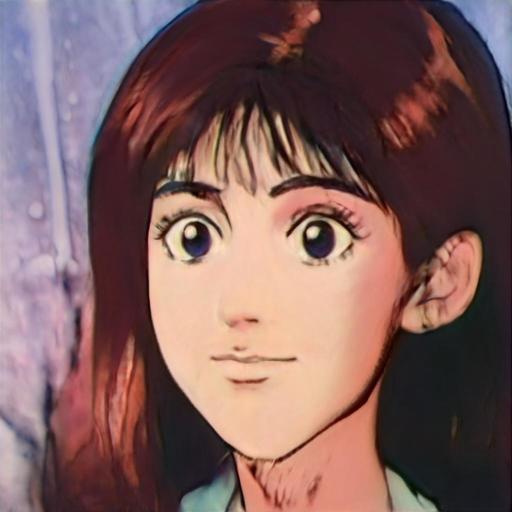}        
                \tabularnewline				
				Input & Inversion & Smile  & Age & Input & Inversion & Smile  & Pose
		\end{tabular}
		}
\caption{Image editing via our StyleGAN Inversion method. For each example, we show from left to right: the input image, the inverted image, and the edited images using off-the-shelf editing techniques. Our method enables high-quality reconstruction and editing for out-of-domain cartoon images.}
\label{fig:teaser}
\end{teaserfigure}

\maketitle

\pagestyle{plain}

\section{Introduction}
In recent years, Generative Adversarial Networks (GANs) \cite{Goodfellow2014} have revolutionized unconditional image synthesis. State-of-the-art models, especially StyleGAN \cite{Karras2019,Karras2020,Karras2020_2}, can now generate realistic and visually-appealing images in various domains. Furthermore, the intermediate latent space of StyleGAN, which is obtained from the input latent space through a mapping network, has been demonstrated as holding the disentanglement property. Based on this property, there emerges numerous models \cite{Karras2020,Shen2020,Hussein2020,Abdal2019,Richardson2021,Tov2021,Shen2021,Patashnik2021,Wang2021_4,Roich2021,Chong2021,Shukor2021,Wu2021_2} for StyleGAN inversion and real image manipulation. 

The target of StyleGAN inversion is to invert an input image into StyleGAN's latent space. Typically, there are two types of latent spaces for StyleGAN inversion. One is StyleGAN's native latent space $\mathcal{W}$ \cite{Shen2020,Jahanian2020,Tewari2020,Erik2020,Abdal2021}, where the style code is a 512-dimensional vector, and the other is an extended latent space $\mathcal{W}+$ \cite{Abdal2019,Abdal2020,Zhu2020,Richardson2021,Tov2021}, where the style code consists of 18 different 512-dimensional vectors. It has been shown \cite{Zhu2020_2,Tov2021,Alaluf2021,Roich2021} that the original space $\mathcal{W}$ is more editable while the extended space $\mathcal{W}+$ is more expressive.

Recently, Roich \etal \cite{Roich2021} propose Pivotal Tuning Inversion (PTI), a two-step approach that achieves significant progress towards reconstruction and editability in StyleGAN inversion. PTI first inverts the input image into a latent code, called pivot code, using an existing per-image optimization inversion method \cite{Karras2020}, and then slightly alters the generator so that the input image can be accurately mapped to the pivot code. It is revealed \cite{Roich2021} that the quality of the pivot code is of crucial importance to the final inversion. To decrease the distortion of the pivot code, one simple method is to apply more training steps during the first step. However, we observe that the per-image optimization method can easily cause overfitting and thereby lead to poor editability. 

\begin{figure*}
\centering
\begin{tabular}{c}
\includegraphics[width=.99\textwidth]{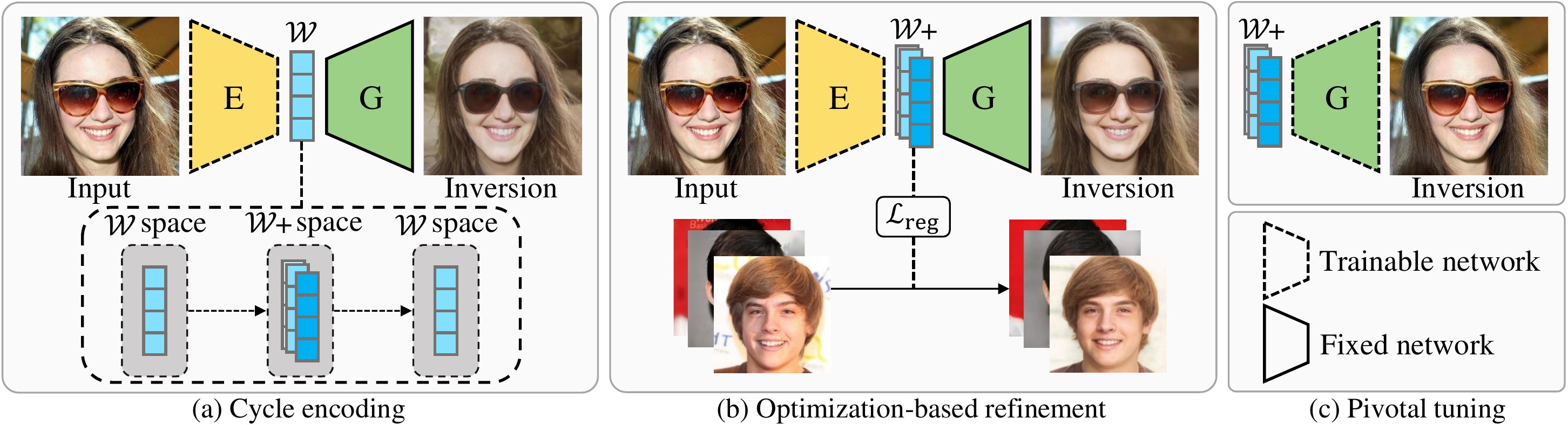}
\end{tabular}
\caption{The framework of our StyleGAN inversion method. (a) Cycle encoding: the encoder is trained in varying spaces according to a cycle scheme: $\mathcal{W}$$\rightarrow$$\mathcal{W}+$$\rightarrow$$\mathcal{W}$. (b) Optimization-based refinement: the encoder is iteratively updated towards the input image where a regularization term is introduced to prevent overfitting. (c) Pivotal tuning: the generator is slightly tuned so that the input image can be accurately mapped to the latent code.}
\label{fig:framework}
\end{figure*}

Instead of using the per-image optimization method, we propose an encoder-based method, named cycle encoding, for a high-quality pivot code. The idea is to preserve the properties of both \W and \Wp spaces, by progressively training an encoder according to a cycle scheme: $\mathcal{W}$$\rightarrow$$\mathcal{W}+$$\rightarrow$$\mathcal{W}$. Specifically, the encoder starts the training in the \W space, then gradually shifts the space from $\mathcal{W}$ to $\mathcal{W}+$, and finally shifts it back from $\mathcal{W}+$ to $\mathcal{W}$. We demonstrate that our method yields lower distortion, higher editability, and less inference time compared to PTI.

Recent evidence \cite{Zhu2020,Guan2020} reveals that the hybrid method exploits the advantages of both encoder-based and optimization-based methods. Inspired by this, we also propose to refine the pivot code obtained in cycle encoding by applying an optimization-based method. Different from existing per-image optimization methods \cite{Abdal2019,Karras2020} which optimize the latent code directly, our method refines the reconstruction by updating the encoder towards the input image so that an additional regularization term can be used to prevent overfitting. This refinement mechanism decreases the distortion at the cost of a subtle degradation in editability. Figure \ref{fig:framework} shows the framework of our approach.

We compare our method with several state-of-the-art StyleGAN inversion methods through qualitative and quantitative evaluation, and demonstrate that our method outperforms these methods in both reconstruction and editability. In Figure \ref{fig:teaser}, we show that our method enables high-quality reconstruction and editing even for out-of-domain cartoon images. Our code is available at https://github.com/xudonmao/CycleEncoding.

\section{Related Work}

\subsection{Latent Space Embedding}
The high visual quality of GAN synthesis \cite{Karras2018,Brock2018,Karras2019,Karras2020} has prompted the researchers to study the latent space of GAN. One fundamental task is GAN inversion \cite{Nguyen2016,Zhu2016}, where a given image is inverted into the latent space of GAN. In general, GAN inversion methods can typically be divided into three categories \cite{Xia2021}: (1) optimization-based methods \cite{Lipton2017,Creswell2018,Raj2019,Abdal2019,Collins2020,Tewari2020_2,Menon2020,Huh2020,Abdal2020,Karras2020,Futschik2021} which directly optimize over the latent code, (2) encoder-based methods \cite{Perarnau2016,Luo2017,Guan2020,Bau2019,Pidhorskyi2020,Chai2021,Alaluf2021_2,Richardson2021,Nitzan2020,Tov2021,Wei2021,Wang2021_3,Kim2021} which learn an encoder to map the input image into the latent space, and (3) hybrid methods \cite{Zhu2016,Baylies2019,Zhu2020,Pan2020,Zhang2021} which combine the above two methods. Typically, optimization-based methods achieve lower distortion but take a substantially longer time for computation compared to encoder-based methods. Specifically, Abdal \etal \cite{Abdal2019} optimize the latent code in the extended $\mathcal{W}+$ space and show that even out-of-domain images can be reconstructed. Karras \etal \cite{Karras2020} perform the optimization over not only the latent code in the original \W space but also the stochastic noise inputs of the StyleGAN generator. Richardson \etal \cite{Richardson2021} introduce a feature pyramid network architecture for the encoder which encodes the input image into the $\mathcal{W}+$ space. Roich \etal \cite{Roich2021} propose to first invert the input image into a latent code using the optimization-based method and then slightly alter the generator such that the input image can be accurately mapped to the latent code.

\subsection{Latent Space Manipulation}
To edit a real image, one may first invert the image into the latent space, and then perform the latent space manipulation techniques. Numerous methods have been proposed to find semantically meaningful directions in the latent space of GANs, where semantic directions can be determined through fully-supervised approaches \cite{Goetschalckx2019,Shen2020,Viazovetskyi2020,Wu2021,Abdal2021,Zhuang2021}, self-supervised approaches \cite{Jahanian2020,Tewari2020,Plumerault2020,Spingarn-Eliezer2021}, or unsupervised approaches \cite{Erik2020,Voynov2020,Cherepkov2021,Wang2021,Shen2021,Wang2021_2,Li2021}. Specifically, Jahanian \etal \cite{Jahanian2020} find semantic directions for camera motion and color transformation in a self-supervised manner. Shen \etal \cite{Shen2020} use binary facial attribute labels to determine semantic directions. H{\"a}rk{\"o}nen \etal \cite{Erik2020} show that using principal component analysis can identify meaningful semantic directions in an unsupervised manner. 

\subsection{Distortion-editability Tradeoff}
The $\mathcal{W}+$ space is superior in achieving low distortion because it is an enlarged space and thus more expressive. However, recent works \cite{Zhu2020_2,Tov2021,Roich2021} show that the \W space obtains better editability than the \Wp space, since StyleGAN is originally trained on this space. Tov \etal \cite{Tov2021} analyze the distortion-editability tradeoff and present an encoder-based method to balance the tradeoff. Zhu \etal \cite{Zhu2020_2} introduce a new normalized space and a regularization term to address the distortion-editability tradeoff. Roich \etal \cite{Roich2021} mitigate the distortion-editability tradeoff by combining the editability of the \W space with an accurate reconstruction technique which slightly alters the generator.

\section{Analysis of Pivotal Tuning Inversion}
\label{sec:analysis}
\subsection{Pivotal Tuning Inversion}
PTI \cite{Roich2021} is a two-step method for StyleGAN inversion. Different from previous methods that find the latent code within the StyleGAN's latent space, PTI augments the latent space by slightly altering the generator. Specifically, in the first step, PTI inverts the input image $x$ into a latent code $w_p \in \mathcal{W}$, called pivot code, using an existing optimization-based method \cite{Karras2020} which optimizes the following objective:
\begin{align}
\label{eq:optimization}
w_p = \underset{w, n}{\arg\min} \mathcal{L}_{\text{LPIPS}}(x, G(w, n)) + \lambda_{n}\mathcal{L}_{n}(n),
\end{align}
where $G(w, n)$ is the generated image by the generator $G$, $\mathcal{L}_{\text{LPIPS}}$ is the LPIPS perceptual loss \cite{Zhang2018}, $n$ is a noise vector, $\mathcal{L}_{n}$ is a noise regularization term, and $\lambda_{n}$ controls the weight of $\mathcal{L}_{n}$.
In the second step, the generator $G$ is tuned so that the input image $x$ can be accurately mapped to the pivot code $w_p$ by optimizing:
\begin{align}
\label{eq:pti}
\mathcal{L}_{\text{PTI}}(x) =  \mathcal{L}_{\text{LPIPS}}(x, G(w_p)) + \lambda_{\text{L2}}\mathcal{L}_{\text{L2}}(x, G(w_p)),
\end{align}
where $\mathcal{L}_{\text{L2}}$ is the pixel-wise L2 loss and $\lambda_{\text{L2}}$ controls the loss weight.

\begin{figure}
\setlength{\tabcolsep}{0.5pt}
\centering
{\small
	\renewcommand{\arraystretch}{0.5}
    \begin{tabular}{c c c c c c c}
        \raisebox{0.18in}{\rotatebox[origin=t]{90}{PTI}}&
        \includegraphics[width=0.075\textwidth]{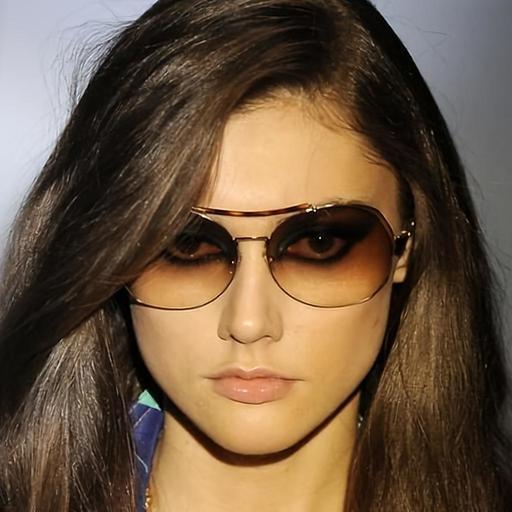}&
        \includegraphics[width=0.075\textwidth]{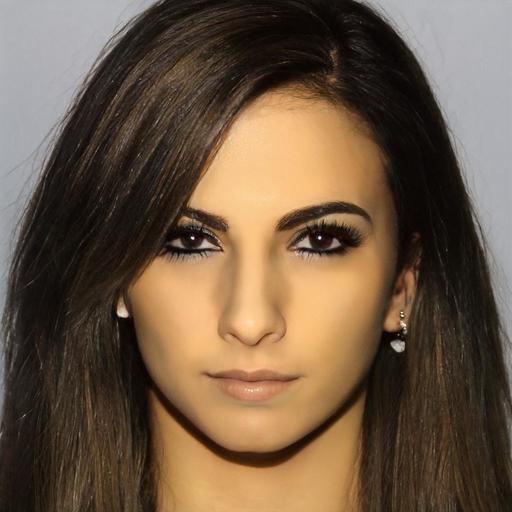}&        
        \includegraphics[width=0.075\textwidth]{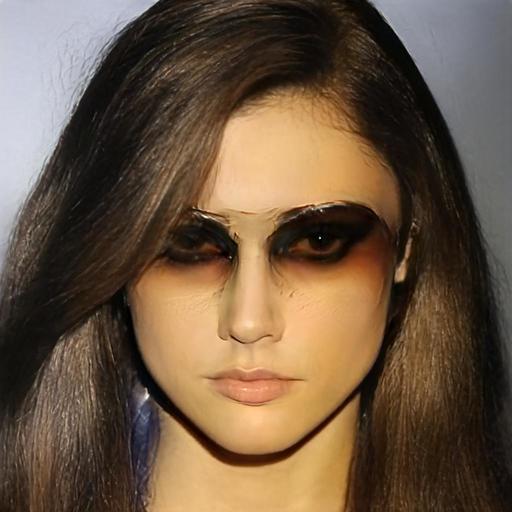}&        
        \includegraphics[width=0.075\textwidth]{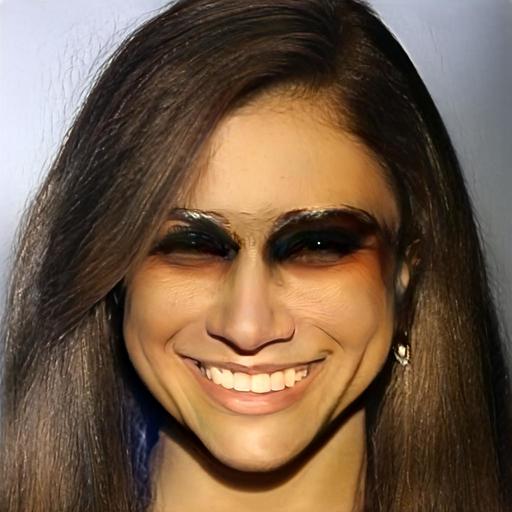}&
        \includegraphics[width=0.075\textwidth]{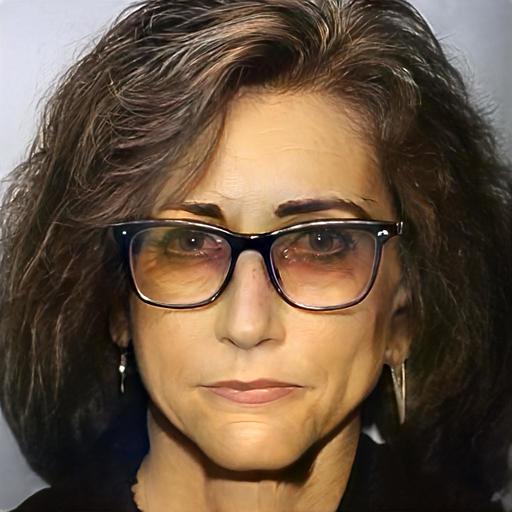}&
        \includegraphics[width=0.075\textwidth]{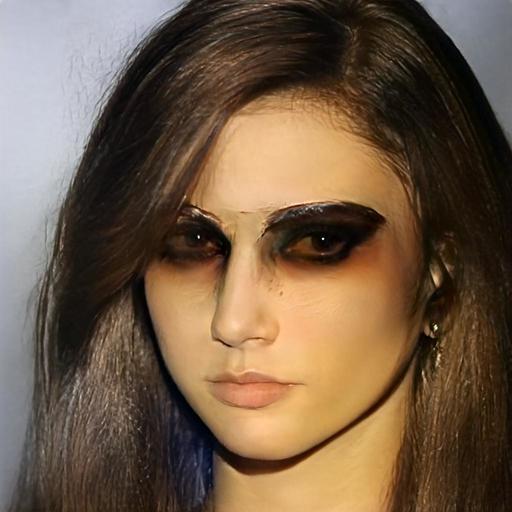} 
		\\
        \raisebox{0.17in}{\rotatebox[origin=t]{90}{PTI++}}&
		\includegraphics[width=0.075\textwidth]{images/analysis/28872_real.jpg}&
        \includegraphics[width=0.075\textwidth]{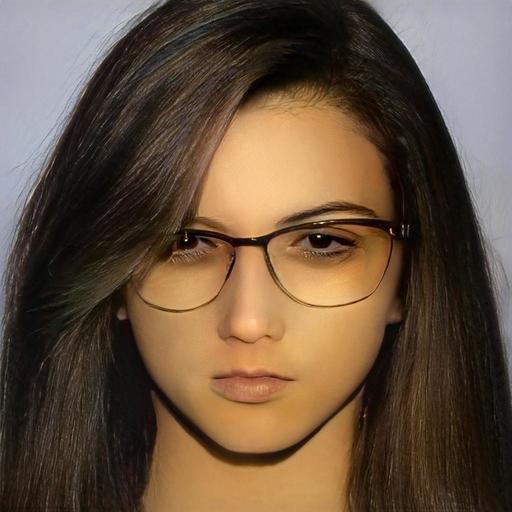}&        
        \includegraphics[width=0.075\textwidth]{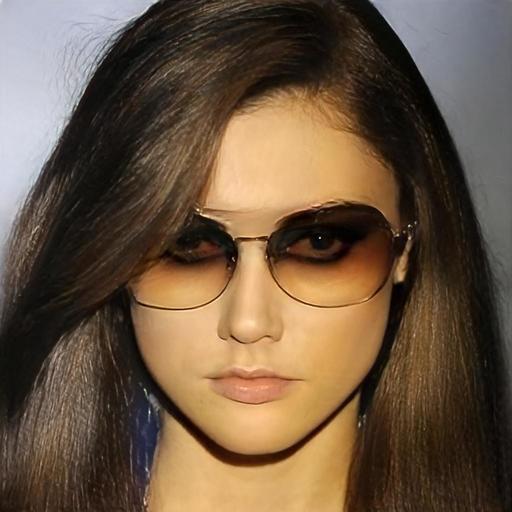}&        
        \includegraphics[width=0.075\textwidth]{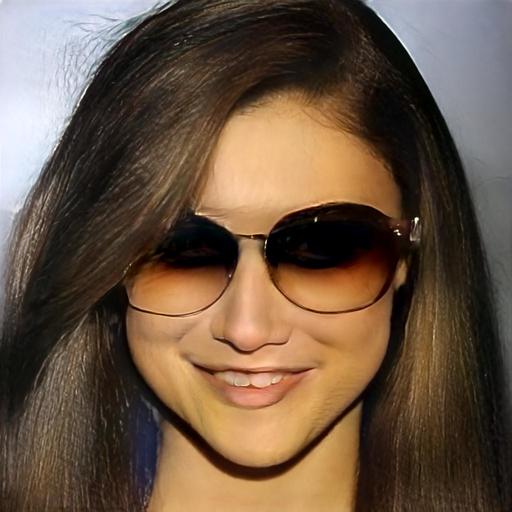}&
        \includegraphics[width=0.075\textwidth]{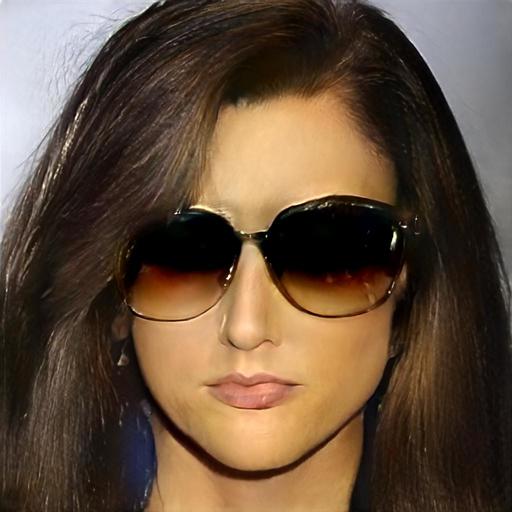}&
        \includegraphics[width=0.075\textwidth]{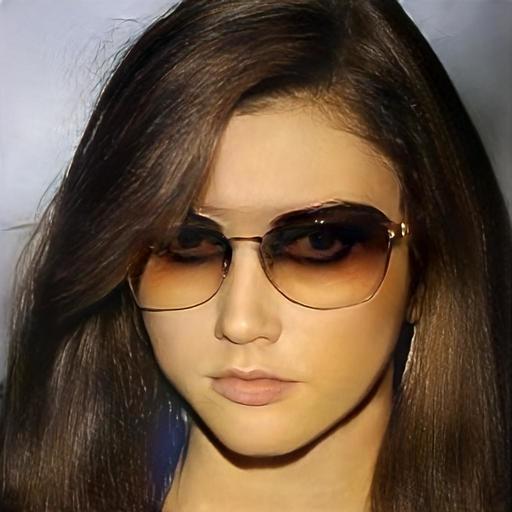} 
		\\
		& Input & First Inv. & Final Inv. & Smile & Age & Pose
    \end{tabular}
    }
	\caption{PTI++ denotes the model that is trained with more training steps in the first step (Eq. \ref{eq:optimization}). PTI++ achieves low distortion but poor editability.}
    \label{fig:analysis}
\end{figure}

\subsection{Distortion-Editability Tradeoff}
Although the distortion can be diminished significantly during the second step of PTI, the distortion in the first step is of crucial importance to the final inversion. Figure \ref{fig:analysis} shows an example, and one can see that lower distortion in the first step (Column 2) leads to lower distortion of the final inversion (Column 3). One simple method to decrease the distortion is to apply more training steps in the first step (Eq. \ref{eq:optimization}). However, we observe that applying more training steps will decrease the editability of the final inversion, as shown in Figure \ref{fig:analysis}. The reason may be that the per-image optimization method suffers from overfitting when training too many steps on a single image. Therefore, the per-image optimization method for the pivot code can hardly balance the distortion-editability tradeoff. We instead train an encoder using a novel training methodology for the pivot code, resulting in a more accurate, more editable, and faster inversion.

\section{Method}
Our target is to improve the quality of the pivot code in PTI. We adopt the encoder-based method to infer the pivot code, because the per-image optimization method can hardly balance the distortion-editability tradeoff as discussed in Section \ref{sec:analysis}. Recent studies \cite{Zhu2020_2,Tov2021,Alaluf2021} reveal that the extended space \Wp is more expressive while the original space \W is more editable. The key idea of our method is to exploit the advantages of both \Wp and \W spaces by training an encoder in both spaces. We propose a novel training methodology that gradually changes the output space of the encoder according to a cycle scheme: $\mathcal{W}$$\rightarrow$$\mathcal{W}+$$\rightarrow$$\mathcal{W}$. To further decrease the distortion, we then follow the hybrid method \cite{Zhu2020,Guan2020} to refine the pivot code obtained from the encoder, by iteratively updating the encoder towards the input image where a regularization term is introduced to alleviate the overfitting problem. Finally, using the latent code obtained from cycle encoding or from the refinement as the pivot code, the generator is slightly tuned so that the input image can be accurately mapped to the pivot code. Figure \ref{fig:framework} shows the framework of our method.

\begin{figure}
    \setlength{\tabcolsep}{0.5pt}
    \centering
        {\small
        \begin{tabular}{c c c c c}
                \includegraphics[width=0.091\textwidth]{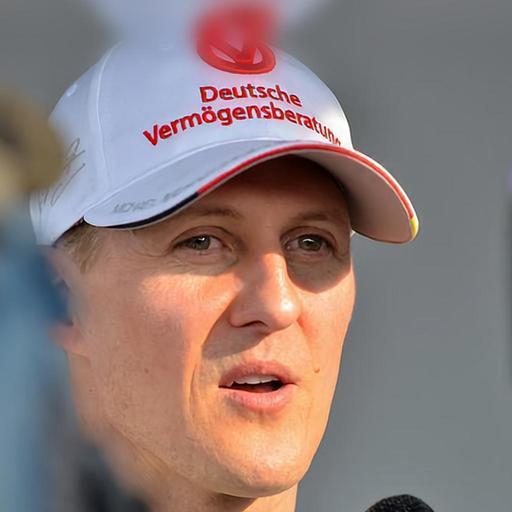}&
                \includegraphics[width=0.091\textwidth]{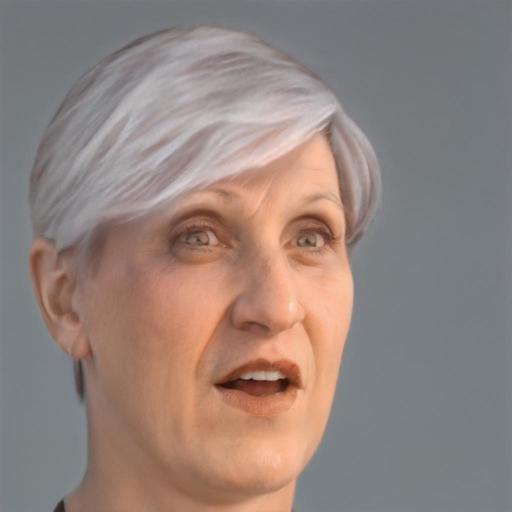}&
                \includegraphics[width=0.091\textwidth]{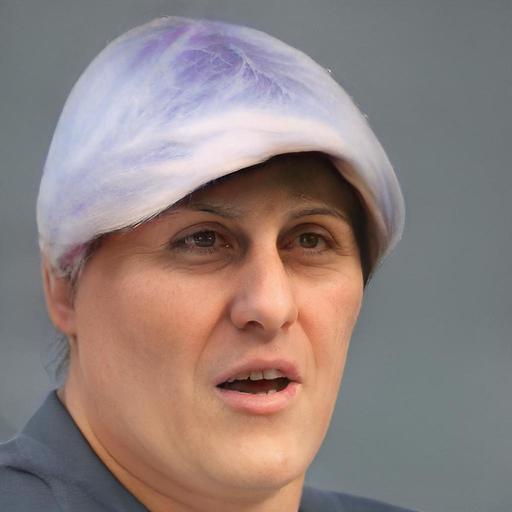}&
                \includegraphics[width=0.091\textwidth]{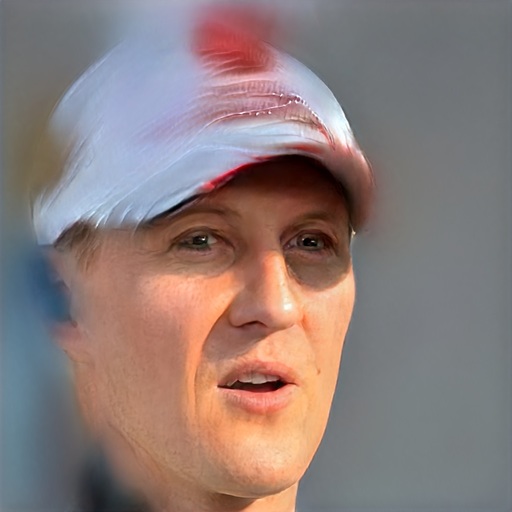}&
                \includegraphics[width=0.091\textwidth]{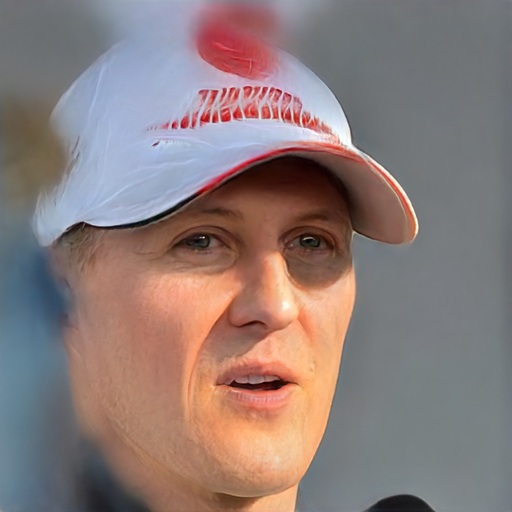}        
                \tabularnewline
                \includegraphics[width=0.091\textwidth]{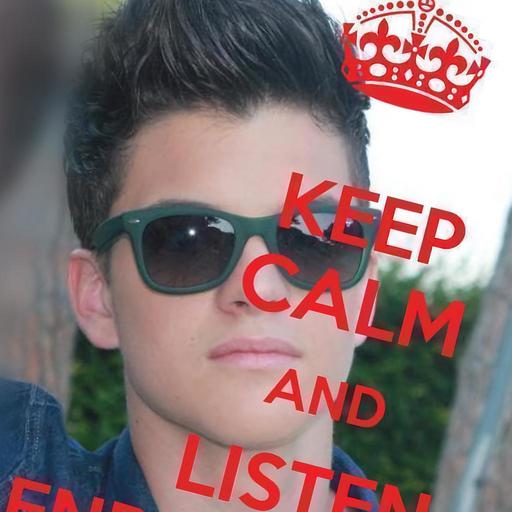}&
                \includegraphics[width=0.091\textwidth]{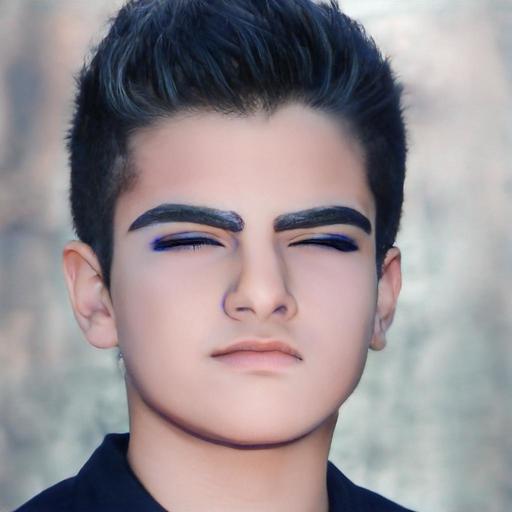}&
                \includegraphics[width=0.091\textwidth]{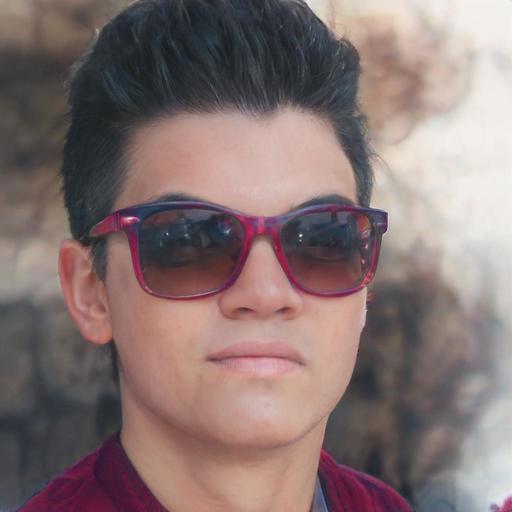}&
                \includegraphics[width=0.091\textwidth]{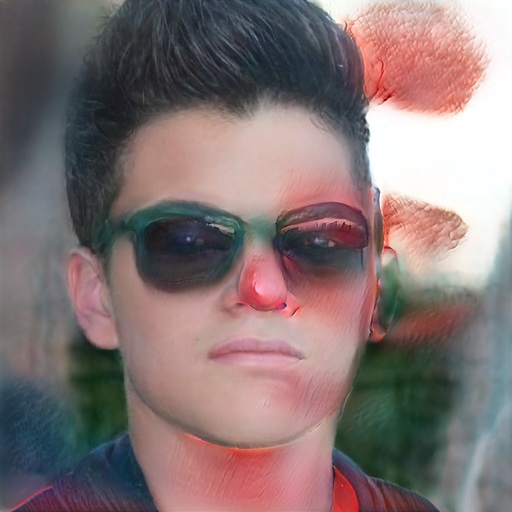}&
                \includegraphics[width=0.091\textwidth]{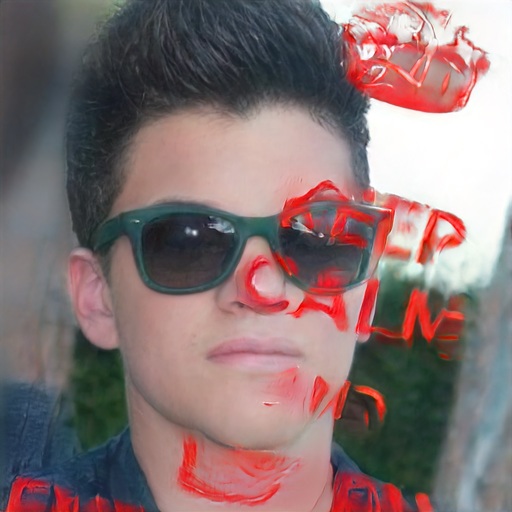}        
                \tabularnewline
                \includegraphics[width=0.091\textwidth]{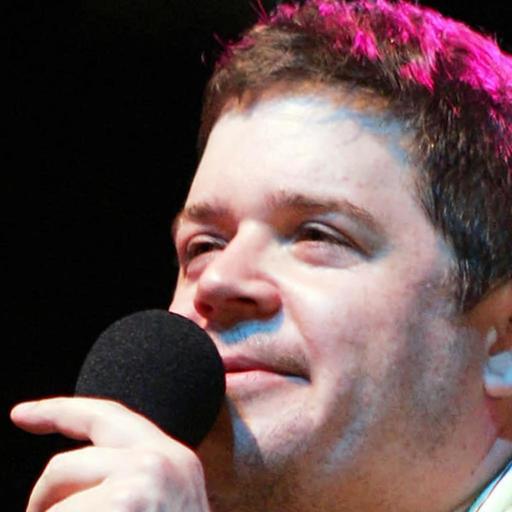}&
                \includegraphics[width=0.091\textwidth]{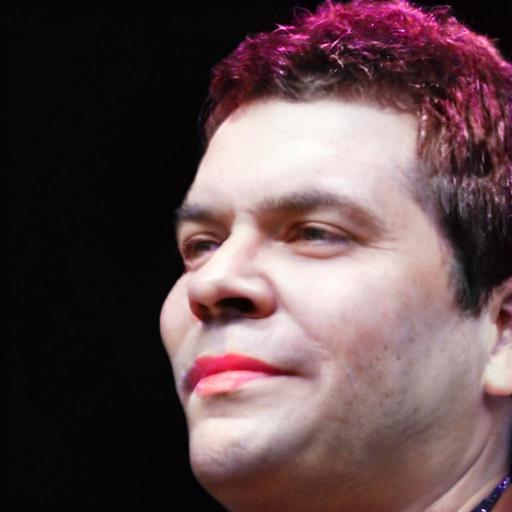}&
                \includegraphics[width=0.091\textwidth]{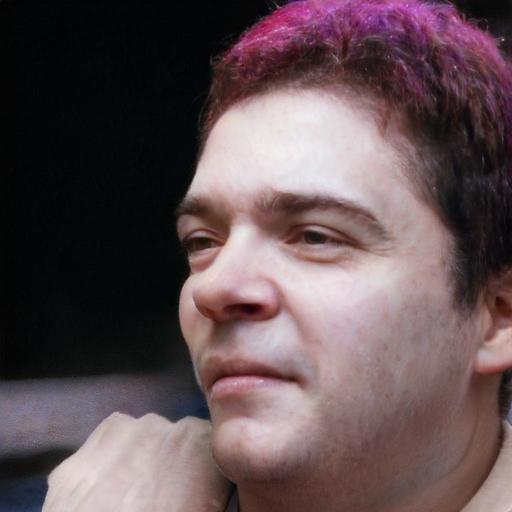}&
                \includegraphics[width=0.091\textwidth]{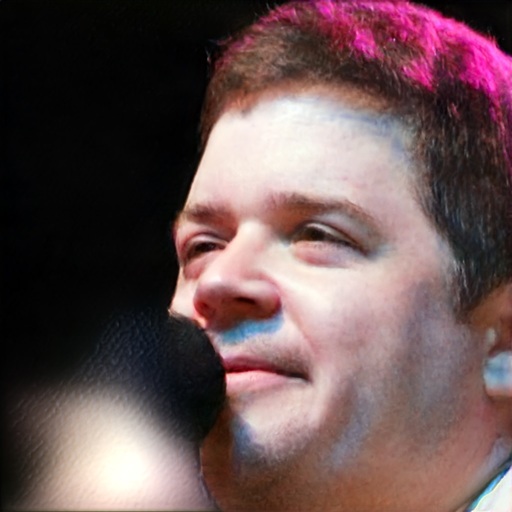}&
                \includegraphics[width=0.091\textwidth]{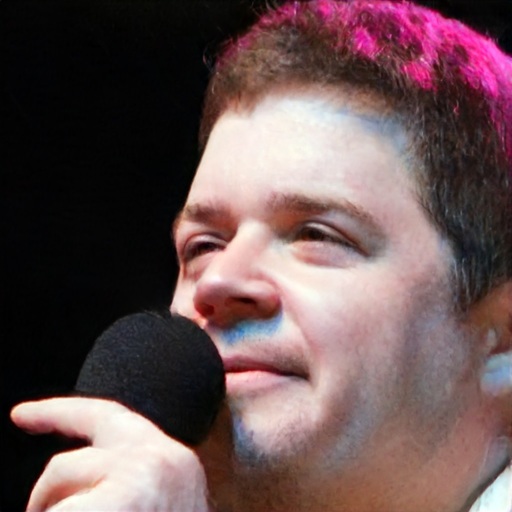}
                \tabularnewline
                Input & SG2 & e4e & PTI & Ours
		\end{tabular}
		}
\caption{Reconstruction quality comparison on CelebA-HQ.}
\label{fig:reconstruction_celeba}
\end{figure}

\subsection{Cycle Encoding}
\label{sec:cycle_encoding}
In cycle encoding, we seek to achieve optimal editability for the pivot code. To this end, we select the \W space as the final state for cycle encoding, as the \W space provides better editability than the \Wp space \cite{Tov2021}. Nevertheless, the \Wp space is more expressive and induces less distortion. To exploit both advantages of the \W and \Wp spaces, we propose to train an encoder in varying spaces according to a cycle scheme: $\mathcal{W}$$\rightarrow$$\mathcal{W}+$$\rightarrow$$\mathcal{W}$. Starting and ending with the \W space provide high editability for the pivot code, and shifting to the \Wp space provides high expressiveness for the pivot code. For $\mathcal{W}$$\rightarrow$$\mathcal{W}+$, we follow \cite{Tov2021} to sequentially allow the latent vectors to be different to the first latent vector. For $\mathcal{W}+$$\rightarrow$$\mathcal{W}$, it is more crucial to the final inversion, as it determines the final state of the pivot code. We propose a new progressive training methodology for $\mathcal{W}+$$\rightarrow$$\mathcal{W}$, which changes the space more smoothly. In the following, we detail the training mechanism that consists of two steps:

\subsubsection{$\mathcal{W}$$\rightarrow$$\mathcal{W}+$}
\label{sec:w2wp}
\hfill \break
We follow the e4e model \cite{Tov2021} to control the output space of the encoder by using the delta regularization loss which measures the difference between each latent vector. Formally, let $E(x)=(w_0, w_1, ..., $ $w_{N - 1})=(w_0, w_0 + \Delta_1, ..., w_0 + \Delta_{N - 1})$ denote the output of the encoder, where $N$ is the number of the latent vectors and $\Delta_i$ is the offset from the first latent vector $w_0$. The delta regularization loss is defined as:
\begin{equation}
\label{eq:delta}
    \mathcal{L_{\text{delta}}}(x) = \sum_{i=1}^{N-1}|| \Delta_i ||_2.
\end{equation}
The encoder starts the training in the \W space by setting $\forall i: \Delta_i=0$, and then gradually shifts the space from \W to \Wp by training $\Delta_i$ sequentially every $T_0$ iterations. 

\subsubsection{$\mathcal{W}+$$\rightarrow$$\mathcal{W}$}
\label{sec:wp2w}
\hfill \break
In this step, we also utilize the delta regularization loss to control the shift from \Wp to \W. We empirically find that smoothly changing the output space of the encoder is beneficial to the editability of the pivot code. To this end, we propose a new progressive training methodology that shifts the space from \Wp to \W more smoothly. Specifically, we first gradually increase the weight of the delta regularization loss by a factor $\beta$ every $T_1$ iterations. A large weight of the delta regularization loss enforces $\Delta_i$ to be very close to 0. After this procedure, the output space of the encoder lies close to the \W space with a small variance. Then, we set from $\Delta_{N-1}=0$ to $\Delta_1=0$ sequentially every $T_2$ iterations. Finally, the output space of the encoder ends at the \W space with $\forall i: \Delta_i=0$.

Increasing the weight of the delta regularization loss can be viewed as a ``soft'' operation, and setting $\Delta_i=0$ can be viewed as a ``hard'' operation. In short, our progressive training methodology first performs the soft operation and then performs the hard operation. Compared with directly performing the hard operation, our method changes the output space of the encoder more smoothly and favors better property preservation of the \W and \Wp spaces. 

The overall objective of cycle encoding is:
\begin{align}
\begin{split}
\label{eq:e4e_loss}
    \mathcal{L}(x) = & \lambda_{\text{L2}}\mathcal{L}_{\text{L2}}(x) + \lambda_{\text{lpips}}\mathcal{L}_{\text{LPIPS}}(x) + \lambda_{\text{id}}\mathcal{L}_{\text{ID}}(x) + \\
	& \lambda_{\text{adv}}\mathcal{L}_{\text{adv}}(x) + \lambda_{\text{delta}}\mathcal{L}_{\text{delta}}(x),
\end{split}
\end{align}
where $\mathcal{L}_{\text{ID}}$ is the identity loss \cite{Deng2019}, $\mathcal{L}_{\text{adv}}$ \cite{Nitzan2020,Tov2021} is an adversarial loss to encourage $w_i$ to lie close to the true distribution of \W, and $\lambda$ controls the weight of each loss.

\begin{figure}
    \setlength{\tabcolsep}{1pt}
    \centering
        {\small
        \begin{tabular}{c c c c c}
                \includegraphics[width=0.091\textwidth]{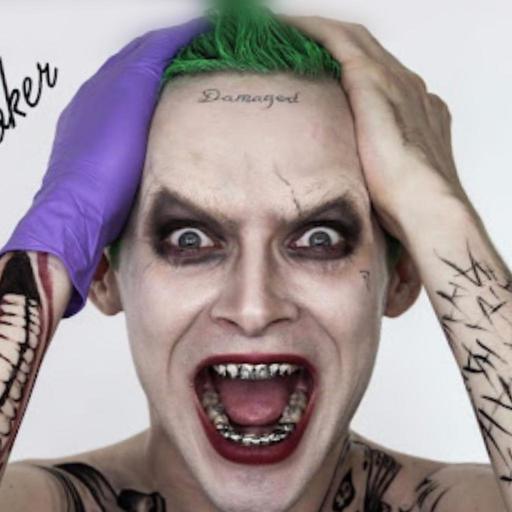}&
                \includegraphics[width=0.091\textwidth]{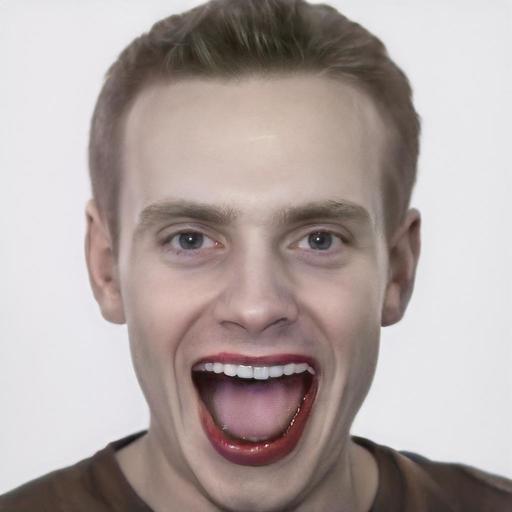}&
                \includegraphics[width=0.091\textwidth]{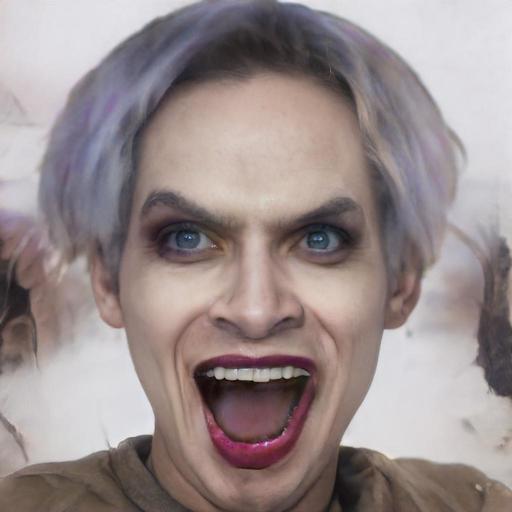}&
                \includegraphics[width=0.091\textwidth]{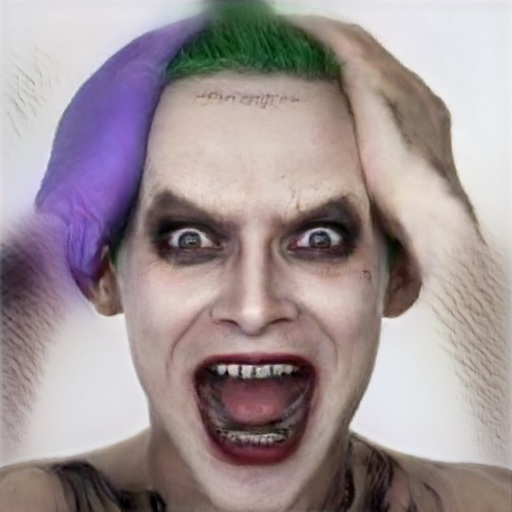}&
                \includegraphics[width=0.091\textwidth]{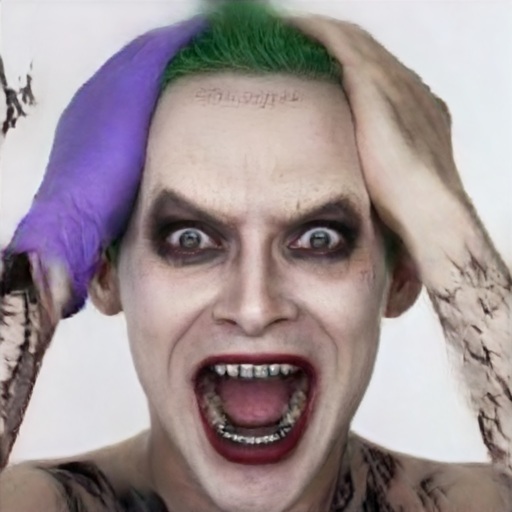}        
                \tabularnewline
				\includegraphics[width=0.091\textwidth]{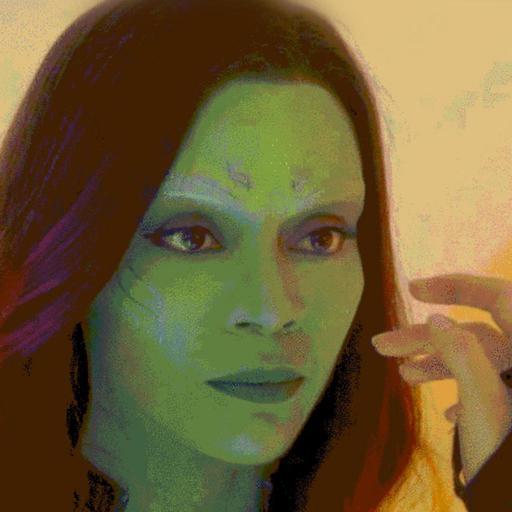}&
                \includegraphics[width=0.091\textwidth]{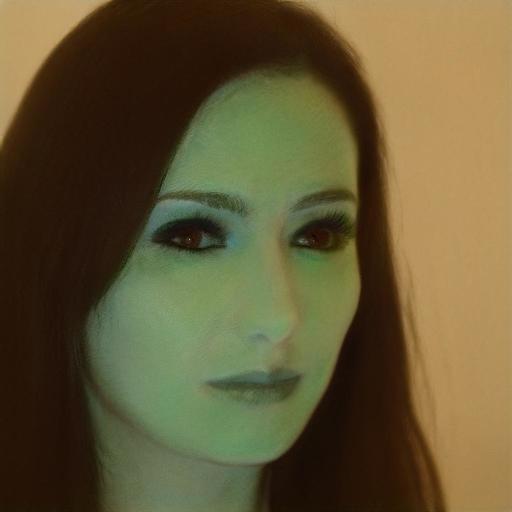}&
                \includegraphics[width=0.091\textwidth]{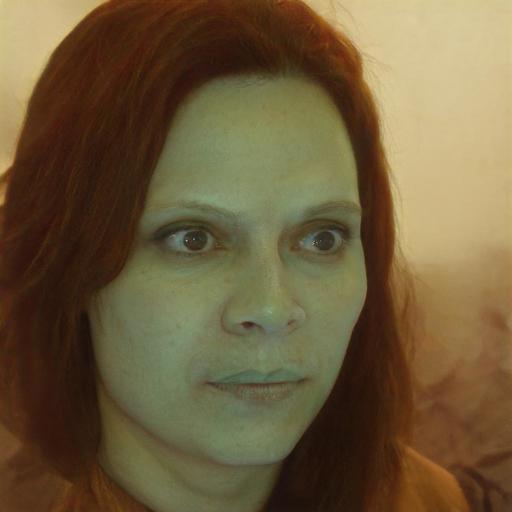}&
                \includegraphics[width=0.091\textwidth]{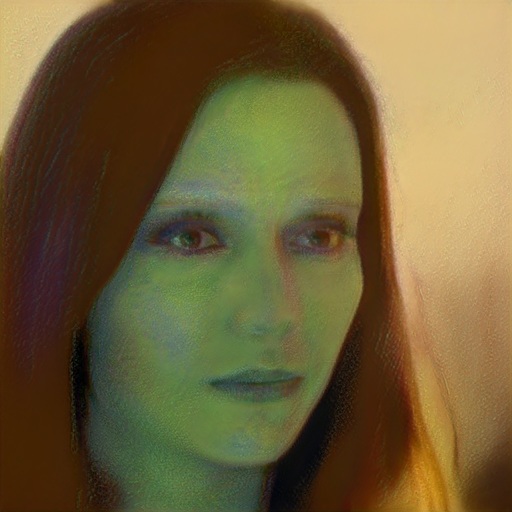}&
                \includegraphics[width=0.091\textwidth]{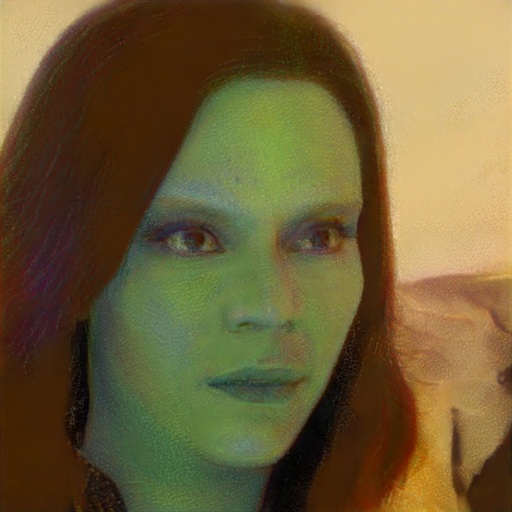}
                \tabularnewline
				\includegraphics[width=0.091\textwidth]{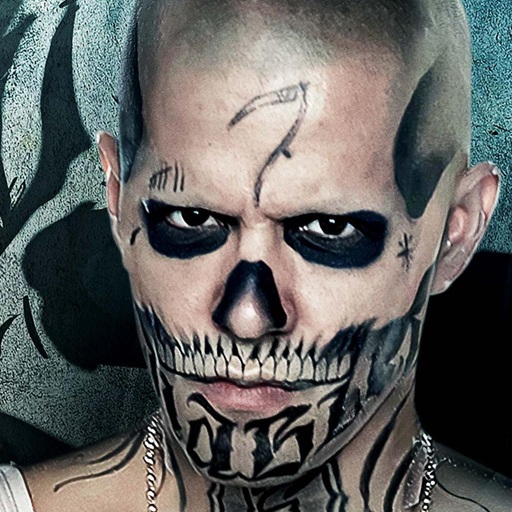}&
                \includegraphics[width=0.091\textwidth]{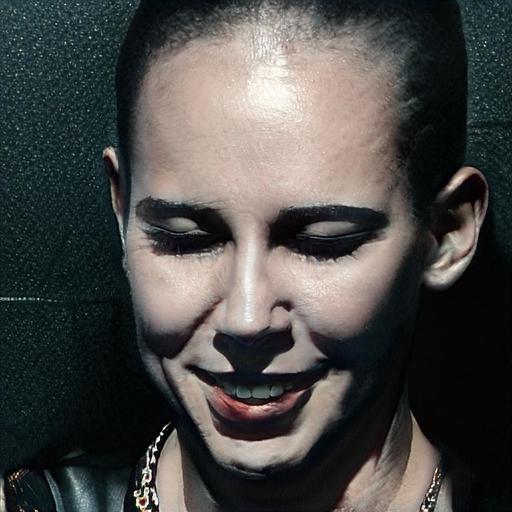}&
                \includegraphics[width=0.091\textwidth]{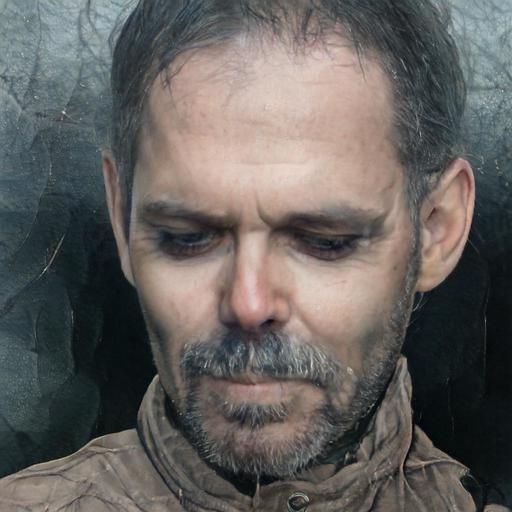}&
                \includegraphics[width=0.091\textwidth]{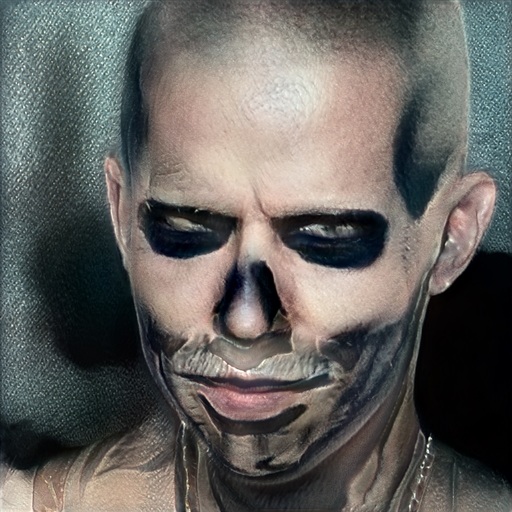}&
                \includegraphics[width=0.091\textwidth]{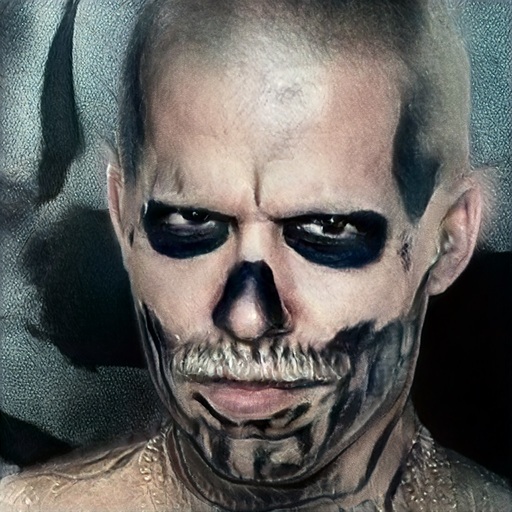}        
                \tabularnewline
                Input & SG2 & e4e & PTI & Ours
		\end{tabular}
		}
\caption{Reconstruction quality comparison using famous character images.}
\label{fig:reconstruction_characters}
\end{figure}

\subsection{Decreasing Distortion via Optimization}
\label{sec:optimization}
The hybrid method \cite{Zhu2020,Guan2020} has proved the effectiveness of utilizing the optimization-based method to refine the output of the encoder. Inspired by this, we also explore applying optimization to the result of cycle encoding to further decrease the distortion. However, as we discussed in Section \ref{sec:analysis}, optimizing over a single image is prone to overfitting, which then decreases the editability. Instead of optimizing the latent code, our method iteratively updates the encoder towards the input image where a regularization term is introduced to alleviate the overfitting problem. At each iteration, we randomly sample $M$ images $x^*$ from the training set and apply the following regularization term:
\begin{align}
\label{eq:refinement_reg}
\mathcal{L}_{\text{reg}}(x^*) =  \frac{1}{M}\sum^{M-1}_{i=0}\mathcal{L}_{\text{S}}(x_i^*, G(E(x_i^*))),
\end{align}
where $\mathcal{L}_{\text{S}} = \lambda_{\text{L2}}\mathcal{L}_{\text{L2}} + \lambda_{\text{lpips}}\mathcal{L}_{\text{LPIPS}} + \lambda_{\text{id}}\mathcal{L}_{\text{ID}}$. Then, the overall objective of this optimization step is:
\begin{align}
\label{eq:refinement_full}
\mathcal{L}_{\text{ref}}(x, x^*) =  \mathcal{L}_{\text{S}}(x) + \lambda_{\text{reg}} \mathcal{L}_{\text{reg}}(x^*),
\end{align}
where $x$ is the input image and $\lambda_{\text{reg}}$ controls the loss weight. Moreover, we apply this optimization in the \Wp space, as we empirically find that learning in the \Wp space in this step is cost-effective in decreasing the distortion with a subtle degradation in editability. The reason may be that slight and local changes to the pivot code can be applied without damaging its editing capability.

Note that this optimization step is optional in practice, and we recommend applying this step for challenging images as skipping this step reduces the inference time. Furthermore, a small number (15 in our experiments) of iterations is sufficient, since the reconstruction quality of cycle encoding is already quite good. 

\begin{figure}
    \setlength{\tabcolsep}{1pt}
    \centering
        {\small
        \begin{tabular}{c c c c c}
                \includegraphics[width=0.091\textwidth]{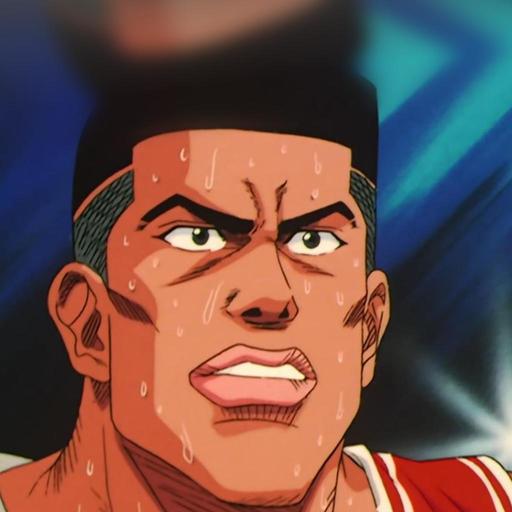}&
                \includegraphics[width=0.091\textwidth]{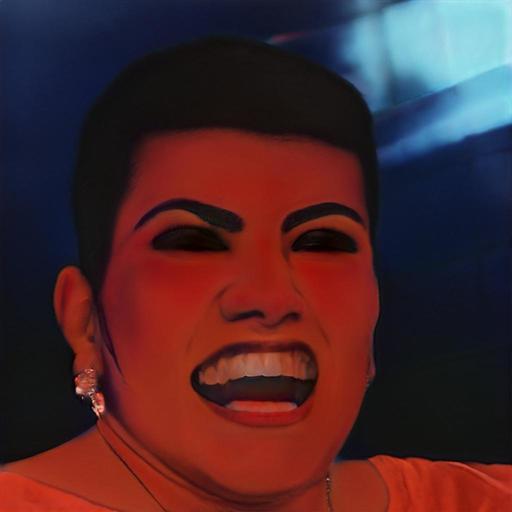}&
                \includegraphics[width=0.091\textwidth]{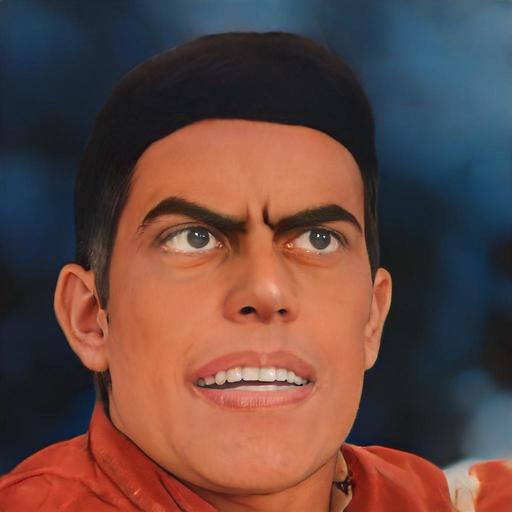}&
                \includegraphics[width=0.091\textwidth]{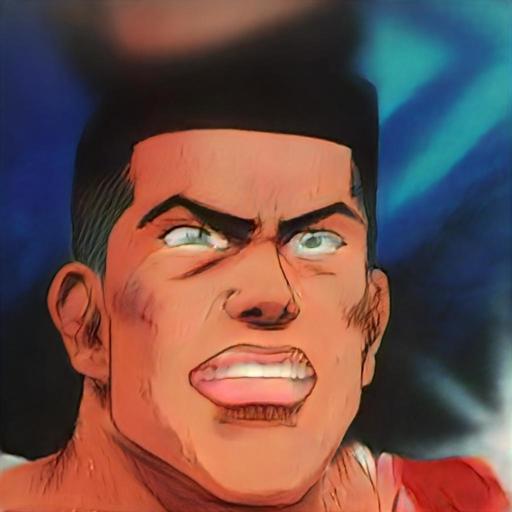}&
                \includegraphics[width=0.091\textwidth]{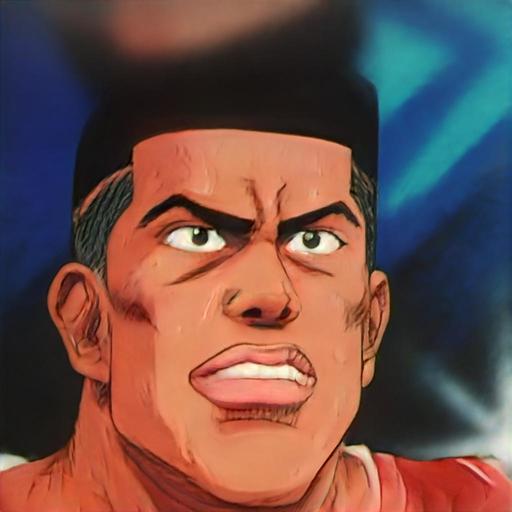}        
                \tabularnewline
				\includegraphics[width=0.091\textwidth]{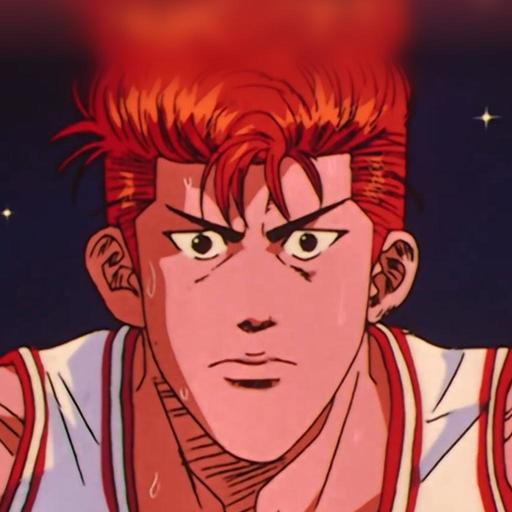}&
                \includegraphics[width=0.091\textwidth]{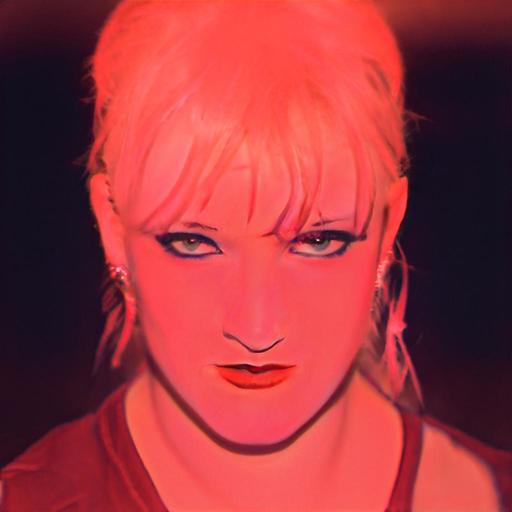}&
                \includegraphics[width=0.091\textwidth]{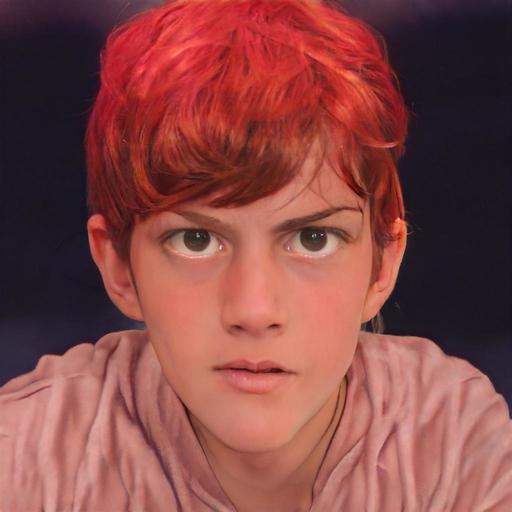}&
                \includegraphics[width=0.091\textwidth]{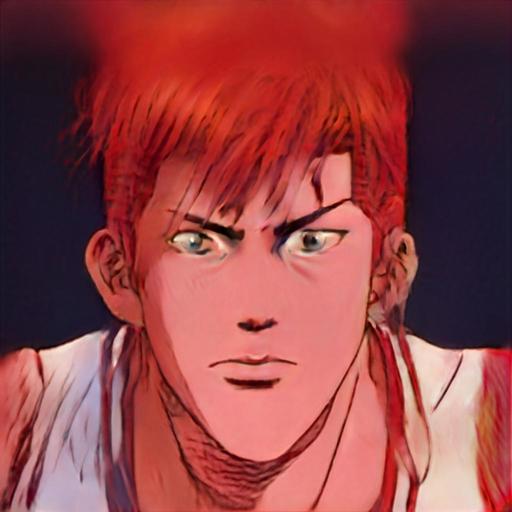}&
                \includegraphics[width=0.091\textwidth]{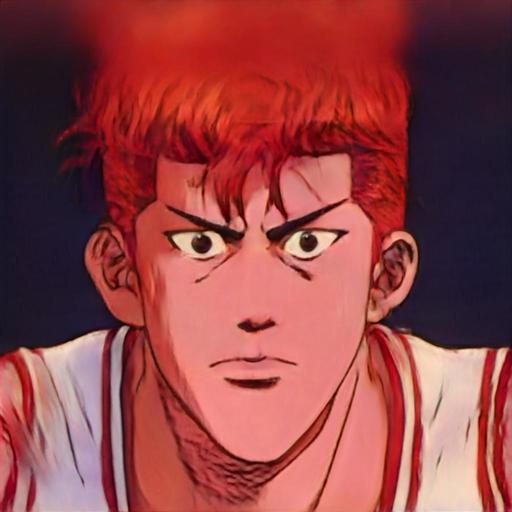}   
				\tabularnewline
				\includegraphics[width=0.091\textwidth]{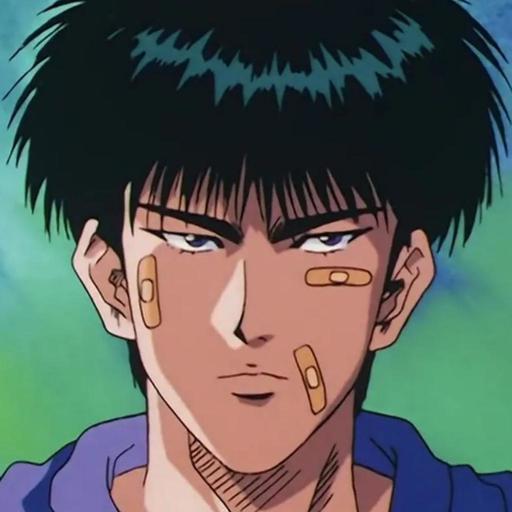}&
                \includegraphics[width=0.091\textwidth]{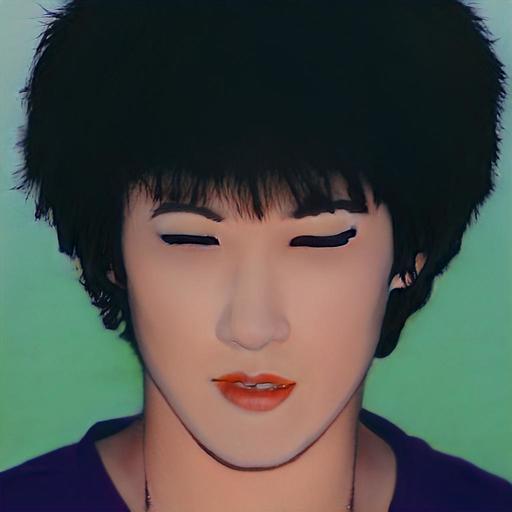}&
                \includegraphics[width=0.091\textwidth]{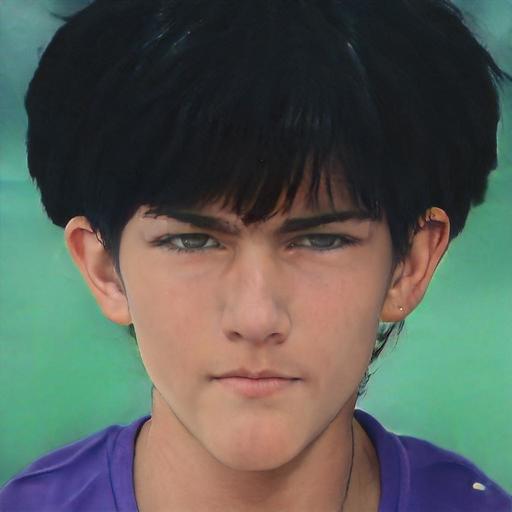}&
                \includegraphics[width=0.091\textwidth]{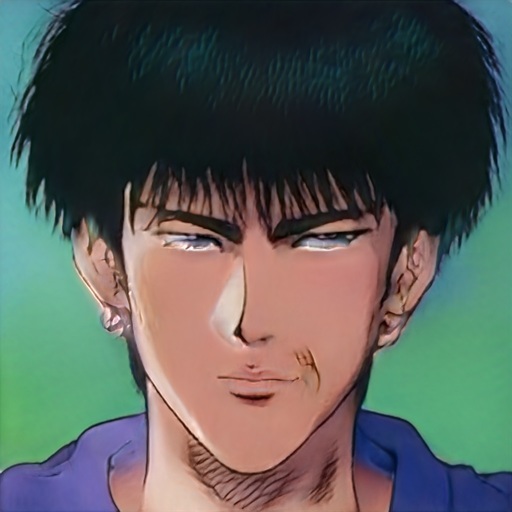}&
                \includegraphics[width=0.091\textwidth]{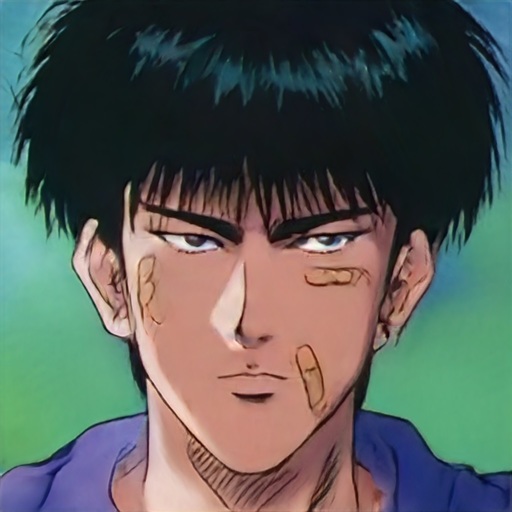}        
                \tabularnewline
                Input & SG2 & e4e & PTI & Ours
		\end{tabular}
		}
\caption{Reconstruction quality comparison using cartoon images.}
\label{fig:reconstruction_cartoon}
\end{figure}

\begin{figure}
\setlength{\tabcolsep}{1pt}
\centering
{\small
    \begin{tabular}{c c c c c c c}
        \raisebox{0.15in}{\rotatebox[origin=t]{90}{Input}}&
        \includegraphics[width=0.072\textwidth]{images/reconstruction/28064_real.jpg}&
        \includegraphics[width=0.072\textwidth]{images/reconstruction/28865_real.jpg}&        
        \includegraphics[width=0.072\textwidth]{images/reconstruction/41_real.jpg}&
        \includegraphics[width=0.072\textwidth]{images/reconstruction/167_real.jpg}&
        \includegraphics[width=0.072\textwidth]{images/reconstruction/239_real.jpg}&
        \includegraphics[width=0.072\textwidth]{images/reconstruction/232_real.jpg} 
		\tabularnewline
        \raisebox{0.15in}{\rotatebox[origin=t]{90}{PTI}}&
        \includegraphics[width=0.072\textwidth]{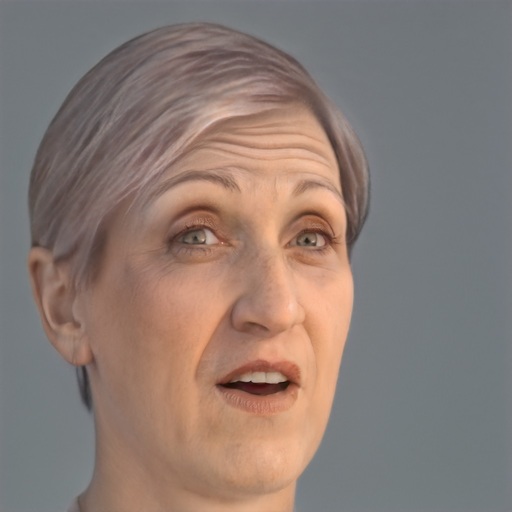}&
        \includegraphics[width=0.072\textwidth]{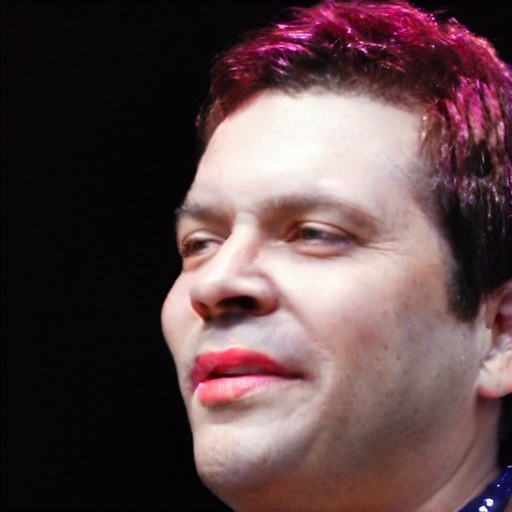}&        
        \includegraphics[width=0.072\textwidth]{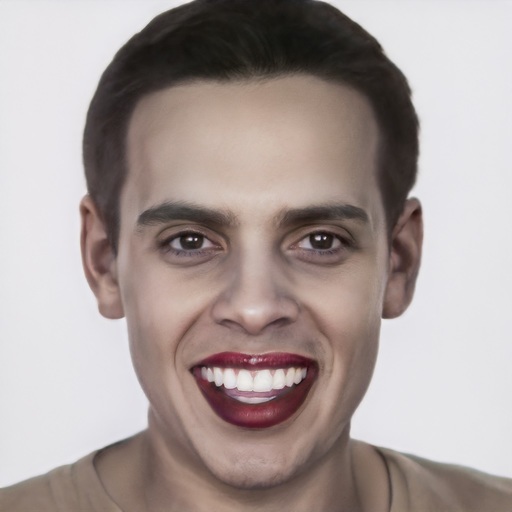}&
        \includegraphics[width=0.072\textwidth]{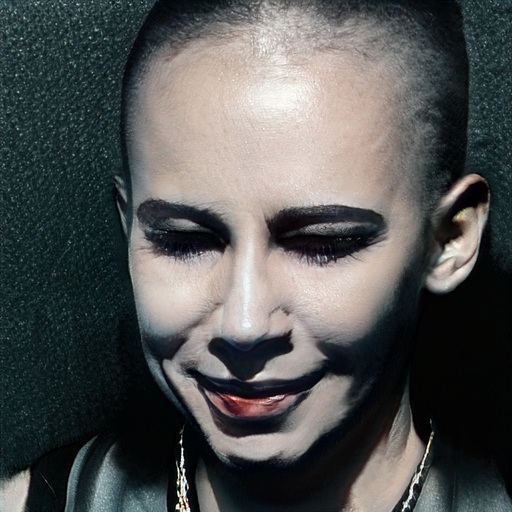}&
        \includegraphics[width=0.072\textwidth]{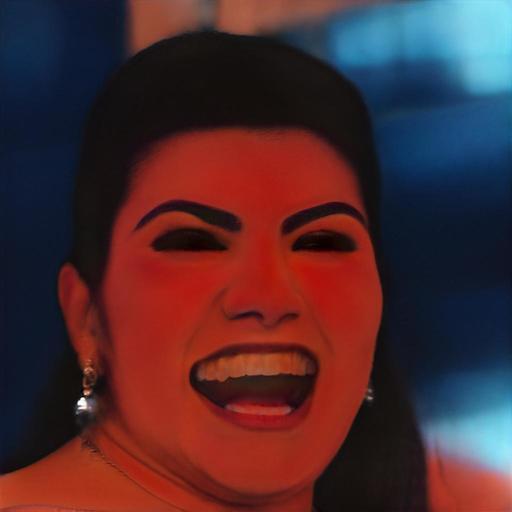}&
        \includegraphics[width=0.072\textwidth]{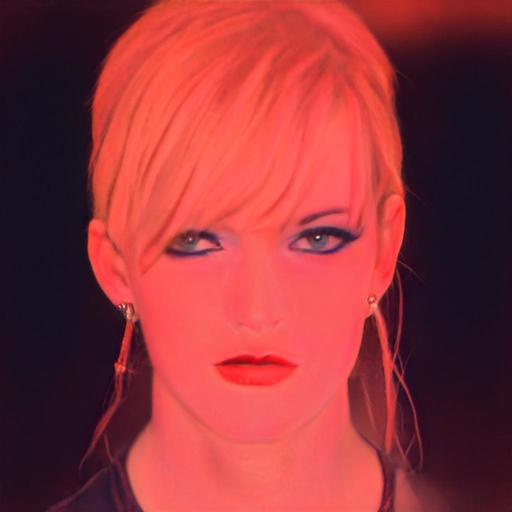} 
		\tabularnewline
        \raisebox{0.15in}{\rotatebox[origin=t]{90}{Ours}}&
        \includegraphics[width=0.072\textwidth]{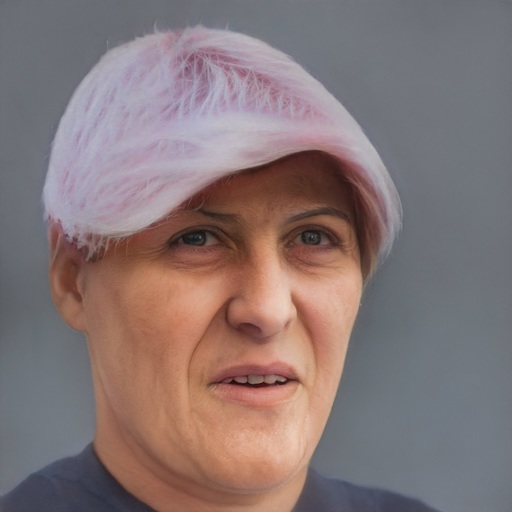}&
        \includegraphics[width=0.072\textwidth]{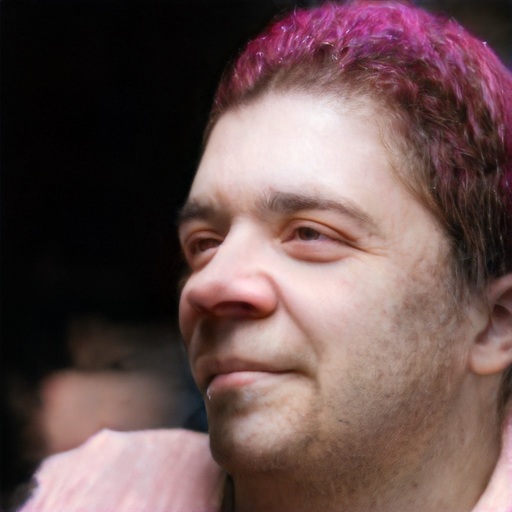}&        
        \includegraphics[width=0.072\textwidth]{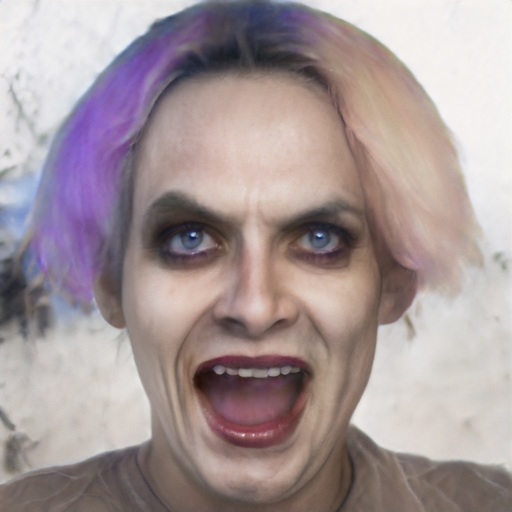}&
        \includegraphics[width=0.072\textwidth]{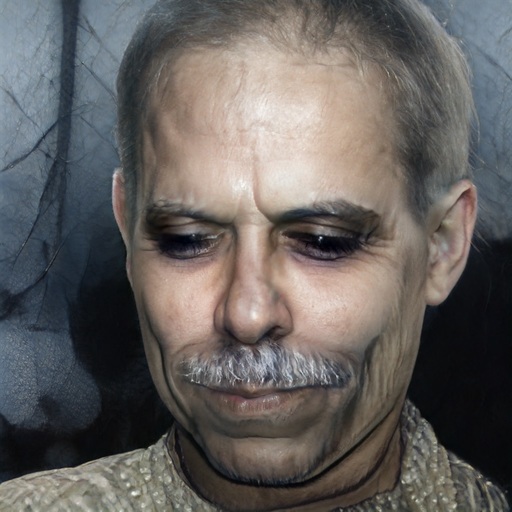}&
        \includegraphics[width=0.072\textwidth]{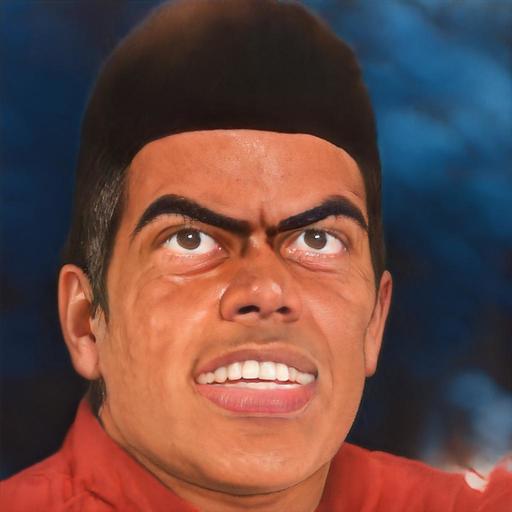}&
        \includegraphics[width=0.072\textwidth]{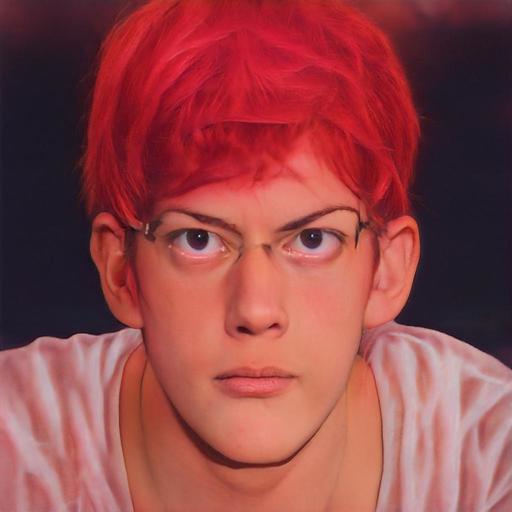} 
    \end{tabular}
    }
	\caption{Reconstruction quality comparison of the pivot code.}
    \label{fig:reconstruction_first_step}
\end{figure}

\section{Experiments}
In this section, we evaluate the effectiveness of our inversion method in terms of reconstruction and editing quality. For all experiments, we use a StyleGAN2 \cite{Karras2020} generator pre-trained on the FFHQ \cite{Karras2019} dataset.

\noindent{\textbf{Datasets}}. 
We train our model on the FFHQ \cite{Karras2019} dataset which contains 70,000 facial images. The CelebA-HQ \cite{Karras2018} test set is used for evaluation. Furthermore, we collect 200 challenging facial images from the web for evaluation, including famous character images and cartoon images.

\noindent{\textbf{Baselines}}. 
We compare our method with four well-known inversion methods: SG2 \cite{Karras2020}, SG2$\mathcal{W}+$ \cite{Abdal2019}, e4e \cite{Tov2021}, and PTI \cite{Roich2021}. SG2 and SG2$\mathcal{W}+$ are optimization-based methods that invert the input images into the $\mathcal{W}$ and $\mathcal{W}+$ spaces, respectively. The e4e model is an encoder-based method that extends the psp model \cite{Richardson2021} by encouraging the latent codes close to $\mathcal{W}$. PTI is a two-step method that first infers the latent code using SG2 and then slightly alters the generator to fit the latent code. The qualitative comparison to SG2$\mathcal{W}+$ is provided in the Supplementary Materials due to the limited space.

\begin{table}
	\centering
	\caption{Quantitative reconstruction quality on CelebA-HQ.}	
	\vspace{-3pt}
	\begin{tabular}{L{2cm}P{1.1cm}P{1.1cm}P{1.0cm}P{1.2cm}}
	\toprule
	Method&Identity$\uparrow$&LPIPS$\downarrow$&MSE$\downarrow$&Runtime$\downarrow$\\
	\midrule
	SG2 \cite{Karras2020} &
	\multicolumn{1}{c}{$0.193$} & 
	\multicolumn{1}{c}{$0.373$} &
	\multicolumn{1}{c}{$0.099$} &
	\multicolumn{1}{c}{$48.91$} \\
	SG2$\mathcal{W}+$ \cite{Abdal2019} &
	\multicolumn{1}{c}{$0.667$} & 
	\multicolumn{1}{c}{$0.279$} &
	\multicolumn{1}{c}{$0.041$} &
	\multicolumn{1}{c}{$162.3$} \\
	e4e \cite{Tov2021} &
	\multicolumn{1}{c}{$0.494$} & 
	\multicolumn{1}{c}{$0.392$} &
	\multicolumn{1}{c}{$0.053$} &
	\multicolumn{1}{c}{$\bf{0.089}$} \\
	PTI \cite{Roich2021} & 
	\multicolumn{1}{c}{$0.845$} &
	\multicolumn{1}{c}{$0.101$} &
	\multicolumn{1}{c}{$0.013$} &
	\multicolumn{1}{c}{$102.2$} \\
	\midrule
	\textbf{Cycle} &
	\multicolumn{1}{c}{$0.855$} &
	\multicolumn{1}{c}{$0.095$} &
	\multicolumn{1}{c}{$\bf{0.011}$} &
	\multicolumn{1}{c}{$67.47$} \\
	\textbf{Cycle+Optim.} &
	\multicolumn{1}{c}{$\bf{0.866}$} &
	\multicolumn{1}{c}{$\bf{0.093}$} &
	\multicolumn{1}{c}{$\bf{0.011}$} &
	\multicolumn{1}{c}{$80.81$} \\
	\bottomrule
	\end{tabular}
	\label{table:reconstruction}
	\vspace{-2pt}
\end{table}

\begin{table}
	\centering
	\caption{Quantitative reconstruction quality on 200 challenging images collected from the web.}
	\vspace{-3pt}
	\begin{tabular}{L{2cm}P{1.1cm}P{1.1cm}P{1.0cm}P{1.2cm}}
	\toprule
	Method&Identity$\uparrow$&LPIPS$\downarrow$&MSE$\downarrow$&Runtime$\downarrow$\\
	\midrule
	$\text{PTI}$ \cite{Roich2021}  &
	\multicolumn{1}{c}{$0.774$} &
	\multicolumn{1}{c}{$0.146$} &
	\multicolumn{1}{c}{$0.022$} &
	\multicolumn{1}{c}{$102.7$} \\
	$\textbf{Cycle}$ &
	\multicolumn{1}{c}{$0.820$} &
	\multicolumn{1}{c}{$0.120$} &
	\multicolumn{1}{c}{$0.018$} &
	\multicolumn{1}{c}{$67.70$} \\
	$\textbf{Cycle+Optim.}$  &
	\multicolumn{1}{c}{$\bf{0.843}$} &
	\multicolumn{1}{c}{$\bf{0.116}$} &
	\multicolumn{1}{c}{$\bf{0.017}$} &
	\multicolumn{1}{c}{$\bf{81.09}$} \\
	\bottomrule
	\end{tabular}
	\label{table:reconstruction_challenging}
\end{table}

\begin{table}
	\centering
	\caption{Quantitative reconstruction quality of the pivot code on CelebA-HQ.}
	\vspace{-3pt}
	\begin{tabular}{L{2cm}P{1.1cm}P{1.1cm}P{1.0cm}P{1.2cm}}
	\toprule
	Method&Identity$\uparrow$&LPIPS$\downarrow$&MSE$\downarrow$&Runtime$\downarrow$\\
	\midrule
	$\text{PTI}$ \cite{Roich2021}  &
	\multicolumn{1}{c}{$0.179$} &
	\multicolumn{1}{c}{$0.377$} &
	\multicolumn{1}{c}{$0.099$} &
	\multicolumn{1}{c}{$35.85$} \\
	$\textbf{Cycle}$ &
	\multicolumn{1}{c}{$0.343$} &
	\multicolumn{1}{c}{$0.415$} &
	\multicolumn{1}{c}{$0.066$} &
	\multicolumn{1}{c}{$\bf{0.043}$} \\
	$\textbf{Cycle+Optim.}$  &
	\multicolumn{1}{c}{$\bf{0.514}$} &
	\multicolumn{1}{c}{$\bf{0.377}$} &
	\multicolumn{1}{c}{$\bf{0.047}$} &
	\multicolumn{1}{c}{$13.34$} \\
	\bottomrule
	\end{tabular}
	\label{table:reconstruction_pivot}
\end{table}

\subsection{Implementation Details}
In the step of $\mathcal{W}$$\rightarrow$$\mathcal{W}+$, we train the encoder for 500K iterations using the same hyperparameters as described in \cite{Tov2021}. In the step of $\mathcal{W}+$$\rightarrow$$\mathcal{W}$, we train the encoder for 250K iterations. Specifically, in the first 150K iterations, we gradually increase $\lambda_{\text{delta}}$ (Eq. \ref{eq:e4e_loss}) by 20\% and decrease $\lambda_{\text{adv}}$ (Eq. \ref{eq:e4e_loss}) by 50\% every 10K iterations. Then, we set from $\Delta_{N-1}=0$ to $\Delta_1=0$ sequentially every 4K iterations. For the regularized refinement step, we randomly sample 7 images from the FFHQ training set and update the encoder for 15 iterations. For the loss weights in Eq. \ref{eq:refinement_reg}-\ref{eq:refinement_full}, we set $\lambda_{\text{L2}}=1$, $\lambda_{\text{lpips}}=0.8$, $\lambda_{\text{id}}=0.1$, and $\lambda_{\text{reg}}=1$. For the pivotal tuning step, we use the same hyperparameters as described in \cite{Roich2021}. For the quantitative experiments, we follow \cite{Roich2021} to evaluate the models on the first 1000 samples from the CelebA-HQ test set. All experiments are performed using a single Nvidia GeForce RTX 3090 GPU.

\begin{figure}
\setlength{\tabcolsep}{1pt}
\centering
{\small
    \begin{tabular}{c c c c c c}
        \raisebox{0.2in}{\rotatebox[origin=t]{90}{Smile}}&
        \includegraphics[width=0.09\textwidth]{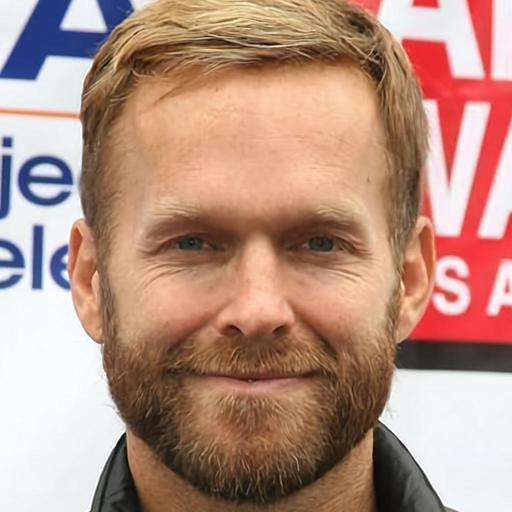}&
        \includegraphics[width=0.09\textwidth]{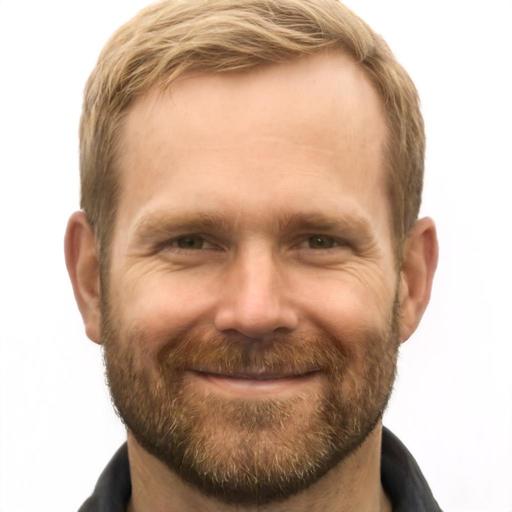}&        
        \includegraphics[width=0.09\textwidth]{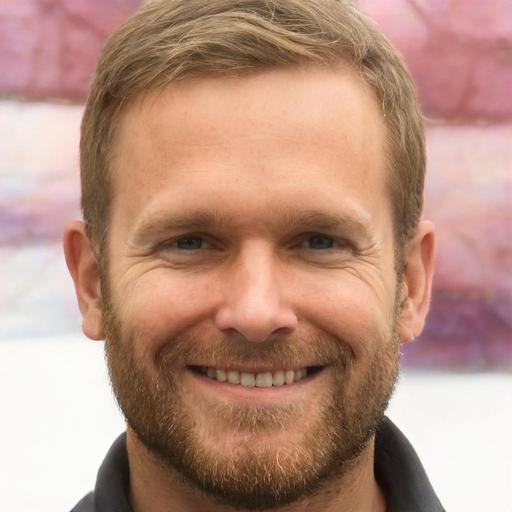}&
        \includegraphics[width=0.09\textwidth]{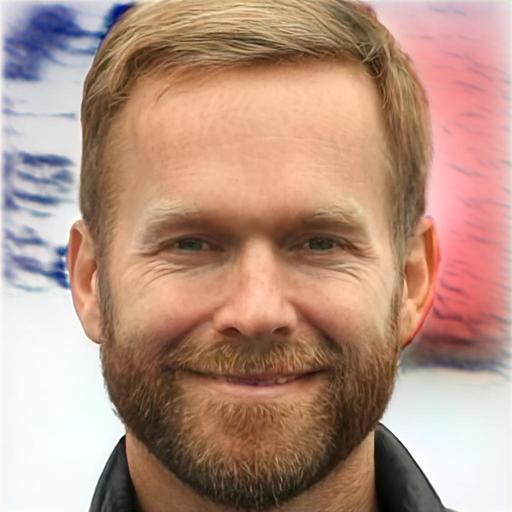}&
        \includegraphics[width=0.09\textwidth]{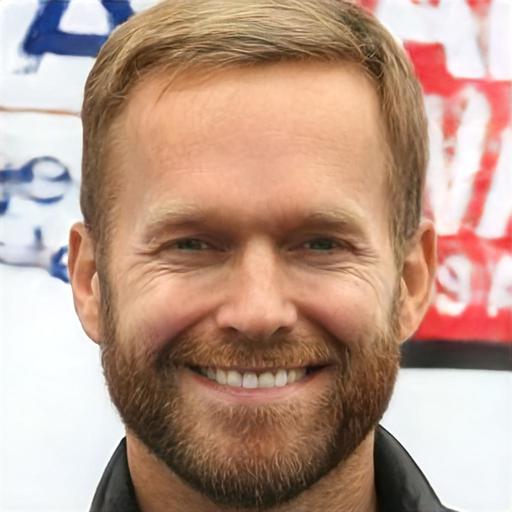} 
		\tabularnewline
		\raisebox{0.2in}{\rotatebox[origin=t]{90}{Pose}}&
        \includegraphics[width=0.09\textwidth]{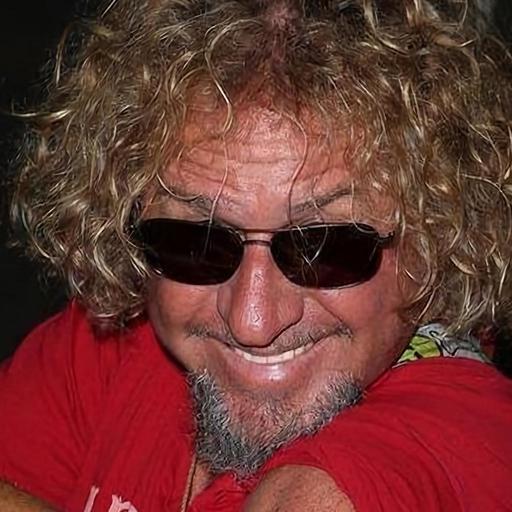}&
        \includegraphics[width=0.09\textwidth]{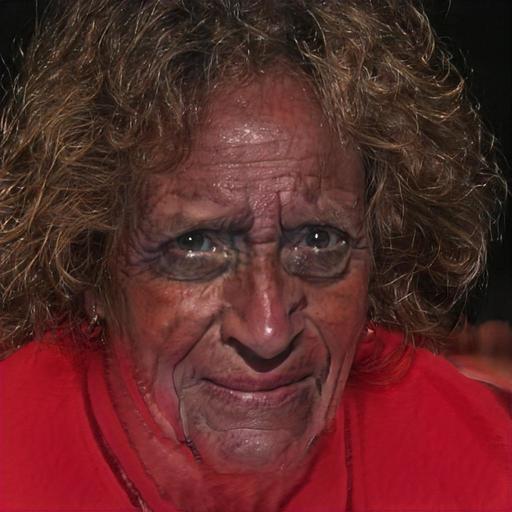}&        
        \includegraphics[width=0.09\textwidth]{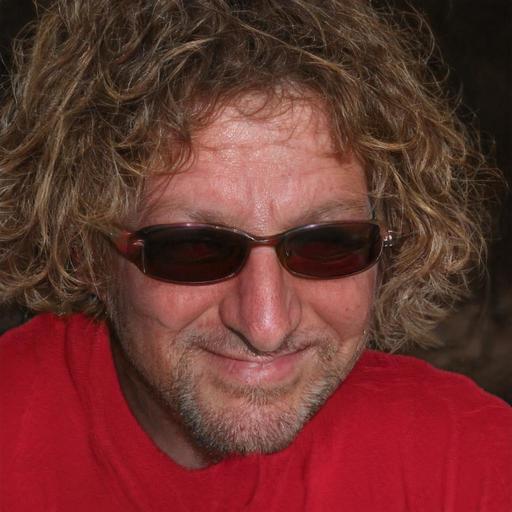}&
        \includegraphics[width=0.09\textwidth]{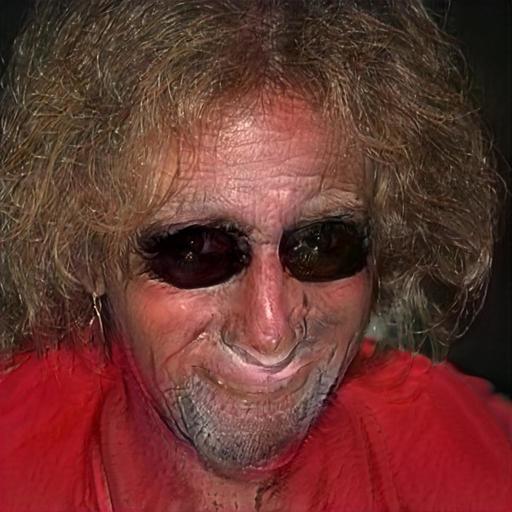}&
        \includegraphics[width=0.09\textwidth]{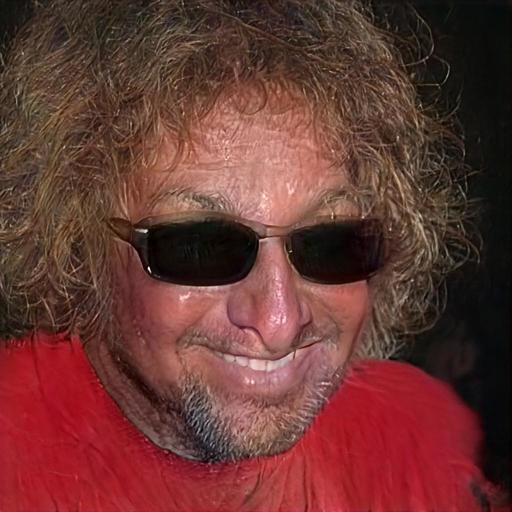} 
		\tabularnewline
        \raisebox{0.2in}{\rotatebox[origin=t]{90}{Age}}&
        \includegraphics[width=0.09\textwidth]{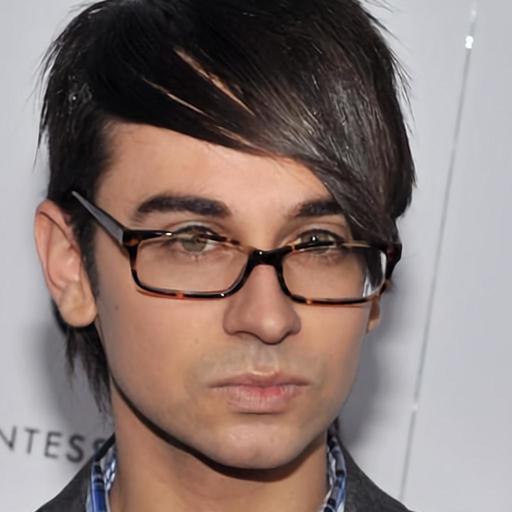}&
        \includegraphics[width=0.09\textwidth]{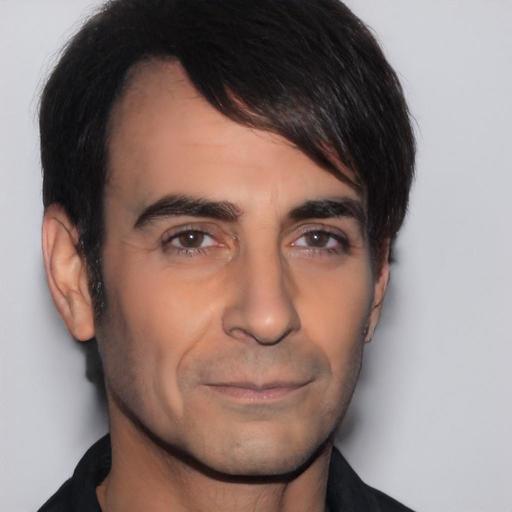}&        
        \includegraphics[width=0.09\textwidth]{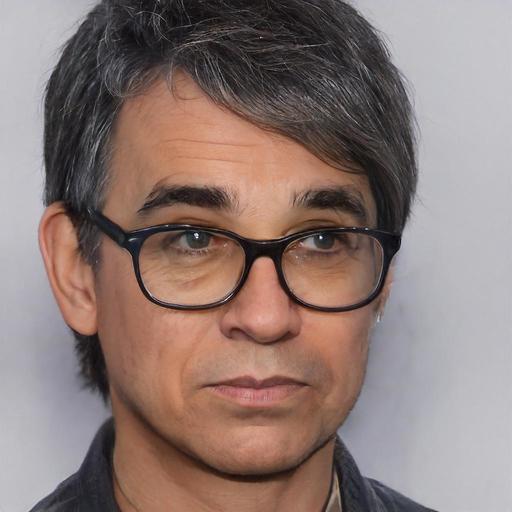}&
        \includegraphics[width=0.09\textwidth]{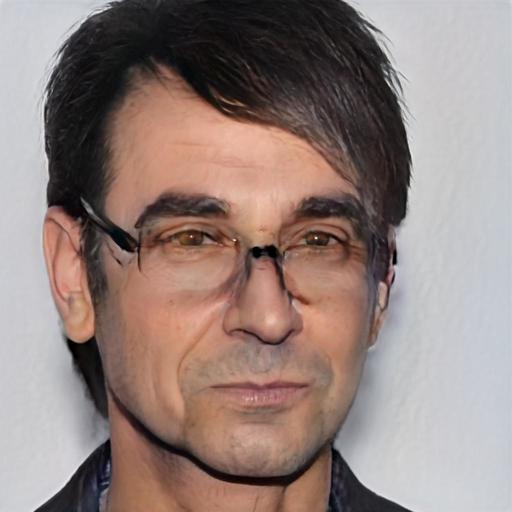}&
        \includegraphics[width=0.09\textwidth]{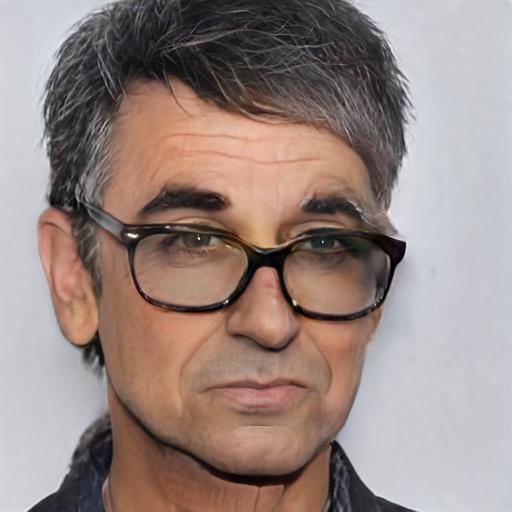} 
		\tabularnewline
		& Input & SG2 & e4e & PTI & Ours
    \end{tabular}
    }
	\caption{Editing quality comparison on CelebA-HQ. In each example, the editing is performed using the same editing weight.}
    \label{fig:editing_celeba}
\end{figure}

\begin{figure}
\setlength{\tabcolsep}{1pt}
\centering
{\small
    \begin{tabular}{c c c c c c}
		\raisebox{0.2in}{\rotatebox[origin=t]{90}{Smile}}&
        \includegraphics[width=0.09\textwidth]{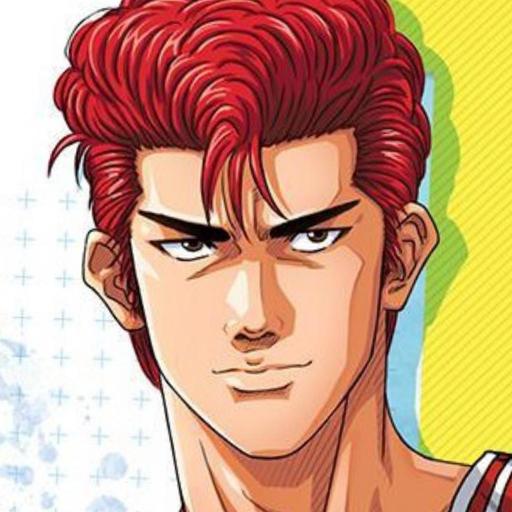}&
        \includegraphics[width=0.09\textwidth]{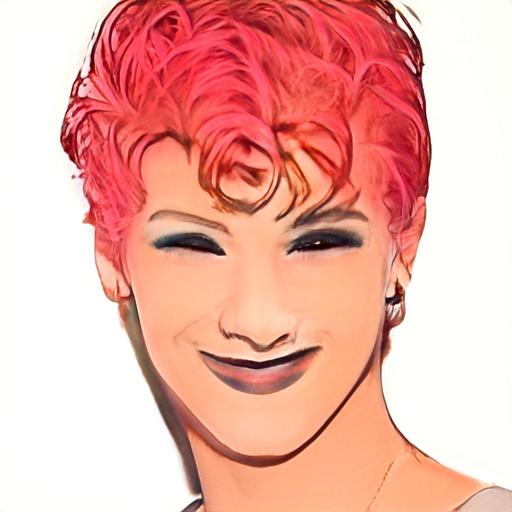}&        
        \includegraphics[width=0.09\textwidth]{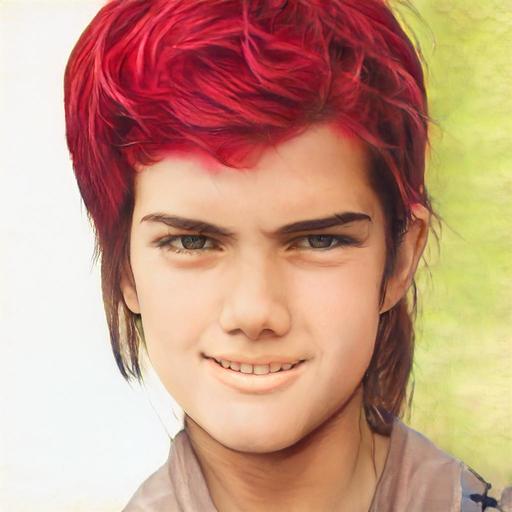}&
        \includegraphics[width=0.09\textwidth]{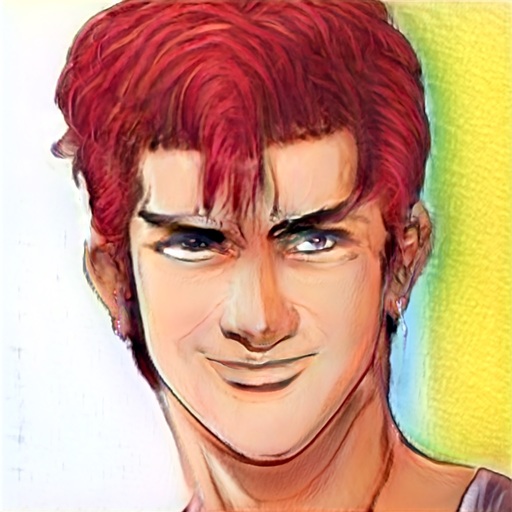}&
        \includegraphics[width=0.09\textwidth]{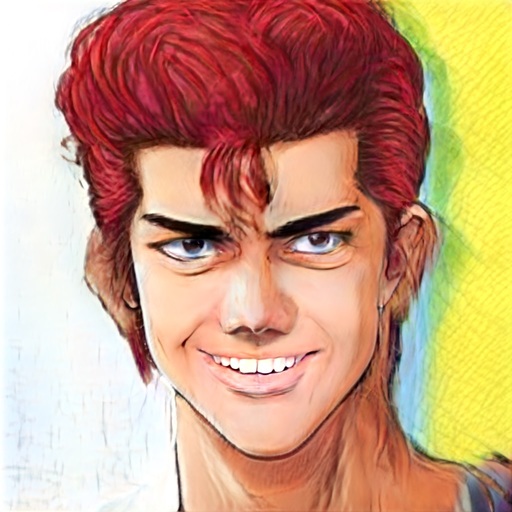} 
		\tabularnewline
        \raisebox{0.2in}{\rotatebox[origin=t]{90}{Pose}}&
        \includegraphics[width=0.09\textwidth]{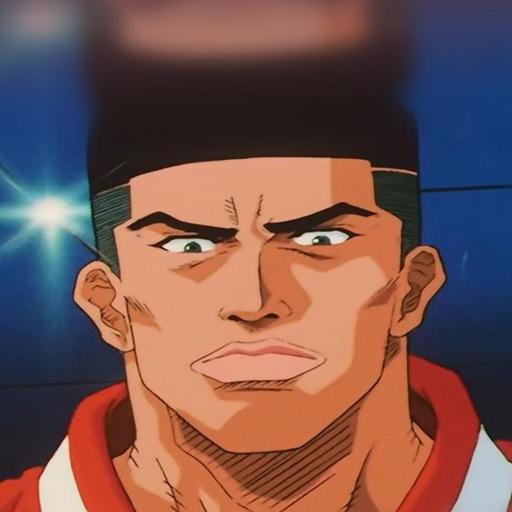}&
        \includegraphics[width=0.09\textwidth]{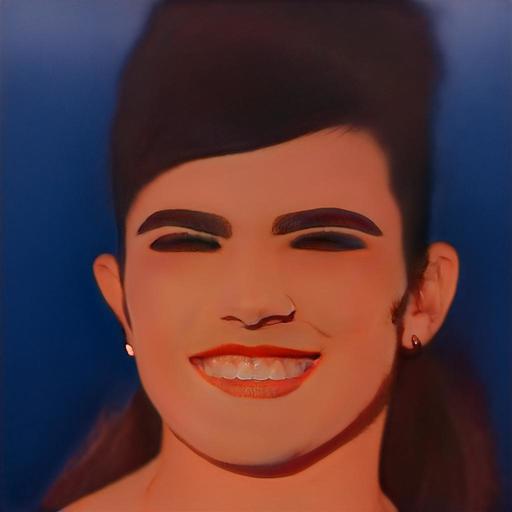}&        
        \includegraphics[width=0.09\textwidth]{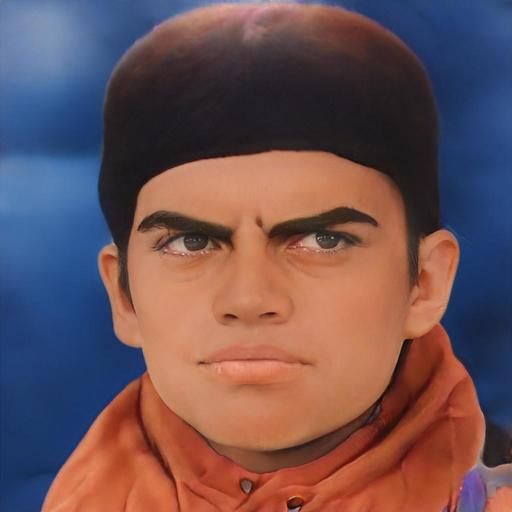}&
        \includegraphics[width=0.09\textwidth]{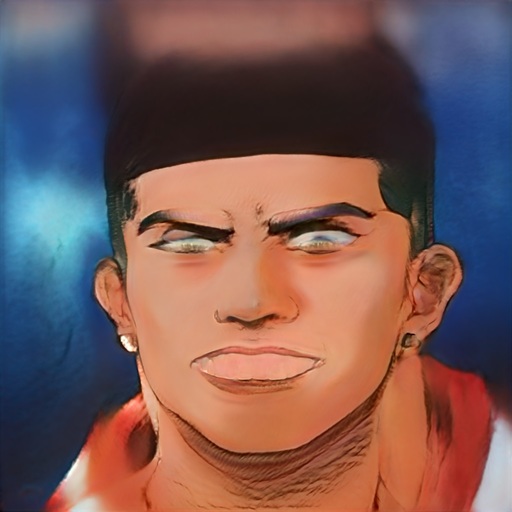}&
        \includegraphics[width=0.09\textwidth]{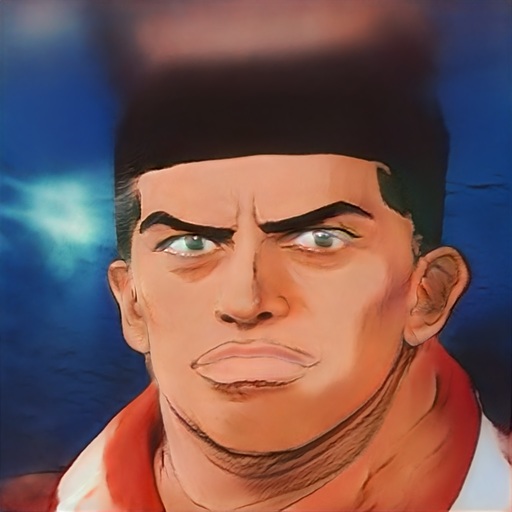} 
		\tabularnewline
        \raisebox{0.2in}{\rotatebox[origin=t]{90}{Age}}&
        \includegraphics[width=0.09\textwidth]{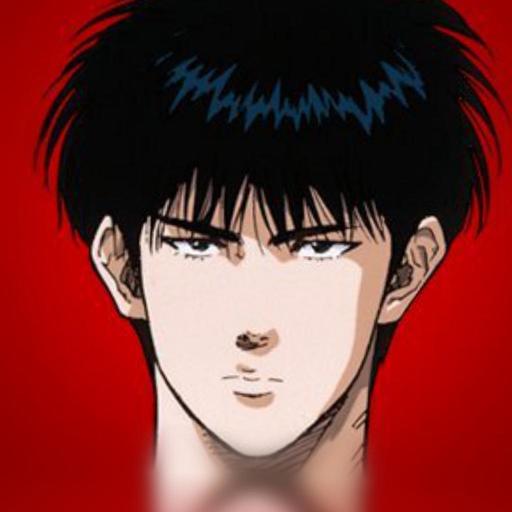}&
        \includegraphics[width=0.09\textwidth]{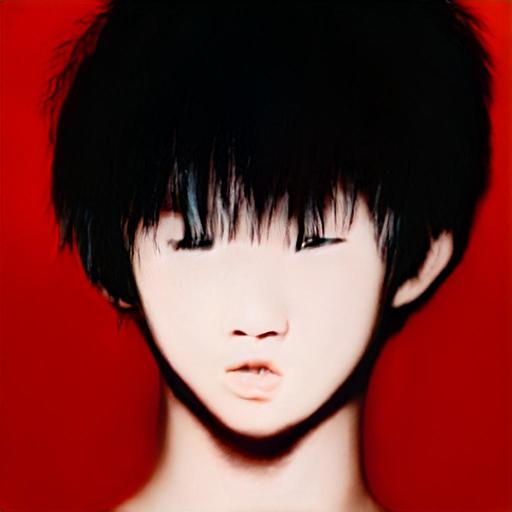}&        
        \includegraphics[width=0.09\textwidth]{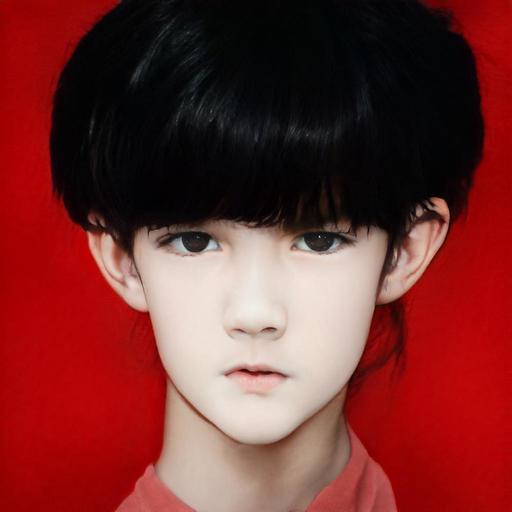}&
        \includegraphics[width=0.09\textwidth]{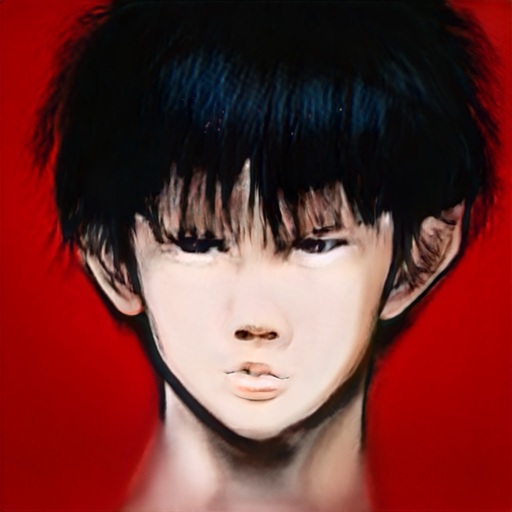}&
        \includegraphics[width=0.09\textwidth]{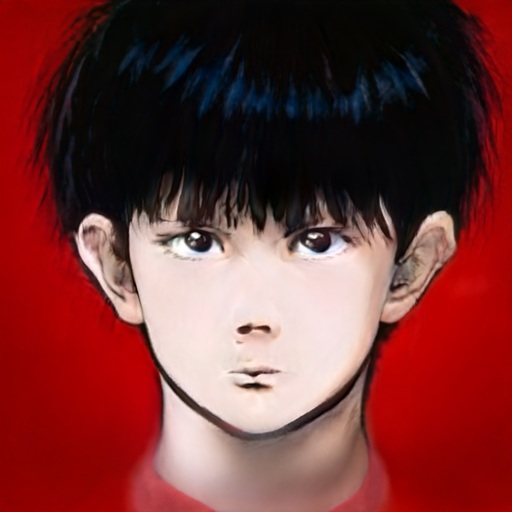} 
		\tabularnewline
		& Input & SG2 & e4e & PTI & Ours
    \end{tabular}
    }
	\caption{Editing quality comparison using cartoon images. In each example, the editing is performed using the same editing weight.}
    \label{fig:editing_cartoon}
\end{figure}

\subsection{Reconstruction Quality}
\begin{figure}
\setlength{\tabcolsep}{1pt}
\centering
{\small
    \begin{tabular}{c c c c c c}
        \includegraphics[width=0.072\textwidth]{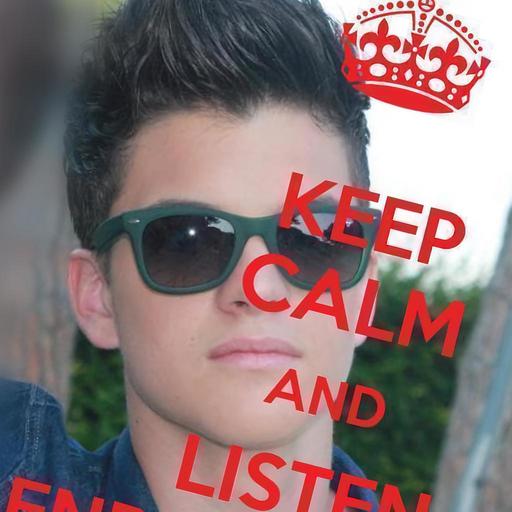}&
        \includegraphics[width=0.072\textwidth]{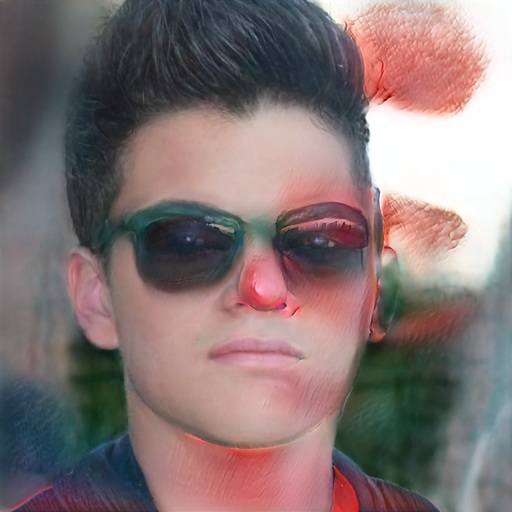}&        
        \includegraphics[width=0.072\textwidth]{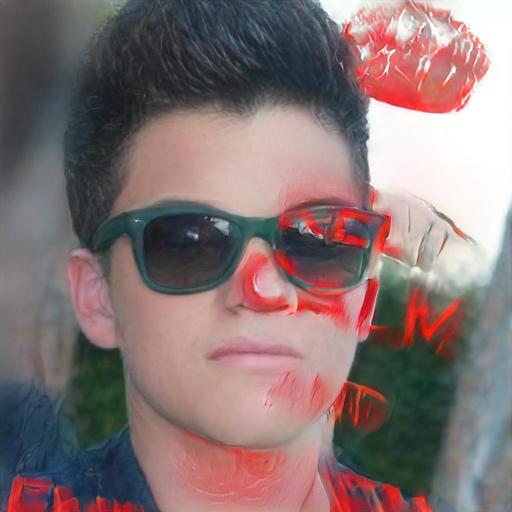}&
        \includegraphics[width=0.072\textwidth]{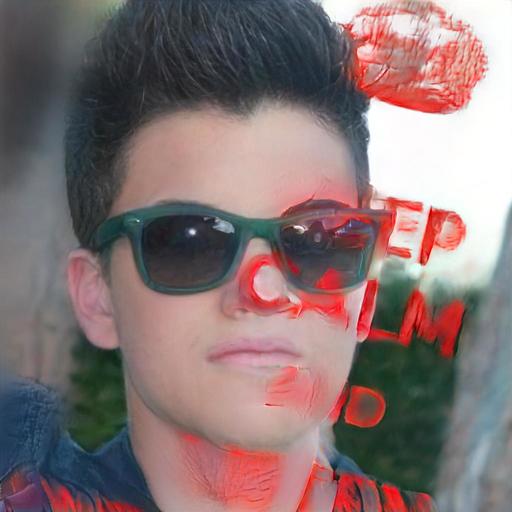}&
        \includegraphics[width=0.072\textwidth]{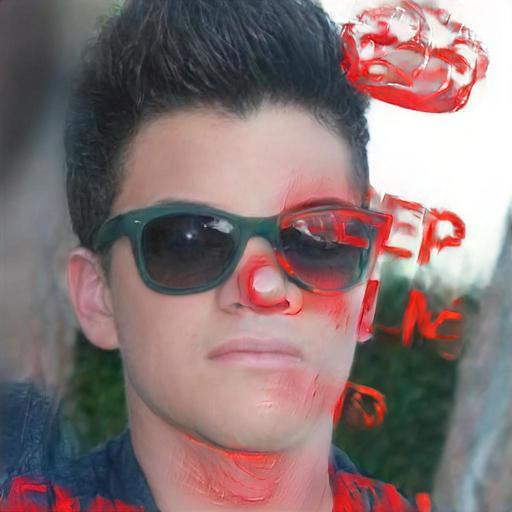}&
        \includegraphics[width=0.072\textwidth]{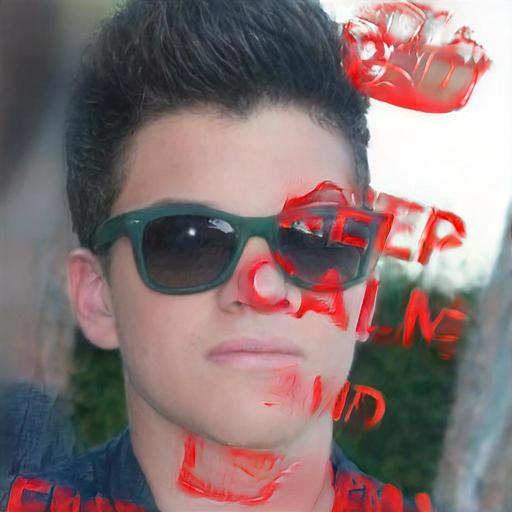} 
		\tabularnewline
        \includegraphics[width=0.072\textwidth]{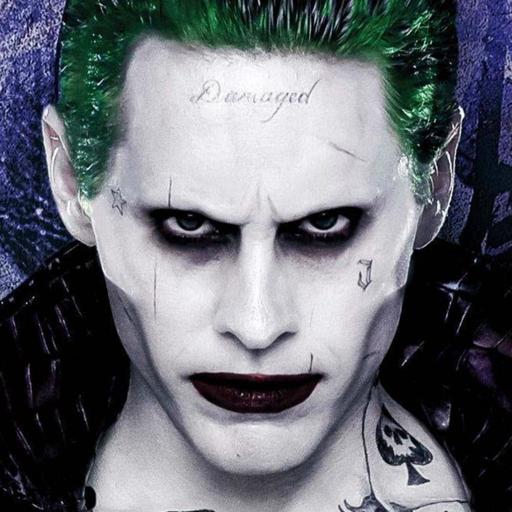}&
        \includegraphics[width=0.072\textwidth]{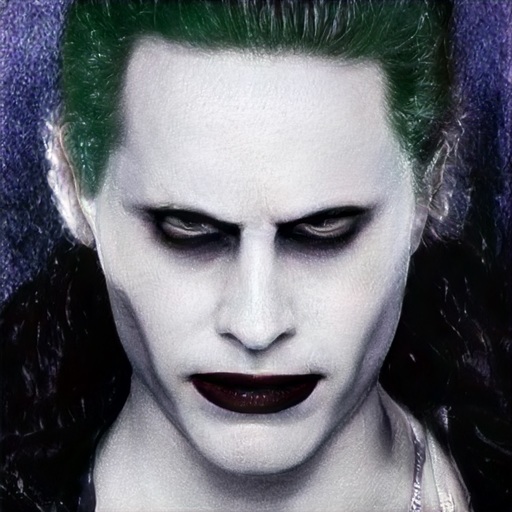}&        
        \includegraphics[width=0.072\textwidth]{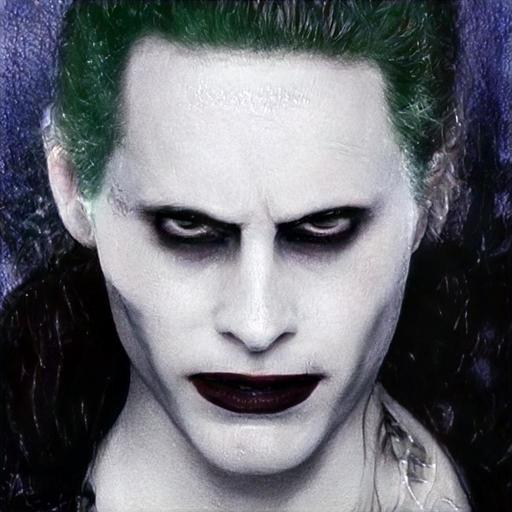}&
        \includegraphics[width=0.072\textwidth]{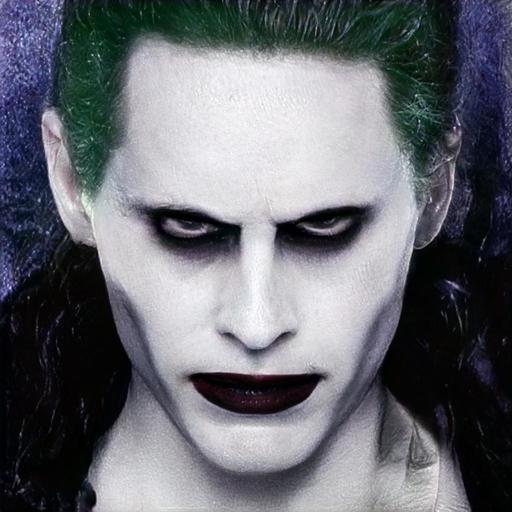}&
        \includegraphics[width=0.072\textwidth]{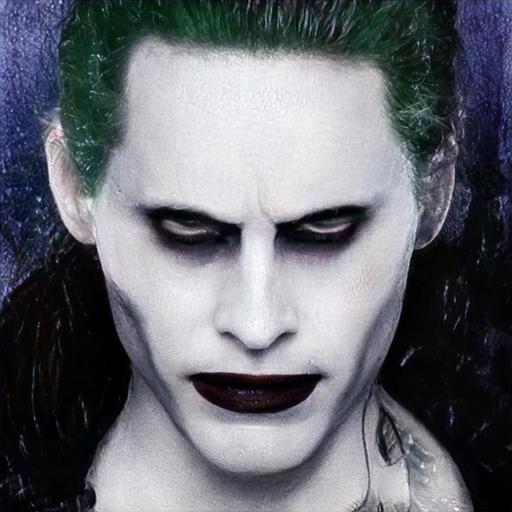}&
        \includegraphics[width=0.072\textwidth]{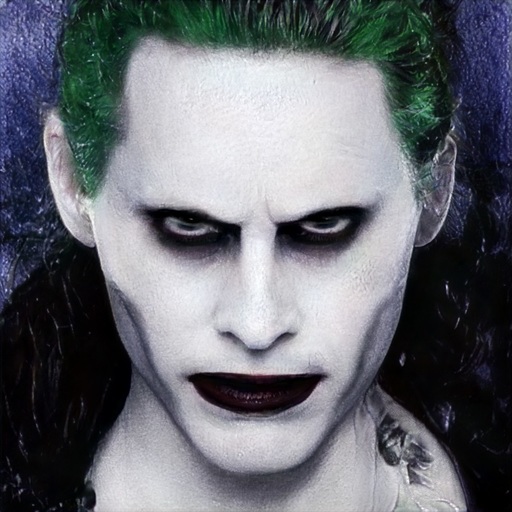} 
		\tabularnewline
        \includegraphics[width=0.072\textwidth]{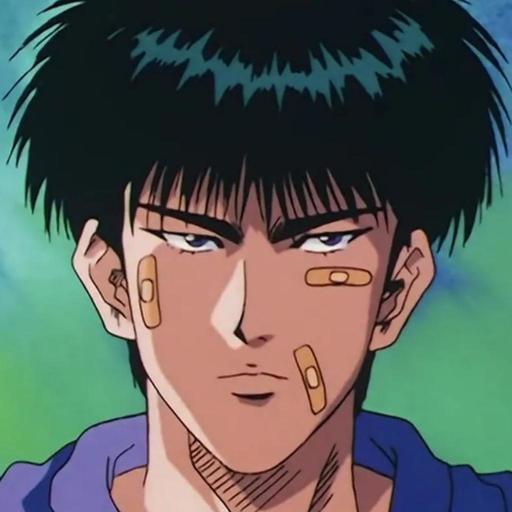}&
        \includegraphics[width=0.072\textwidth]{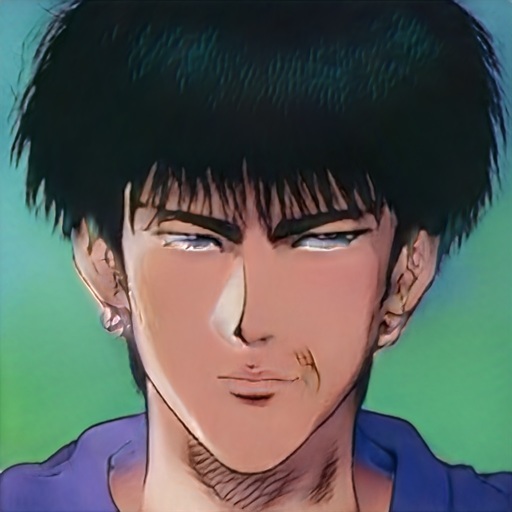}&        
        \includegraphics[width=0.072\textwidth]{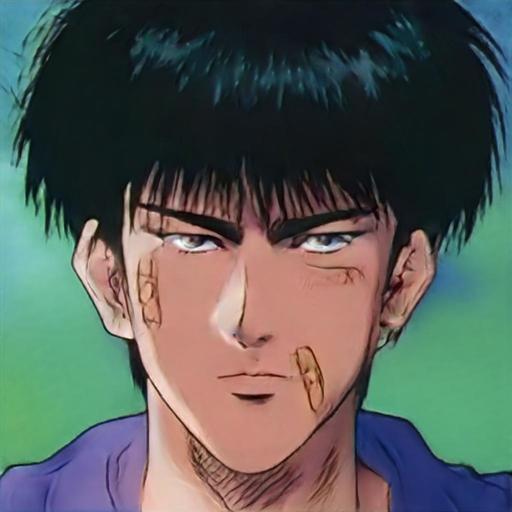}&
        \includegraphics[width=0.072\textwidth]{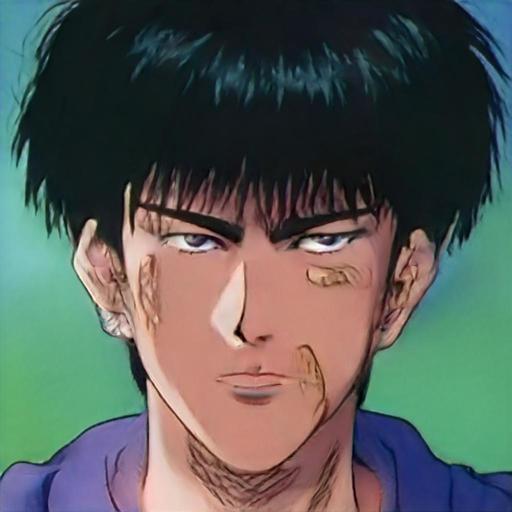}&
        \includegraphics[width=0.072\textwidth]{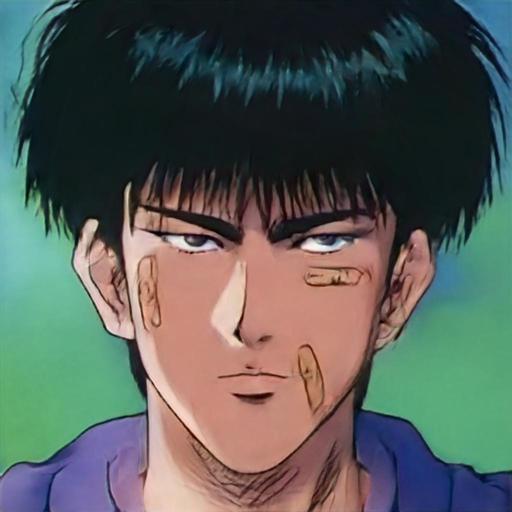}&
        \includegraphics[width=0.072\textwidth]{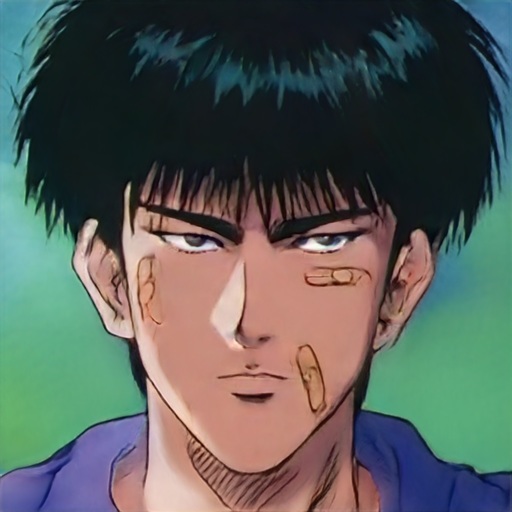}
 		\tabularnewline
		Input & PTI & w/o Optim. & \W & $\mathcal{W}+$$\rightarrow$$\mathcal{W}$  & Full
		\end{tabular}
    }
	\caption{Ablation study. The full model achieves the best reconstruction quality.}
    \label{fig:ablation_optimization1}
\end{figure}

\noindent{\textbf{Qualitative Evaluation}}.
Figures \ref{fig:reconstruction_celeba} to \ref{fig:reconstruction_cartoon} present a visually qualitative comparison of the reconstructed images. The results show that our method achieves superior reconstruction for all images from three different data sources. Figures \ref{fig:reconstruction_celeba} and \ref{fig:reconstruction_characters} show that the reconstruction obtained by our method preserves more accurate details, such as cap (row 1, Figure \ref{fig:reconstruction_celeba}), background (row 2, Figure \ref{fig:reconstruction_celeba}), hand (row 1, Figure \ref{fig:reconstruction_characters}), and makeup (row 3, Figure \ref{fig:reconstruction_characters}). Figure \ref{fig:reconstruction_cartoon} presents the reconstruction of cartoon images. Inverting cartoon images is more challenging as the cartoon images are completely out-of-domain. In Figure \ref{fig:reconstruction_cartoon}, the reconstructed images by PTI tend to be blurry, and some key components (e.g., eye and mouth) are not accurately reconstructed. In contrast, our method successfully reconstructs the cartoon images. Examples of using cycle encoding only (i.e., without applying the optimization step) are provided in Figure \ref{fig:ablation_optimization1}, and one can see that cycle encoding already outperforms PTI in reconstruction. More visual results are provided in the Supplementary Materials.

\noindent{\textbf{Quantitative Evaluation}}. Table \ref{table:reconstruction} presents a quantitative evaluation among different inversion methods. We employ three metrics, including identity similarity score \cite{Huang2020}, LPIPS \cite{Zhang2018}, and Mean Squared Error (MSE)\footnote{The implementation of identity similarity score, LPIPS, and MSE is from https://github.com/eladrich/pixel2style2pixel.}. Following \cite{Richardson2021}, we measure the identity similarity score by using a different face recognition model (Curricularface \cite{Huang2020}) to make the similarity score independent from the loss function (ArcFace \cite{Deng2019}). The results demonstrate that our method achieves the best reconstruction quality in terms of all three metrics. Our method reduces the inference time of PTI from 102.2 seconds to 80.8 seconds. Without applying the optimization step, the inference time can be further reduced to 67.5 seconds. Moreover, we also evaluate the models on 200 challenging images collected from the web. In Table \ref{table:reconstruction_challenging}, we see a substantial improvement over PTI from 0.774 to 0.843 in identity similarity.

\noindent{\textbf{Pivot Code Quality}}.
Here, we evaluate the reconstruction quality of the pivot code. Figure \ref{fig:reconstruction_first_step} presents the reconstructed results of the pivot code using examples in Figures \ref{fig:reconstruction_celeba} to \ref{fig:reconstruction_cartoon}. One can see that our method achieves significantly lower distortion compared to PTI, especially for the challenging cartoon images. The quantitative results in Table \ref{table:reconstruction_pivot} also demonstrate that our method outperforms PTI by a substantial margin.

\begin{table}
	\centering
	\caption{Quantitative editing quality by evaluating the editing magnitude (e.g., rotation angle) when applying the same editing weight $\alpha$.}
	\vspace{-6pt}
	\begin{tabular}{L{2.3cm}P{1.3cm}P{1.3cm}P{1.3cm}}
	\toprule
	Rotation Angle$\uparrow$ & $\alpha=1$ & $\alpha=5$ & $\alpha=10$ \\
	\midrule
	SG2 \cite{Karras2020} &
	\multicolumn{1}{c}{$3.83$} & 
	\multicolumn{1}{c}{$16.70$} &
	\multicolumn{1}{c}{$32.68$} \\
	SG2$\mathcal{W}+$ \cite{Abdal2019} &
	\multicolumn{1}{c}{$2.48$} & 
	\multicolumn{1}{c}{$10.74$} &
	\multicolumn{1}{c}{$21.72$} \\
	e4e \cite{Tov2021} &
	\multicolumn{1}{c}{$3.49$} & 
	\multicolumn{1}{c}{$14.90$} &
	\multicolumn{1}{c}{$29.13$} \\
	PTI \cite{Roich2021} & 
	\multicolumn{1}{c}{$3.44$} &
	\multicolumn{1}{c}{$16.64$} &
	\multicolumn{1}{c}{$33.09$} \\
	\midrule
	\textbf{Ours} &
	\multicolumn{1}{c}{$\bf{4.63}$} &
	\multicolumn{1}{c}{$\bf{22.20}$} &
	\multicolumn{1}{c}{$\bf{42.54}$} \\
	\bottomrule
	\end{tabular}
	\label{table:editing_angle}
	\vspace{-5pt}
\end{table}

\begin{table}
	\centering
	\caption{Quantitative editing quality by evaluating the identity similarity when applying the same editing magnitude (e.g., rotation angle).}
	\vspace{-6pt}
	\begin{tabular}{L{2cm}P{1.4cm}P{1.4cm}P{1.4cm}}
	\toprule
	Identity$\uparrow$ & Angle$\pm$5 & Angle$\pm$10 & Angle$\pm$15 \\
	\midrule
	SG2 \cite{Karras2020} &
	\multicolumn{1}{c}{$0.188$} & 
	\multicolumn{1}{c}{$0.179$} &
	\multicolumn{1}{c}{$0.166$} \\
	SG2$\mathcal{W}+$ \cite{Abdal2019} &
	\multicolumn{1}{c}{$0.621$} & 
	\multicolumn{1}{c}{$0.538$} &
	\multicolumn{1}{c}{$0.454$} \\
	e4e \cite{Tov2021} &
	\multicolumn{1}{c}{$0.487$} & 
	\multicolumn{1}{c}{$0.471$} &
	\multicolumn{1}{c}{$0.442$} \\
	PTI \cite{Roich2021} & 
	\multicolumn{1}{c}{$0.778$} & 
	\multicolumn{1}{c}{$0.673$} &
	\multicolumn{1}{c}{$0.565$} \\
	\midrule
	\textbf{Ours} &
	\multicolumn{1}{c}{$\bf{0.812}$} & 
	\multicolumn{1}{c}{$\bf{0.728}$} &
	\multicolumn{1}{c}{$\bf{0.638}$} \\
	\bottomrule
	\end{tabular}
	\label{table:editing_similarity}
\end{table}

\begin{figure*}
\setlength{\tabcolsep}{1.5pt}
\centering
{
    \begin{tabular}{c c c c c c}
        \raisebox{0.5in}{\rotatebox[origin=t]{90}{Smile}}&
        \includegraphics[width=0.17\textwidth]{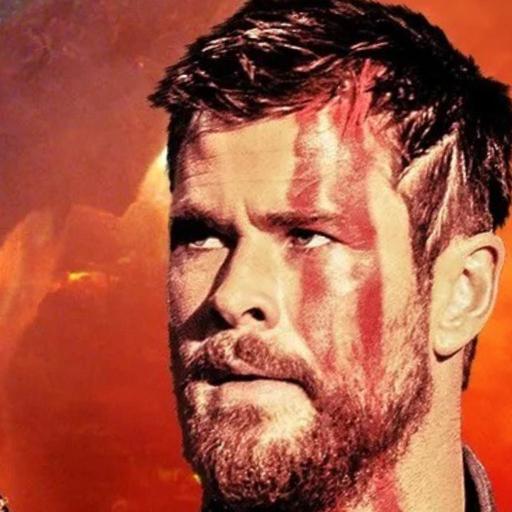}&
        \includegraphics[width=0.17\textwidth]{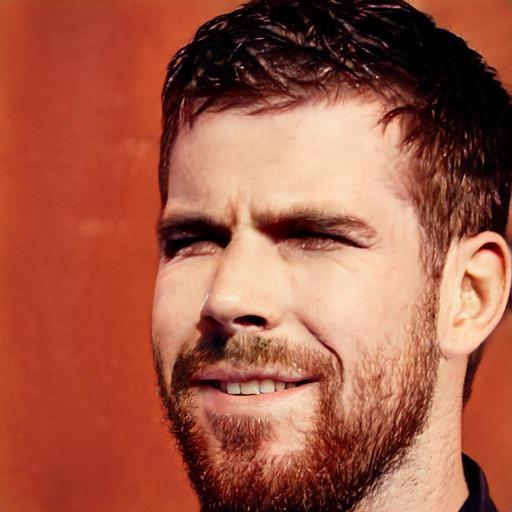}&        
        \includegraphics[width=0.17\textwidth]{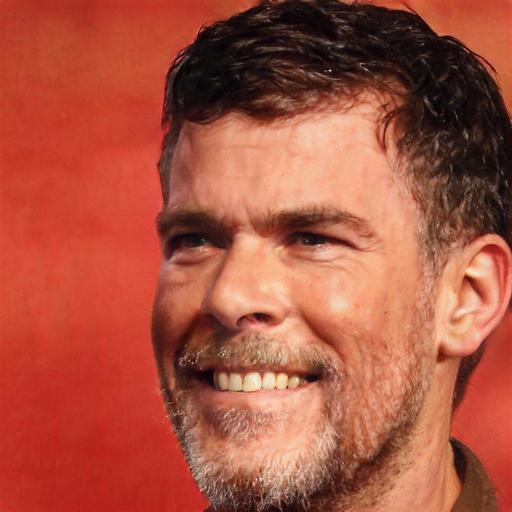}&
        \includegraphics[width=0.17\textwidth]{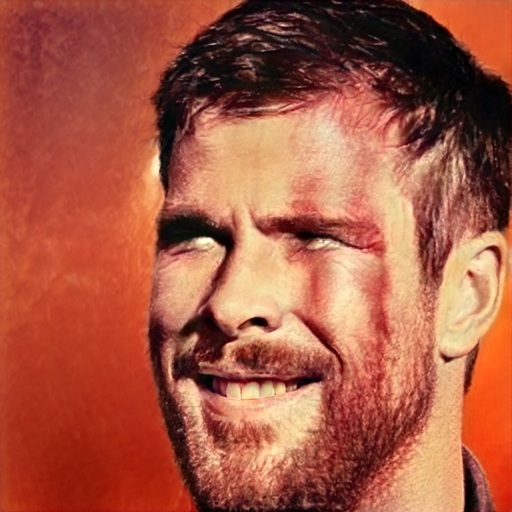}&
        \includegraphics[width=0.17\textwidth]{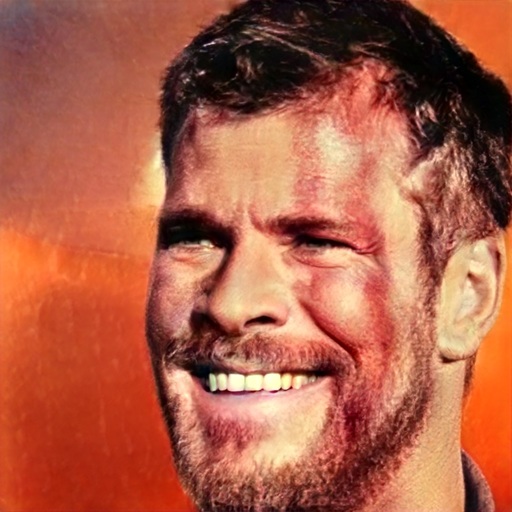} 
		\tabularnewline
        \raisebox{0.5in}{\rotatebox[origin=t]{90}{Pose}}&
        \includegraphics[width=0.17\textwidth]{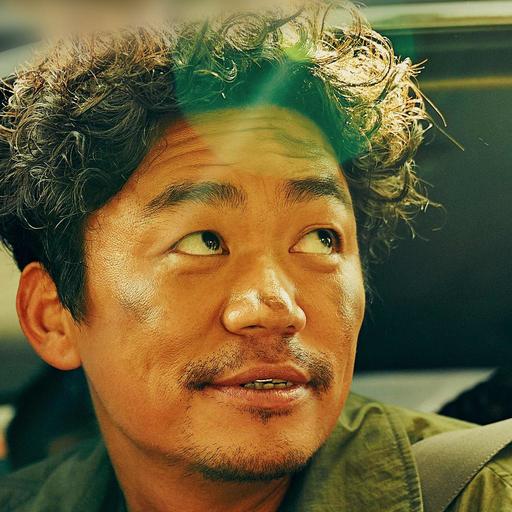}&
        \includegraphics[width=0.17\textwidth]{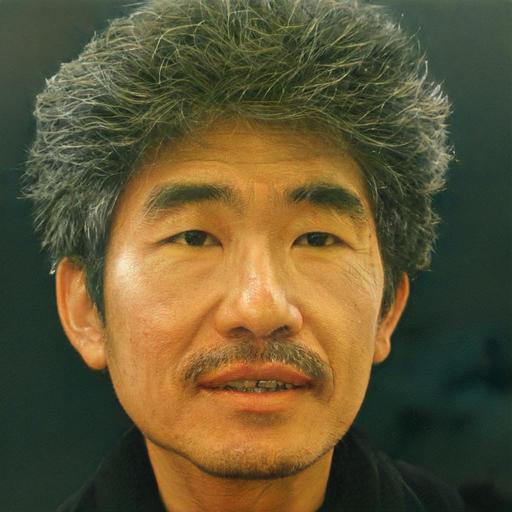}&        
        \includegraphics[width=0.17\textwidth]{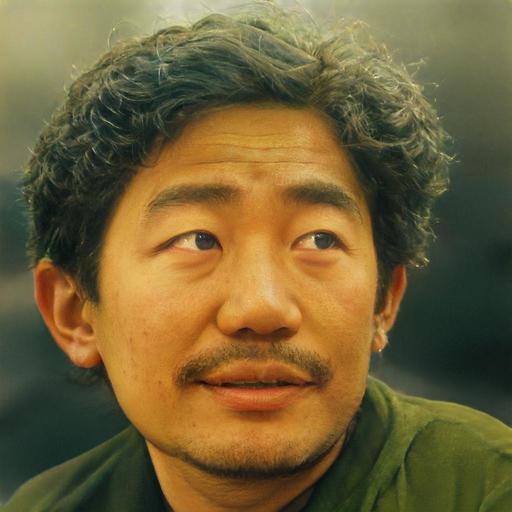}&
        \includegraphics[width=0.17\textwidth]{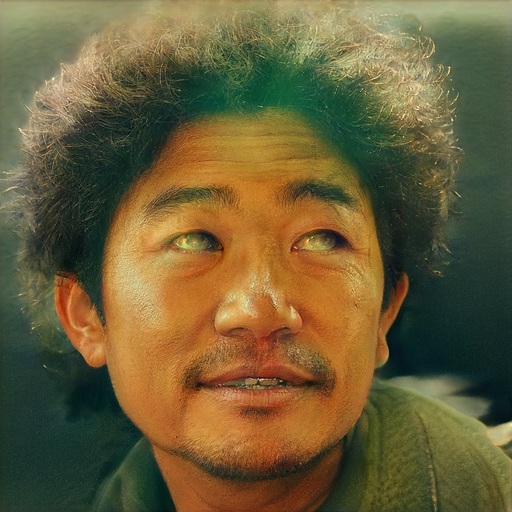}&
        \includegraphics[width=0.17\textwidth]{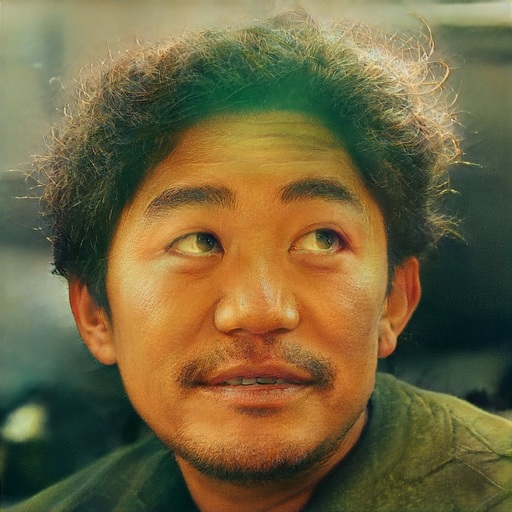} 
		\tabularnewline
        \raisebox{0.5in}{\rotatebox[origin=t]{90}{+Age}}&
        \includegraphics[width=0.17\textwidth]{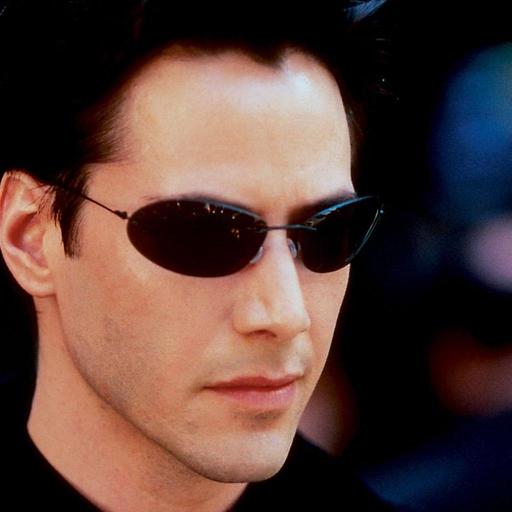}&
        \includegraphics[width=0.17\textwidth]{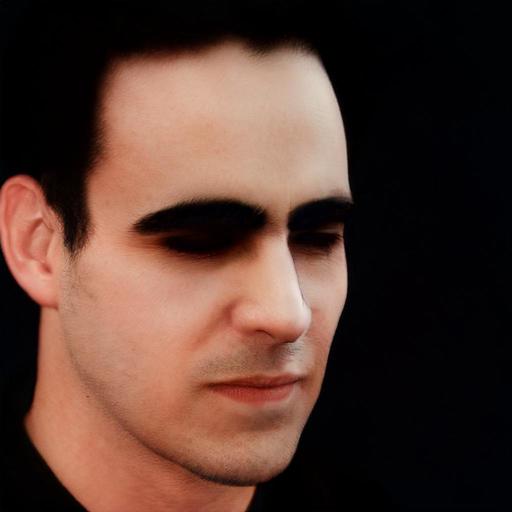}&        
        \includegraphics[width=0.17\textwidth]{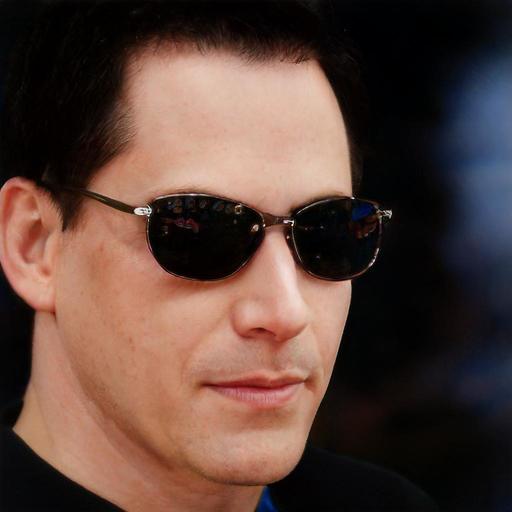}&
        \includegraphics[width=0.17\textwidth]{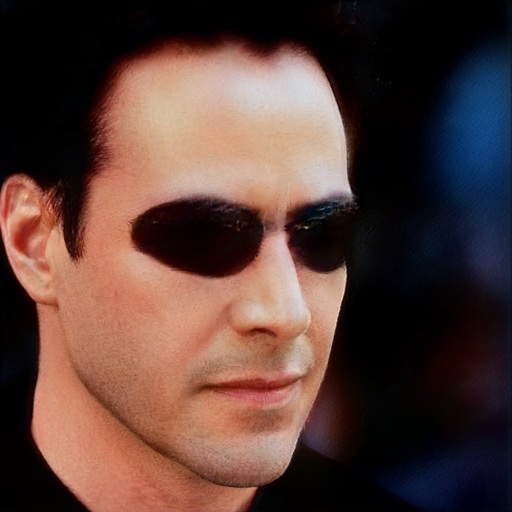}&
        \includegraphics[width=0.17\textwidth]{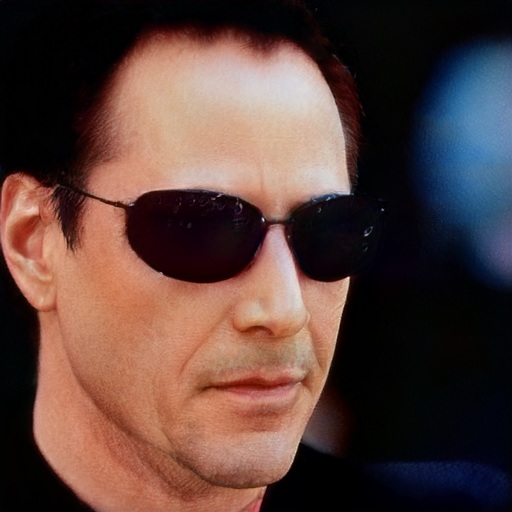} 
		\tabularnewline
		\raisebox{0.5in}{\rotatebox[origin=t]{90}{-Age}}&
        \includegraphics[width=0.17\textwidth]{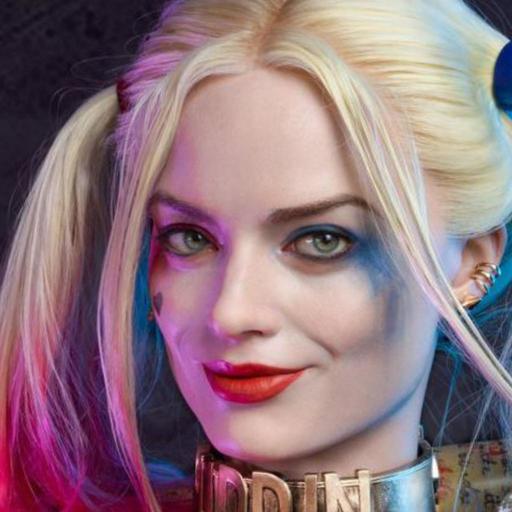}&
        \includegraphics[width=0.17\textwidth]{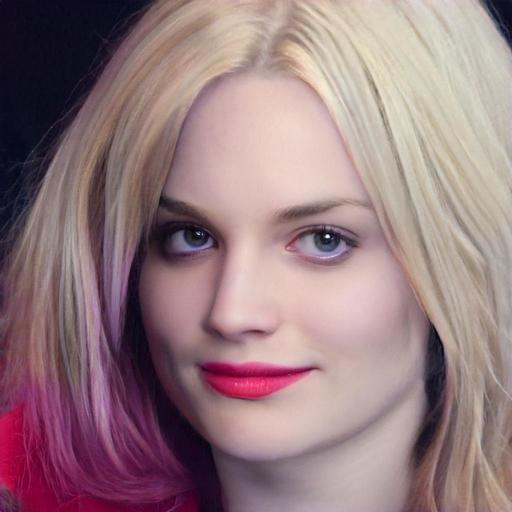}&        
        \includegraphics[width=0.17\textwidth]{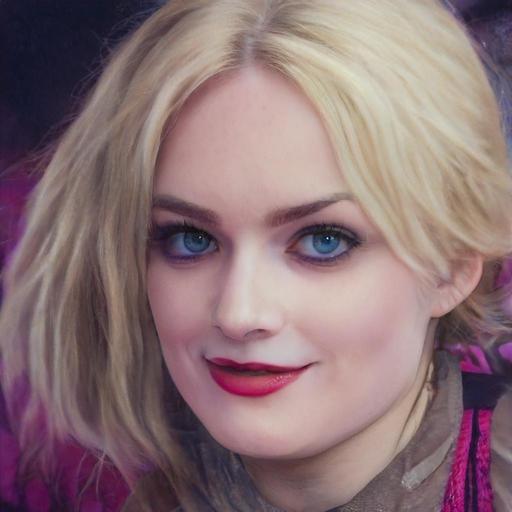}&
        \includegraphics[width=0.17\textwidth]{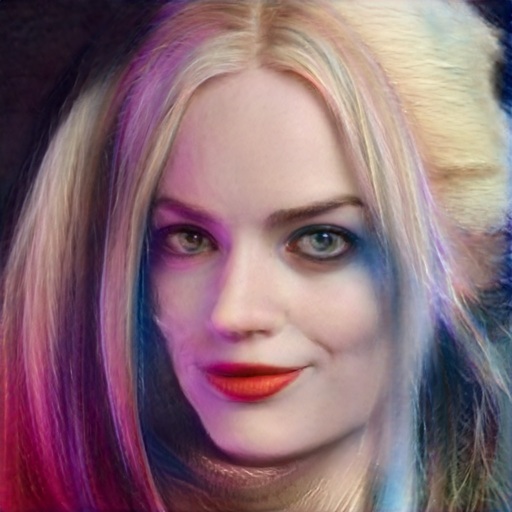}&
        \includegraphics[width=0.17\textwidth]{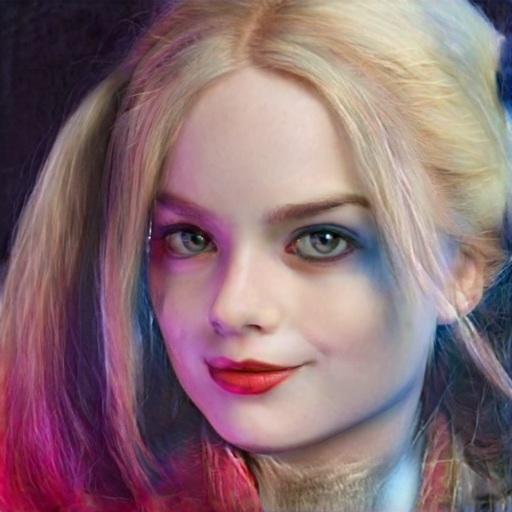} 
		\tabularnewline
		& Input & SG2 & e4e & PTI & Ours
    \end{tabular}
    }
	\caption{Editing quality comparison using famous character images. In each example, the editing is performed using the same editing weight.}
    \label{fig:editing_characters}
\end{figure*}

\subsection{Editing Quality}
In addition to reconstruction quality, editing quality is another important target for GAN inversion, as the major motivation for GAN inversion is the downstream editing task. Several works \cite{Zhu2020_2, Tov2021, Roich2021} have discussed the editing quality, and they show that the native $\mathcal{W}$ space is superior to the extended $\mathcal{W}+$ space in terms of editability. A high editability is expected that given the inverted latent code, one can edit it and obtain the desired editing magnitude with a small reconstruction accuracy drop. In the following experiments, we use the popular editing method, InterfaceGAN \cite{Shen2020}, for latent-based editing. InterfaceGAN edits the original latent code $z$ with $z_{edit} = z+\alpha n$, where $\alpha$ is the editing weight and $n$ is a unit normal vector, corresponding to a semantic direction. One can see that the reconstruction quality will decrease as the editing weight $\alpha$ increases, since $\alpha n$ will dominate the value of $z_{edit}$ when $\alpha$ is large. Thus we also expect to obtain the desired editing magnitude with a small editing weight $\alpha$. Based on the above observations, we follow \cite{Roich2021} to evaluate the editing quality by using two metrics: \textit{editing magnitude} when applying the same editing weight, and \textit{identity preservation} when applying the same editing magnitude.

\noindent{\textbf{Qualitative Evaluation}}.
In Figures \ref{fig:editing_celeba} to \ref{fig:editing_characters}, we provide a qualitative comparison of different methods in editing images. To compare the editing magnitude, in each example, the edited images by different methods are obtained using the same editing weight $\alpha$. As can be seen in Figures \ref{fig:editing_celeba} and \ref{fig:editing_characters}, our method achieves the largest editing magnitude for all the examples, and our method provides the best visually-pleasing editing results and most accurately preserves the identity of the input images. For example, PTI loses the identity and details, such as background (row 1, Figure \ref{fig:editing_celeba}), appearance (row 2, Figure \ref{fig:editing_celeba}), eyes (row 2, Figure \ref{fig:editing_characters}), and glasses (row 3, Figure \ref{fig:editing_characters}). SG2 and e4e achieve visually-pleasing editing quality but lose the identity of the input images. We also investigate the performance of our method on more challenging out-of-domain cartoon images in Figure \ref{fig:editing_cartoon}. The edited images by PTI tend to be blurry. Our method is the only one that successfully reconstructs and edits the cartoon images. For more editing results, more editing directions, and editing results using StyleClip \cite{Patashnik2021}, see the Supplementary Materials.

\noindent{\textbf{Quantitative Evaluation}}.
Tables \ref{table:editing_angle} and \ref{table:editing_similarity} present a quantitative comparison of different methods. As aforementioned, we use \textit{editing magnitude} and \textit{identity preservation} to measure the editing quality. We follow \cite{Zhu2020_2,Roich2021} to use the pose editing operation for this evaluation, as evaluating the rotation angle is more accurate than the other operations. Microsoft Face API is used to calculate the rotation angle. As shown in Table \ref{table:editing_angle}, our method induces the largest rotation angle compared to the baselines, especially when the editing weight is large. In Table \ref{table:editing_similarity}, our method also achieves the best score in terms of identity similarity, which indicates that our method most accurately preserves the identity of the original images when performing the same editing magnitude. For the case of Angle$\pm$15, we see a substantial editing quality improvement over PTI from 0.565 to 0.638.

\begin{figure}
\setlength{\tabcolsep}{2pt}
\centering
{\small
    \begin{tabular}{c c c c c c}
        \raisebox{0.18in}{\rotatebox[origin=t]{90}{w/o Reg.}}&
        \includegraphics[width=0.08\textwidth]{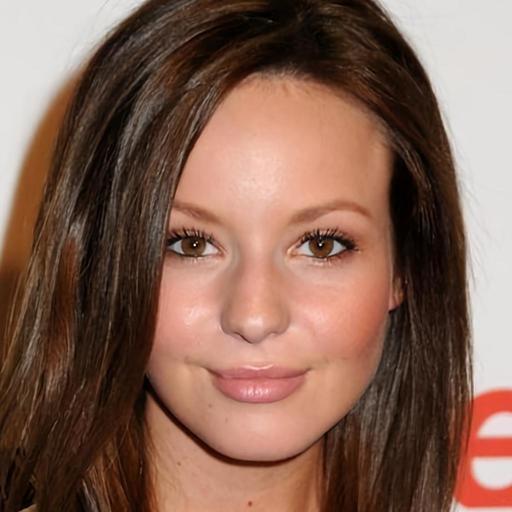}&
        \includegraphics[width=0.08\textwidth]{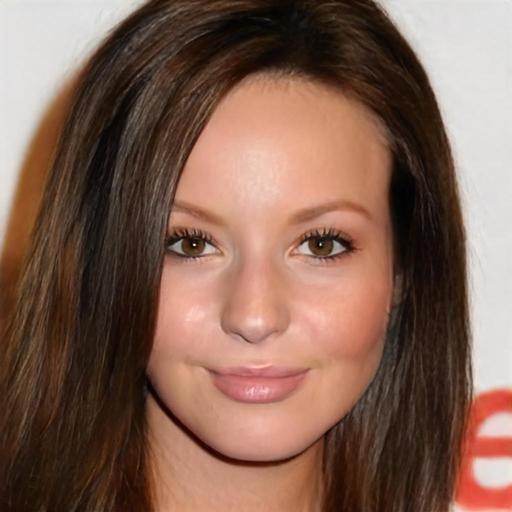}&        
        \includegraphics[width=0.08\textwidth]{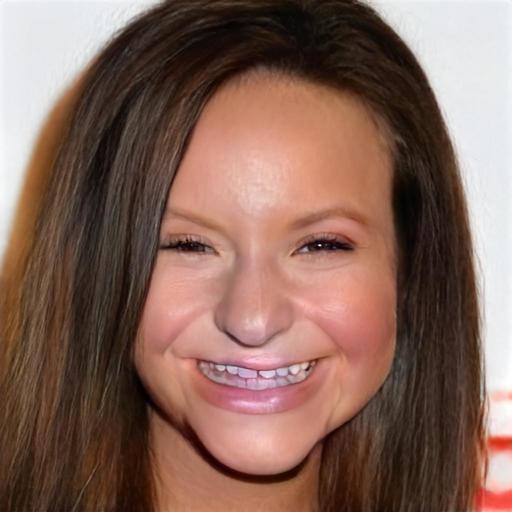}&
        \includegraphics[width=0.08\textwidth]{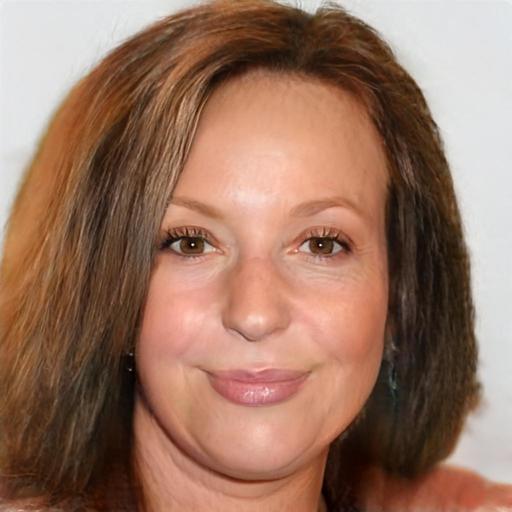}&
        \includegraphics[width=0.08\textwidth]{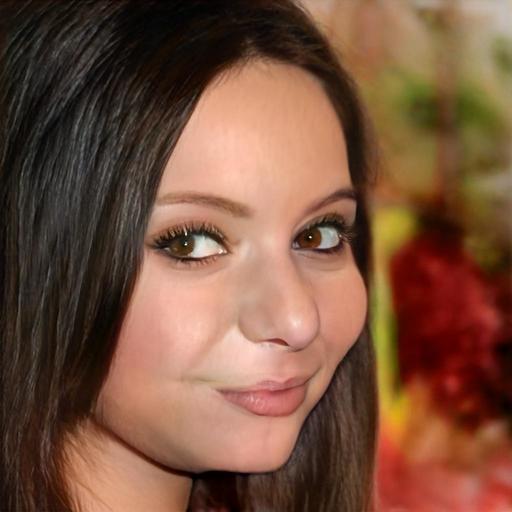} 
		\\
        \raisebox{0.17in}{\rotatebox[origin=t]{90}{w/ Reg.}}&
		\includegraphics[width=0.08\textwidth]{images/ablation/optim/28707_real.jpg}&
        \includegraphics[width=0.08\textwidth]{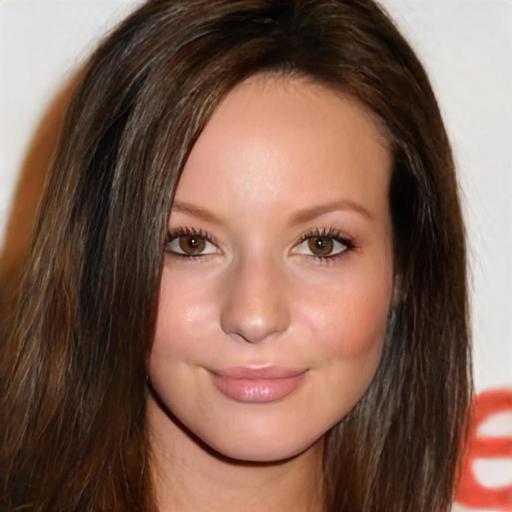}&        
        \includegraphics[width=0.08\textwidth]{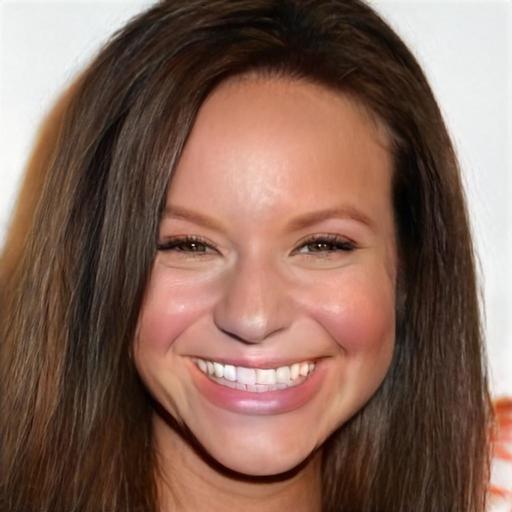}&
        \includegraphics[width=0.08\textwidth]{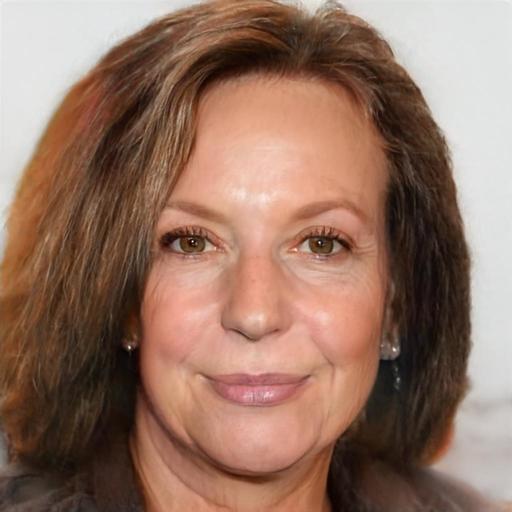}&
        \includegraphics[width=0.08\textwidth]{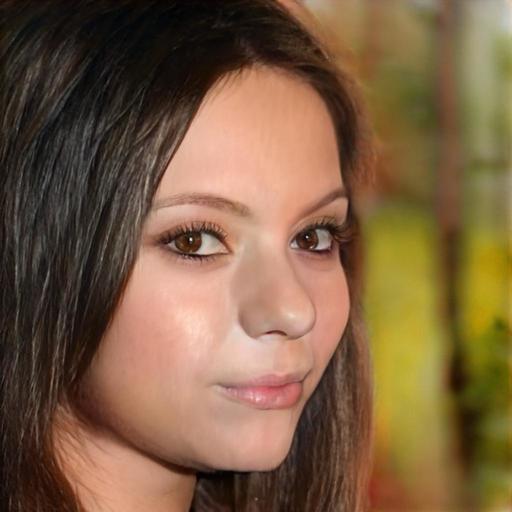} 
		\\
		& Input  & Inversion & Smile & Age & Pose
    \end{tabular}
    }
	\caption{Ablation study of the regularization, where the editing is performed using the same editing weight. Without the regularization, the model yields inferior editability.}
    \label{fig:ablation_optimization2}
\end{figure}

\subsection{Ablation Study}
\label{sec:ablation}
\noindent{\textbf{Optimization-based Refinement}}.
We first perform an ablation study on the optimization step described in Section \ref{sec:optimization}. We compare our model with two variants: skipping the optimization step (denoted as w/o Optim.) and removing the regularization term (w/o Reg.). The quantitative results in Table \ref{table:optimization} demonstrate that the optimization step decreases the distortion with a subtle drop in editability, and the regularization term alleviates the drop in editability. The qualitative examples in Figure \ref{fig:ablation_optimization1} and \ref{fig:ablation_optimization2} show consistent results with the quantitative evaluation. Note that without the optimization step, cycle encoding already outperforms PTI in both reconstruction and editability.

\begin{figure}
\setlength{\tabcolsep}{2pt}
\centering
{\small
    \begin{tabular}{c c c c c c}
        \raisebox{0.18in}{\rotatebox[origin=t]{90}{$\mathcal{W}$$\rightarrow$$\mathcal{W}+$}}&
        \includegraphics[width=0.08\textwidth]{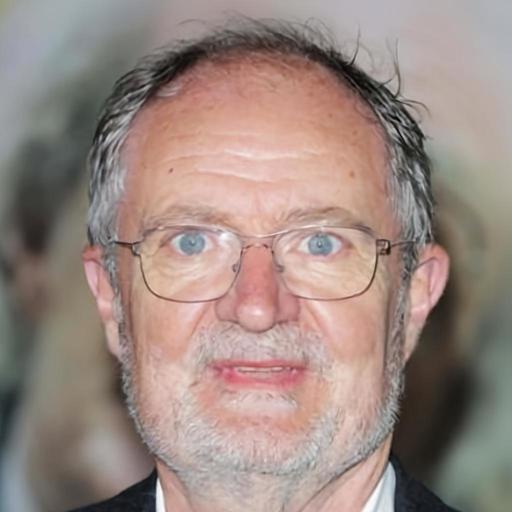}&
        \includegraphics[width=0.08\textwidth]{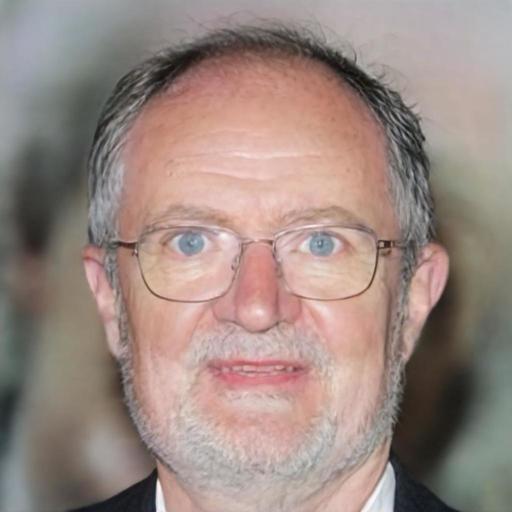}&        
        \includegraphics[width=0.08\textwidth]{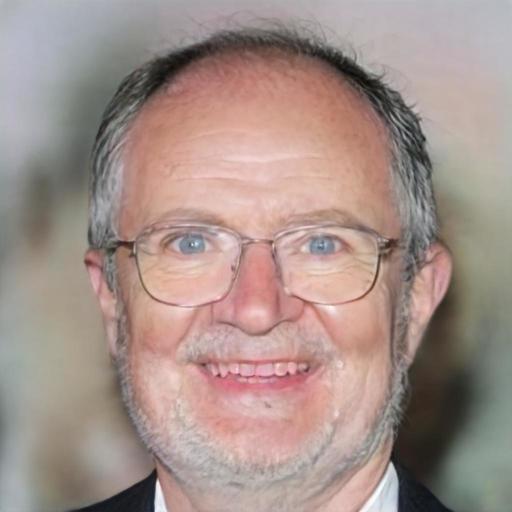}&
        \includegraphics[width=0.08\textwidth]{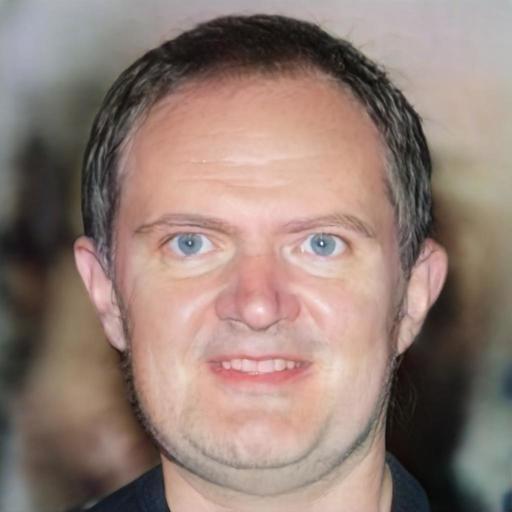}&
        \includegraphics[width=0.08\textwidth]{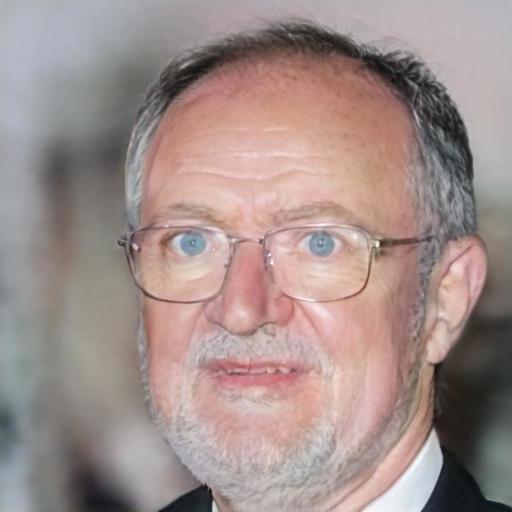} 
		\\
        \raisebox{0.17in}{\rotatebox[origin=t]{90}{Full}}&
		\includegraphics[width=0.08\textwidth]{images/ablation/cycle/28520_real.jpg}&
        \includegraphics[width=0.08\textwidth]{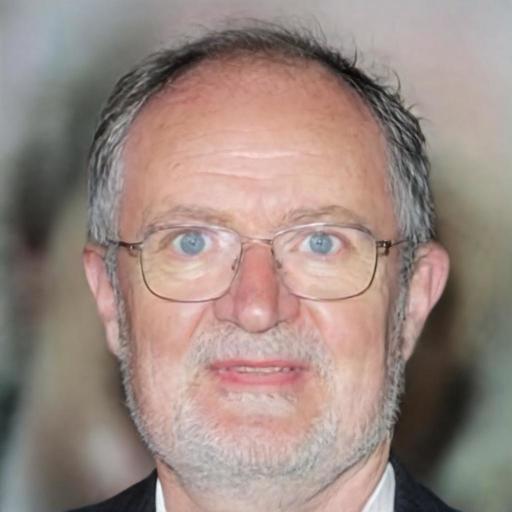}&        
        \includegraphics[width=0.08\textwidth]{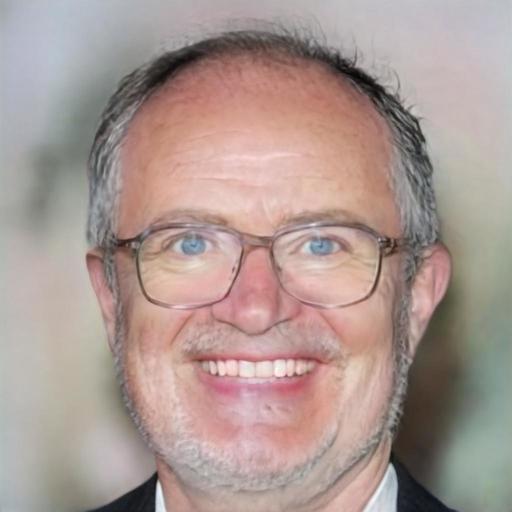}&
        \includegraphics[width=0.08\textwidth]{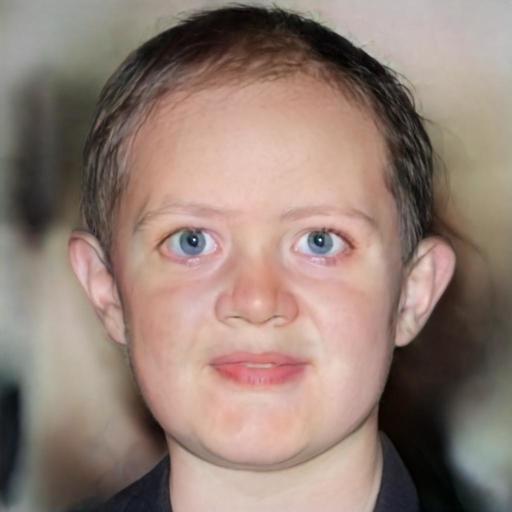}&
        \includegraphics[width=0.08\textwidth]{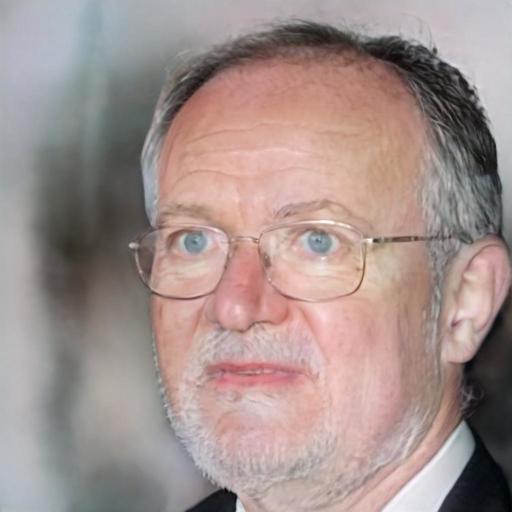} 
		\\
		& Input  & Inversion & Smile & Age & Pose
    \end{tabular}
    }
	\caption{Ablation study of cycle encoding, where the editing is performed using the same editing weight. $\mathcal{W}$$\rightarrow$$\mathcal{W}+$ yields inferior editability. }
    \label{fig:ablation_cycle}
\end{figure}

\noindent{\textbf{Cycle Encoding}}.
We then perform an ablation study on the cycle encoding step. We compare our model with three variants: training the encoder in \W, training the encoder in $\mathcal{W}$$\rightarrow$$\mathcal{W}+$, and training the encoder in $\mathcal{W}+$$\rightarrow$$\mathcal{W}$. The details of the model configurations can be found in the Supplementary Materials, and all the models are trained for the same number of iterations. The quantitative results in Table \ref{table:ablation_cycle} show that $\mathcal{W}$$\rightarrow$$\mathcal{W}+$ achieves the lowest distortion but poorest editability. Figure \ref{fig:ablation_cycle} also demonstrates that cycle encoding achieves superior editability than $\mathcal{W}$$\rightarrow$$\mathcal{W}+$. For $\mathcal{W}+$$\rightarrow$$\mathcal{W}$, we observe a slight degeneration in both reconstruction and editability compared to cycle encoding from Figure \ref{fig:ablation_optimization1} and Table \ref{table:ablation_cycle}. \W behaves poorly in both reconstruction and editability.

\begin{table}
	\centering
	\caption{Ablation study of the optimization step. First row: reconstruction evaluation via identity similarity. Second row: editability evaluation via rotation angle.}
	\begin{tabular}{L{1.3cm}P{1.2cm}P{1.6cm}P{1.2cm}P{1.2cm}}
	\toprule
	 & PTI \cite{Roich2021} & w/o Optim.  & w/o Reg. & Full\\
	\midrule
	Identity$\uparrow$ &
	\multicolumn{1}{c}{$0.845$} &
	\multicolumn{1}{c}{$0.855$} &
	\multicolumn{1}{c}{$0.869$} &
	\multicolumn{1}{c}{$0.866$} \\
	Angle$\uparrow$  &
	\multicolumn{1}{c}{$16.64$} &
	\multicolumn{1}{c}{$22.49$} &
	\multicolumn{1}{c}{$19.95$} &
	\multicolumn{1}{c}{$22.20$} \\
	\bottomrule
	\end{tabular}
	\label{table:optimization}
\end{table}

\begin{table}
	\centering
	\caption{Ablation study of cycle encoding. First row: reconstruction evaluation via identity similarity. Second row: editability evaluation via rotation angle.}
	\begin{tabular}{L{1.1cm}P{1.3cm}P{1.3cm}P{1.3cm}P{1.0cm}}
	\toprule
	 & \W &  $\mathcal{W}$$\rightarrow$$\mathcal{W}+$ & $\mathcal{W}+$$\rightarrow$$\mathcal{W}$ & Full\\
	\midrule
	Identity$\uparrow$ &
	\multicolumn{1}{c}{$0.849$} &
	\multicolumn{1}{c}{$0.875$} &
	\multicolumn{1}{c}{$0.859$} &
	\multicolumn{1}{c}{$0.866$} \\
	Angle$\uparrow$  &
	\multicolumn{1}{c}{$16.33$} &
	\multicolumn{1}{c}{$14.70$} &
	\multicolumn{1}{c}{$21.24$} &
	\multicolumn{1}{c}{$22.20$} \\
	\bottomrule
	\end{tabular}
	\label{table:ablation_cycle}
\end{table}

\section{Discussion and Conclusion}
In this work, we propose a StyleGAN inversion method, named cycle encoding, for a high-quality pivot code in PTI. We demonstrate the superior performance of our method in both reconstruction and editability compared to several state-of-the-art methods. Our method even enables high-quality reconstruction and editing for out-of-domain cartoon images. Although our model successfully reconstructs the presented cartoon examples, it can hardly reconstruct more cartoonized images. In the future, we plan to develop a model with stronger generalization ability, so as to invert more challenging images from different domains such as cartoons and sketches.

\section{Acknowledgement}
This work is supported by the National Natural Science Foundation of China (No.62176223).

\clearpage

\bibliographystyle{ACM-Reference-Format}
\bibliography{ref}


\begin{thebibliography}{65}


\ifx \showCODEN    \undefined \def \showCODEN     #1{\unskip}     \fi
\ifx \showDOI      \undefined \def \showDOI       #1{#1}\fi
\ifx \showISBNx    \undefined \def \showISBNx     #1{\unskip}     \fi
\ifx \showISBNxiii \undefined \def \showISBNxiii  #1{\unskip}     \fi
\ifx \showISSN     \undefined \def \showISSN      #1{\unskip}     \fi
\ifx \showLCCN     \undefined \def \showLCCN      #1{\unskip}     \fi
\ifx \shownote     \undefined \def \shownote      #1{#1}          \fi
\ifx \showarticletitle \undefined \def \showarticletitle #1{#1}   \fi
\ifx \showURL      \undefined \def \showURL       {\relax}        \fi
\providecommand\bibfield[2]{#2}
\providecommand\bibinfo[2]{#2}
\providecommand\natexlab[1]{#1}
\providecommand\showeprint[2][]{arXiv:#2}

\bibitem[\protect\citeauthoryear{Abdal, Qin, and Wonka}{Abdal
  et~al\mbox{.}}{2019}]%
        {Abdal2019}
\bibfield{author}{\bibinfo{person}{Rameen Abdal}, \bibinfo{person}{Yipeng Qin},
  {and} \bibinfo{person}{Peter Wonka}.} \bibinfo{year}{2019}\natexlab{}.
\newblock \showarticletitle{Image2stylegan: How to embed images into the
  stylegan latent space?}. In \bibinfo{booktitle}{\emph{Proceedings of
  International Conference on Computer Vision}}.
\newblock


\bibitem[\protect\citeauthoryear{Abdal, Qin, and Wonka}{Abdal
  et~al\mbox{.}}{2020}]%
        {Abdal2020}
\bibfield{author}{\bibinfo{person}{Rameen Abdal}, \bibinfo{person}{Yipeng Qin},
  {and} \bibinfo{person}{Peter Wonka}.} \bibinfo{year}{2020}\natexlab{}.
\newblock \showarticletitle{Image2StyleGAN++: How to Edit the Embedded
  Images?}. In \bibinfo{booktitle}{\emph{Proceedings of IEEE Conference on
  Computer Vision and Pattern Recognition}}.
\newblock


\bibitem[\protect\citeauthoryear{Abdal, Zhu, Mitra, and Wonka}{Abdal
  et~al\mbox{.}}{2021}]%
        {Abdal2021}
\bibfield{author}{\bibinfo{person}{Rameen Abdal}, \bibinfo{person}{Peihao Zhu},
  \bibinfo{person}{Niloy~J. Mitra}, {and} \bibinfo{person}{Peter Wonka}.}
  \bibinfo{year}{2021}\natexlab{}.
\newblock \showarticletitle{StyleFlow: Attribute-Conditioned Exploration of
  StyleGAN-Generated Images Using Conditional Continuous Normalizing Flows}.
\newblock \bibinfo{journal}{\emph{ACM Transactions on Graphics}}
  \bibinfo{volume}{40}, \bibinfo{number}{3} (\bibinfo{year}{2021}).
\newblock


\bibitem[\protect\citeauthoryear{{Abu Hussein}, {Tirer}, and {Giryes}}{{Abu
  Hussein} et~al\mbox{.}}{2020}]%
        {Hussein2020}
\bibfield{author}{\bibinfo{person}{Shady {Abu Hussein}}, \bibinfo{person}{Tom
  {Tirer}}, {and} \bibinfo{person}{Raja {Giryes}}.}
  \bibinfo{year}{2020}\natexlab{}.
\newblock \showarticletitle{Image-Adaptive GAN based Reconstruction}. In
  \bibinfo{booktitle}{\emph{Proceedings of AAAI Conference}}.
\newblock


\bibitem[\protect\citeauthoryear{Alaluf, Patashnik, and Cohen-Or}{Alaluf
  et~al\mbox{.}}{2021a}]%
        {Alaluf2021_2}
\bibfield{author}{\bibinfo{person}{Yuval Alaluf}, \bibinfo{person}{Or
  Patashnik}, {and} \bibinfo{person}{Daniel Cohen-Or}.}
  \bibinfo{year}{2021}\natexlab{a}.
\newblock \showarticletitle{Only a Matter of Style: Age Transformation Using a
  Style-Based Regression Model}.
\newblock \bibinfo{journal}{\emph{arXiv preprint arXiv:2102.02754}}
  (\bibinfo{year}{2021}).
\newblock


\bibitem[\protect\citeauthoryear{Alaluf, Patashnik, and Cohen{-}Or}{Alaluf
  et~al\mbox{.}}{2021b}]%
        {Alaluf2021}
\bibfield{author}{\bibinfo{person}{Yuval Alaluf}, \bibinfo{person}{Or
  Patashnik}, {and} \bibinfo{person}{Daniel Cohen{-}Or}.}
  \bibinfo{year}{2021}\natexlab{b}.
\newblock \showarticletitle{ReStyle: {A} Residual-Based StyleGAN Encoder via
  Iterative Refinement}.
\newblock \bibinfo{journal}{\emph{arXiv preprint arXiv:2104.02699}}
  (\bibinfo{year}{2021}).
\newblock


\bibitem[\protect\citeauthoryear{Bau, Strobelt, Peebles, Wulff, Zhou, Zhu, and
  Torralba}{Bau et~al\mbox{.}}{2019}]%
        {Bau2019}
\bibfield{author}{\bibinfo{person}{David Bau}, \bibinfo{person}{Hendrik
  Strobelt}, \bibinfo{person}{William Peebles}, \bibinfo{person}{Jonas Wulff},
  \bibinfo{person}{Bolei Zhou}, \bibinfo{person}{Jun-Yan Zhu}, {and}
  \bibinfo{person}{Antonio Torralba}.} \bibinfo{year}{2019}\natexlab{}.
\newblock \showarticletitle{Semantic photo manipulation with a generative image
  prior}.
\newblock \bibinfo{journal}{\emph{ACM Transactions on Graphics}}
  \bibinfo{volume}{38}, \bibinfo{number}{4} (\bibinfo{year}{2019}).
\newblock


\bibitem[\protect\citeauthoryear{Baylies}{Baylies}{2019}]%
        {Baylies2019}
\bibfield{author}{\bibinfo{person}{Baylies}.} \bibinfo{year}{2019}\natexlab{}.
\newblock \bibinfo{title}{https://github.com/pbaylies/stylegan-encoder}.
\newblock
\newblock


\bibitem[\protect\citeauthoryear{Brock, Donahue, and Simonyan}{Brock
  et~al\mbox{.}}{2018}]%
        {Brock2018}
\bibfield{author}{\bibinfo{person}{Andrew Brock}, \bibinfo{person}{Jeff
  Donahue}, {and} \bibinfo{person}{Karen Simonyan}.}
  \bibinfo{year}{2018}\natexlab{}.
\newblock \showarticletitle{Large Scale GAN Training for High Fidelity Natural
  Image Synthesis}. In \bibinfo{booktitle}{\emph{Proceedings of International
  Conference on Learning Representations}}.
\newblock


\bibitem[\protect\citeauthoryear{Chai, Wulff, and Isola}{Chai
  et~al\mbox{.}}{2021}]%
        {Chai2021}
\bibfield{author}{\bibinfo{person}{Lucy Chai}, \bibinfo{person}{Jonas Wulff},
  {and} \bibinfo{person}{Phillip Isola}.} \bibinfo{year}{2021}\natexlab{}.
\newblock \showarticletitle{Using latent space regression to analyze and
  leverage compositionality in GANs}. In \bibinfo{booktitle}{\emph{Proceedings
  of International Conference on Learning Representations}}.
\newblock


\bibitem[\protect\citeauthoryear{Cherepkov, Voynov, and Babenko}{Cherepkov
  et~al\mbox{.}}{2021}]%
        {Cherepkov2021}
\bibfield{author}{\bibinfo{person}{Anton Cherepkov}, \bibinfo{person}{Andrey
  Voynov}, {and} \bibinfo{person}{Artem Babenko}.}
  \bibinfo{year}{2021}\natexlab{}.
\newblock \showarticletitle{Navigating the GAN Parameter Space for Semantic
  Image Editing}. In \bibinfo{booktitle}{\emph{Proceedings of IEEE Conference
  on Computer Vision and Pattern Recognition}}.
\newblock


\bibitem[\protect\citeauthoryear{Chong, Lee, and Forsyth}{Chong
  et~al\mbox{.}}{2021}]%
        {Chong2021}
\bibfield{author}{\bibinfo{person}{Min~Jin Chong}, \bibinfo{person}{Hsin-Ying
  Lee}, {and} \bibinfo{person}{David Forsyth}.}
  \bibinfo{year}{2021}\natexlab{}.
\newblock \showarticletitle{StyleGAN of All Trades: Image Manipulation with
  Only Pretrained StyleGAN}.
\newblock \bibinfo{journal}{\emph{arXiv preprint arXiv:2111.01619}}
  (\bibinfo{year}{2021}).
\newblock


\bibitem[\protect\citeauthoryear{Collins, Bala, Price, and Susstrunk}{Collins
  et~al\mbox{.}}{2020}]%
        {Collins2020}
\bibfield{author}{\bibinfo{person}{Edo Collins}, \bibinfo{person}{Raja Bala},
  \bibinfo{person}{Bob Price}, {and} \bibinfo{person}{Sabine Susstrunk}.}
  \bibinfo{year}{2020}\natexlab{}.
\newblock \showarticletitle{Editing in Style: Uncovering the Local Semantics of
  GANs}. In \bibinfo{booktitle}{\emph{Proceedings of IEEE Conference on
  Computer Vision and Pattern Recognition}}.
\newblock


\bibitem[\protect\citeauthoryear{Creswell and Bharath}{Creswell and
  Bharath}{2018}]%
        {Creswell2018}
\bibfield{author}{\bibinfo{person}{Antonia Creswell} {and}
  \bibinfo{person}{Anil~Anthony Bharath}.} \bibinfo{year}{2018}\natexlab{}.
\newblock \showarticletitle{Inverting the generator of a generative adversarial
  network}.
\newblock \bibinfo{journal}{\emph{IEEE Transactions on Neural Networks and
  Learning Systems}} \bibinfo{volume}{30}, \bibinfo{number}{7}
  (\bibinfo{year}{2018}), \bibinfo{pages}{1967--1974}.
\newblock


\bibitem[\protect\citeauthoryear{Deng, Guo, Xue, and Zafeiriou}{Deng
  et~al\mbox{.}}{2019}]%
        {Deng2019}
\bibfield{author}{\bibinfo{person}{Jiankang Deng}, \bibinfo{person}{Jia Guo},
  \bibinfo{person}{Niannan Xue}, {and} \bibinfo{person}{Stefanos Zafeiriou}.}
  \bibinfo{year}{2019}\natexlab{}.
\newblock \showarticletitle{Arcface: Additive angular margin loss for deep face
  recognition}. In \bibinfo{booktitle}{\emph{Proceedings of IEEE Conference on
  Computer Vision and Pattern Recognition}}.
\newblock


\bibitem[\protect\citeauthoryear{Futschik, Luk\'{a}\v{c}, Shechtman, and
  S\'{y}kora}{Futschik et~al\mbox{.}}{2021}]%
        {Futschik2021}
\bibfield{author}{\bibinfo{person}{David Futschik}, \bibinfo{person}{Michal
  Luk\'{a}\v{c}}, \bibinfo{person}{Eli Shechtman}, {and}
  \bibinfo{person}{Daniel S\'{y}kora}.} \bibinfo{year}{2021}\natexlab{}.
\newblock \showarticletitle{Real Image Inversion via Segments}.
\newblock \bibinfo{journal}{\emph{arXiv preprint arXiv:2110.06269}}
  (\bibinfo{year}{2021}).
\newblock


\bibitem[\protect\citeauthoryear{Goetschalckx, Andonian, Oliva, and
  Isola}{Goetschalckx et~al\mbox{.}}{2019}]%
        {Goetschalckx2019}
\bibfield{author}{\bibinfo{person}{Lore Goetschalckx}, \bibinfo{person}{Alex
  Andonian}, \bibinfo{person}{Aude Oliva}, {and} \bibinfo{person}{Phillip
  Isola}.} \bibinfo{year}{2019}\natexlab{}.
\newblock \showarticletitle{GANalyze: Toward Visual Definitions of Cognitive
  Image Properties}. In \bibinfo{booktitle}{\emph{Proceedings of International
  Conference on Computer Vision}}.
\newblock


\bibitem[\protect\citeauthoryear{Goodfellow, Pouget-Abadie, Mirza, Xu,
  Warde-Farley, Ozair, Courville, and Bengio}{Goodfellow et~al\mbox{.}}{2014}]%
        {Goodfellow2014}
\bibfield{author}{\bibinfo{person}{Ian Goodfellow}, \bibinfo{person}{Jean
  Pouget-Abadie}, \bibinfo{person}{Mehdi Mirza}, \bibinfo{person}{Bing Xu},
  \bibinfo{person}{David Warde-Farley}, \bibinfo{person}{Sherjil Ozair},
  \bibinfo{person}{Aaron Courville}, {and} \bibinfo{person}{Yoshua Bengio}.}
  \bibinfo{year}{2014}\natexlab{}.
\newblock \showarticletitle{Generative Adversarial Nets}. In
  \bibinfo{booktitle}{\emph{Proceedings of Neural Information Processing
  Systems}}, Vol.~\bibinfo{volume}{27}.
\newblock


\bibitem[\protect\citeauthoryear{Guan, Tai, Ni, Zhu, Huang, and Yang}{Guan
  et~al\mbox{.}}{2020}]%
        {Guan2020}
\bibfield{author}{\bibinfo{person}{Shanyan Guan}, \bibinfo{person}{Ying Tai},
  \bibinfo{person}{Bingbing Ni}, \bibinfo{person}{Feida Zhu},
  \bibinfo{person}{Feiyue Huang}, {and} \bibinfo{person}{Xiaokang Yang}.}
  \bibinfo{year}{2020}\natexlab{}.
\newblock \showarticletitle{Collaborative Learning for Faster StyleGAN
  Embedding}.
\newblock \bibinfo{journal}{\emph{arXiv preprint arXiv:2007.01758}}
  (\bibinfo{year}{2020}).
\newblock


\bibitem[\protect\citeauthoryear{H{\"a}rk{\"o}nen, Hertzmann, Lehtinen, and
  Paris}{H{\"a}rk{\"o}nen et~al\mbox{.}}{2020}]%
        {Erik2020}
\bibfield{author}{\bibinfo{person}{Erik H{\"a}rk{\"o}nen},
  \bibinfo{person}{Aaron Hertzmann}, \bibinfo{person}{Jaakko Lehtinen}, {and}
  \bibinfo{person}{Sylvain Paris}.} \bibinfo{year}{2020}\natexlab{}.
\newblock \showarticletitle{GANSpace: Discovering Interpretable GAN Controls}.
  In \bibinfo{booktitle}{\emph{Proceedings of Neural Information Processing
  Systems}}.
\newblock


\bibitem[\protect\citeauthoryear{Huang, Wang, Tai, Liu, Shen, Li, Li, and
  Huang}{Huang et~al\mbox{.}}{2020}]%
        {Huang2020}
\bibfield{author}{\bibinfo{person}{Yuge Huang}, \bibinfo{person}{Yuhan Wang},
  \bibinfo{person}{Ying Tai}, \bibinfo{person}{Xiaoming Liu},
  \bibinfo{person}{Pengcheng Shen}, \bibinfo{person}{Shaoxin Li},
  \bibinfo{person}{Jilin Li}, {and} \bibinfo{person}{Feiyue Huang}.}
  \bibinfo{year}{2020}\natexlab{}.
\newblock \showarticletitle{CurricularFace: Adaptive Curriculum Learning Loss
  for Deep Face Recognition}. In \bibinfo{booktitle}{\emph{Proceedings of IEEE
  Conference on Computer Vision and Pattern Recognition}}.
\newblock


\bibitem[\protect\citeauthoryear{Huh, Zhang, Zhu, Paris, and Hertzmann}{Huh
  et~al\mbox{.}}{2020}]%
        {Huh2020}
\bibfield{author}{\bibinfo{person}{Minyoung Huh}, \bibinfo{person}{Richard
  Zhang}, \bibinfo{person}{Jun-Yan Zhu}, \bibinfo{person}{Sylvain Paris}, {and}
  \bibinfo{person}{Aaron Hertzmann}.} \bibinfo{year}{2020}\natexlab{}.
\newblock \showarticletitle{Transforming and Projecting Images into
  Class-conditional Generative Networks}. In
  \bibinfo{booktitle}{\emph{Proceedings of European Conference on Computer
  Vision}}.
\newblock


\bibitem[\protect\citeauthoryear{Jahanian, Chai, and Isola}{Jahanian
  et~al\mbox{.}}{2020}]%
        {Jahanian2020}
\bibfield{author}{\bibinfo{person}{Ali Jahanian}, \bibinfo{person}{Lucy Chai},
  {and} \bibinfo{person}{Phillip Isola}.} \bibinfo{year}{2020}\natexlab{}.
\newblock \showarticletitle{On the "steerability" of generative adversarial
  networks}. In \bibinfo{booktitle}{\emph{Proceedings of International
  Conference on Learning Representations}}.
\newblock


\bibitem[\protect\citeauthoryear{Karras, Aila, Laine, and Lehtinen}{Karras
  et~al\mbox{.}}{2018}]%
        {Karras2018}
\bibfield{author}{\bibinfo{person}{Tero Karras}, \bibinfo{person}{Timo Aila},
  \bibinfo{person}{Samuli Laine}, {and} \bibinfo{person}{Jaakko Lehtinen}.}
  \bibinfo{year}{2018}\natexlab{}.
\newblock \showarticletitle{Progressive growing of gans for improved quality,
  stability, and variation}. In \bibinfo{booktitle}{\emph{Proceedings of
  International Conference on Learning Representations}}.
\newblock


\bibitem[\protect\citeauthoryear{Karras, Aittala, Hellsten, Laine, Lehtinen,
  and Aila}{Karras et~al\mbox{.}}{2020a}]%
        {Karras2020_2}
\bibfield{author}{\bibinfo{person}{Tero Karras}, \bibinfo{person}{Miika
  Aittala}, \bibinfo{person}{Janne Hellsten}, \bibinfo{person}{Samuli Laine},
  \bibinfo{person}{Jaakko Lehtinen}, {and} \bibinfo{person}{Timo Aila}.}
  \bibinfo{year}{2020}\natexlab{a}.
\newblock \showarticletitle{Training Generative Adversarial Networks with
  Limited Data}. In \bibinfo{booktitle}{\emph{Proceedings of Neural Information
  Processing Systems}}.
\newblock


\bibitem[\protect\citeauthoryear{Karras, Laine, and Aila}{Karras
  et~al\mbox{.}}{2019}]%
        {Karras2019}
\bibfield{author}{\bibinfo{person}{Tero Karras}, \bibinfo{person}{Samuli
  Laine}, {and} \bibinfo{person}{Timo Aila}.} \bibinfo{year}{2019}\natexlab{}.
\newblock \showarticletitle{A style-based generator architecture for generative
  adversarial networks}. In \bibinfo{booktitle}{\emph{Proceedings of IEEE
  Conference on Computer Vision and Pattern Recognition}}.
\newblock


\bibitem[\protect\citeauthoryear{Karras, Laine, Aittala, Hellsten, Lehtinen,
  and Aila}{Karras et~al\mbox{.}}{2020b}]%
        {Karras2020}
\bibfield{author}{\bibinfo{person}{Tero Karras}, \bibinfo{person}{Samuli
  Laine}, \bibinfo{person}{Miika Aittala}, \bibinfo{person}{Janne Hellsten},
  \bibinfo{person}{Jaakko Lehtinen}, {and} \bibinfo{person}{Timo Aila}.}
  \bibinfo{year}{2020}\natexlab{b}.
\newblock \showarticletitle{Analyzing and improving the image quality of
  stylegan}. In \bibinfo{booktitle}{\emph{Proceedings of IEEE Conference on
  Computer Vision and Pattern Recognition}}.
\newblock


\bibitem[\protect\citeauthoryear{Kim, Choi, Kim, Yoo, and Uh}{Kim
  et~al\mbox{.}}{2021}]%
        {Kim2021}
\bibfield{author}{\bibinfo{person}{Hyunsu Kim}, \bibinfo{person}{Yunjey Choi},
  \bibinfo{person}{Junho Kim}, \bibinfo{person}{Sungjoo Yoo}, {and}
  \bibinfo{person}{Youngjung Uh}.} \bibinfo{year}{2021}\natexlab{}.
\newblock \showarticletitle{Exploiting Spatial Dimensions of Latent in GAN for
  Real-time Image Editing}. In \bibinfo{booktitle}{\emph{Proceedings of IEEE
  Conference on Computer Vision and Pattern Recognition}}.
\newblock


\bibitem[\protect\citeauthoryear{Li, Liu, Wei, Zhang, Wu, Xu, and Wong}{Li
  et~al\mbox{.}}{2021}]%
        {Li2021}
\bibfield{author}{\bibinfo{person}{Guanyue Li}, \bibinfo{person}{Yi Liu},
  \bibinfo{person}{Xiwen Wei}, \bibinfo{person}{Yang Zhang},
  \bibinfo{person}{Si Wu}, \bibinfo{person}{Yong Xu}, {and}
  \bibinfo{person}{Hau-San Wong}.} \bibinfo{year}{2021}\natexlab{}.
\newblock \showarticletitle{Discovering Density-Preserving Latent Space Walks
  in GANs for Semantic Image Transformations}. In
  \bibinfo{booktitle}{\emph{Proceedings of ACM International Conference on
  Multimedia}}.
\newblock


\bibitem[\protect\citeauthoryear{Lipton and Tripathi}{Lipton and
  Tripathi}{2017}]%
        {Lipton2017}
\bibfield{author}{\bibinfo{person}{Zachary~C Lipton} {and}
  \bibinfo{person}{Subarna Tripathi}.} \bibinfo{year}{2017}\natexlab{}.
\newblock \showarticletitle{Precise recovery of latent vectors from generative
  adversarial networks}. In \bibinfo{booktitle}{\emph{Proceedings of
  International Conference on Learning Representations Workshops}}.
\newblock


\bibitem[\protect\citeauthoryear{Luo, Xu, Tang, and Lv}{Luo
  et~al\mbox{.}}{2017}]%
        {Luo2017}
\bibfield{author}{\bibinfo{person}{Junyu Luo}, \bibinfo{person}{Yong Xu},
  \bibinfo{person}{Chenwei Tang}, {and} \bibinfo{person}{Jiancheng Lv}.}
  \bibinfo{year}{2017}\natexlab{}.
\newblock \showarticletitle{Learning Inverse Mapping by AutoEncoder Based
  Generative Adversarial Nets}. In \bibinfo{booktitle}{\emph{Proceedings of
  Neural Information Processing Systems}}.
\newblock


\bibitem[\protect\citeauthoryear{Menon, Damian, Hu, Ravi, and Rudin}{Menon
  et~al\mbox{.}}{2020}]%
        {Menon2020}
\bibfield{author}{\bibinfo{person}{Sachit Menon}, \bibinfo{person}{Alexandru
  Damian}, \bibinfo{person}{Shijia Hu}, \bibinfo{person}{Nikhil Ravi}, {and}
  \bibinfo{person}{Cynthia Rudin}.} \bibinfo{year}{2020}\natexlab{}.
\newblock \showarticletitle{PULSE: Self-Supervised Photo Upsampling via Latent
  Space Exploration of Generative Models}. In
  \bibinfo{booktitle}{\emph{Proceedings of IEEE Conference on Computer Vision
  and Pattern Recognition}}.
\newblock


\bibitem[\protect\citeauthoryear{Nguyen, Dosovitskiy, Yosinski, Brox, and
  Clune}{Nguyen et~al\mbox{.}}{2016}]%
        {Nguyen2016}
\bibfield{author}{\bibinfo{person}{Anh Nguyen}, \bibinfo{person}{Alexey
  Dosovitskiy}, \bibinfo{person}{Jason Yosinski}, \bibinfo{person}{Thomas
  Brox}, {and} \bibinfo{person}{Jeff Clune}.} \bibinfo{year}{2016}\natexlab{}.
\newblock \showarticletitle{Synthesizing the preferred inputs for neurons in
  neural networks via deep generator networks}. In
  \bibinfo{booktitle}{\emph{Proceedings of Neural Information Processing
  Systems}}.
\newblock


\bibitem[\protect\citeauthoryear{Nitzan, Bermano, Li, and Cohen-Or}{Nitzan
  et~al\mbox{.}}{2020}]%
        {Nitzan2020}
\bibfield{author}{\bibinfo{person}{Yotam Nitzan}, \bibinfo{person}{A. Bermano},
  \bibinfo{person}{Yangyan Li}, {and} \bibinfo{person}{D. Cohen-Or}.}
  \bibinfo{year}{2020}\natexlab{}.
\newblock \showarticletitle{Face identity disentanglement via latent space
  mapping}.
\newblock \bibinfo{journal}{\emph{ACM Transactions on Graphics}}
  \bibinfo{volume}{39} (\bibinfo{year}{2020}), \bibinfo{pages}{1 -- 14}.
\newblock


\bibitem[\protect\citeauthoryear{Pan, Zhan, Dai, Lin, Loy, and Luo}{Pan
  et~al\mbox{.}}{2020}]%
        {Pan2020}
\bibfield{author}{\bibinfo{person}{Xingang Pan}, \bibinfo{person}{Xiaohang
  Zhan}, \bibinfo{person}{Bo Dai}, \bibinfo{person}{Dahua Lin},
  \bibinfo{person}{Chen~Change Loy}, {and} \bibinfo{person}{Ping Luo}.}
  \bibinfo{year}{2020}\natexlab{}.
\newblock \showarticletitle{Exploiting deep generative prior for versatile
  image restoration and manipulation}. In \bibinfo{booktitle}{\emph{Proceedings
  of European Conference on Computer Vision}}.
\newblock


\bibitem[\protect\citeauthoryear{Patashnik, Wu, Shechtman, Cohen-Or, and
  Lischinski}{Patashnik et~al\mbox{.}}{2021}]%
        {Patashnik2021}
\bibfield{author}{\bibinfo{person}{Or Patashnik}, \bibinfo{person}{Zongze Wu},
  \bibinfo{person}{Eli Shechtman}, \bibinfo{person}{Daniel Cohen-Or}, {and}
  \bibinfo{person}{Dani Lischinski}.} \bibinfo{year}{2021}\natexlab{}.
\newblock \showarticletitle{StyleCLIP: Text-Driven Manipulation of StyleGAN
  Imagery}. In \bibinfo{booktitle}{\emph{Proceedings of International
  Conference on Computer Vision}}.
\newblock


\bibitem[\protect\citeauthoryear{Perarnau, Van De~Weijer, Raducanu, and
  {\'A}lvarez}{Perarnau et~al\mbox{.}}{2016}]%
        {Perarnau2016}
\bibfield{author}{\bibinfo{person}{Guim Perarnau}, \bibinfo{person}{Joost Van
  De~Weijer}, \bibinfo{person}{Bogdan Raducanu}, {and} \bibinfo{person}{Jose~M
  {\'A}lvarez}.} \bibinfo{year}{2016}\natexlab{}.
\newblock \showarticletitle{Invertible conditional gans for image editing}.
\newblock \bibinfo{journal}{\emph{arXiv preprint arXiv:1611.06355}}
  (\bibinfo{year}{2016}).
\newblock


\bibitem[\protect\citeauthoryear{Pidhorskyi, Adjeroh, and Doretto}{Pidhorskyi
  et~al\mbox{.}}{2020}]%
        {Pidhorskyi2020}
\bibfield{author}{\bibinfo{person}{Stanislav Pidhorskyi},
  \bibinfo{person}{Donald Adjeroh}, {and} \bibinfo{person}{Gianfranco
  Doretto}.} \bibinfo{year}{2020}\natexlab{}.
\newblock \showarticletitle{Adversarial Latent Autoencoders}. In
  \bibinfo{booktitle}{\emph{Proceedings of IEEE Conference on Computer Vision
  and Pattern Recognition}}.
\newblock


\bibitem[\protect\citeauthoryear{Plumerault, Borgne, and Hudelot}{Plumerault
  et~al\mbox{.}}{2020}]%
        {Plumerault2020}
\bibfield{author}{\bibinfo{person}{Antoine Plumerault},
  \bibinfo{person}{Herv{\'e}~Le Borgne}, {and} \bibinfo{person}{C{\'e}line
  Hudelot}.} \bibinfo{year}{2020}\natexlab{}.
\newblock \showarticletitle{Controlling generative models with continuous
  factors of variations}. In \bibinfo{booktitle}{\emph{Proceedings of
  International Conference on Learning Representations}}.
\newblock


\bibitem[\protect\citeauthoryear{Raj, Li, and Bresler}{Raj
  et~al\mbox{.}}{2019}]%
        {Raj2019}
\bibfield{author}{\bibinfo{person}{Ankit Raj}, \bibinfo{person}{Yuqi Li}, {and}
  \bibinfo{person}{Yoram Bresler}.} \bibinfo{year}{2019}\natexlab{}.
\newblock \showarticletitle{GAN-based Projector for Faster Recovery with
  Convergence Guarantees in Linear Inverse Problems}. In
  \bibinfo{booktitle}{\emph{Proceedings of International Conference on Computer
  Vision}}.
\newblock


\bibitem[\protect\citeauthoryear{Richardson, Alaluf, Patashnik, Nitzan, Azar,
  Shapiro, and Cohen-Or}{Richardson et~al\mbox{.}}{2021}]%
        {Richardson2021}
\bibfield{author}{\bibinfo{person}{Elad Richardson}, \bibinfo{person}{Yuval
  Alaluf}, \bibinfo{person}{Or Patashnik}, \bibinfo{person}{Yotam Nitzan},
  \bibinfo{person}{Yaniv Azar}, \bibinfo{person}{Stav Shapiro}, {and}
  \bibinfo{person}{Daniel Cohen-Or}.} \bibinfo{year}{2021}\natexlab{}.
\newblock \showarticletitle{Encoding in Style: a StyleGAN Encoder for
  Image-to-Image Translation}. In \bibinfo{booktitle}{\emph{Proceedings of IEEE
  Conference on Computer Vision and Pattern Recognition}}.
\newblock


\bibitem[\protect\citeauthoryear{Roich, Mokady, Bermano, and Cohen-Or}{Roich
  et~al\mbox{.}}{2021}]%
        {Roich2021}
\bibfield{author}{\bibinfo{person}{Daniel Roich}, \bibinfo{person}{Ron Mokady},
  \bibinfo{person}{Amit~H. Bermano}, {and} \bibinfo{person}{Daniel Cohen-Or}.}
  \bibinfo{year}{2021}\natexlab{}.
\newblock \showarticletitle{Pivotal Tuning for Latent-based Editing of Real
  Images}.
\newblock \bibinfo{journal}{\emph{arXiv preprint arXiv:2106.05744}}
  (\bibinfo{year}{2021}).
\newblock


\bibitem[\protect\citeauthoryear{Shen, Gu, Tang, and Zhou}{Shen
  et~al\mbox{.}}{2020}]%
        {Shen2020}
\bibfield{author}{\bibinfo{person}{Yujun Shen}, \bibinfo{person}{Jinjin Gu},
  \bibinfo{person}{Xiaoou Tang}, {and} \bibinfo{person}{Bolei Zhou}.}
  \bibinfo{year}{2020}\natexlab{}.
\newblock \showarticletitle{Interpreting the latent space of gans for semantic
  face editing}. In \bibinfo{booktitle}{\emph{Proceedings of IEEE Conference on
  Computer Vision and Pattern Recognition}}.
\newblock


\bibitem[\protect\citeauthoryear{Shen and Zhou}{Shen and Zhou}{2021}]%
        {Shen2021}
\bibfield{author}{\bibinfo{person}{Yujun Shen} {and} \bibinfo{person}{Bolei
  Zhou}.} \bibinfo{year}{2021}\natexlab{}.
\newblock \showarticletitle{Closed-Form Factorization of Latent Semantics in
  GANs}. In \bibinfo{booktitle}{\emph{Proceedings of IEEE Conference on
  Computer Vision and Pattern Recognition}}.
\newblock


\bibitem[\protect\citeauthoryear{Shukor, Yao, Damodaran, and Hellier}{Shukor
  et~al\mbox{.}}{2021}]%
        {Shukor2021}
\bibfield{author}{\bibinfo{person}{Mustafa Shukor}, \bibinfo{person}{Xu Yao},
  \bibinfo{person}{Bharath~Bhushan Damodaran}, {and} \bibinfo{person}{Pierre
  Hellier}.} \bibinfo{year}{2021}\natexlab{}.
\newblock \showarticletitle{Semantic and Geometric Unfolding of StyleGAN Latent
  Space}.
\newblock \bibinfo{journal}{\emph{arXiv preprint arXiv:2107.04481}}
  (\bibinfo{year}{2021}).
\newblock


\bibitem[\protect\citeauthoryear{Spingarn-Eliezer, Banner, and
  Michaeli}{Spingarn-Eliezer et~al\mbox{.}}{2021}]%
        {Spingarn-Eliezer2021}
\bibfield{author}{\bibinfo{person}{Nurit Spingarn-Eliezer},
  \bibinfo{person}{Ron Banner}, {and} \bibinfo{person}{Tomer Michaeli}.}
  \bibinfo{year}{2021}\natexlab{}.
\newblock \showarticletitle{GAN Steerability without optimization}. In
  \bibinfo{booktitle}{\emph{Proceedings of International Conference on Learning
  Representations}}.
\newblock


\bibitem[\protect\citeauthoryear{Tewari, Elgharib, Bharaj, Bernard, Seidel,
  P{\'e}rez, Zollh{\"o}fer, and Theobalt}{Tewari et~al\mbox{.}}{2020a}]%
        {Tewari2020}
\bibfield{author}{\bibinfo{person}{Ayush Tewari}, \bibinfo{person}{Mohamed
  Elgharib}, \bibinfo{person}{Gaurav Bharaj}, \bibinfo{person}{Florian
  Bernard}, \bibinfo{person}{Hans-Peter Seidel}, \bibinfo{person}{Patrick
  P{\'e}rez}, \bibinfo{person}{Michael Zollh{\"o}fer}, {and}
  \bibinfo{person}{Christian Theobalt}.} \bibinfo{year}{2020}\natexlab{a}.
\newblock \showarticletitle{StyleRig: Rigging StyleGAN for 3D Control over
  Portrait Images}.
\newblock \bibinfo{journal}{\emph{arXiv preprint arXiv:2004.00121}}
  (\bibinfo{year}{2020}).
\newblock


\bibitem[\protect\citeauthoryear{Tewari, Elgharib, R., Bernard, Seidel, Pérez,
  Zollhöfer, and Theobalt}{Tewari et~al\mbox{.}}{2020b}]%
        {Tewari2020_2}
\bibfield{author}{\bibinfo{person}{Ayush Tewari}, \bibinfo{person}{Mohamed
  Elgharib}, \bibinfo{person}{Mallikarjun~B R.}, \bibinfo{person}{Florian
  Bernard}, \bibinfo{person}{Hans-Peter Seidel}, \bibinfo{person}{Patrick
  Pérez}, \bibinfo{person}{Michael Zollhöfer}, {and}
  \bibinfo{person}{Christian Theobalt}.} \bibinfo{year}{2020}\natexlab{b}.
\newblock \showarticletitle{PIE: Portrait Image Embedding for Semantic
  Control}.
\newblock \bibinfo{journal}{\emph{ACM Transactions on Graphics}}
  \bibinfo{volume}{39}, \bibinfo{number}{6} (\bibinfo{year}{2020}).
\newblock


\bibitem[\protect\citeauthoryear{Tov, Alaluf, Nitzan, Patashnik, and
  Cohen-Or}{Tov et~al\mbox{.}}{2021}]%
        {Tov2021}
\bibfield{author}{\bibinfo{person}{Omer Tov}, \bibinfo{person}{Yuval Alaluf},
  \bibinfo{person}{Yotam Nitzan}, \bibinfo{person}{Or Patashnik}, {and}
  \bibinfo{person}{Daniel Cohen-Or}.} \bibinfo{year}{2021}\natexlab{}.
\newblock \showarticletitle{Designing an Encoder for StyleGAN Image
  Manipulation}.
\newblock \bibinfo{journal}{\emph{ACM Transactions on Graphics}}
  \bibinfo{volume}{40}, \bibinfo{number}{4} (\bibinfo{year}{2021}).
\newblock


\bibitem[\protect\citeauthoryear{Viazovetskyi, Ivashkin, and
  Kashin}{Viazovetskyi et~al\mbox{.}}{2020}]%
        {Viazovetskyi2020}
\bibfield{author}{\bibinfo{person}{Yuri Viazovetskyi},
  \bibinfo{person}{Vladimir Ivashkin}, {and} \bibinfo{person}{Evgeny Kashin}.}
  \bibinfo{year}{2020}\natexlab{}.
\newblock \showarticletitle{StyleGAN2 Distillation for Feed-forward Image
  Manipulation}. In \bibinfo{booktitle}{\emph{Proceedings of European
  Conference on Computer Vision}}.
\newblock


\bibitem[\protect\citeauthoryear{Voynov and Babenko}{Voynov and
  Babenko}{2020}]%
        {Voynov2020}
\bibfield{author}{\bibinfo{person}{Andrey Voynov} {and} \bibinfo{person}{Artem
  Babenko}.} \bibinfo{year}{2020}\natexlab{}.
\newblock \showarticletitle{Unsupervised discovery of interpretable directions
  in the gan latent space}. In \bibinfo{booktitle}{\emph{Proceedings of
  International Conference on Machine Learning}}.
\newblock


\bibitem[\protect\citeauthoryear{Wang and Ponce}{Wang and Ponce}{2021}]%
        {Wang2021}
\bibfield{author}{\bibinfo{person}{Binxu Wang} {and} \bibinfo{person}{Carlos~R.
  Ponce}.} \bibinfo{year}{2021}\natexlab{}.
\newblock \showarticletitle{The Geometry of Deep Generative Image Models and
  its Applications}. In \bibinfo{booktitle}{\emph{Proceedings of International
  Conference on Learning Representations}}.
\newblock


\bibitem[\protect\citeauthoryear{Wang, Lin, Hoi, and Miao}{Wang
  et~al\mbox{.}}{2021a}]%
        {Wang2021_4}
\bibfield{author}{\bibinfo{person}{Hao Wang}, \bibinfo{person}{Guosheng Lin},
  \bibinfo{person}{Steven C.~H. Hoi}, {and} \bibinfo{person}{Chunyan Miao}.}
  \bibinfo{year}{2021}\natexlab{a}.
\newblock \showarticletitle{Cycle-Consistent Inverse GAN for Text-to-Image
  Synthesis}. In \bibinfo{booktitle}{\emph{Proceedings of ACM International
  Conference on Multimedia}}.
\newblock


\bibitem[\protect\citeauthoryear{Wang, Yu, and Fritz}{Wang
  et~al\mbox{.}}{2021b}]%
        {Wang2021_2}
\bibfield{author}{\bibinfo{person}{Hui-Po Wang}, \bibinfo{person}{Ning Yu},
  {and} \bibinfo{person}{Mario Fritz}.} \bibinfo{year}{2021}\natexlab{b}.
\newblock \showarticletitle{Hijack-GAN: Unintended-Use of Pretrained, Black-Box
  GANs}. In \bibinfo{booktitle}{\emph{Proceedings of IEEE Conference on
  Computer Vision and Pattern Recognition}}.
\newblock


\bibitem[\protect\citeauthoryear{Wang, Zhang, Fan, Wang, and Chen}{Wang
  et~al\mbox{.}}{2021c}]%
        {Wang2021_3}
\bibfield{author}{\bibinfo{person}{Tengfei Wang}, \bibinfo{person}{Yong Zhang},
  \bibinfo{person}{Yanbo Fan}, \bibinfo{person}{Jue Wang}, {and}
  \bibinfo{person}{Qifeng Chen}.} \bibinfo{year}{2021}\natexlab{c}.
\newblock \showarticletitle{High-Fidelity GAN Inversion for Image Attribute
  Editing}.
\newblock \bibinfo{journal}{\emph{arXiv preprint arXiv:2109.06590}}
  (\bibinfo{year}{2021}).
\newblock


\bibitem[\protect\citeauthoryear{Wei, Chen, Zhou, Liao, Zhang, Yuan, Hua, and
  Yu}{Wei et~al\mbox{.}}{2021}]%
        {Wei2021}
\bibfield{author}{\bibinfo{person}{Tianyi Wei}, \bibinfo{person}{Dongdong
  Chen}, \bibinfo{person}{Wenbo Zhou}, \bibinfo{person}{Jing Liao},
  \bibinfo{person}{Weiming Zhang}, \bibinfo{person}{Lu Yuan},
  \bibinfo{person}{Gang Hua}, {and} \bibinfo{person}{Nenghai Yu}.}
  \bibinfo{year}{2021}\natexlab{}.
\newblock \showarticletitle{A Simple Baseline for StyleGAN Inversion}.
\newblock \bibinfo{journal}{\emph{arXiv preprint arXiv:2104.07661}}
  (\bibinfo{year}{2021}).
\newblock


\bibitem[\protect\citeauthoryear{Wu, Lischinski, and Shechtman}{Wu
  et~al\mbox{.}}{2021a}]%
        {Wu2021}
\bibfield{author}{\bibinfo{person}{Zongze Wu}, \bibinfo{person}{Dani
  Lischinski}, {and} \bibinfo{person}{Eli Shechtman}.}
  \bibinfo{year}{2021}\natexlab{a}.
\newblock \showarticletitle{StyleSpace Analysis: Disentangled Controls for
  StyleGAN Image Generation}. In \bibinfo{booktitle}{\emph{Proceedings of IEEE
  Conference on Computer Vision and Pattern Recognition}}.
\newblock


\bibitem[\protect\citeauthoryear{Wu, Nitzan, Shechtman, and Lischinski}{Wu
  et~al\mbox{.}}{2021b}]%
        {Wu2021_2}
\bibfield{author}{\bibinfo{person}{Zongze Wu}, \bibinfo{person}{Yotam Nitzan},
  \bibinfo{person}{Eli Shechtman}, {and} \bibinfo{person}{Dani Lischinski}.}
  \bibinfo{year}{2021}\natexlab{b}.
\newblock \showarticletitle{StyleAlign: Analysis and Applications of Aligned
  StyleGAN Models}.
\newblock \bibinfo{journal}{\emph{arXiv preprint arXiv:2110.11323}}
  (\bibinfo{year}{2021}).
\newblock


\bibitem[\protect\citeauthoryear{Xia, Zhang, Yang, Xue, Zhou, and Yang}{Xia
  et~al\mbox{.}}{2021}]%
        {Xia2021}
\bibfield{author}{\bibinfo{person}{Weihao Xia}, \bibinfo{person}{Yulun Zhang},
  \bibinfo{person}{Yujiu Yang}, \bibinfo{person}{Jing-Hao Xue},
  \bibinfo{person}{Bolei Zhou}, {and} \bibinfo{person}{Ming-Hsuan Yang}.}
  \bibinfo{year}{2021}\natexlab{}.
\newblock \showarticletitle{GAN Inversion: A Survey}.
\newblock \bibinfo{journal}{\emph{arXiv preprint arXiv:2101.05278}}
  (\bibinfo{year}{2021}).
\newblock


\bibitem[\protect\citeauthoryear{Zhang, Bai, and Gao}{Zhang
  et~al\mbox{.}}{2021}]%
        {Zhang2021}
\bibfield{author}{\bibinfo{person}{Lingyun Zhang}, \bibinfo{person}{Xiuxiu
  Bai}, {and} \bibinfo{person}{Yao Gao}.} \bibinfo{year}{2021}\natexlab{}.
\newblock \showarticletitle{SalS-GAN: Spatially-Adaptive Latent Space in
  StyleGAN for Real Image Embedding}. In \bibinfo{booktitle}{\emph{Proceedings
  of ACM International Conference on Multimedia}}.
\newblock


\bibitem[\protect\citeauthoryear{Zhang, Isola, Efros, Shechtman, and
  Wang}{Zhang et~al\mbox{.}}{2018}]%
        {Zhang2018}
\bibfield{author}{\bibinfo{person}{Richard Zhang}, \bibinfo{person}{Phillip
  Isola}, \bibinfo{person}{Alexei~A. Efros}, \bibinfo{person}{Eli Shechtman},
  {and} \bibinfo{person}{Oliver Wang}.} \bibinfo{year}{2018}\natexlab{}.
\newblock \showarticletitle{The Unreasonable Effectiveness of Deep Features as
  a Perceptual Metric}. In \bibinfo{booktitle}{\emph{Proceedings of IEEE
  Conference on Computer Vision and Pattern Recognition}}.
\newblock


\bibitem[\protect\citeauthoryear{Zhu, Shen, Zhao, and Zhou}{Zhu
  et~al\mbox{.}}{2020b}]%
        {Zhu2020}
\bibfield{author}{\bibinfo{person}{Jiapeng Zhu}, \bibinfo{person}{Yujun Shen},
  \bibinfo{person}{Deli Zhao}, {and} \bibinfo{person}{Bolei Zhou}.}
  \bibinfo{year}{2020}\natexlab{b}.
\newblock \showarticletitle{In-domain gan inversion for real image editing}. In
  \bibinfo{booktitle}{\emph{Proceedings of European Conference on Computer
  Vision}}.
\newblock


\bibitem[\protect\citeauthoryear{Zhu, Kr{\"a}henb{\"u}hl, Shechtman, and
  Efros}{Zhu et~al\mbox{.}}{2016}]%
        {Zhu2016}
\bibfield{author}{\bibinfo{person}{Jun-Yan Zhu}, \bibinfo{person}{Philipp
  Kr{\"a}henb{\"u}hl}, \bibinfo{person}{Eli Shechtman}, {and}
  \bibinfo{person}{Alexei~A Efros}.} \bibinfo{year}{2016}\natexlab{}.
\newblock \showarticletitle{Generative visual manipulation on the natural image
  manifold}. In \bibinfo{booktitle}{\emph{Proceedings of European Conference on
  Computer Vision}}.
\newblock


\bibitem[\protect\citeauthoryear{Zhu, Abdal, Qin, and Wonka}{Zhu
  et~al\mbox{.}}{2020a}]%
        {Zhu2020_2}
\bibfield{author}{\bibinfo{person}{Peihao Zhu}, \bibinfo{person}{Rameen Abdal},
  \bibinfo{person}{Yipeng Qin}, {and} \bibinfo{person}{Peter Wonka}.}
  \bibinfo{year}{2020}\natexlab{a}.
\newblock \showarticletitle{Improved StyleGAN Embedding: Where are the Good
  Latents?}
\newblock \bibinfo{journal}{\emph{arXiv preprint arXiv:2012.09036}}
  (\bibinfo{year}{2020}).
\newblock


\bibitem[\protect\citeauthoryear{Zhuang, Koyejo, and Schwing}{Zhuang
  et~al\mbox{.}}{2021}]%
        {Zhuang2021}
\bibfield{author}{\bibinfo{person}{Peiye Zhuang}, \bibinfo{person}{Oluwasanmi
  Koyejo}, {and} \bibinfo{person}{Alexander~G. Schwing}.}
  \bibinfo{year}{2021}\natexlab{}.
\newblock \showarticletitle{Enjoy Your Editing: Controllable GANs for Image
  Editing via Latent Space Navigation}. In
  \bibinfo{booktitle}{\emph{Proceedings of International Conference on Learning
  Representations}}.
\newblock


\end{thebibliography}


\clearpage
\appendix
\appendixpage

\section{Additional Visual Results}
\begin{figure*}
\setlength{\tabcolsep}{1pt}
\centering
{
	\renewcommand{\arraystretch}{0.5}
    \begin{tabular}{c c c c c c c}
		& Input & SG2 & SG2$\mathcal{W}+$ & e4e & PTI & Ours\\
        \raisebox{0.45in}{\rotatebox[origin=t]{90}{Inversion}}&
        \includegraphics[width=0.16\textwidth]{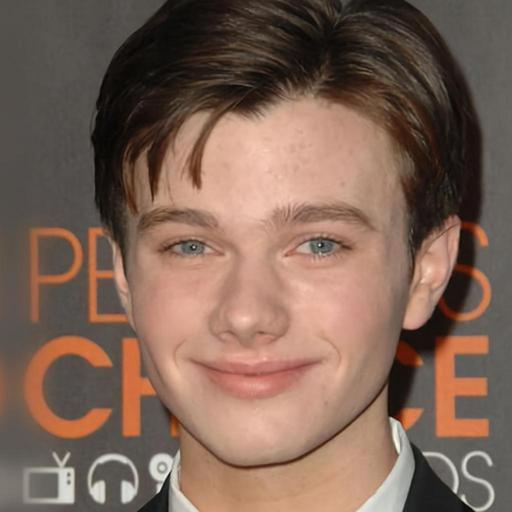}&
        \includegraphics[width=0.16\textwidth]{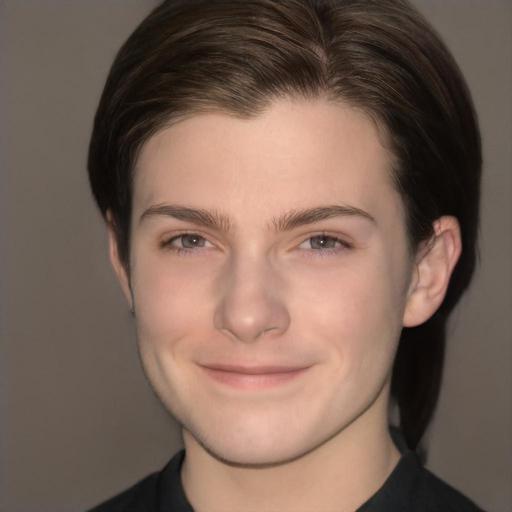}&        
        \includegraphics[width=0.16\textwidth]{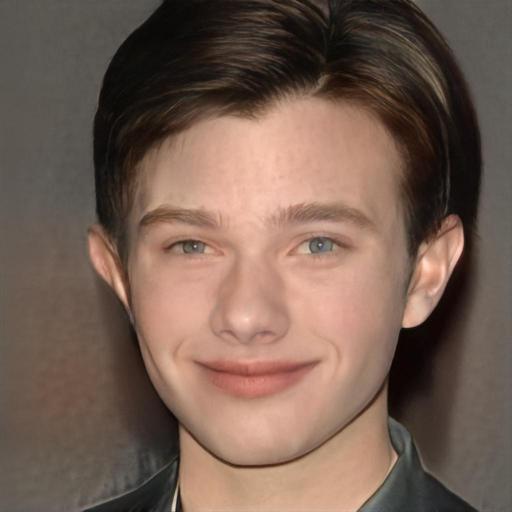}&        
        \includegraphics[width=0.16\textwidth]{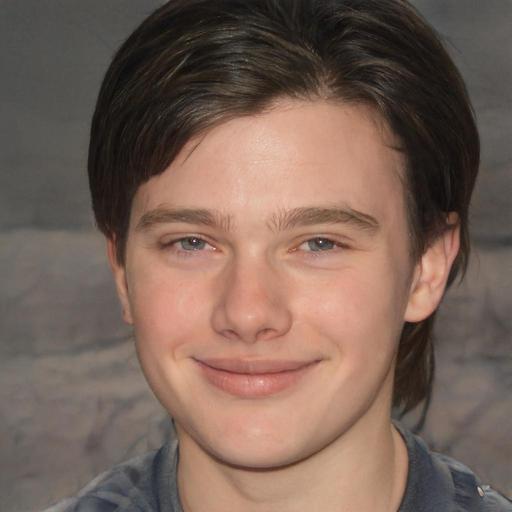}&
        \includegraphics[width=0.16\textwidth]{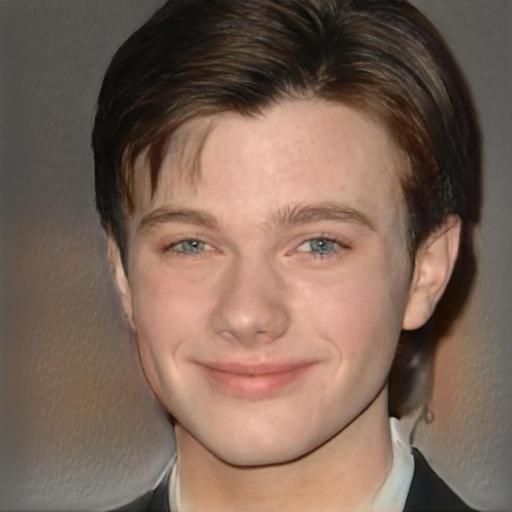}&
        \includegraphics[width=0.16\textwidth]{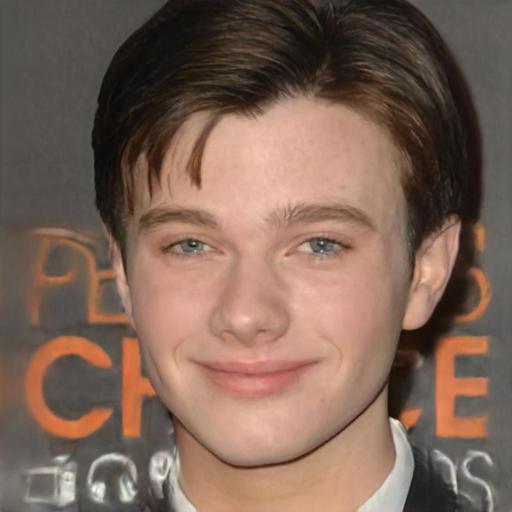} 
		\\
        \raisebox{0.45in}{\rotatebox[origin=t]{90}{Smile}}&&
        \includegraphics[width=0.16\textwidth]{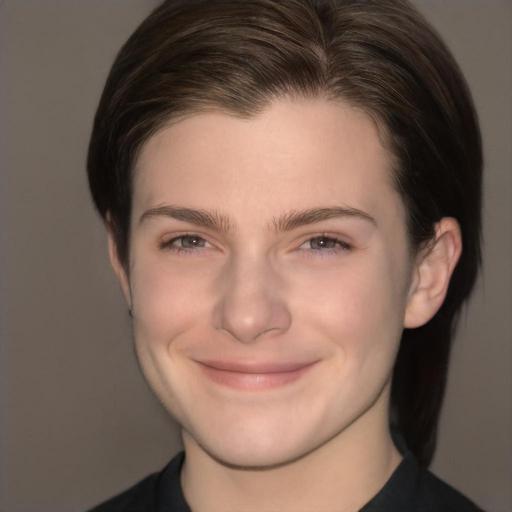}&        
        \includegraphics[width=0.16\textwidth]{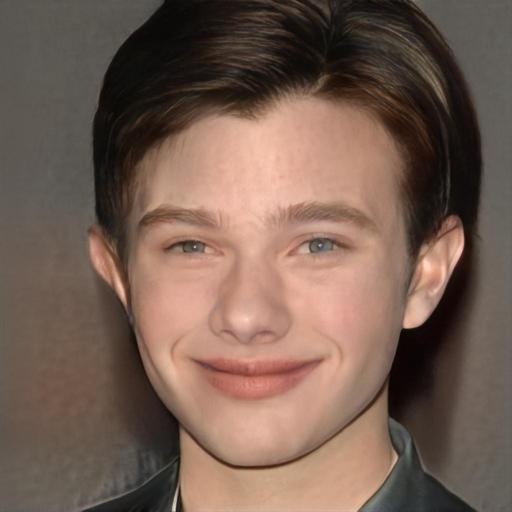}&        
        \includegraphics[width=0.16\textwidth]{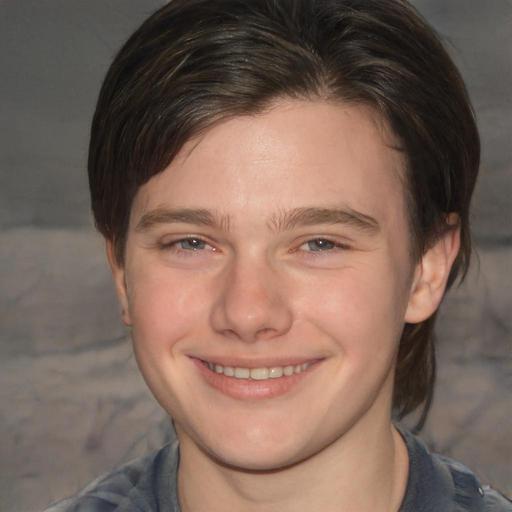}&
        \includegraphics[width=0.16\textwidth]{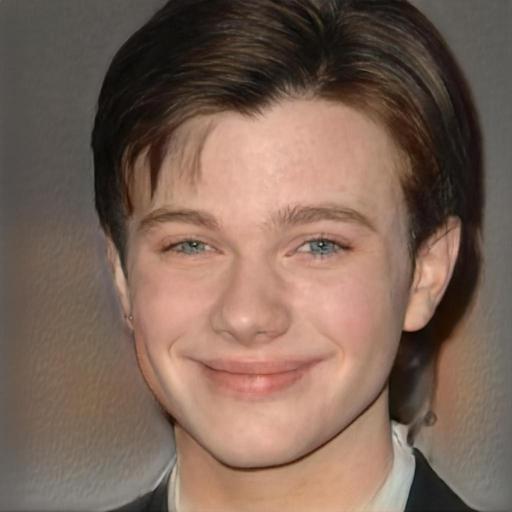}&
        \includegraphics[width=0.16\textwidth]{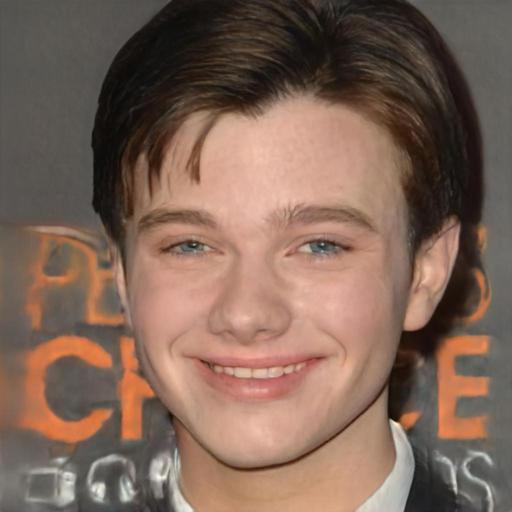} 
		\\[3pt]
        \raisebox{0.45in}{\rotatebox[origin=t]{90}{Inversion}}&
        \includegraphics[width=0.16\textwidth]{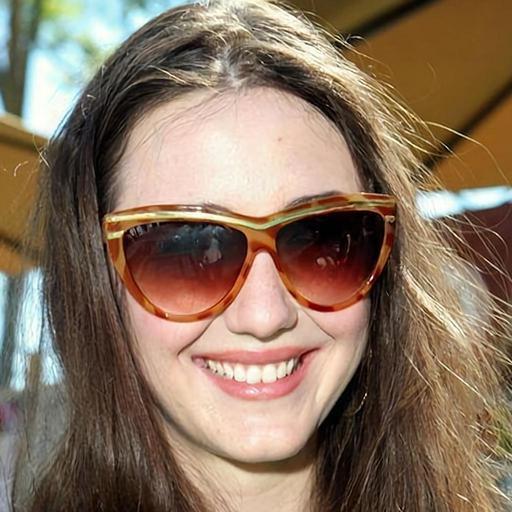}&
        \includegraphics[width=0.16\textwidth]{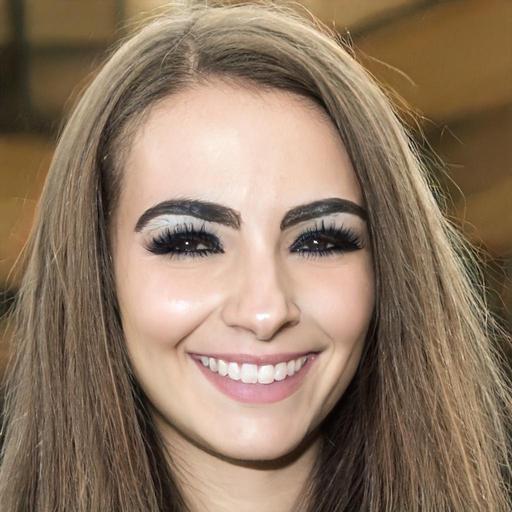}&        
        \includegraphics[width=0.16\textwidth]{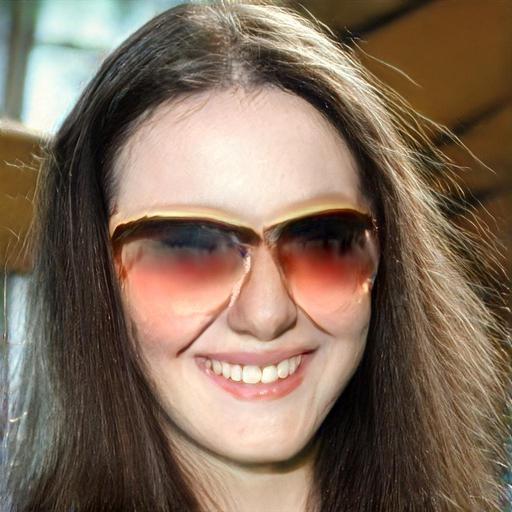}&        
        \includegraphics[width=0.16\textwidth]{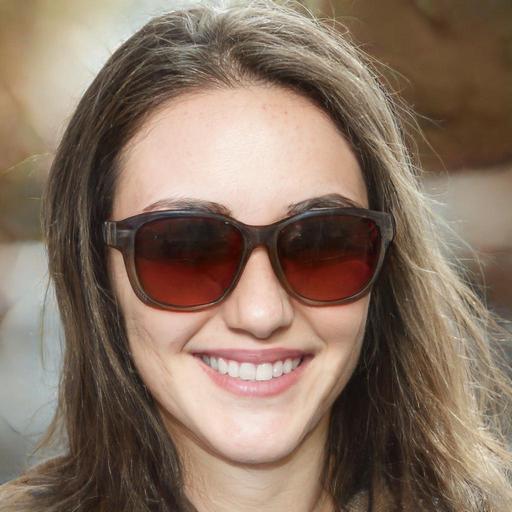}&
        \includegraphics[width=0.16\textwidth]{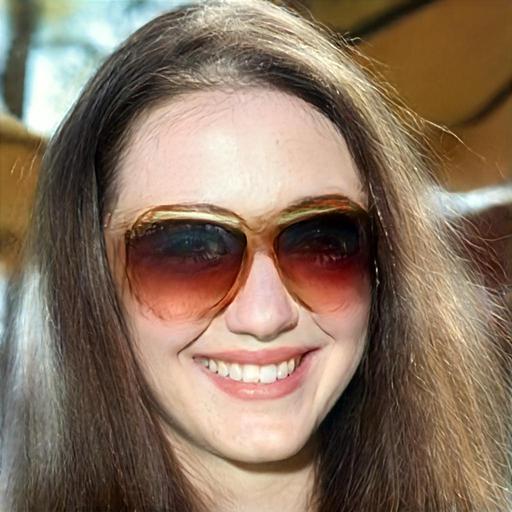}&
        \includegraphics[width=0.16\textwidth]{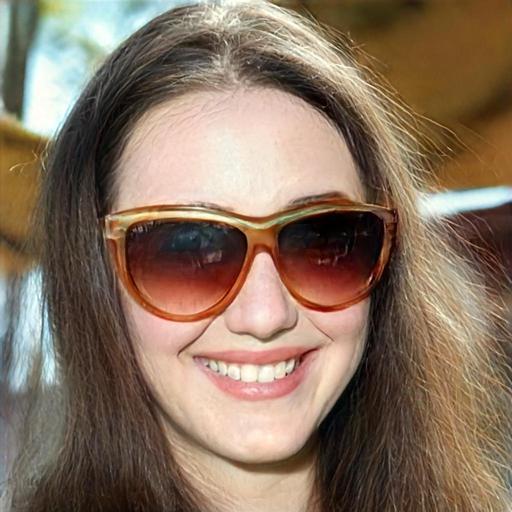} 
		\\
        \raisebox{0.45in}{\rotatebox[origin=t]{90}{Pose}}&&
        \includegraphics[width=0.16\textwidth]{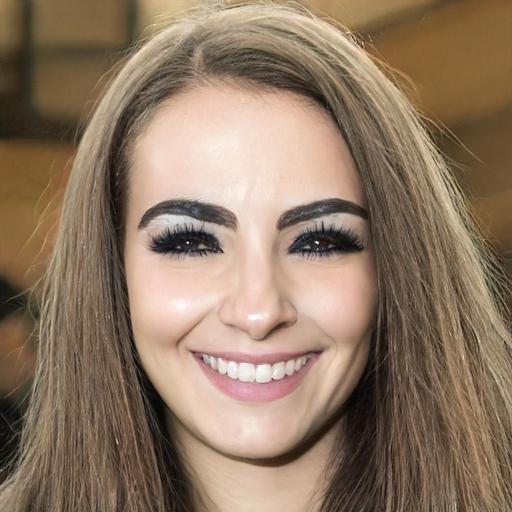}&        
        \includegraphics[width=0.16\textwidth]{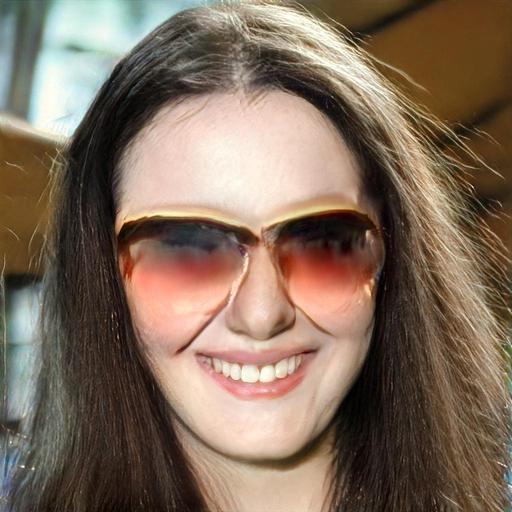}&        
        \includegraphics[width=0.16\textwidth]{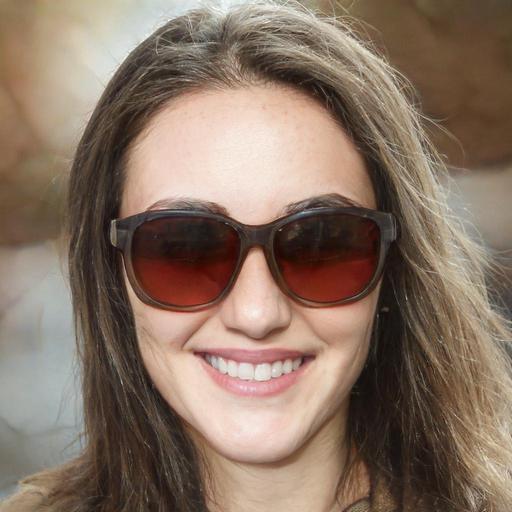}&
        \includegraphics[width=0.16\textwidth]{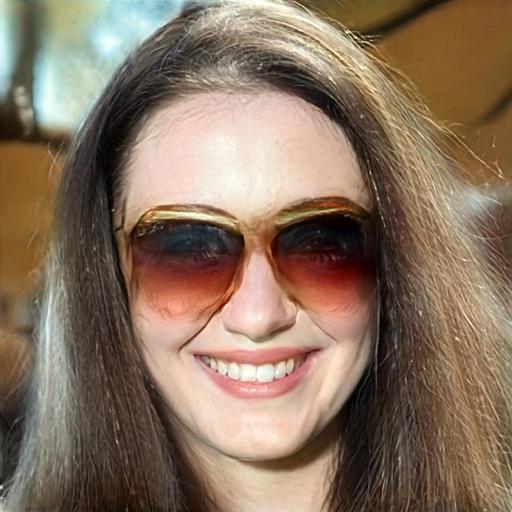}&
        \includegraphics[width=0.16\textwidth]{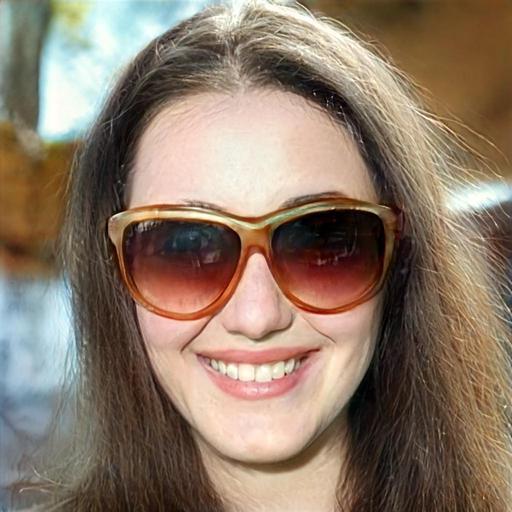} 
		\\[3pt]
        \raisebox{0.45in}{\rotatebox[origin=t]{90}{Inversion}}&
        \includegraphics[width=0.16\textwidth]{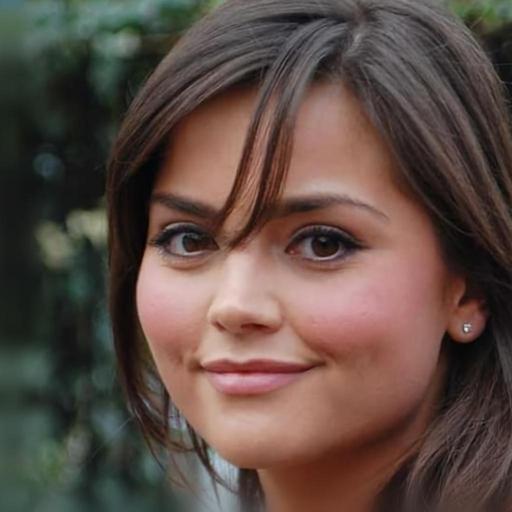}&
        \includegraphics[width=0.16\textwidth]{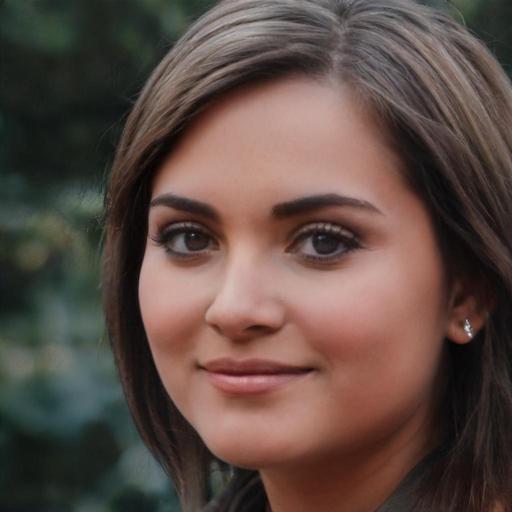}&        
        \includegraphics[width=0.16\textwidth]{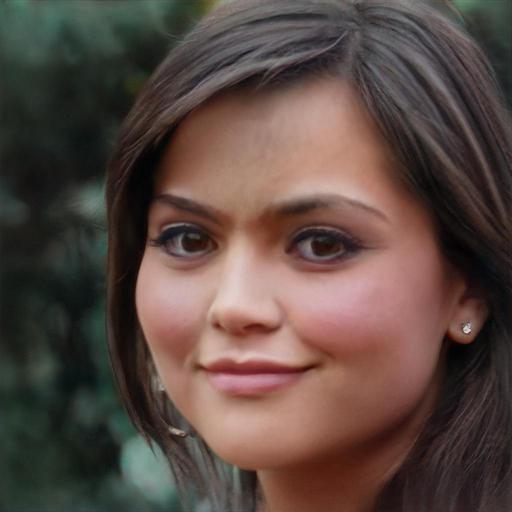}&        
        \includegraphics[width=0.16\textwidth]{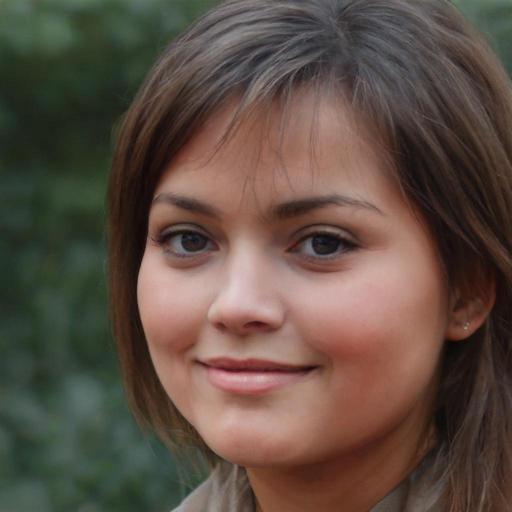}&
        \includegraphics[width=0.16\textwidth]{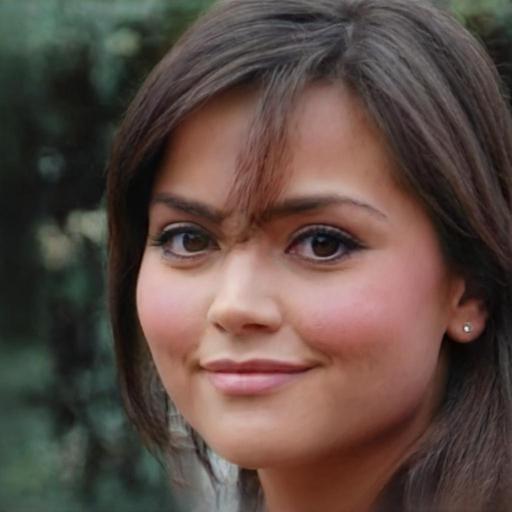}&
        \includegraphics[width=0.16\textwidth]{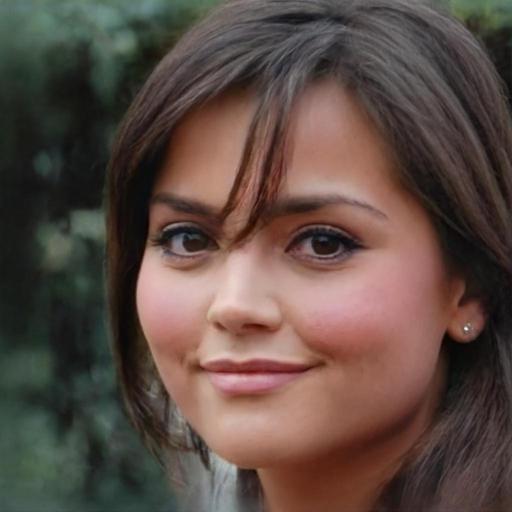} 
		\\
        \raisebox{0.45in}{\rotatebox[origin=t]{90}{Age}}&&
        \includegraphics[width=0.16\textwidth]{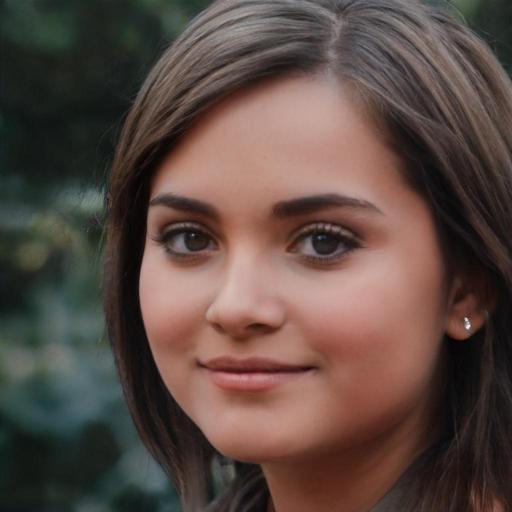}&        
        \includegraphics[width=0.16\textwidth]{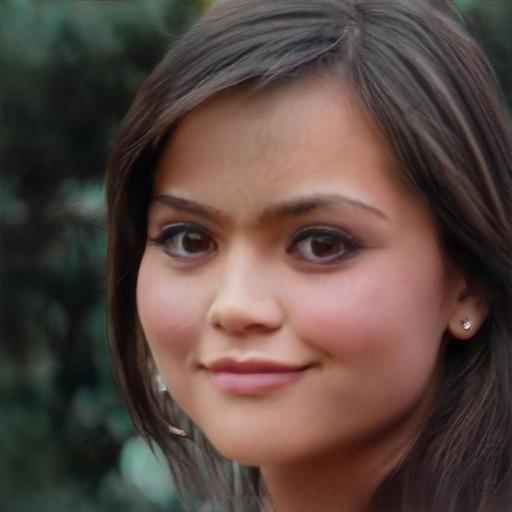}&        
        \includegraphics[width=0.16\textwidth]{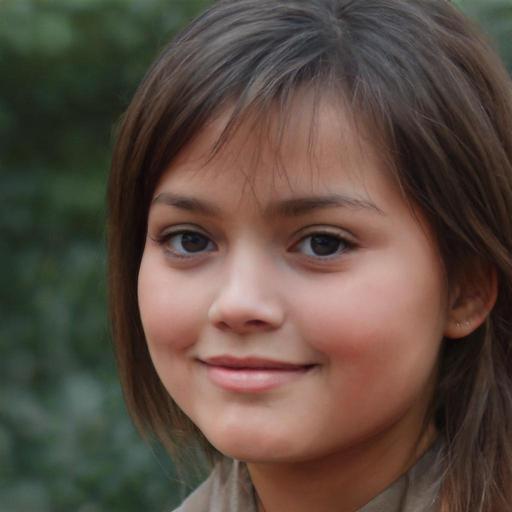}&
        \includegraphics[width=0.16\textwidth]{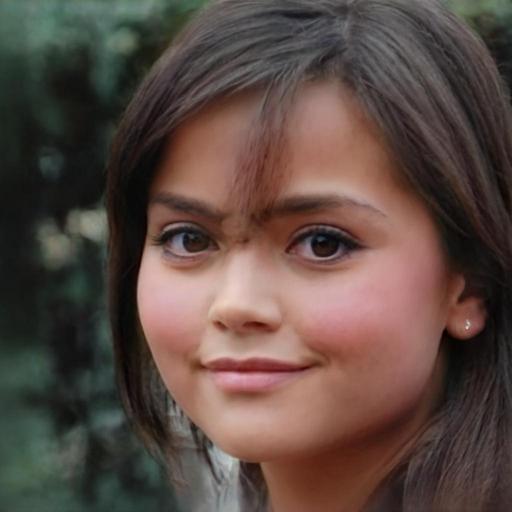}&
        \includegraphics[width=0.16\textwidth]{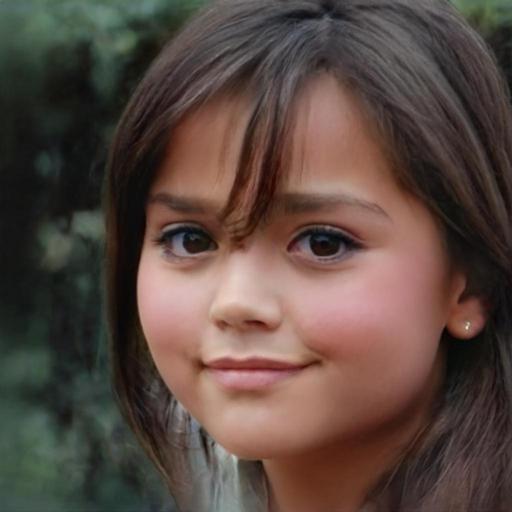} 
		\\
		& Input & SG2 & SG2$\mathcal{W}+$ & e4e & PTI & Ours
    \end{tabular}
    }
	\caption{Reconstruction and editing quality comparison using examples from CelebA-HQ test set. In each example, the editing is performed using the same editing weight.}
    \label{fig:appendix0}
\end{figure*}

\begin{figure*}
\setlength{\tabcolsep}{1pt}
\centering
{
	\renewcommand{\arraystretch}{0.5}
    \begin{tabular}{c c c c c c c}
		& Input & SG2 & SG2$\mathcal{W}+$ & e4e & PTI & Ours\\
        \raisebox{0.45in}{\rotatebox[origin=t]{90}{Inversion}}&
        \includegraphics[width=0.16\textwidth]{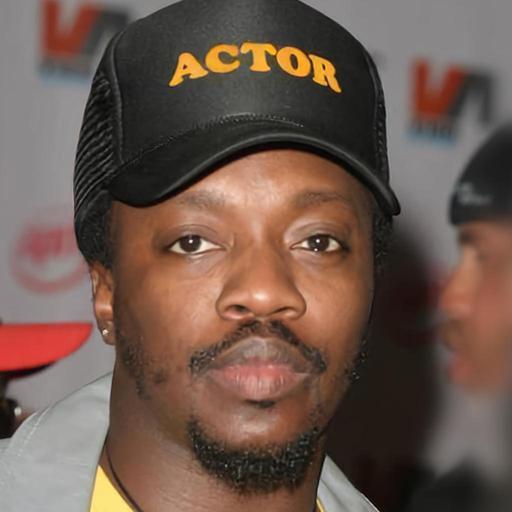}&
        \includegraphics[width=0.16\textwidth]{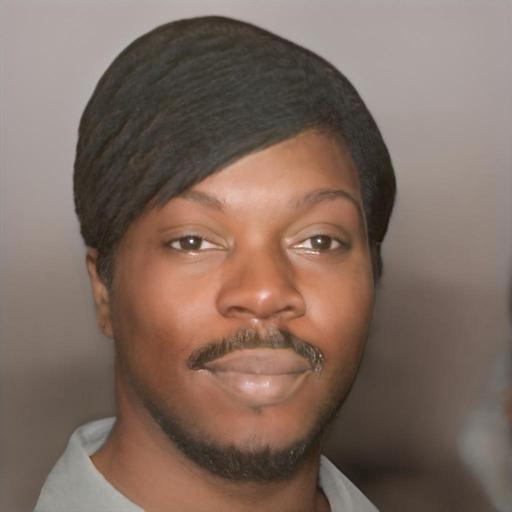}&        
        \includegraphics[width=0.16\textwidth]{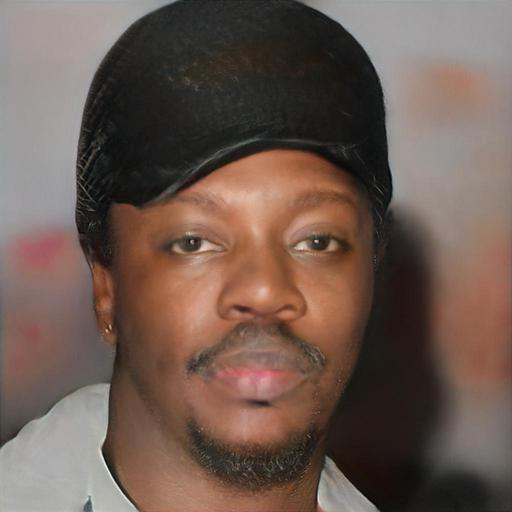}&        
        \includegraphics[width=0.16\textwidth]{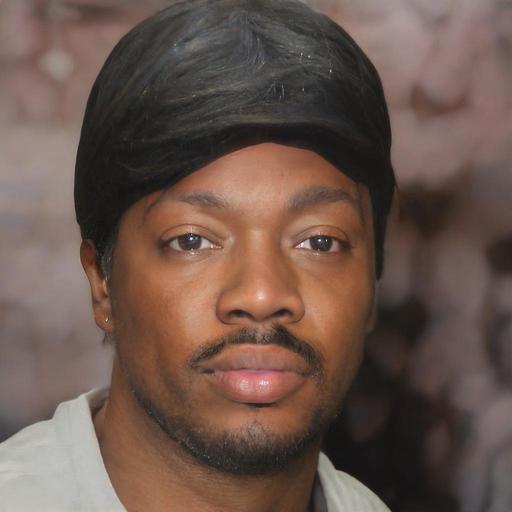}&
        \includegraphics[width=0.16\textwidth]{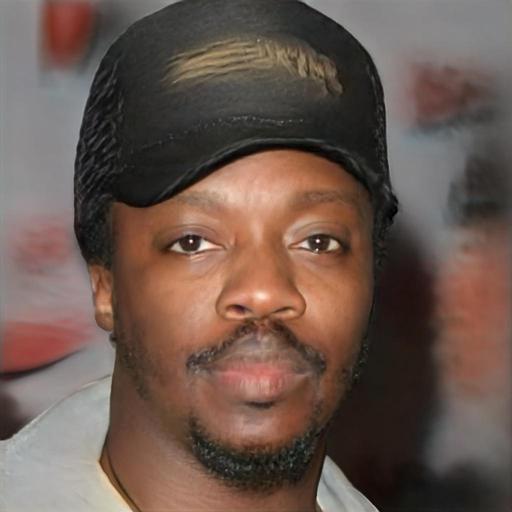}&
        \includegraphics[width=0.16\textwidth]{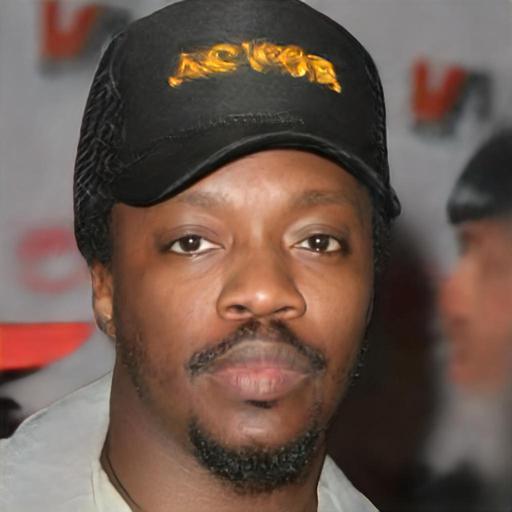} 
		\\
        \raisebox{0.45in}{\rotatebox[origin=t]{90}{Smile}}&&
        \includegraphics[width=0.16\textwidth]{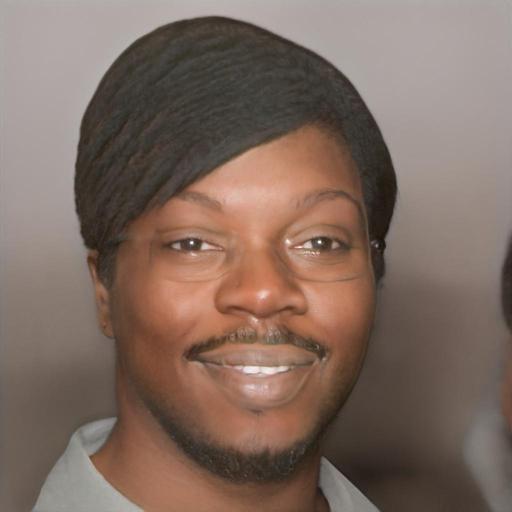}&        
        \includegraphics[width=0.16\textwidth]{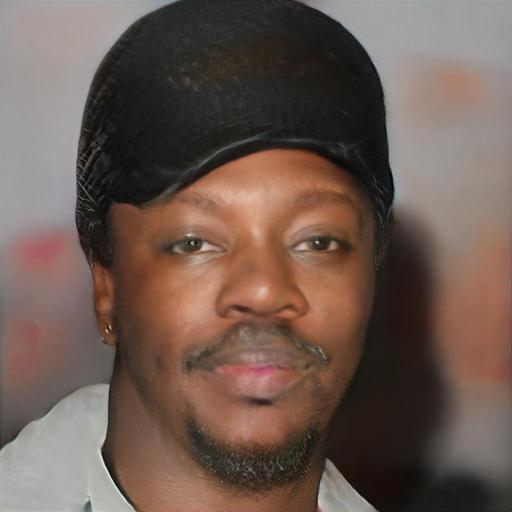}&        
        \includegraphics[width=0.16\textwidth]{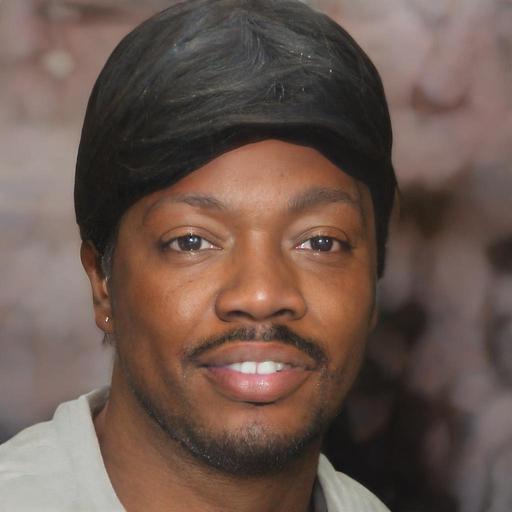}&
        \includegraphics[width=0.16\textwidth]{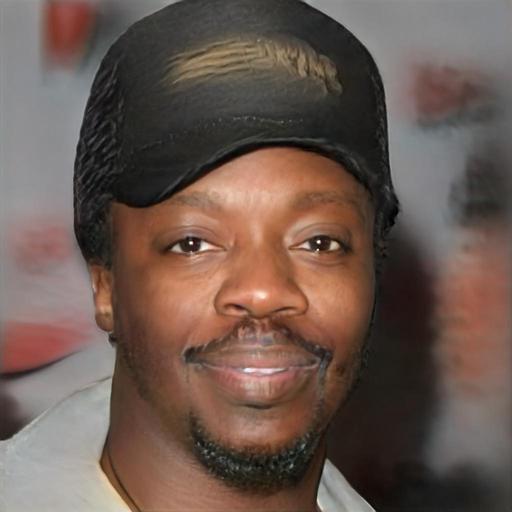}&
        \includegraphics[width=0.16\textwidth]{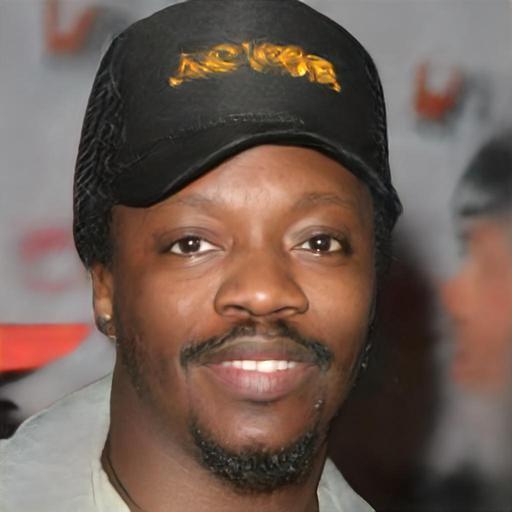} 
		\\[3pt]
        \raisebox{0.45in}{\rotatebox[origin=t]{90}{Inversion}}&
        \includegraphics[width=0.16\textwidth]{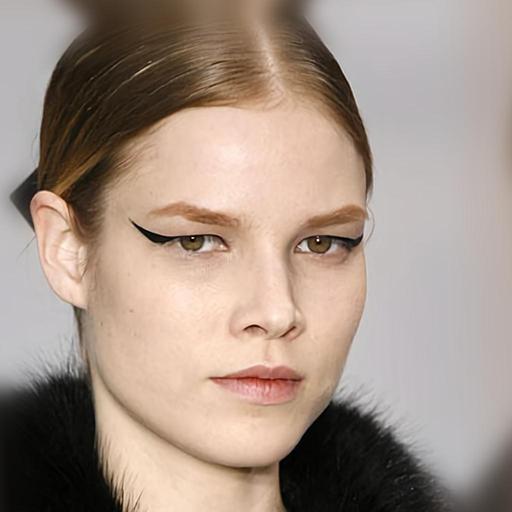}&
        \includegraphics[width=0.16\textwidth]{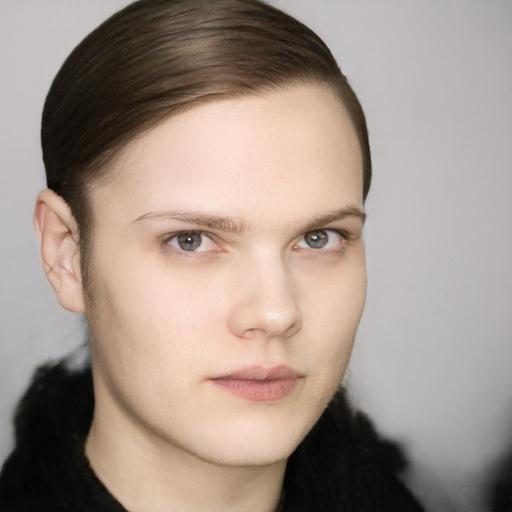}&        
        \includegraphics[width=0.16\textwidth]{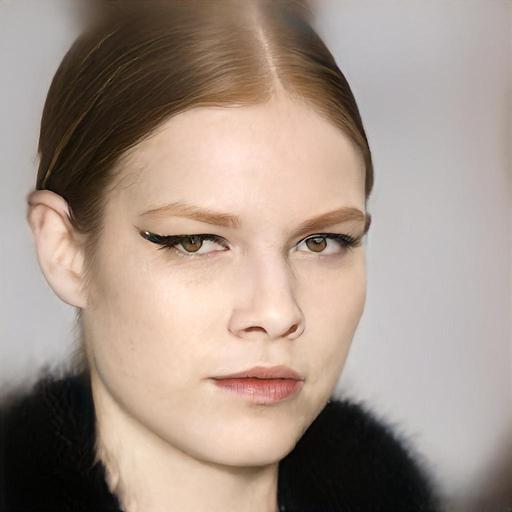}&        
        \includegraphics[width=0.16\textwidth]{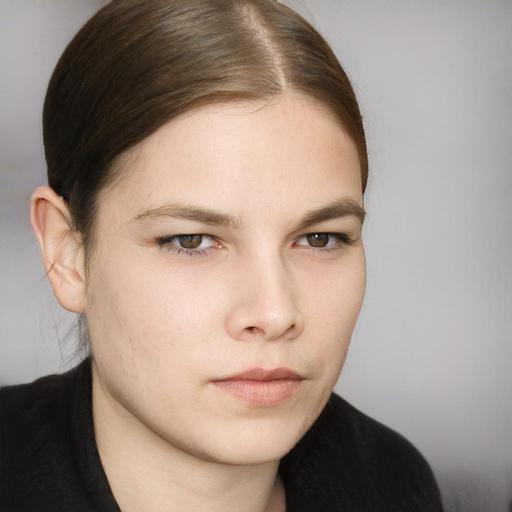}&
        \includegraphics[width=0.16\textwidth]{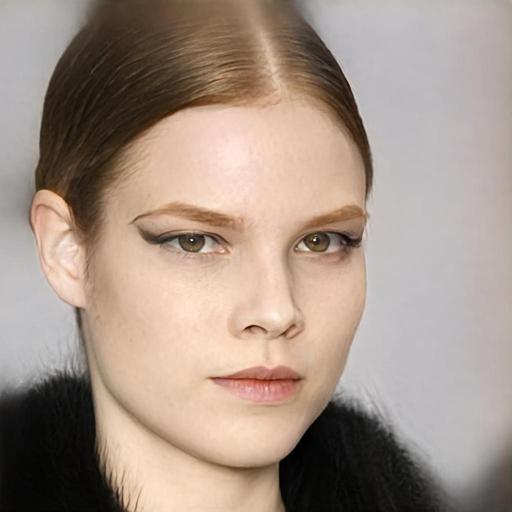}&
        \includegraphics[width=0.16\textwidth]{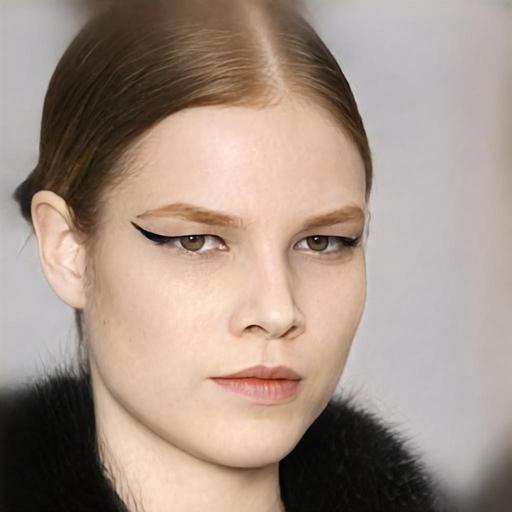} 
		\\
        \raisebox{0.45in}{\rotatebox[origin=t]{90}{Pose}}&&
        \includegraphics[width=0.16\textwidth]{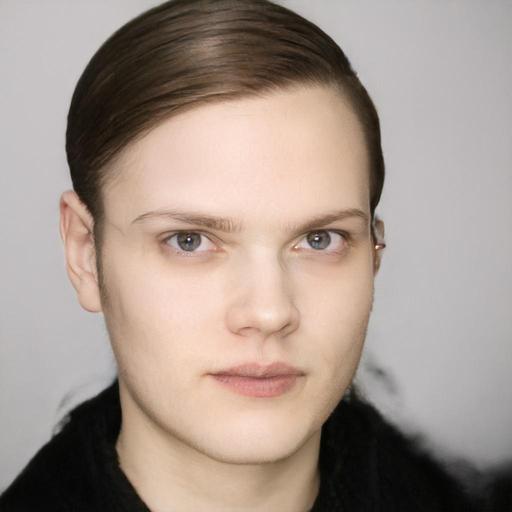}&        
        \includegraphics[width=0.16\textwidth]{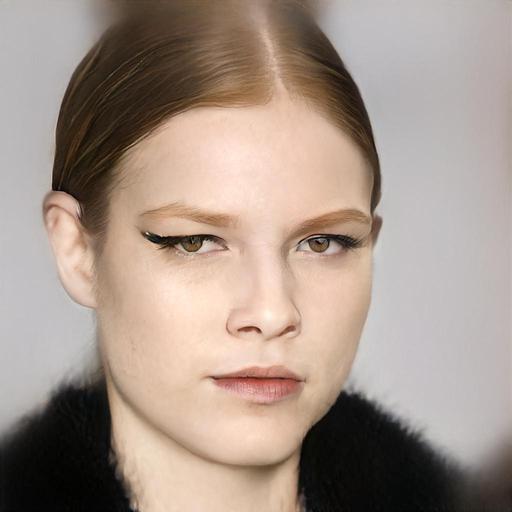}&        
        \includegraphics[width=0.16\textwidth]{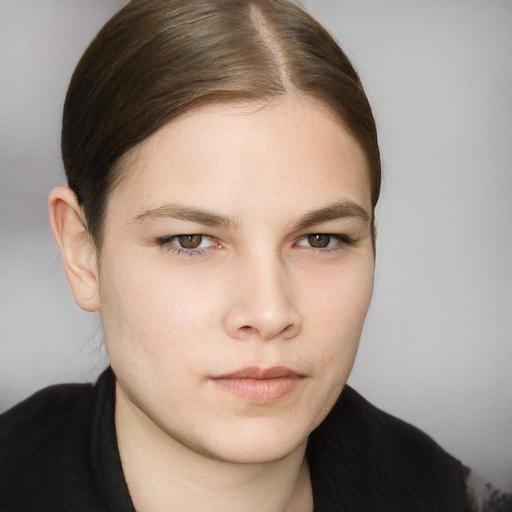}&
        \includegraphics[width=0.16\textwidth]{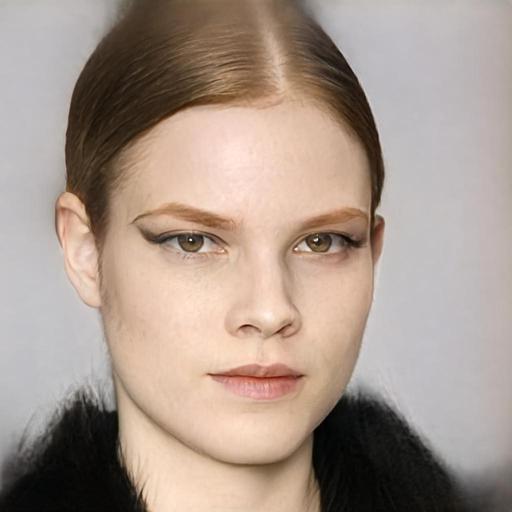}&
        \includegraphics[width=0.16\textwidth]{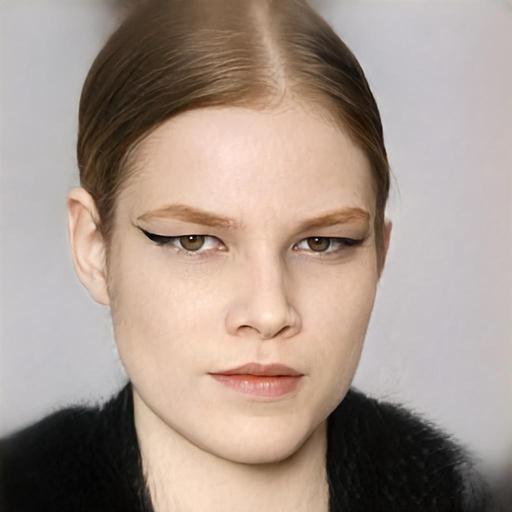} 
		\\[3pt]
        \raisebox{0.45in}{\rotatebox[origin=t]{90}{Inversion}}&
        \includegraphics[width=0.16\textwidth]{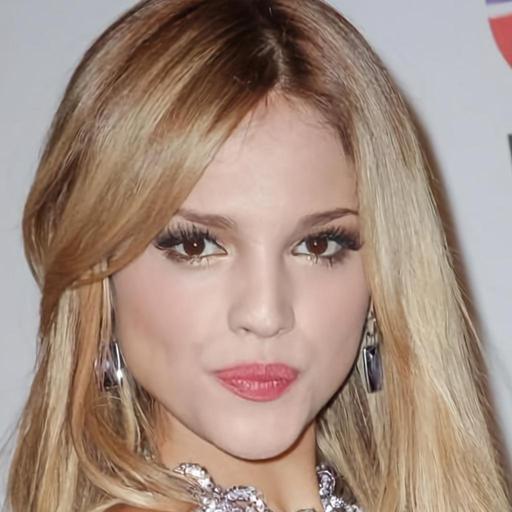}&
        \includegraphics[width=0.16\textwidth]{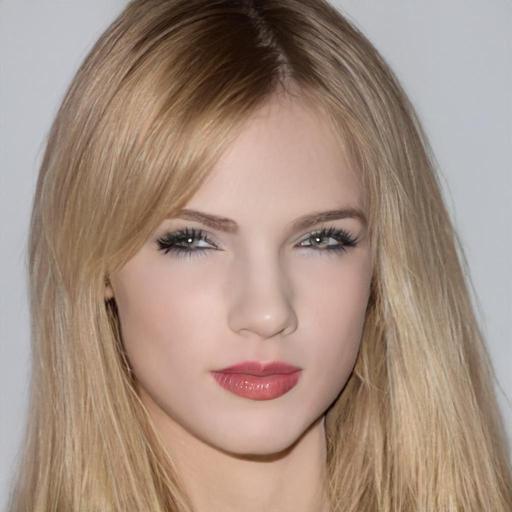}&        
        \includegraphics[width=0.16\textwidth]{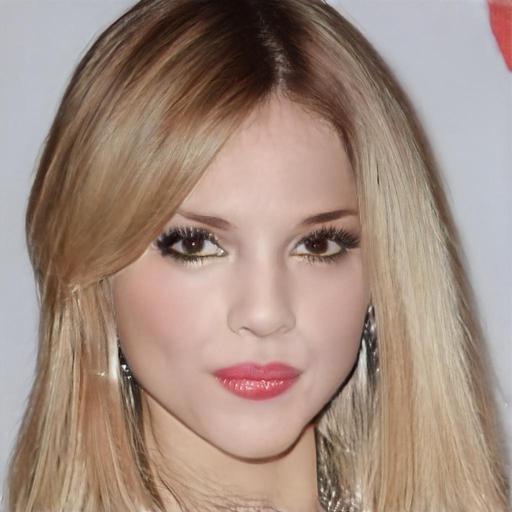}&        
        \includegraphics[width=0.16\textwidth]{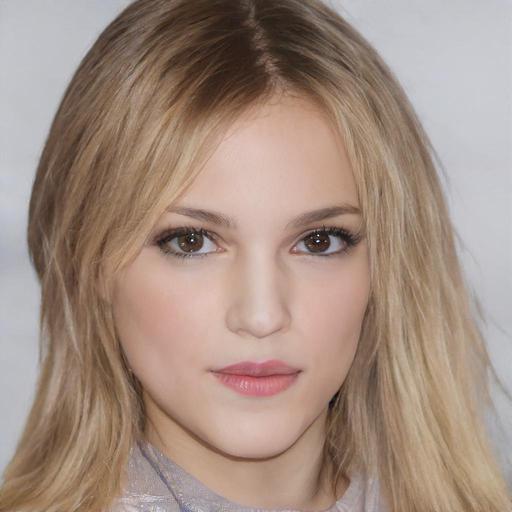}&
        \includegraphics[width=0.16\textwidth]{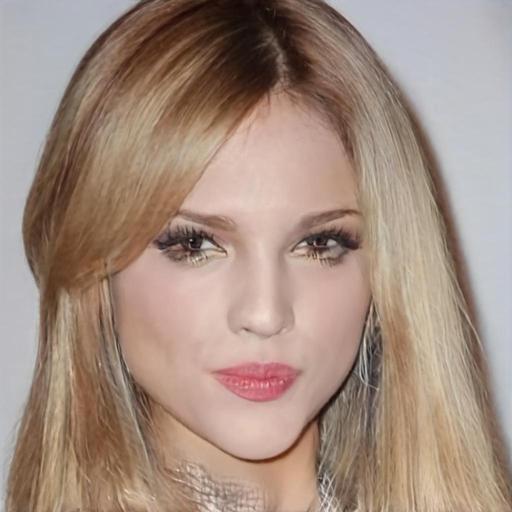}&
        \includegraphics[width=0.16\textwidth]{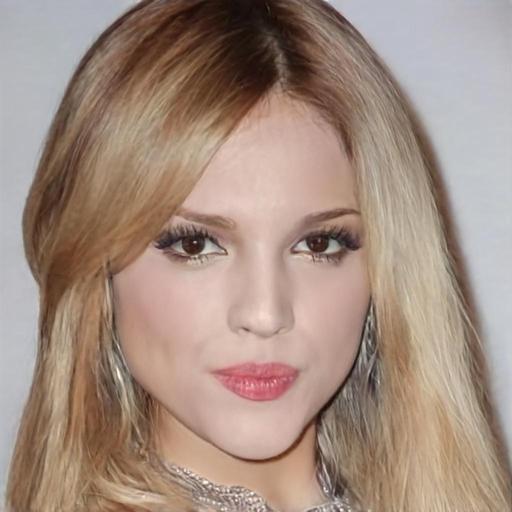} 
		\\
        \raisebox{0.45in}{\rotatebox[origin=t]{90}{Age}}&&
        \includegraphics[width=0.16\textwidth]{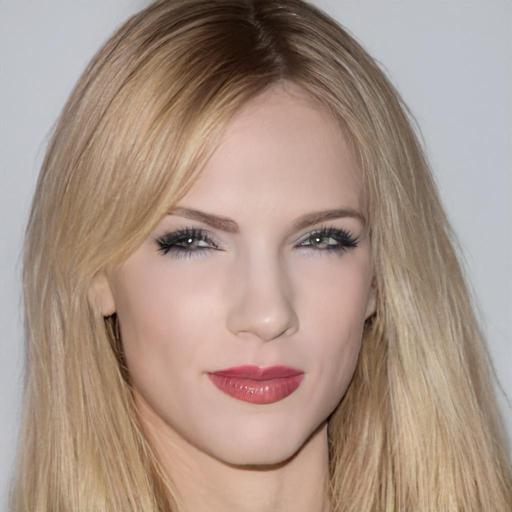}&        
        \includegraphics[width=0.16\textwidth]{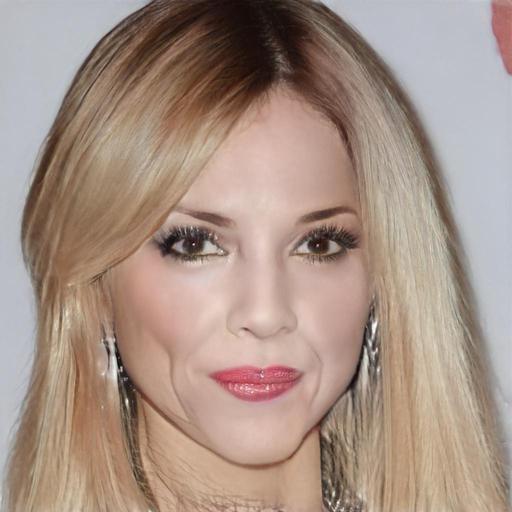}&        
        \includegraphics[width=0.16\textwidth]{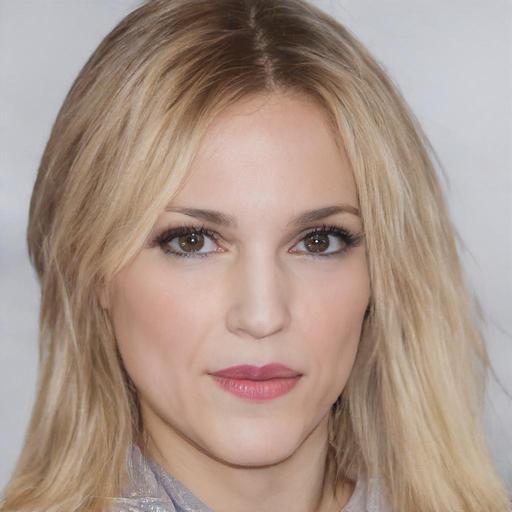}&
        \includegraphics[width=0.16\textwidth]{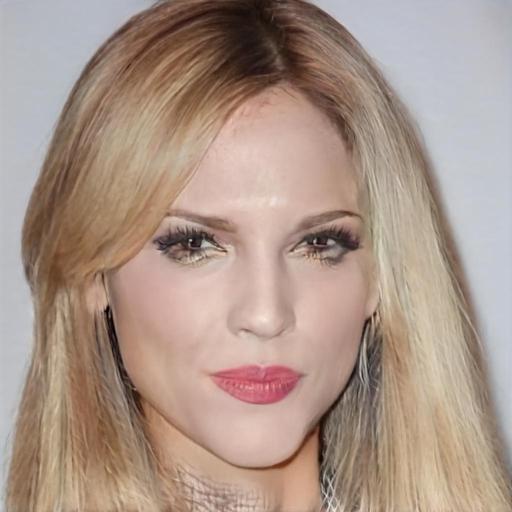}&
        \includegraphics[width=0.16\textwidth]{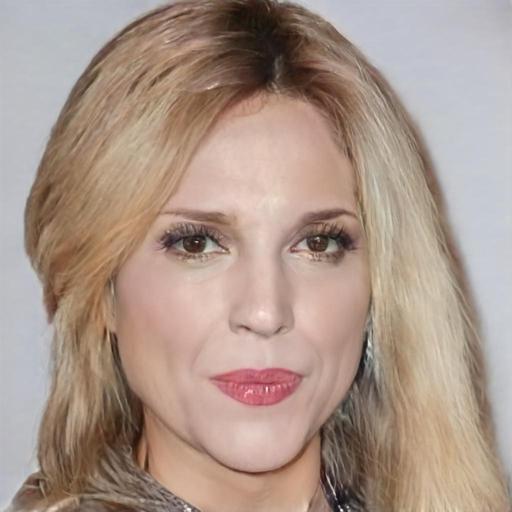} 
		\\
		& Input & SG2 & SG2$\mathcal{W}+$ & e4e & PTI & Ours
    \end{tabular}
    }
	\caption{Reconstruction and editing quality comparison using examples from CelebA-HQ test set. In each example, the editing is performed using the same editing weight.}
    \label{fig:appendix1}
\end{figure*}

\begin{figure*}
\setlength{\tabcolsep}{1pt}
\centering
{
	\renewcommand{\arraystretch}{0.5}
    \begin{tabular}{c c c c c c c}
		& Input & SG2 & SG2$\mathcal{W}+$ & e4e & PTI & Ours\\
        \raisebox{0.45in}{\rotatebox[origin=t]{90}{Inversion}}&
        \includegraphics[width=0.16\textwidth]{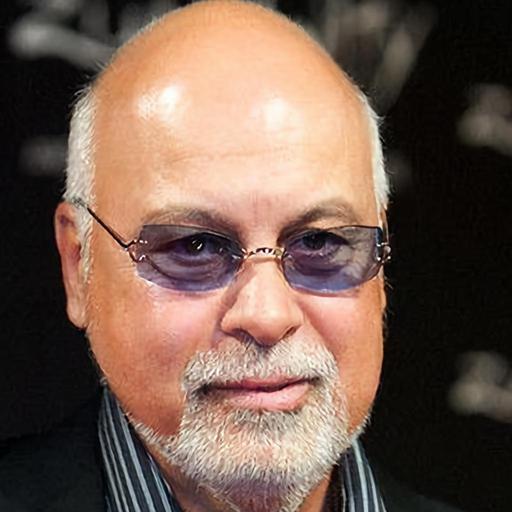}&
        \includegraphics[width=0.16\textwidth]{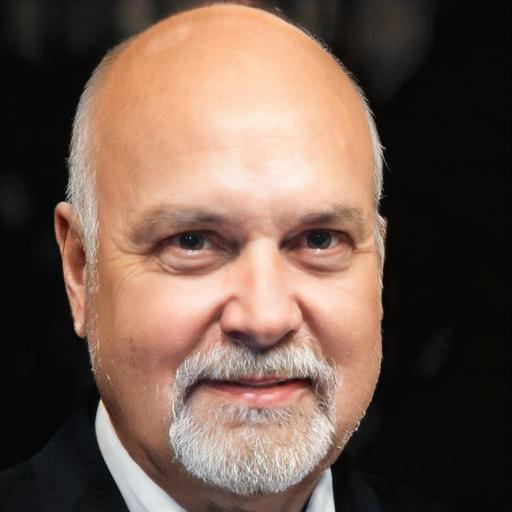}&        
        \includegraphics[width=0.16\textwidth]{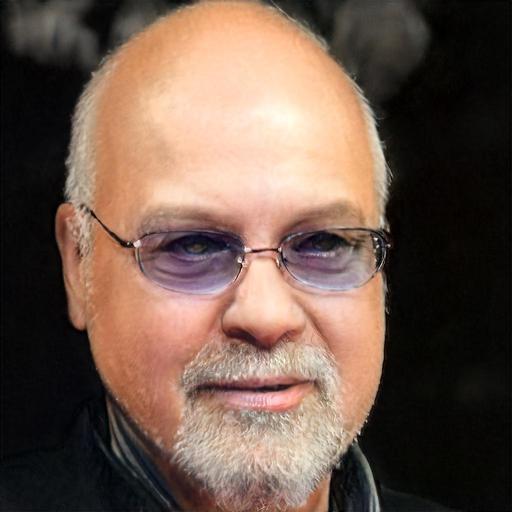}&        
        \includegraphics[width=0.16\textwidth]{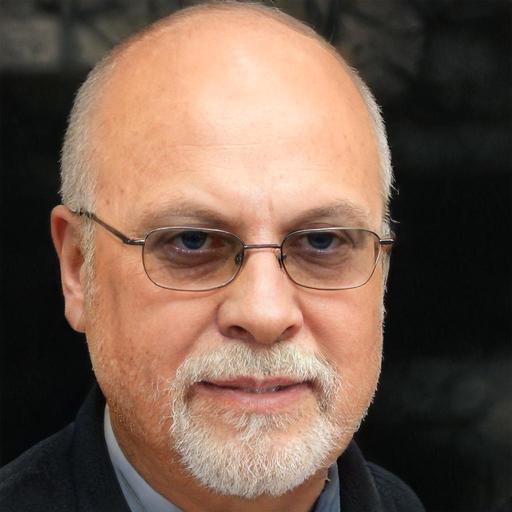}&
        \includegraphics[width=0.16\textwidth]{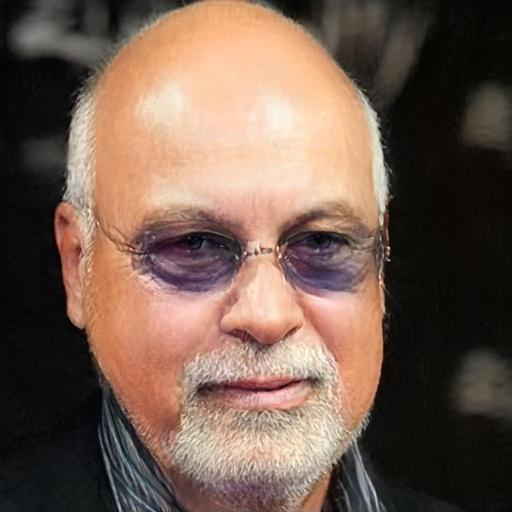}&
        \includegraphics[width=0.16\textwidth]{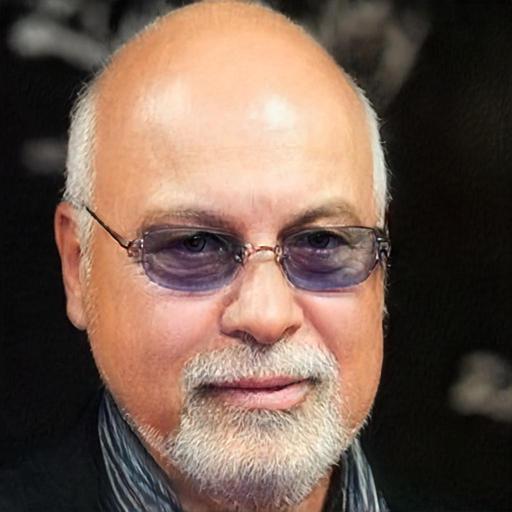} 
		\\
        \raisebox{0.45in}{\rotatebox[origin=t]{90}{Smile}}&&
        \includegraphics[width=0.16\textwidth]{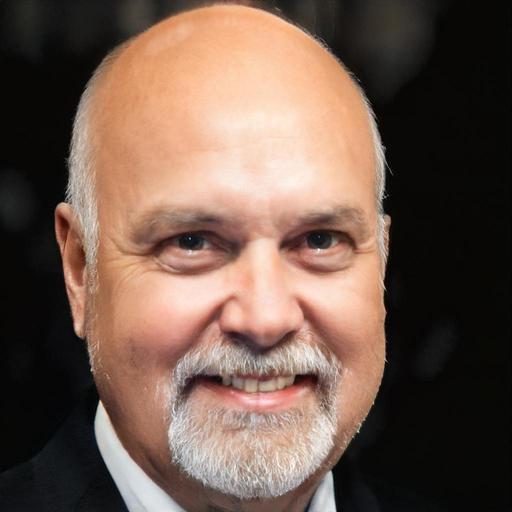}&        
        \includegraphics[width=0.16\textwidth]{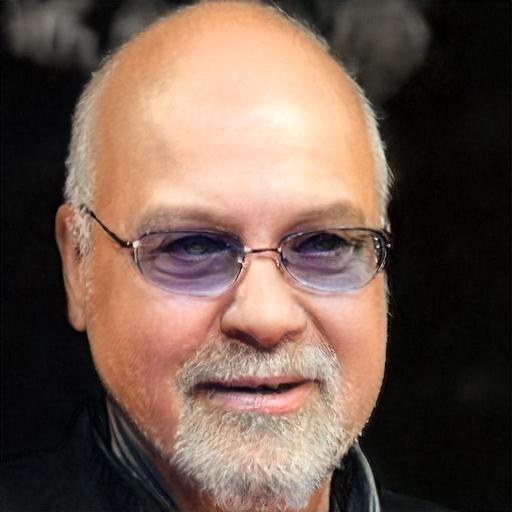}&        
        \includegraphics[width=0.16\textwidth]{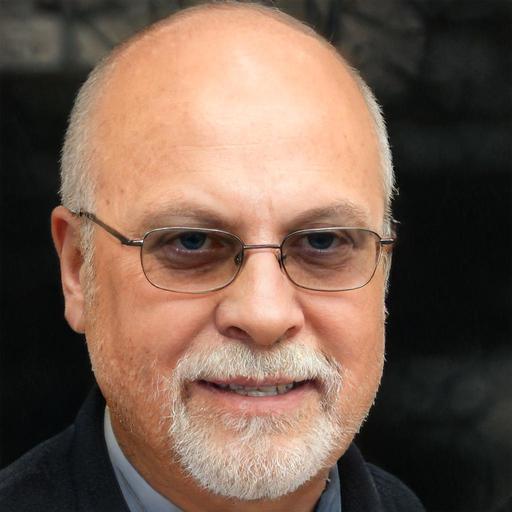}&
        \includegraphics[width=0.16\textwidth]{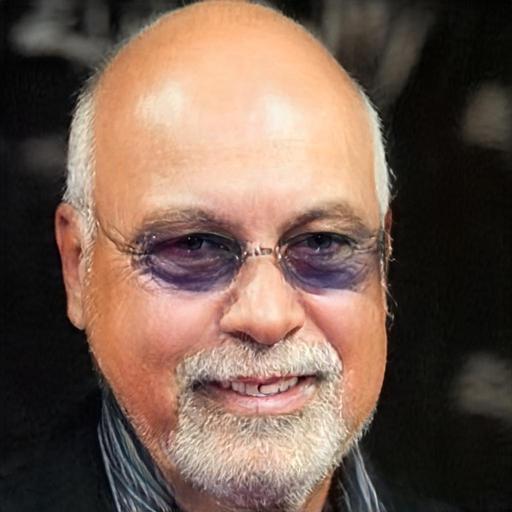}&
        \includegraphics[width=0.16\textwidth]{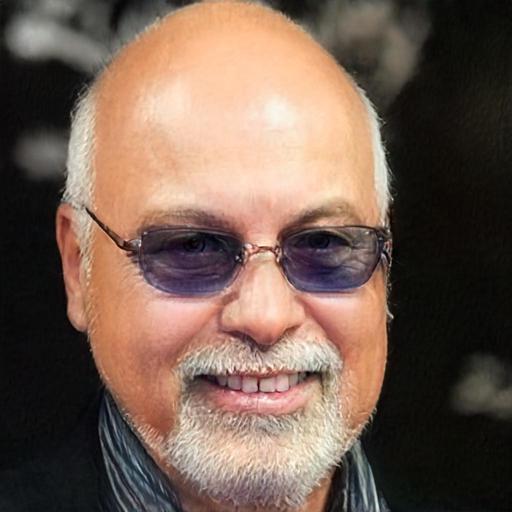} 
		\\[3pt]
        \raisebox{0.45in}{\rotatebox[origin=t]{90}{Inversion}}&
        \includegraphics[width=0.16\textwidth]{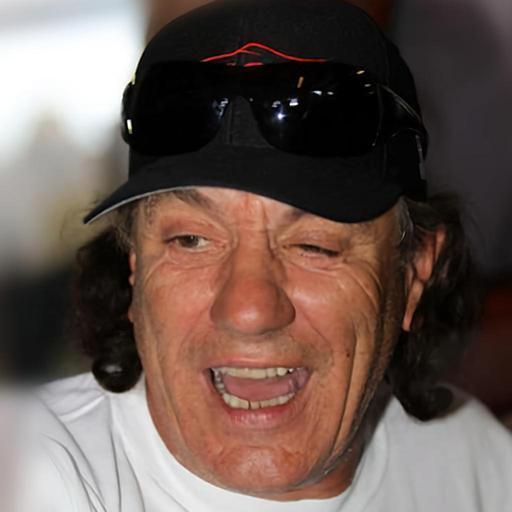}&
        \includegraphics[width=0.16\textwidth]{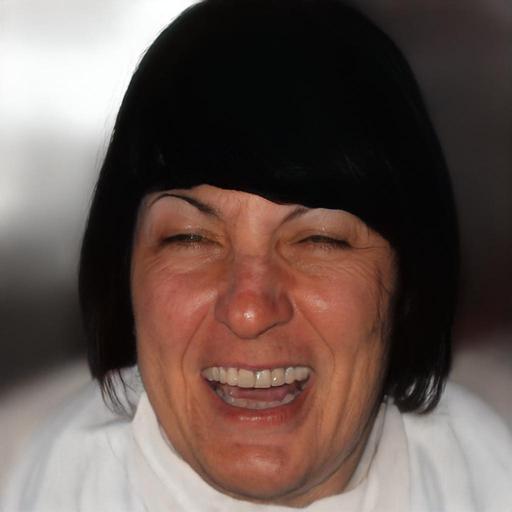}&        
        \includegraphics[width=0.16\textwidth]{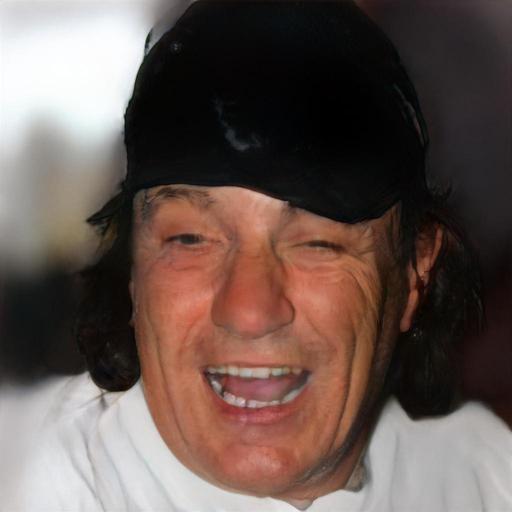}&        
        \includegraphics[width=0.16\textwidth]{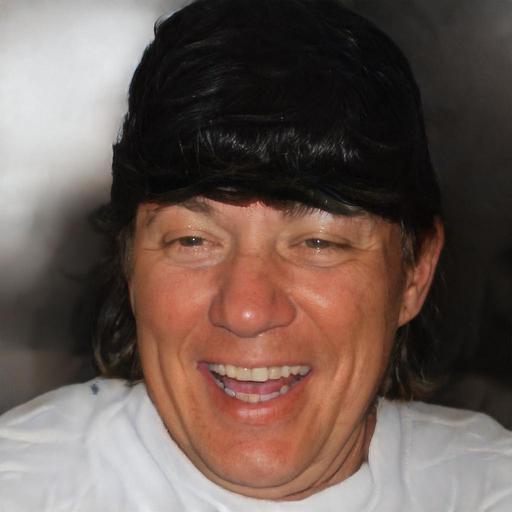}&
        \includegraphics[width=0.16\textwidth]{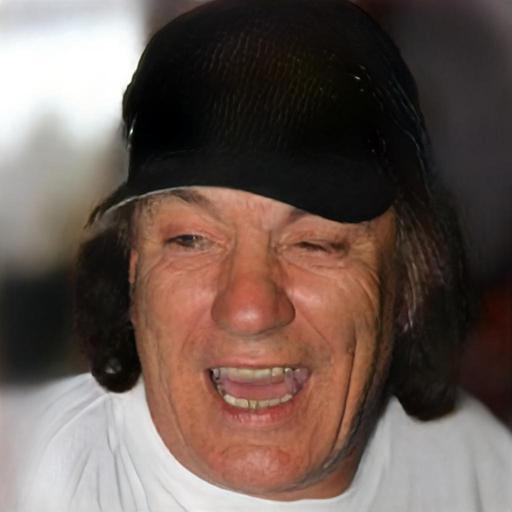}&
        \includegraphics[width=0.16\textwidth]{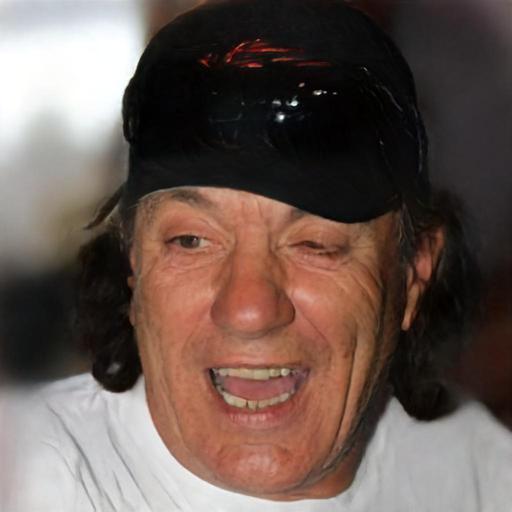} 
		\\
        \raisebox{0.45in}{\rotatebox[origin=t]{90}{Pose}}&&
        \includegraphics[width=0.16\textwidth]{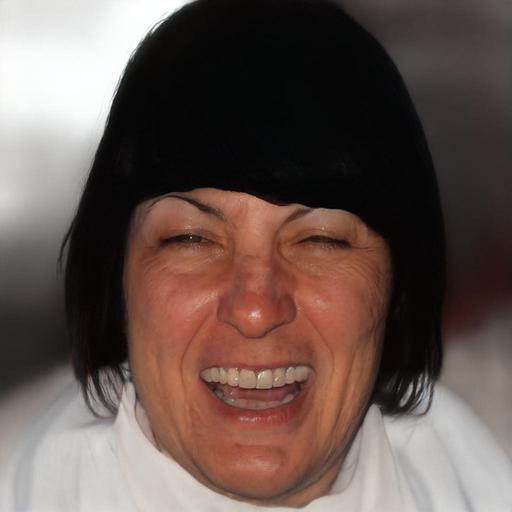}&        
        \includegraphics[width=0.16\textwidth]{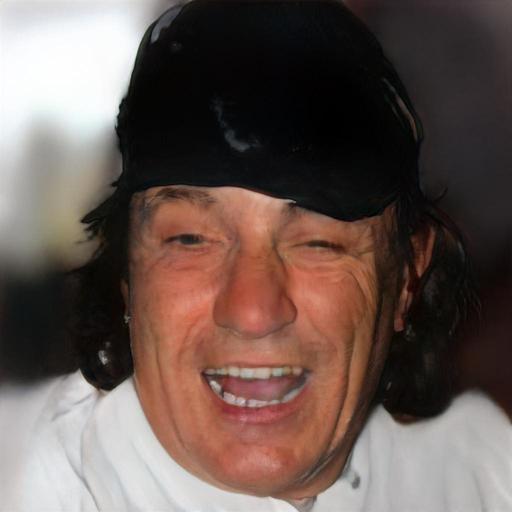}&        
        \includegraphics[width=0.16\textwidth]{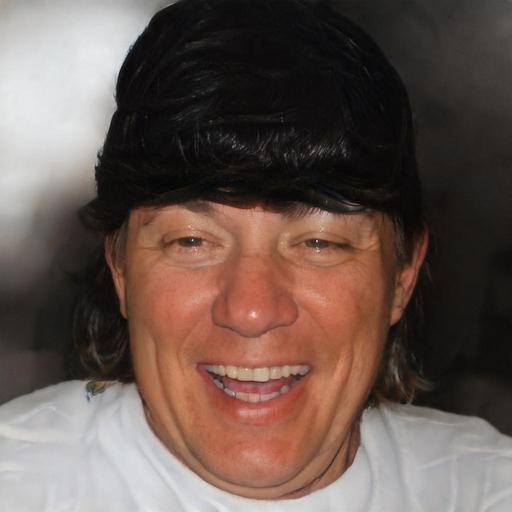}&
        \includegraphics[width=0.16\textwidth]{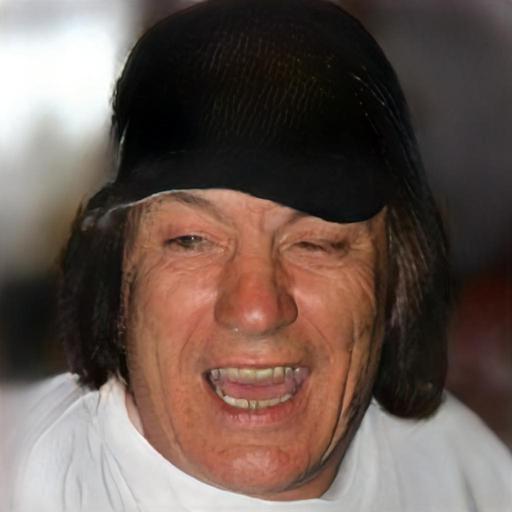}&
        \includegraphics[width=0.16\textwidth]{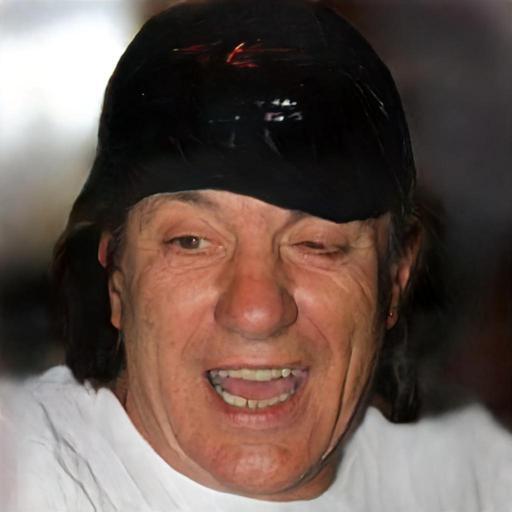} 
		\\[3pt]
        \raisebox{0.45in}{\rotatebox[origin=t]{90}{Inversion}}&
        \includegraphics[width=0.16\textwidth]{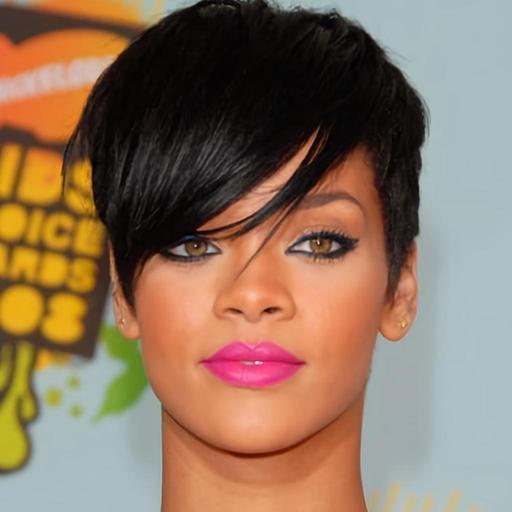}&
        \includegraphics[width=0.16\textwidth]{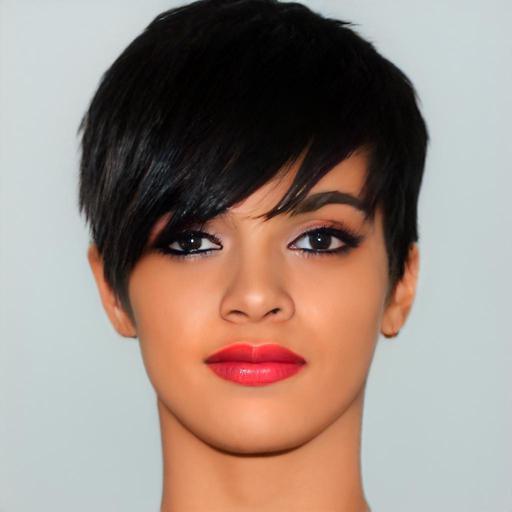}&        
        \includegraphics[width=0.16\textwidth]{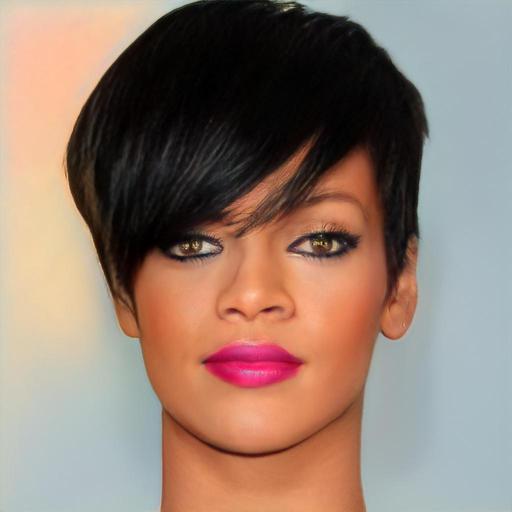}&        
        \includegraphics[width=0.16\textwidth]{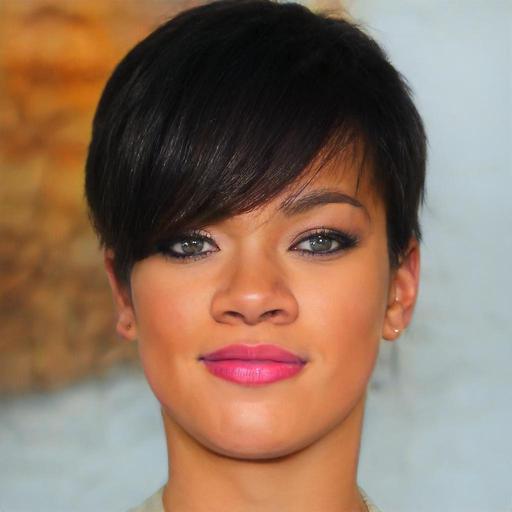}&
        \includegraphics[width=0.16\textwidth]{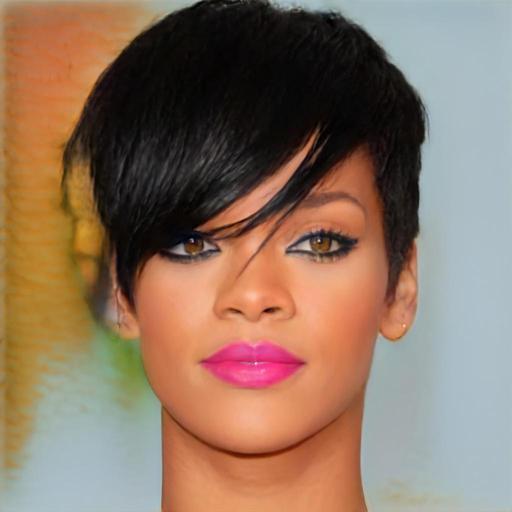}&
        \includegraphics[width=0.16\textwidth]{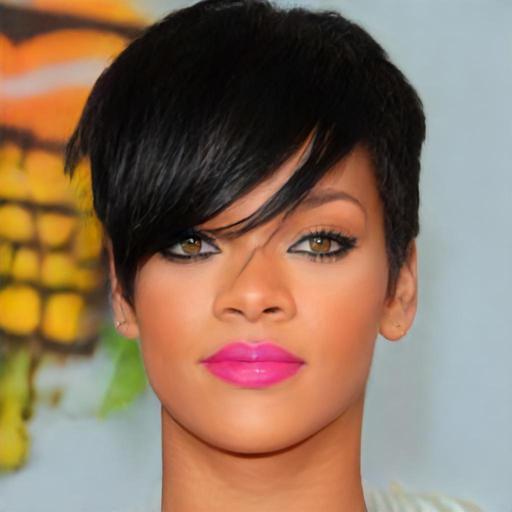} 
		\\
        \raisebox{0.45in}{\rotatebox[origin=t]{90}{Age}}&&
        \includegraphics[width=0.16\textwidth]{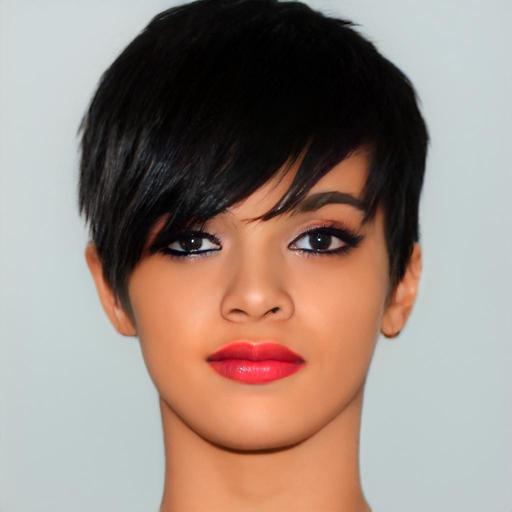}&        
        \includegraphics[width=0.16\textwidth]{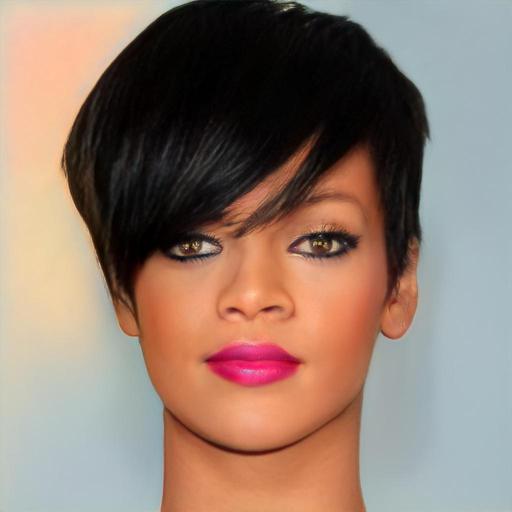}&        
        \includegraphics[width=0.16\textwidth]{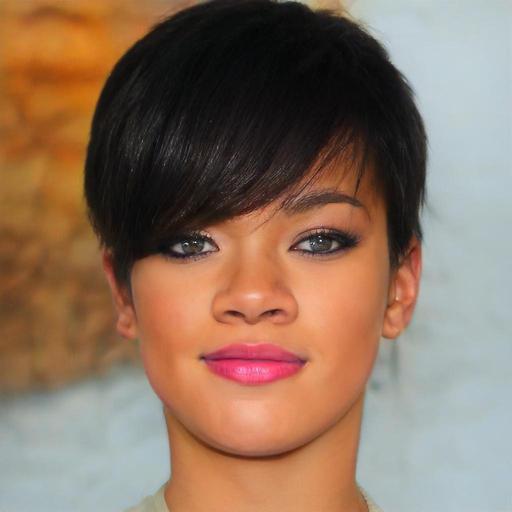}&
        \includegraphics[width=0.16\textwidth]{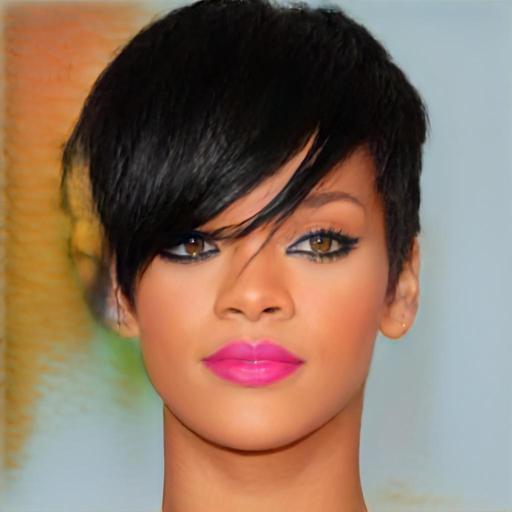}&
        \includegraphics[width=0.16\textwidth]{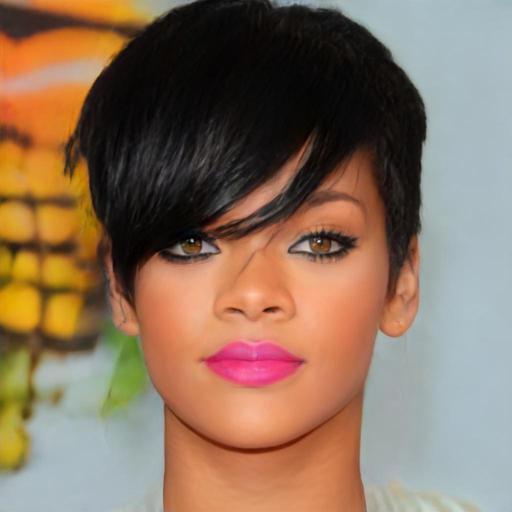} 
		\\
		& Input & SG2 & SG2$\mathcal{W}+$ & e4e & PTI & Ours
    \end{tabular}
    }
	\caption{Reconstruction and editing quality comparison using examples from CelebA-HQ test set. In each example, the editing is performed using the same editing weight.}
    \label{fig:appendix2}
\end{figure*}

\begin{figure*}
\setlength{\tabcolsep}{1pt}
\centering
{
	\renewcommand{\arraystretch}{0.5}
    \begin{tabular}{c c c c c c c}
		& Input & SG2 & SG2$\mathcal{W}+$ & e4e & PTI & Ours\\
        \raisebox{0.45in}{\rotatebox[origin=t]{90}{Inversion}}&
        \includegraphics[width=0.16\textwidth]{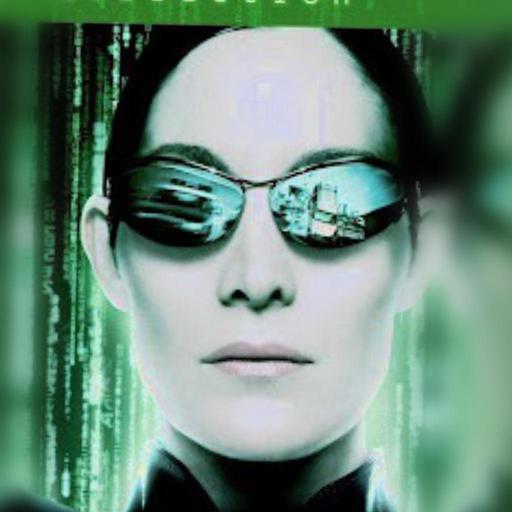}&
        \includegraphics[width=0.16\textwidth]{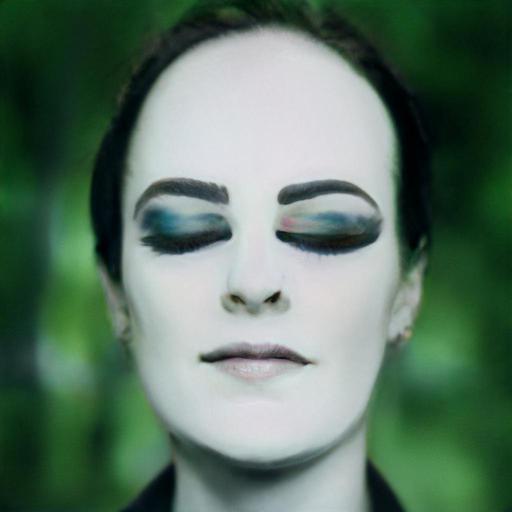}&        
        \includegraphics[width=0.16\textwidth]{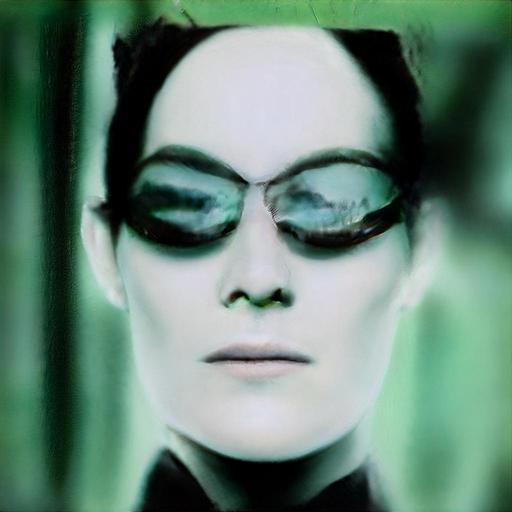}&        
        \includegraphics[width=0.16\textwidth]{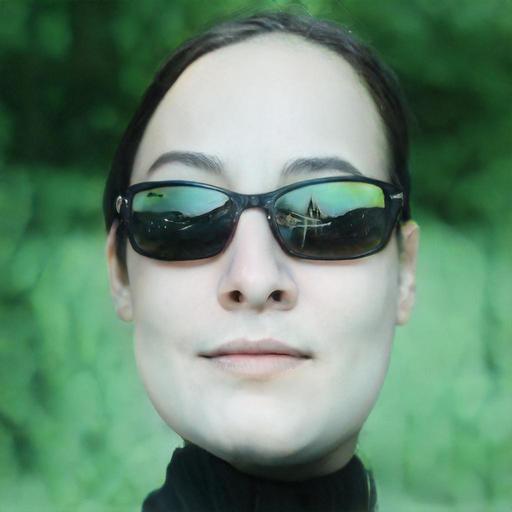}&
        \includegraphics[width=0.16\textwidth]{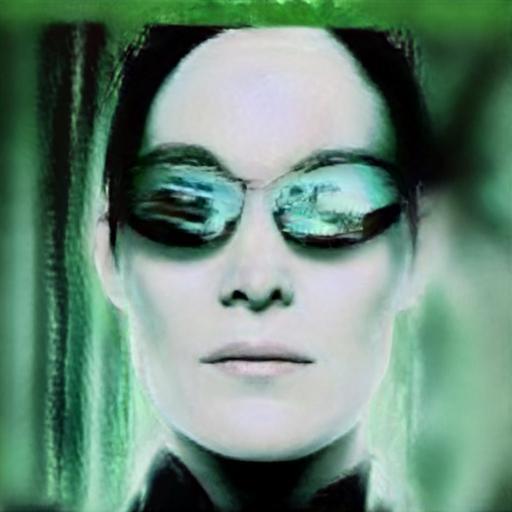}&
        \includegraphics[width=0.16\textwidth]{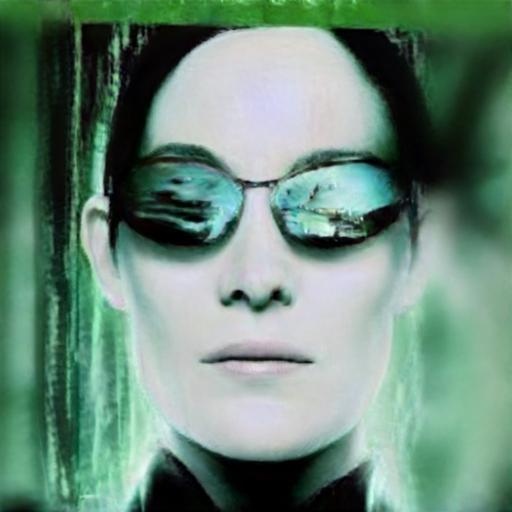} 
		\\
        \raisebox{0.45in}{\rotatebox[origin=t]{90}{Smile}}&&
        \includegraphics[width=0.16\textwidth]{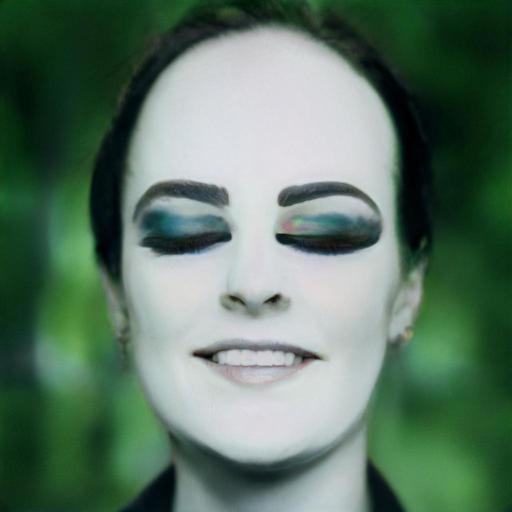}&        
        \includegraphics[width=0.16\textwidth]{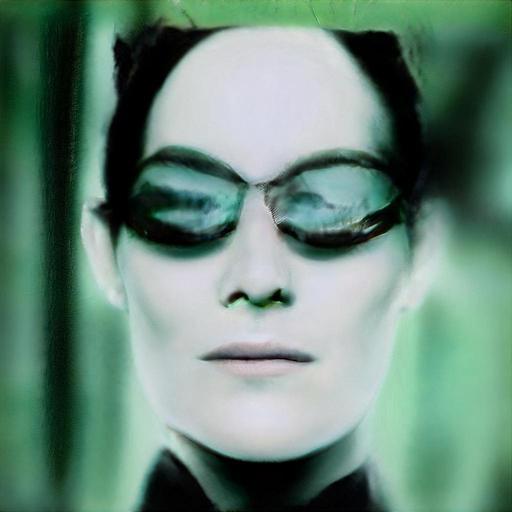}&        
        \includegraphics[width=0.16\textwidth]{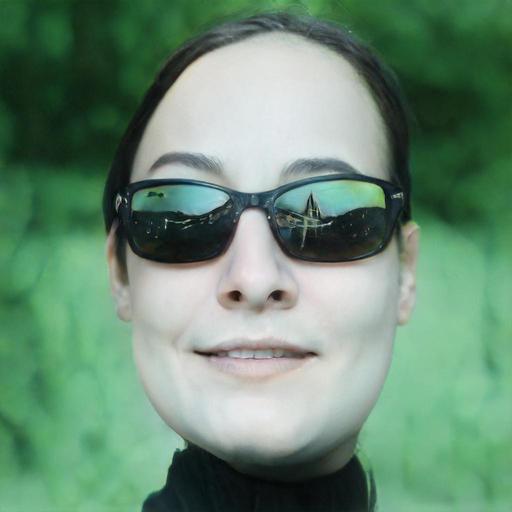}&
        \includegraphics[width=0.16\textwidth]{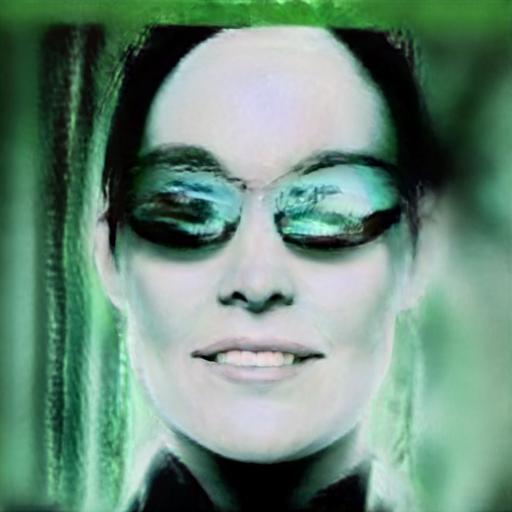}&
        \includegraphics[width=0.16\textwidth]{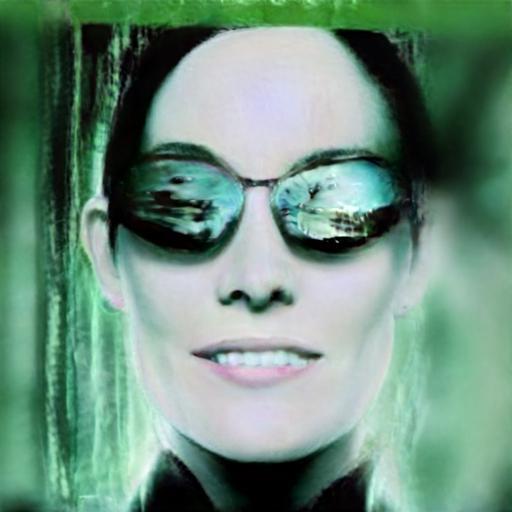} 
		\\[3pt]
        \raisebox{0.45in}{\rotatebox[origin=t]{90}{Inversion}}&
        \includegraphics[width=0.16\textwidth]{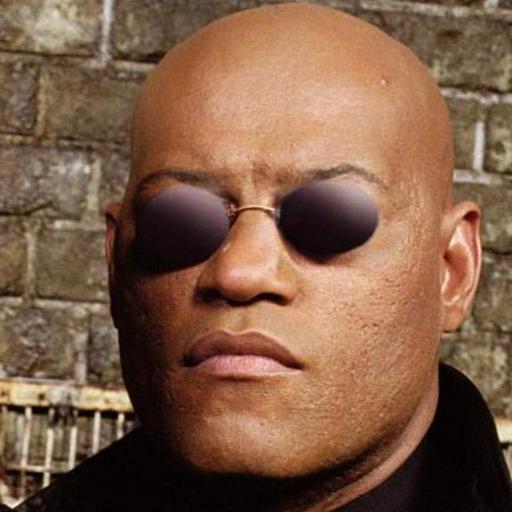}&
        \includegraphics[width=0.16\textwidth]{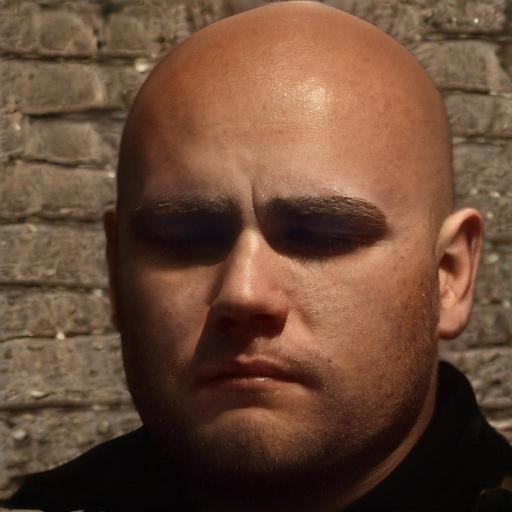}&        
        \includegraphics[width=0.16\textwidth]{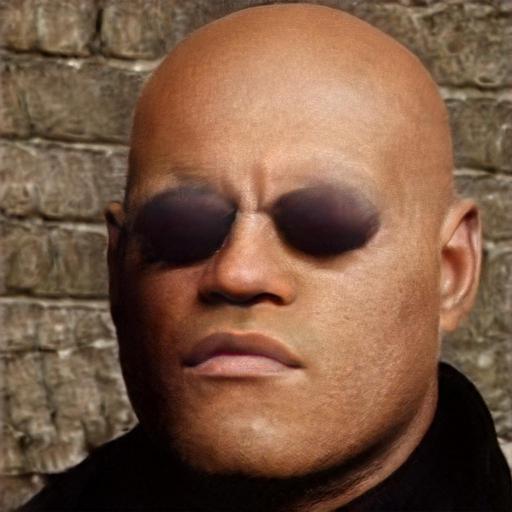}&        
        \includegraphics[width=0.16\textwidth]{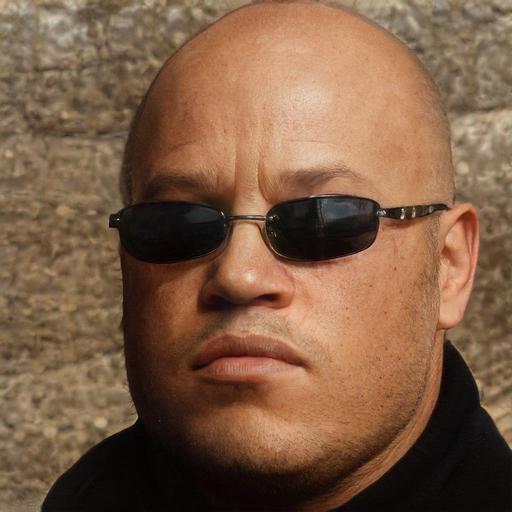}&
        \includegraphics[width=0.16\textwidth]{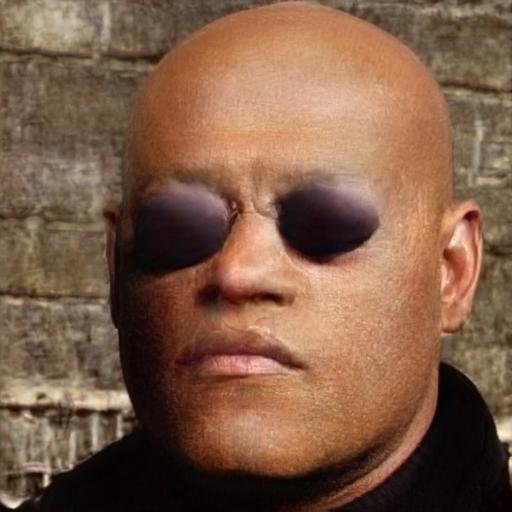}&
        \includegraphics[width=0.16\textwidth]{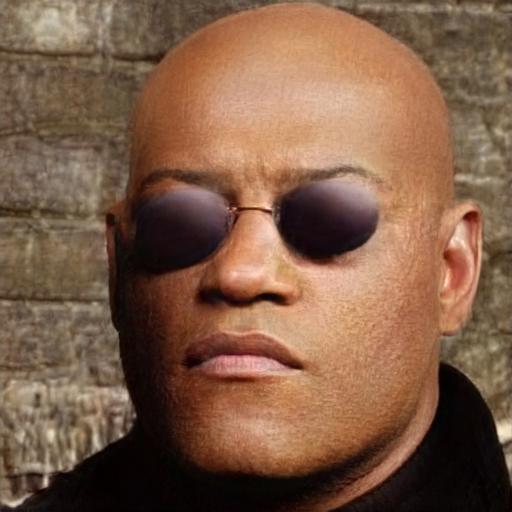} 
		\\
        \raisebox{0.45in}{\rotatebox[origin=t]{90}{Pose}}&&
        \includegraphics[width=0.16\textwidth]{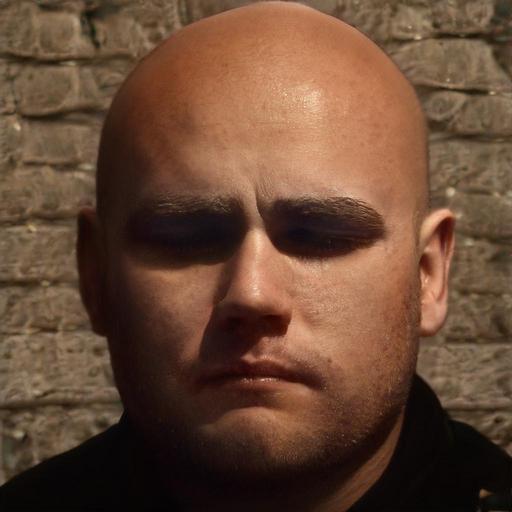}&        
        \includegraphics[width=0.16\textwidth]{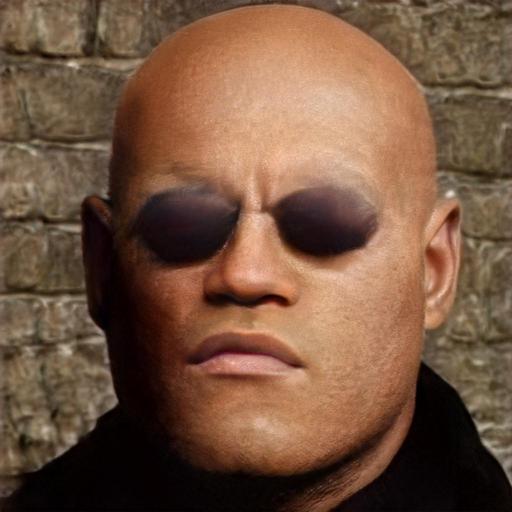}&        
        \includegraphics[width=0.16\textwidth]{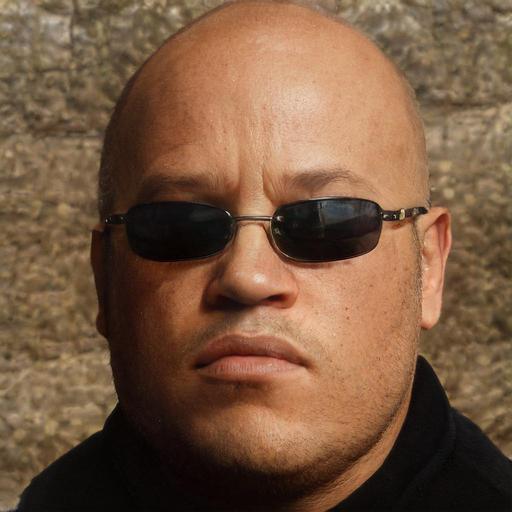}&
        \includegraphics[width=0.16\textwidth]{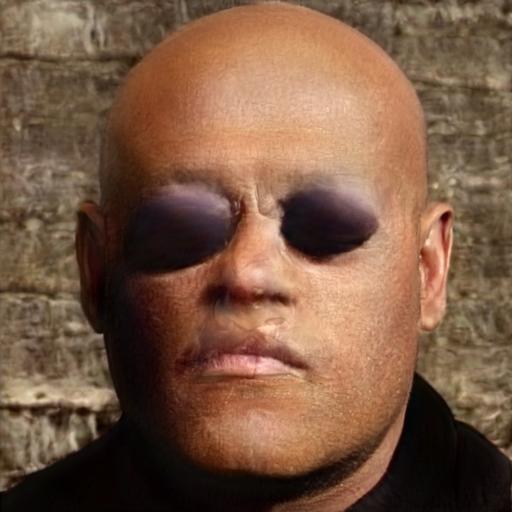}&
        \includegraphics[width=0.16\textwidth]{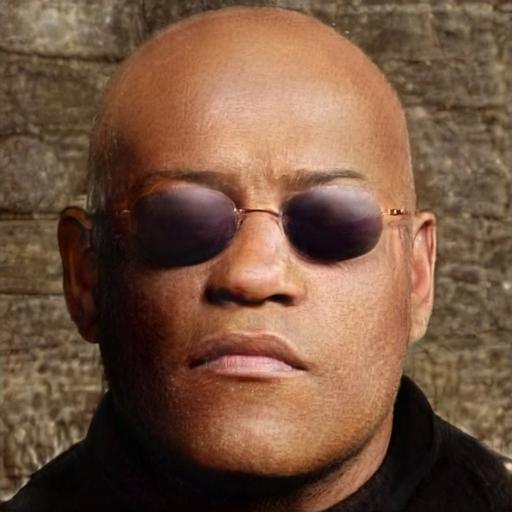} 
		\\[3pt]
        \raisebox{0.45in}{\rotatebox[origin=t]{90}{Inversion}}&
        \includegraphics[width=0.16\textwidth]{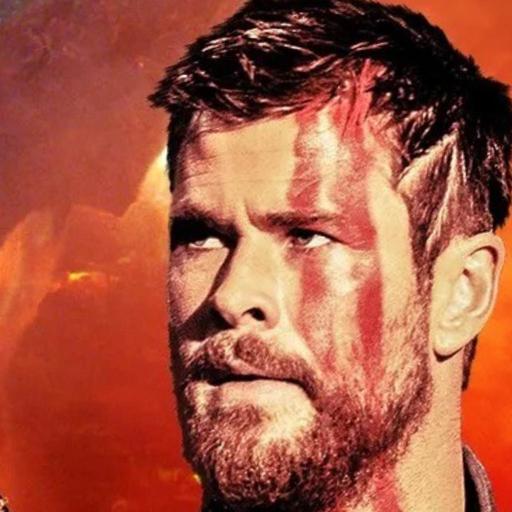}&
        \includegraphics[width=0.16\textwidth]{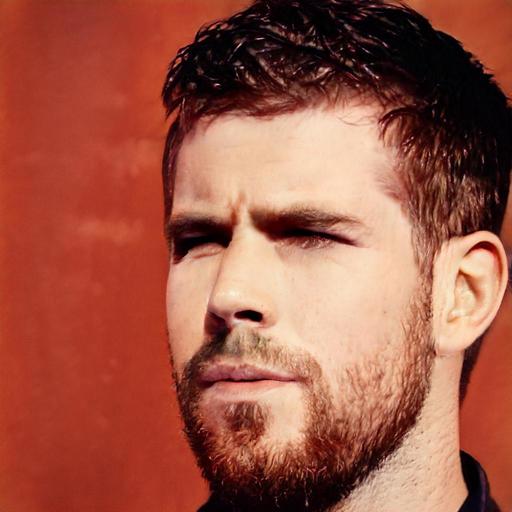}&        
        \includegraphics[width=0.16\textwidth]{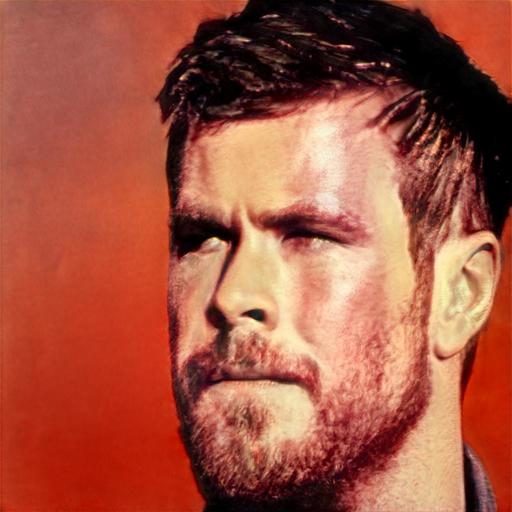}&        
        \includegraphics[width=0.16\textwidth]{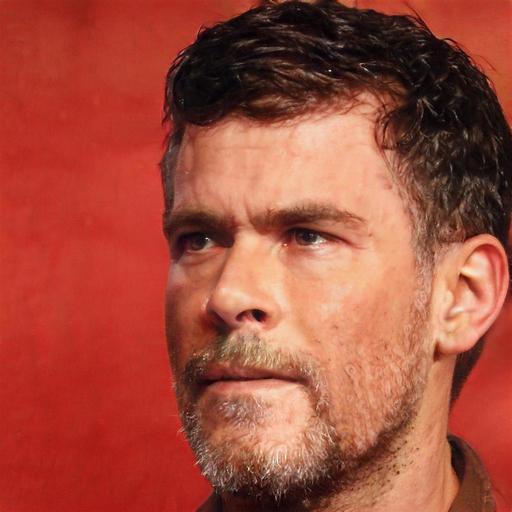}&
        \includegraphics[width=0.16\textwidth]{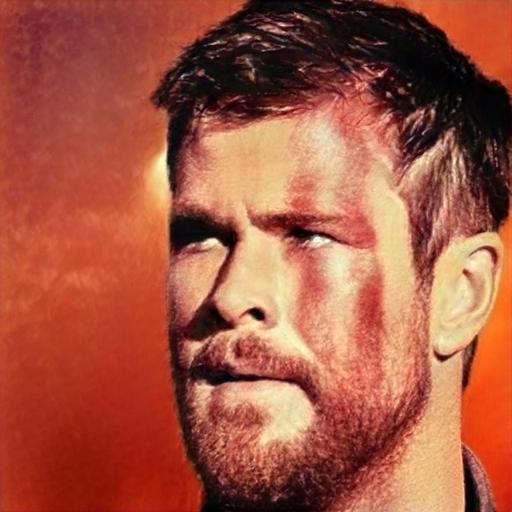}&
        \includegraphics[width=0.16\textwidth]{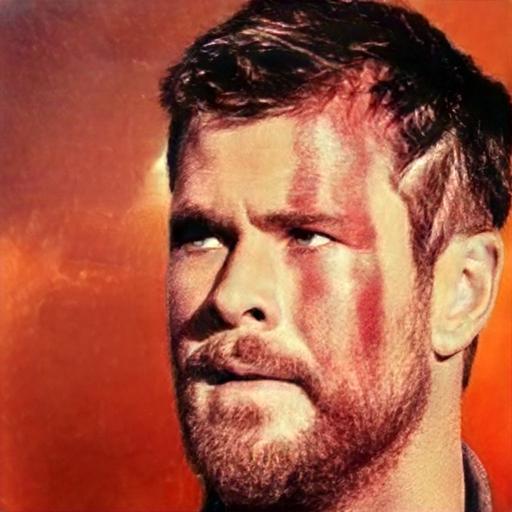} 
		\\
        \raisebox{0.45in}{\rotatebox[origin=t]{90}{Age}}&&
        \includegraphics[width=0.16\textwidth]{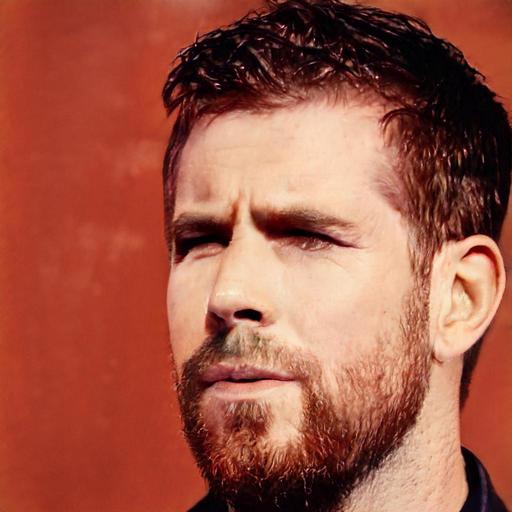}&        
        \includegraphics[width=0.16\textwidth]{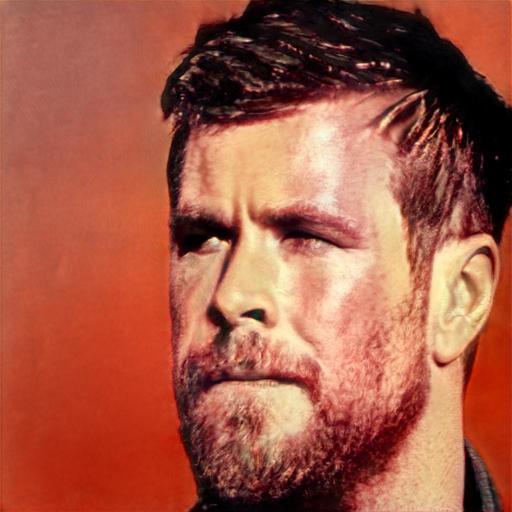}&        
        \includegraphics[width=0.16\textwidth]{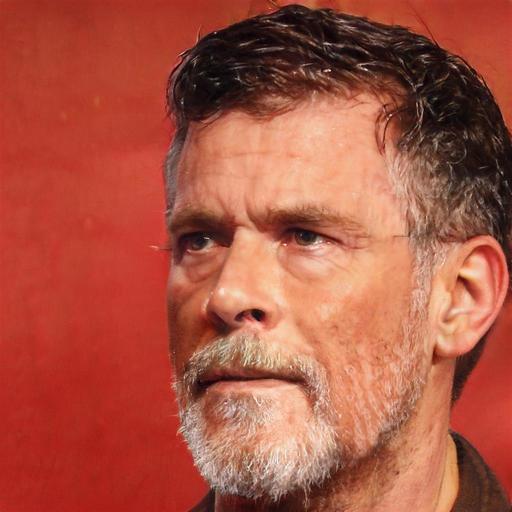}&
        \includegraphics[width=0.16\textwidth]{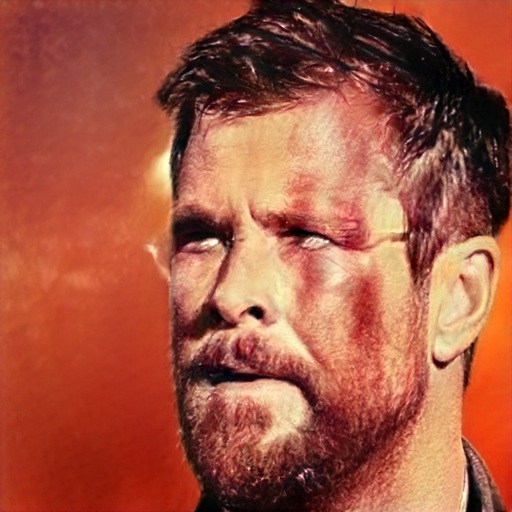}&
        \includegraphics[width=0.16\textwidth]{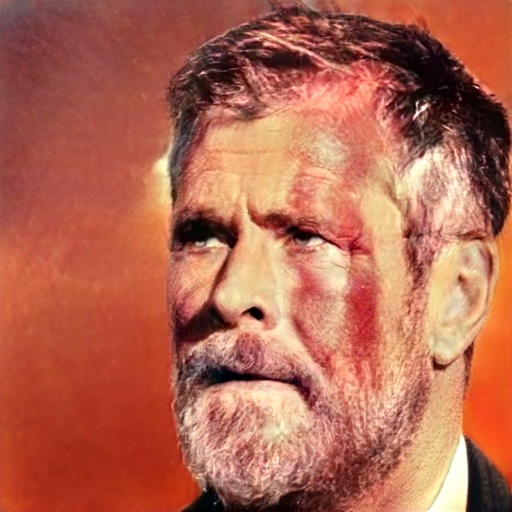} 
		\\
		& Input & SG2 & SG2$\mathcal{W}+$ & e4e & PTI & Ours
    \end{tabular}
    }
	\caption{Reconstruction and editing quality comparison using famous character images collected from the web. In each example, the editing is performed using the same editing weight.}
    \label{fig:appendix3}
\end{figure*}

\begin{figure*}
\setlength{\tabcolsep}{1pt}
\centering
{
	\renewcommand{\arraystretch}{0.5}
    \begin{tabular}{c c c c c c c}
		& Input & SG2 & SG2$\mathcal{W}+$ & e4e & PTI & Ours\\
        \raisebox{0.45in}{\rotatebox[origin=t]{90}{Inversion}}&
        \includegraphics[width=0.16\textwidth]{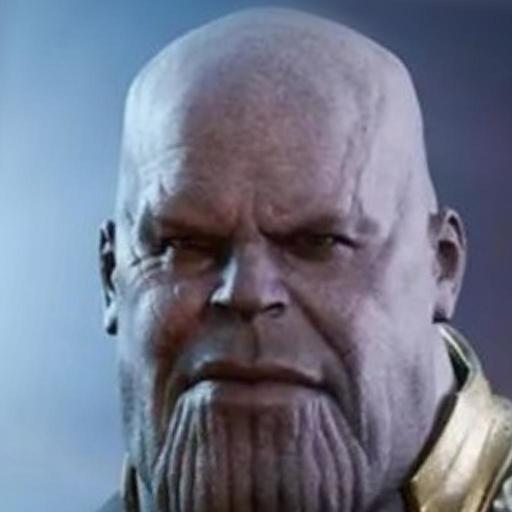}&
        \includegraphics[width=0.16\textwidth]{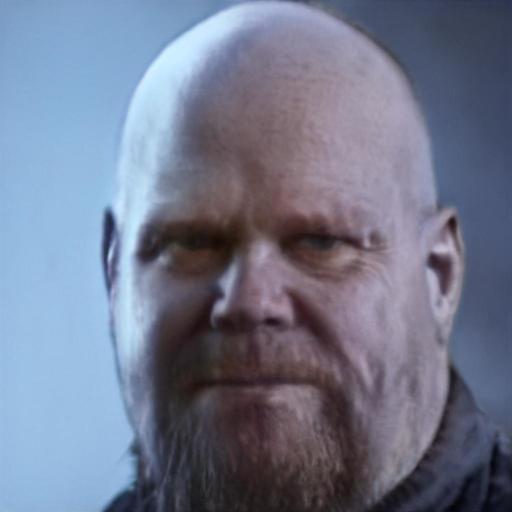}&        
        \includegraphics[width=0.16\textwidth]{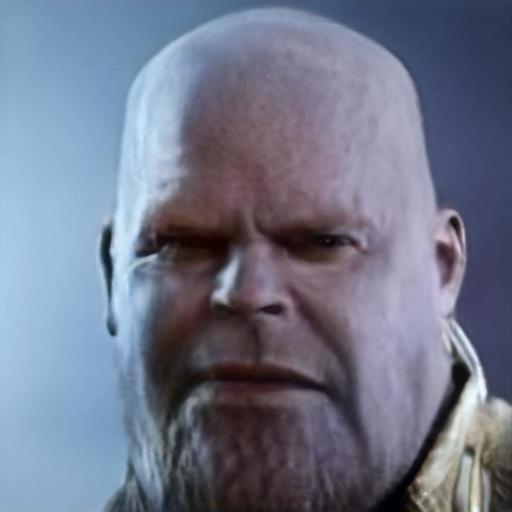}&        
        \includegraphics[width=0.16\textwidth]{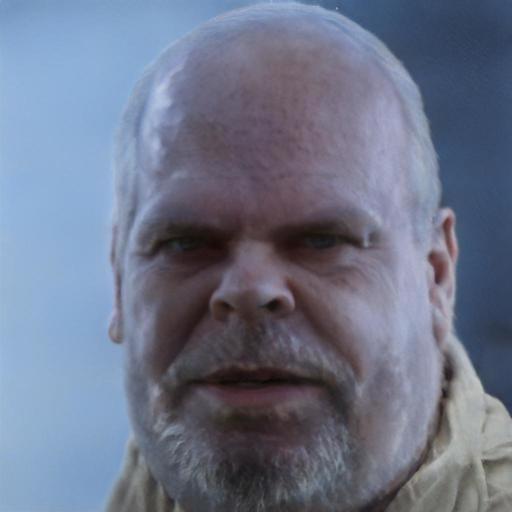}&
        \includegraphics[width=0.16\textwidth]{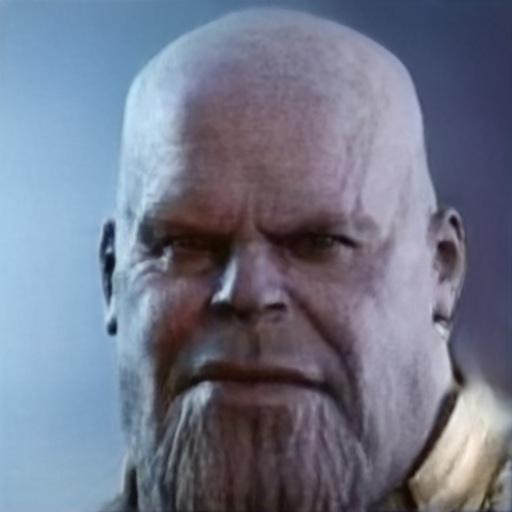}&
        \includegraphics[width=0.16\textwidth]{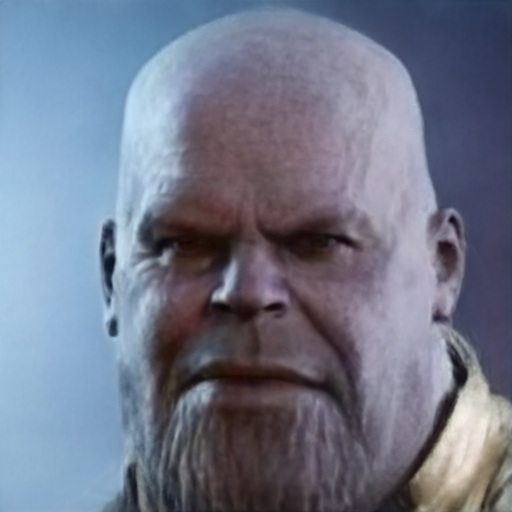} 
		\\
        \raisebox{0.45in}{\rotatebox[origin=t]{90}{Smile}}&&
        \includegraphics[width=0.16\textwidth]{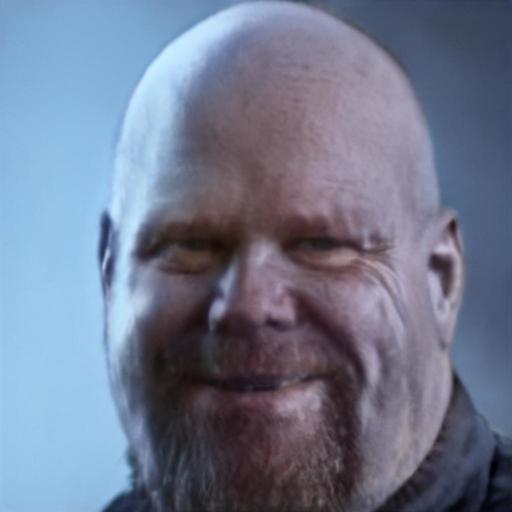}&        
        \includegraphics[width=0.16\textwidth]{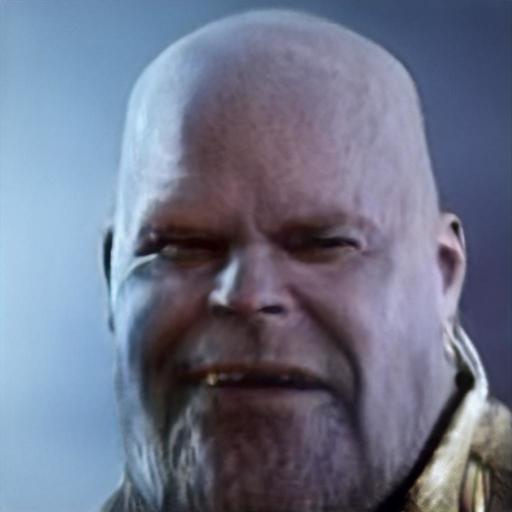}&        
        \includegraphics[width=0.16\textwidth]{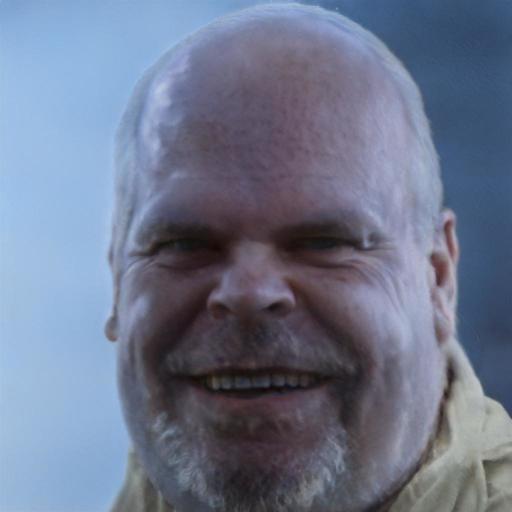}&
        \includegraphics[width=0.16\textwidth]{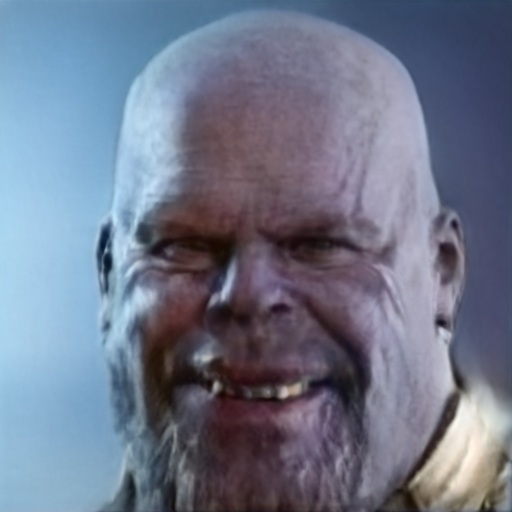}&
        \includegraphics[width=0.16\textwidth]{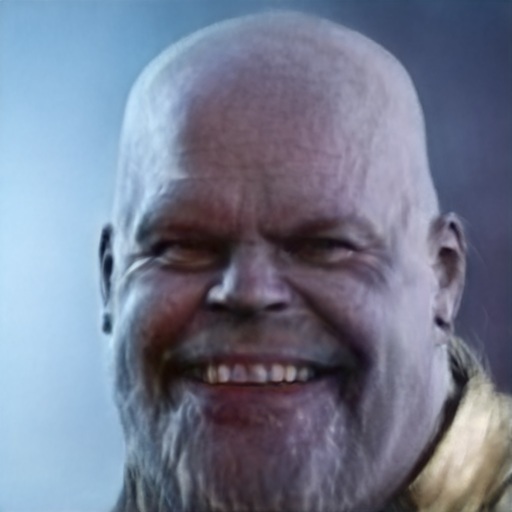} 
		\\[3pt]
        \raisebox{0.45in}{\rotatebox[origin=t]{90}{Inversion}}&
        \includegraphics[width=0.16\textwidth]{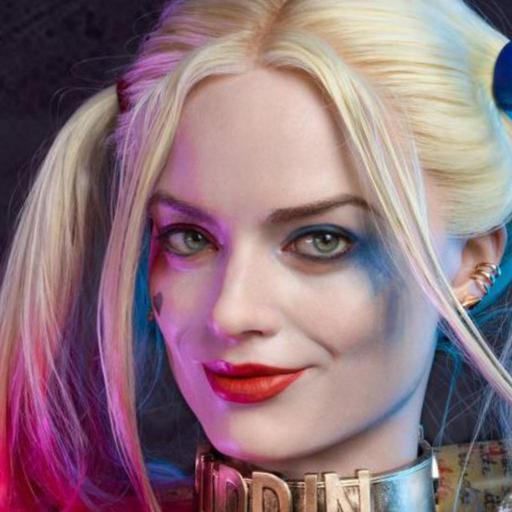}&
        \includegraphics[width=0.16\textwidth]{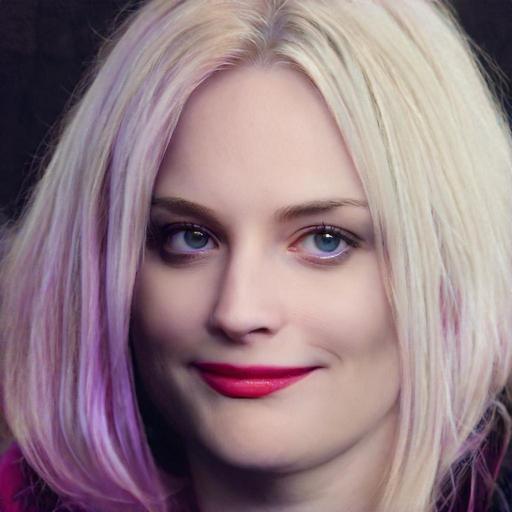}&        
        \includegraphics[width=0.16\textwidth]{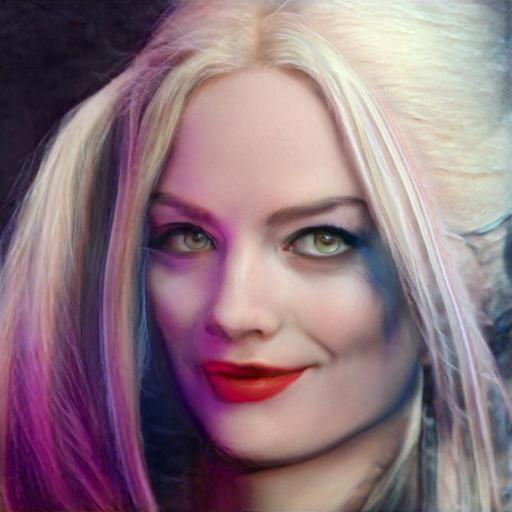}&        
        \includegraphics[width=0.16\textwidth]{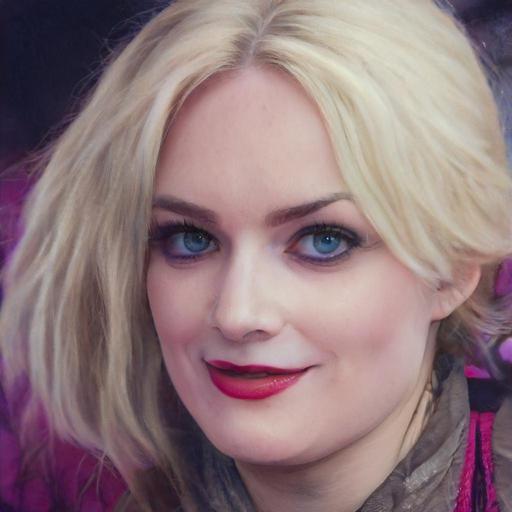}&
        \includegraphics[width=0.16\textwidth]{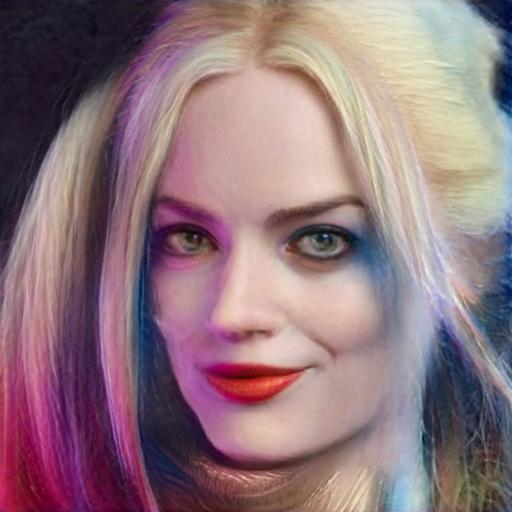}&
        \includegraphics[width=0.16\textwidth]{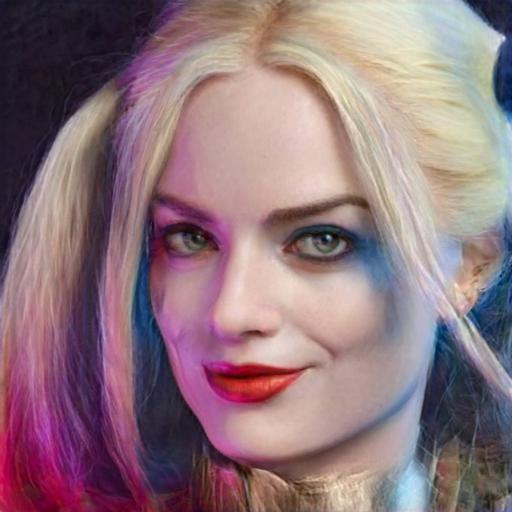} 
		\\
        \raisebox{0.45in}{\rotatebox[origin=t]{90}{Pose}}&&
        \includegraphics[width=0.16\textwidth]{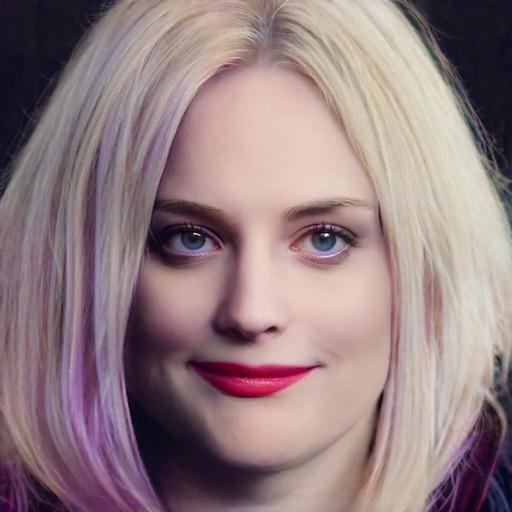}&        
        \includegraphics[width=0.16\textwidth]{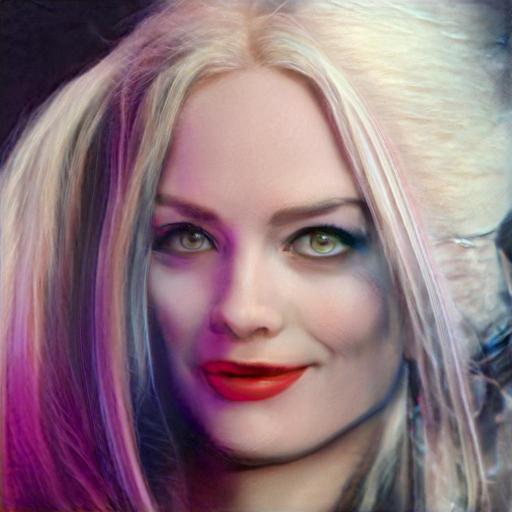}&        
        \includegraphics[width=0.16\textwidth]{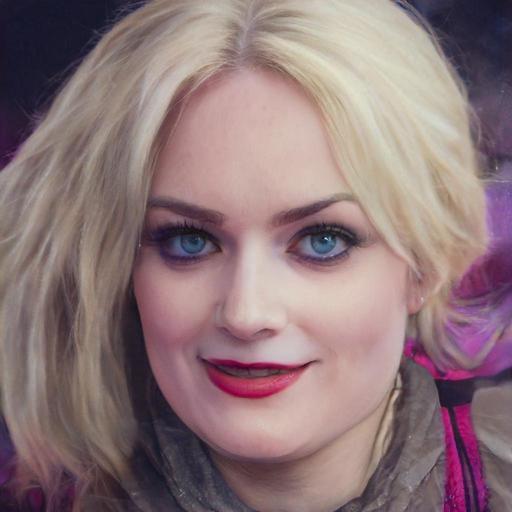}&
        \includegraphics[width=0.16\textwidth]{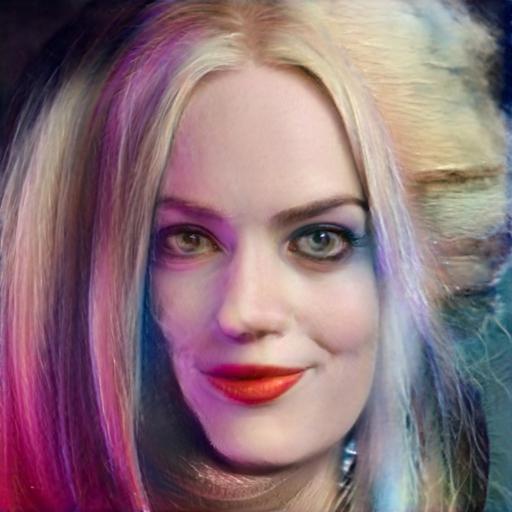}&
        \includegraphics[width=0.16\textwidth]{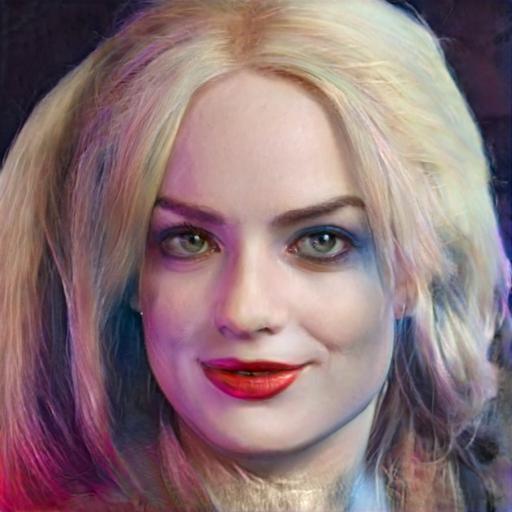} 
		\\[3pt]
        \raisebox{0.45in}{\rotatebox[origin=t]{90}{Inversion}}&
        \includegraphics[width=0.16\textwidth]{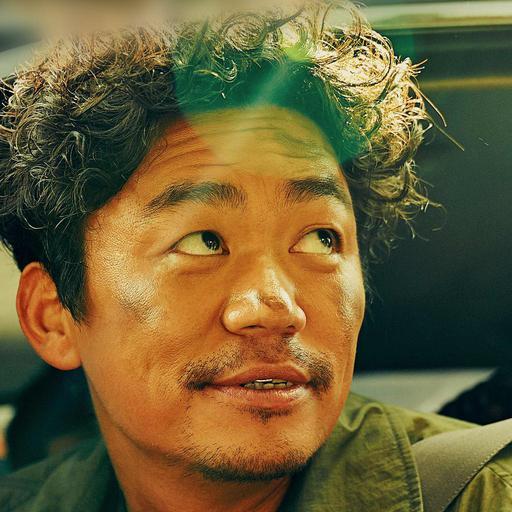}&
        \includegraphics[width=0.16\textwidth]{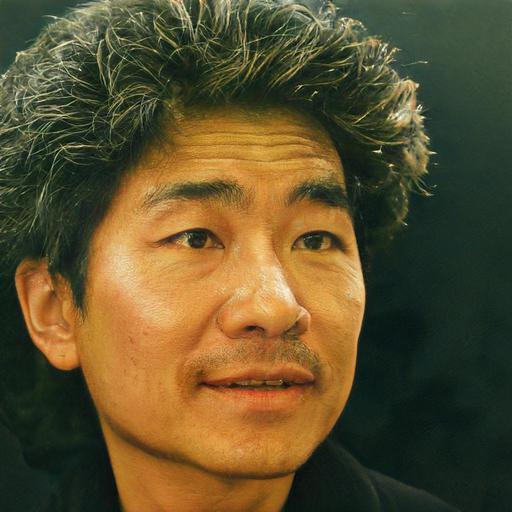}&        
        \includegraphics[width=0.16\textwidth]{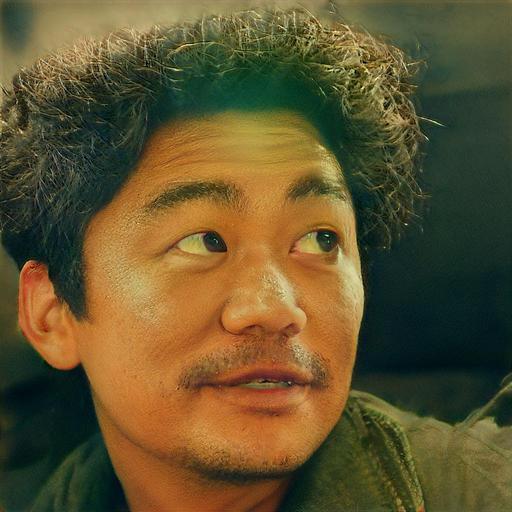}&        
        \includegraphics[width=0.16\textwidth]{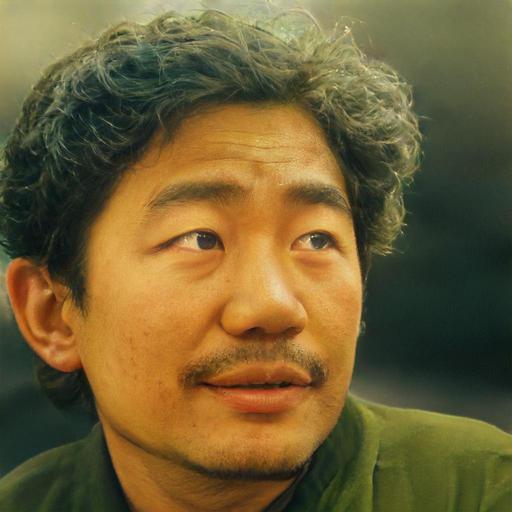}&
        \includegraphics[width=0.16\textwidth]{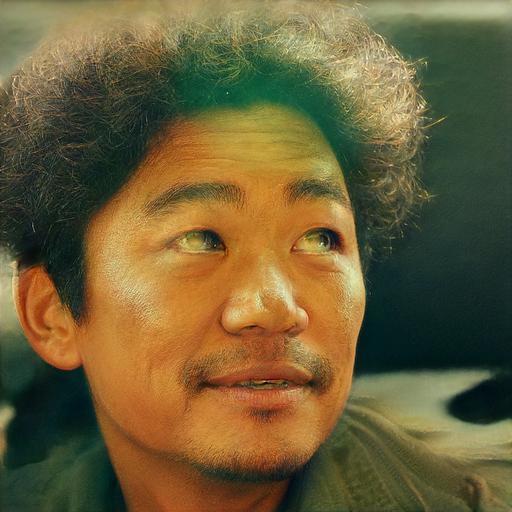}&
        \includegraphics[width=0.16\textwidth]{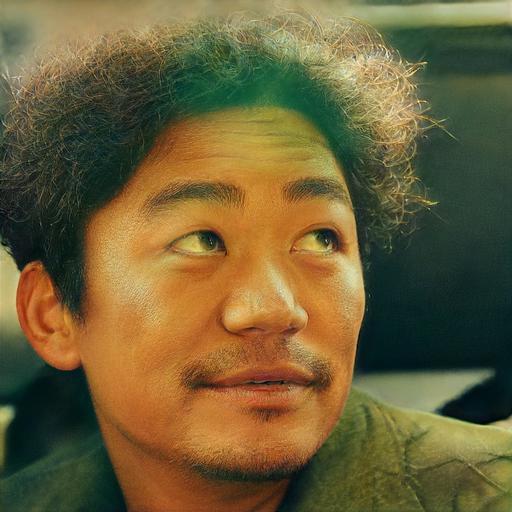} 
		\\
        \raisebox{0.45in}{\rotatebox[origin=t]{90}{Age}}&&
        \includegraphics[width=0.16\textwidth]{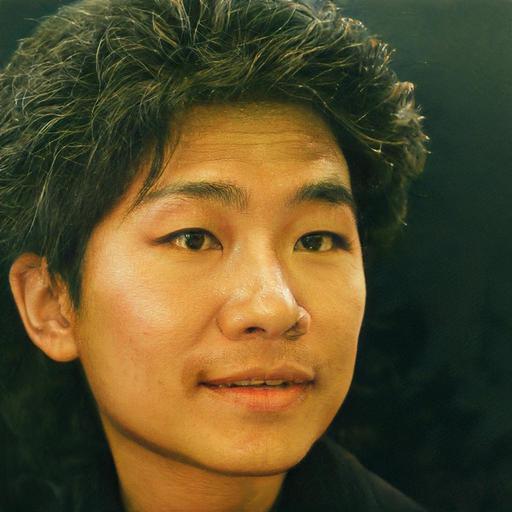}&        
        \includegraphics[width=0.16\textwidth]{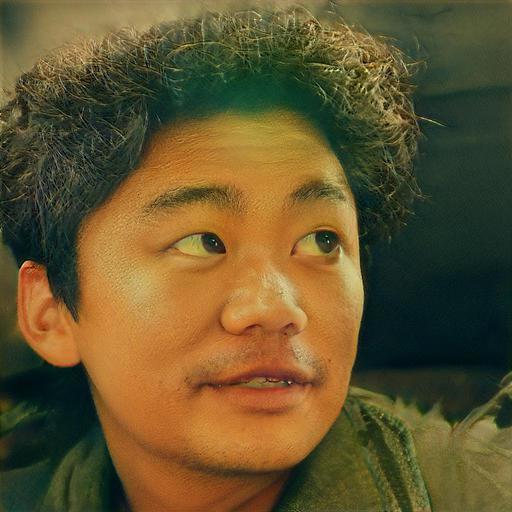}&        
        \includegraphics[width=0.16\textwidth]{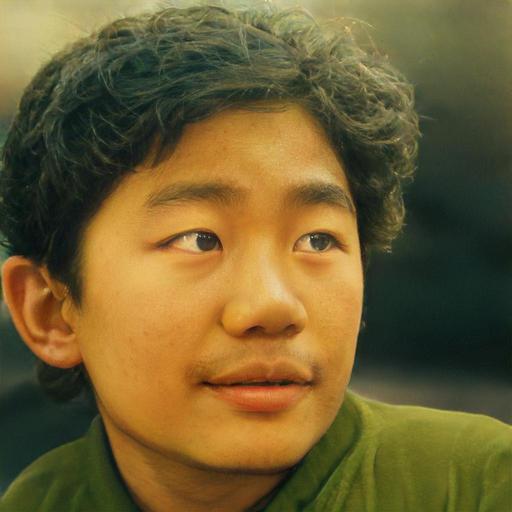}&
        \includegraphics[width=0.16\textwidth]{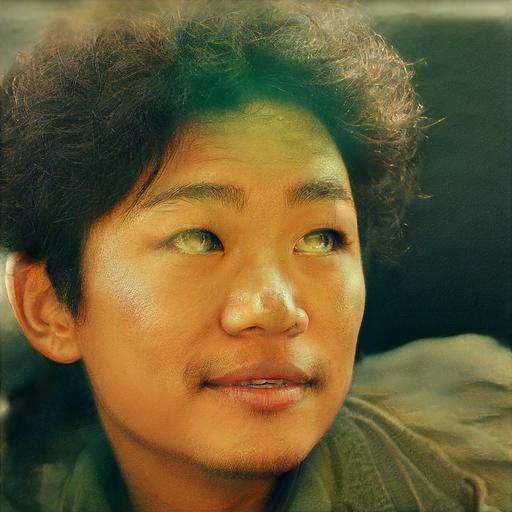}&
        \includegraphics[width=0.16\textwidth]{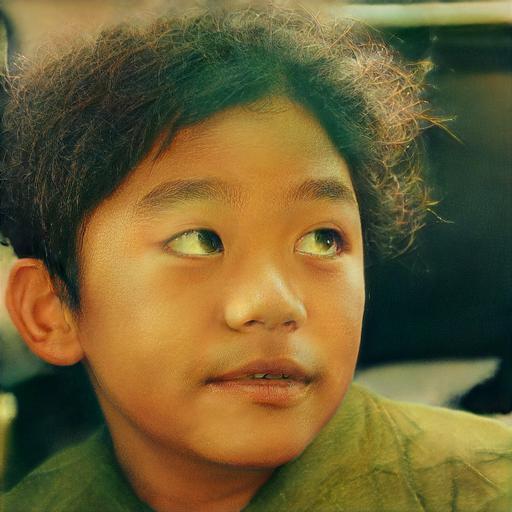} 
		\\
		& Input & SG2 & SG2$\mathcal{W}+$ & e4e & PTI & Ours
    \end{tabular}
    }
	\caption{Reconstruction and editing quality comparison using famous character images collected from the web. In each example, the editing is performed using the same editing weight.}
    \label{fig:appendix4}
\end{figure*}

\begin{figure*}
\setlength{\tabcolsep}{1pt}
\centering
{
	\renewcommand{\arraystretch}{0.5}
    \begin{tabular}{c c c c c c c}
		& Input & SG2 & SG2$\mathcal{W}+$ & e4e & PTI & Ours\\
        \raisebox{0.45in}{\rotatebox[origin=t]{90}{Inversion}}&
        \includegraphics[width=0.16\textwidth]{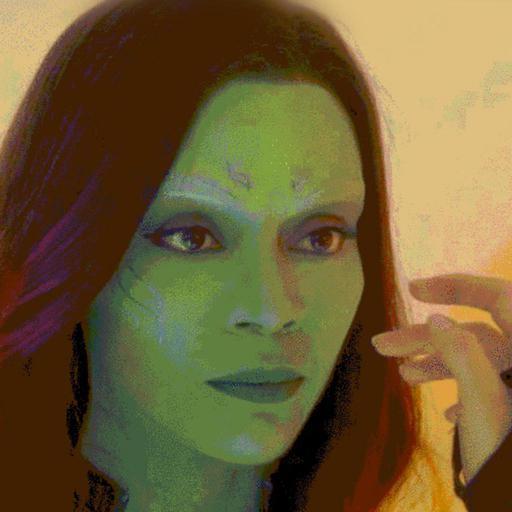}&
        \includegraphics[width=0.16\textwidth]{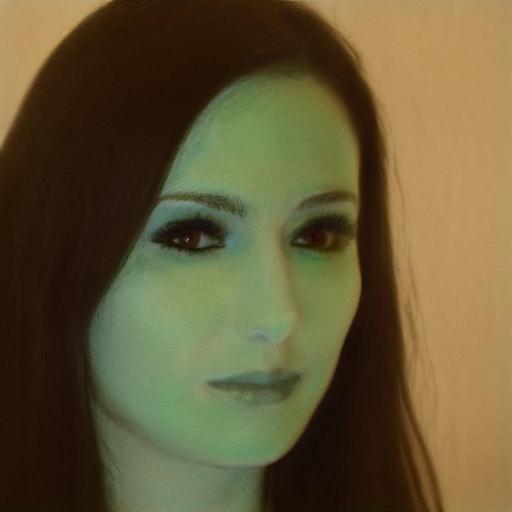}&        
        \includegraphics[width=0.16\textwidth]{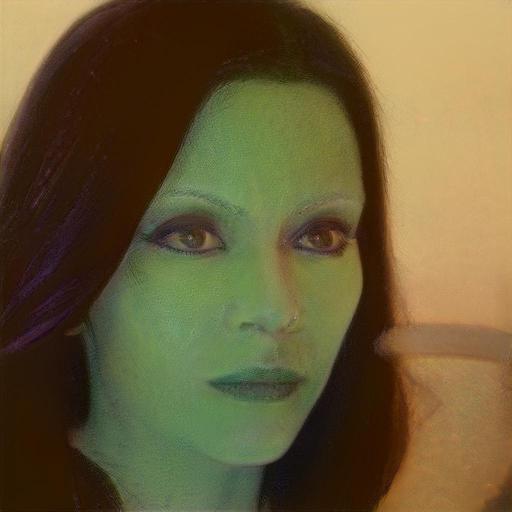}&        
        \includegraphics[width=0.16\textwidth]{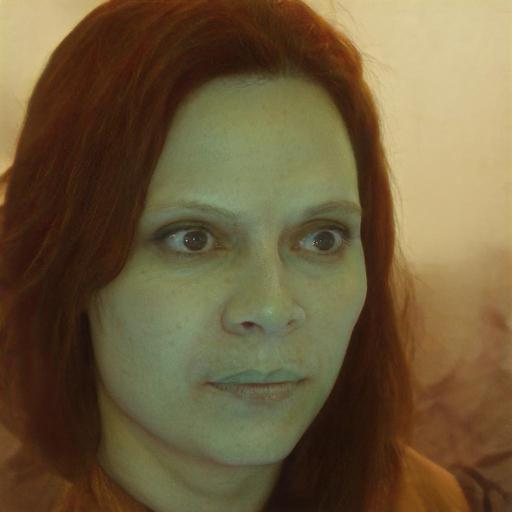}&
        \includegraphics[width=0.16\textwidth]{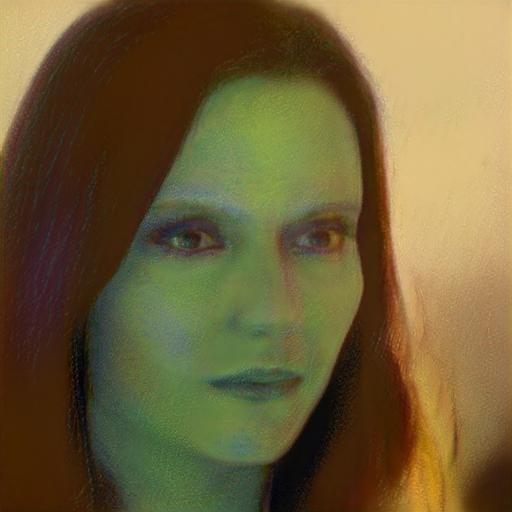}&
        \includegraphics[width=0.16\textwidth]{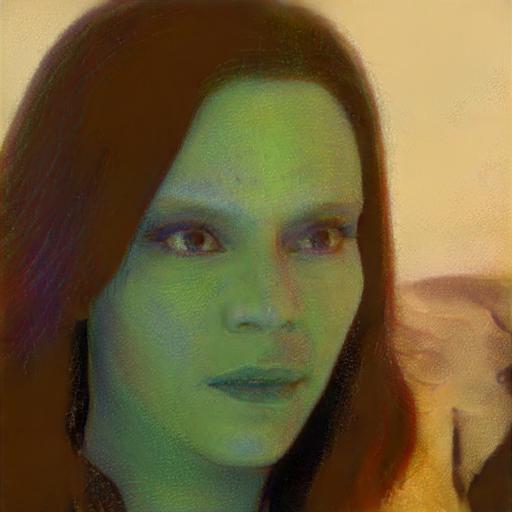} 
		\\
        \raisebox{0.45in}{\rotatebox[origin=t]{90}{Smile}}&&
        \includegraphics[width=0.16\textwidth]{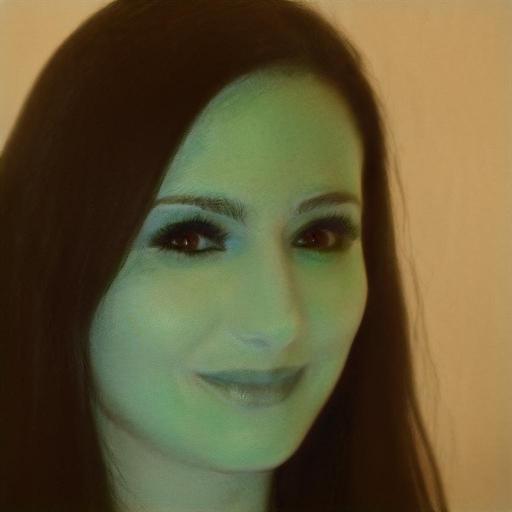}&        
        \includegraphics[width=0.16\textwidth]{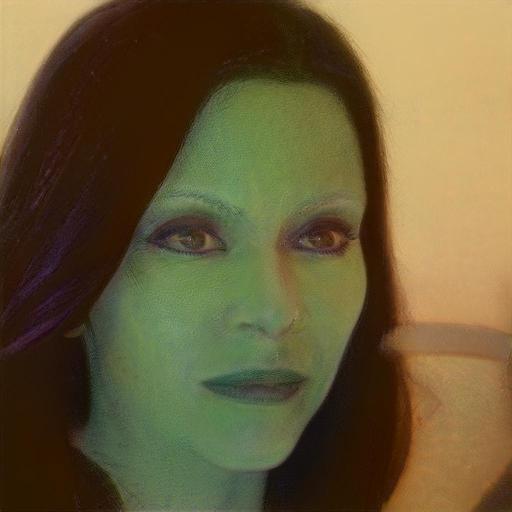}&        
        \includegraphics[width=0.16\textwidth]{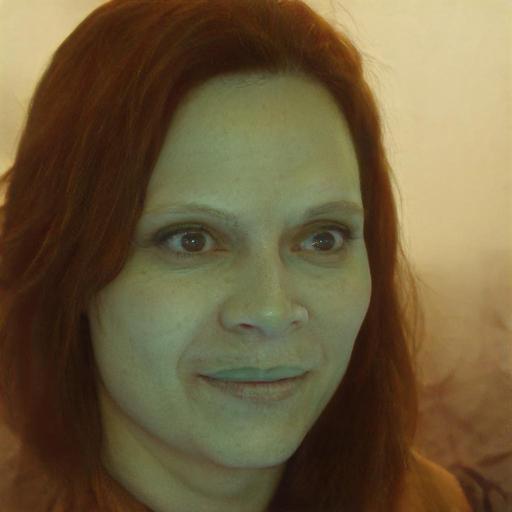}&
        \includegraphics[width=0.16\textwidth]{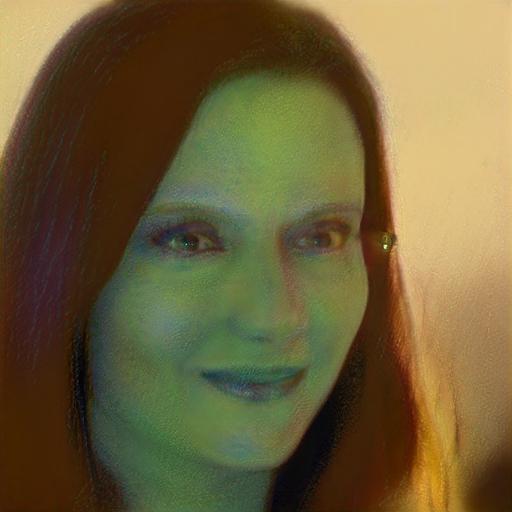}&
        \includegraphics[width=0.16\textwidth]{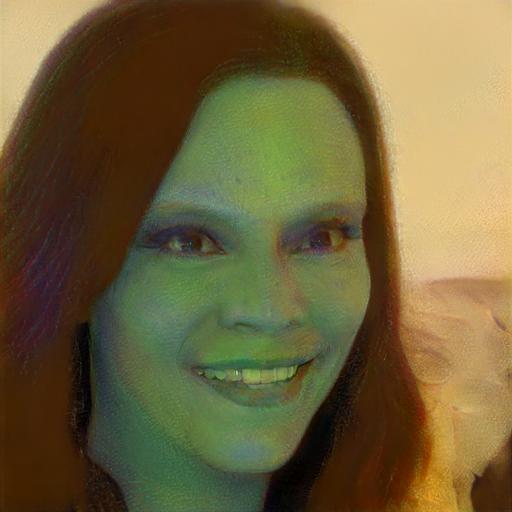} 
		\\[3pt]
        \raisebox{0.45in}{\rotatebox[origin=t]{90}{Inversion}}&
        \includegraphics[width=0.16\textwidth]{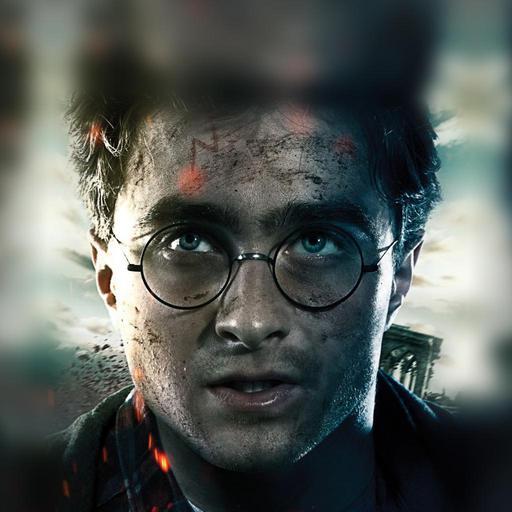}&
        \includegraphics[width=0.16\textwidth]{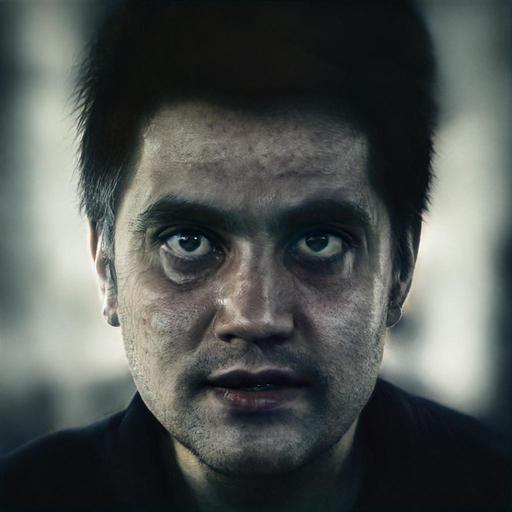}&        
        \includegraphics[width=0.16\textwidth]{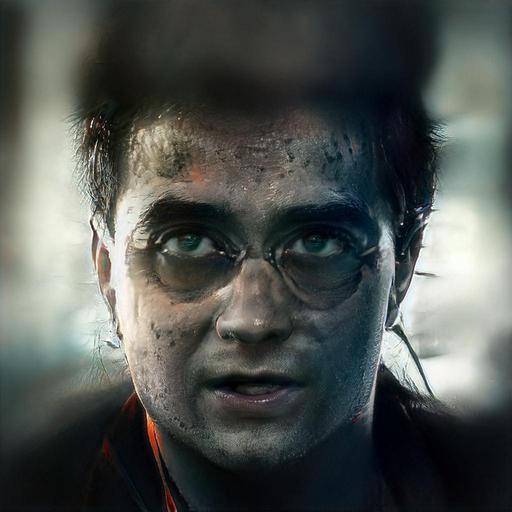}&        
        \includegraphics[width=0.16\textwidth]{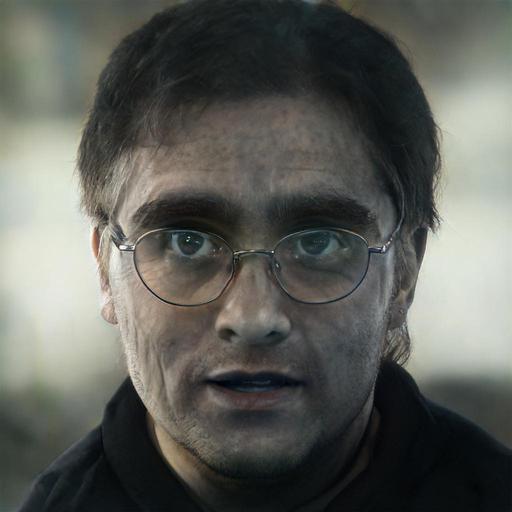}&
        \includegraphics[width=0.16\textwidth]{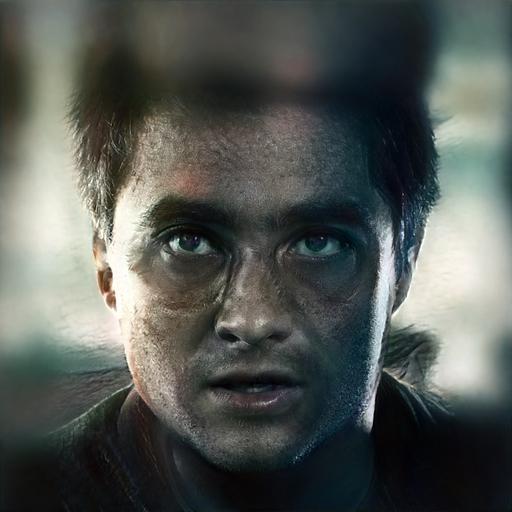}&
        \includegraphics[width=0.16\textwidth]{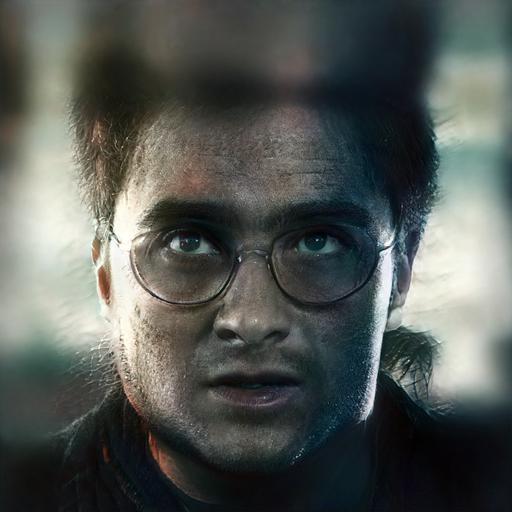} 
		\\
        \raisebox{0.45in}{\rotatebox[origin=t]{90}{Pose}}&&
        \includegraphics[width=0.16\textwidth]{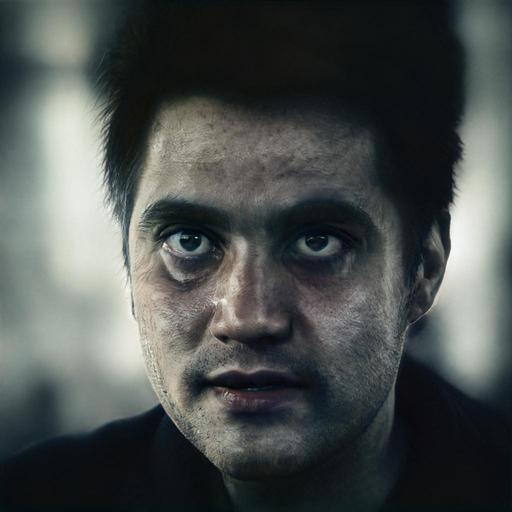}&        
        \includegraphics[width=0.16\textwidth]{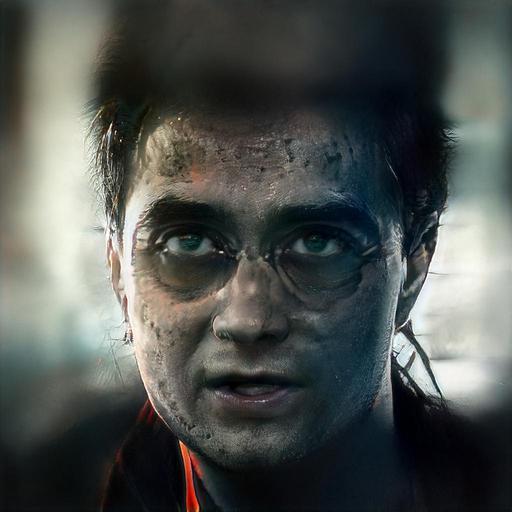}&        
        \includegraphics[width=0.16\textwidth]{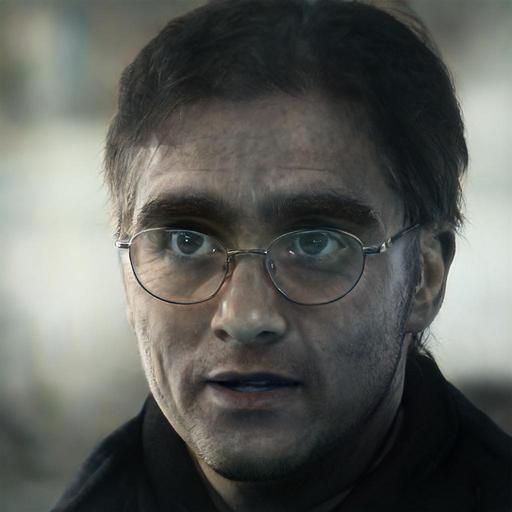}&
        \includegraphics[width=0.16\textwidth]{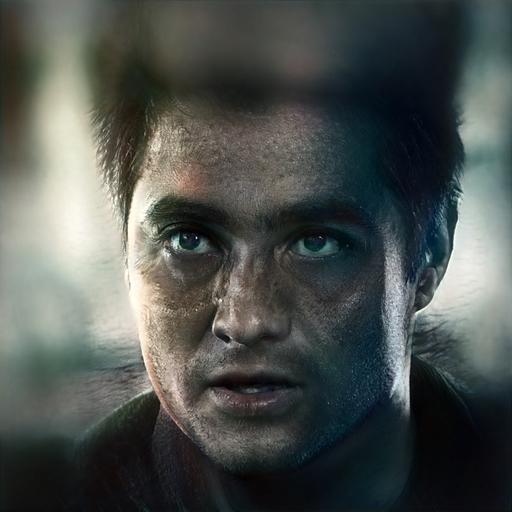}&
        \includegraphics[width=0.16\textwidth]{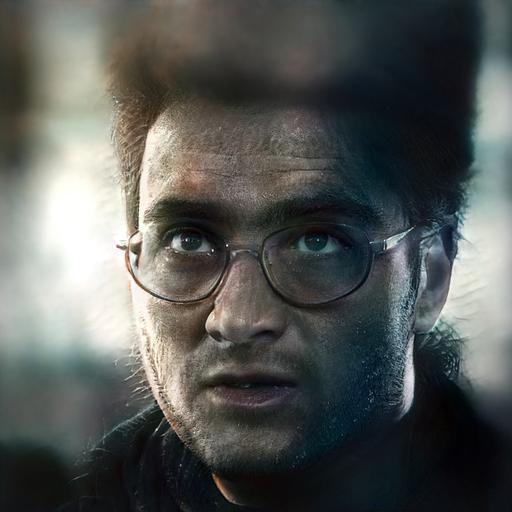} 
		\\[3pt]
        \raisebox{0.45in}{\rotatebox[origin=t]{90}{Inversion}}&
        \includegraphics[width=0.16\textwidth]{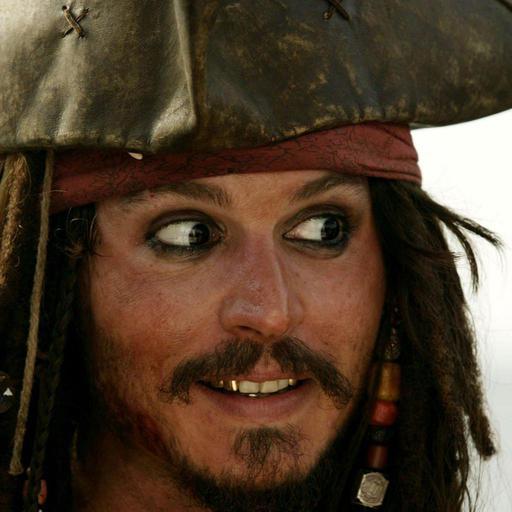}&
        \includegraphics[width=0.16\textwidth]{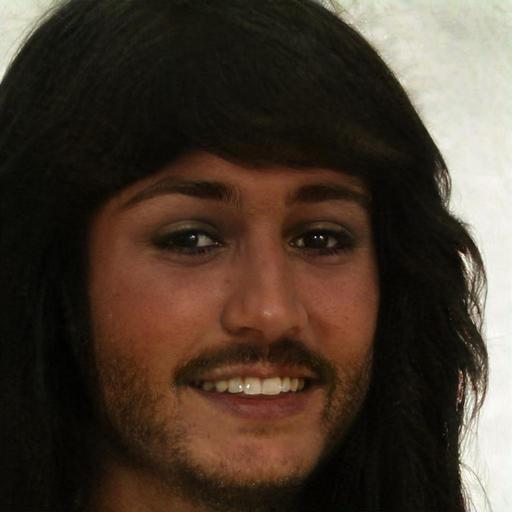}&        
        \includegraphics[width=0.16\textwidth]{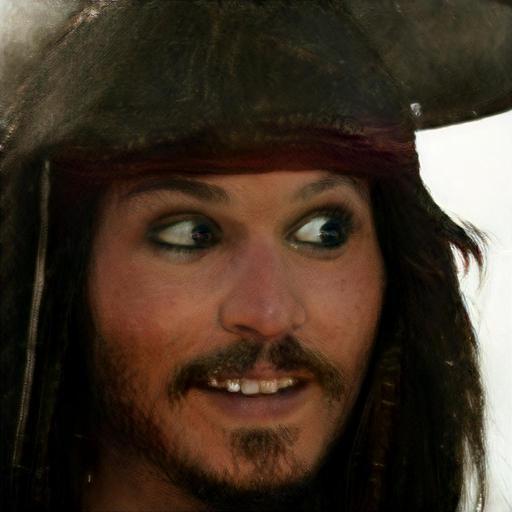}&        
        \includegraphics[width=0.16\textwidth]{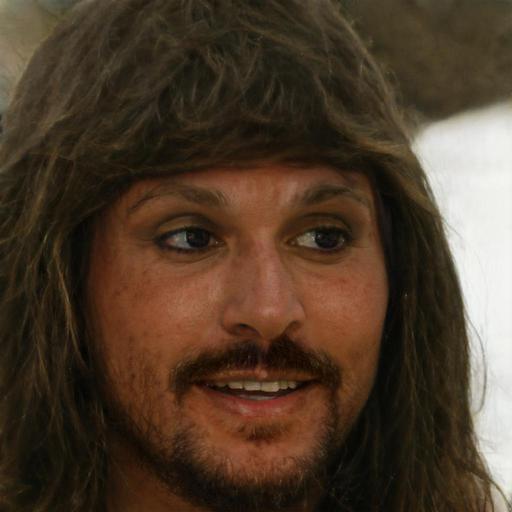}&
        \includegraphics[width=0.16\textwidth]{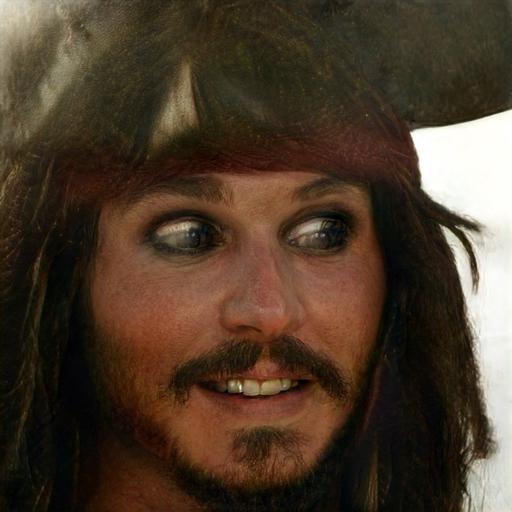}&
        \includegraphics[width=0.16\textwidth]{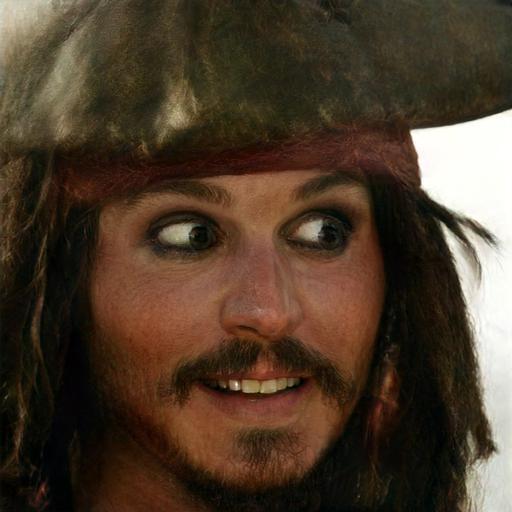} 
		\\
        \raisebox{0.45in}{\rotatebox[origin=t]{90}{Age}}&&
        \includegraphics[width=0.16\textwidth]{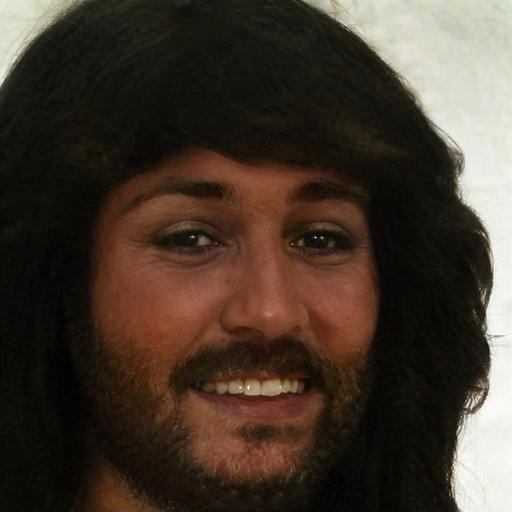}&        
        \includegraphics[width=0.16\textwidth]{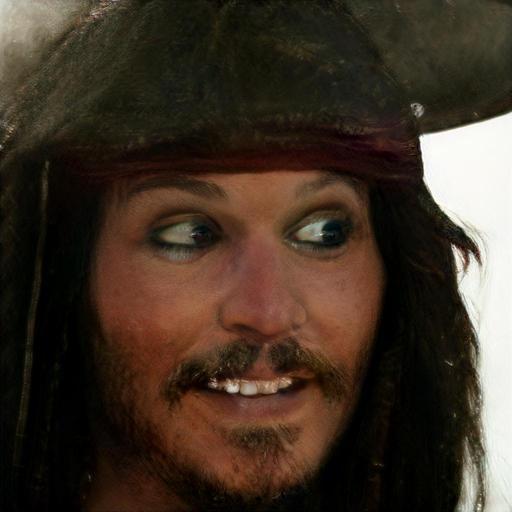}&        
        \includegraphics[width=0.16\textwidth]{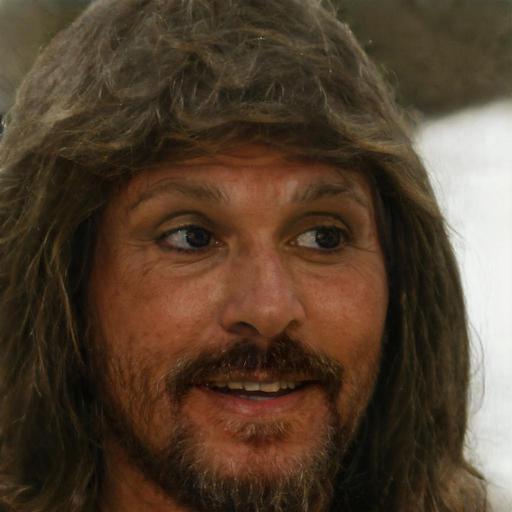}&
        \includegraphics[width=0.16\textwidth]{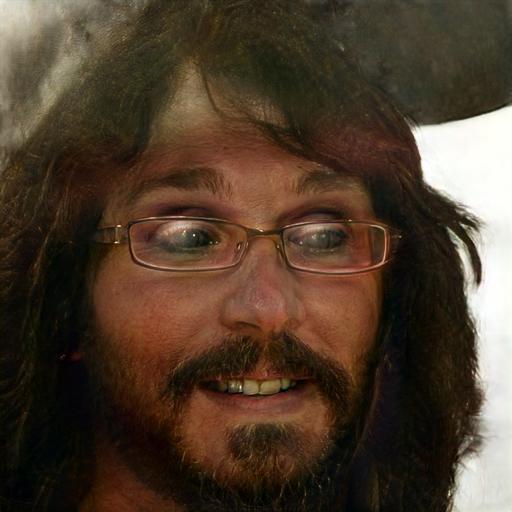}&
        \includegraphics[width=0.16\textwidth]{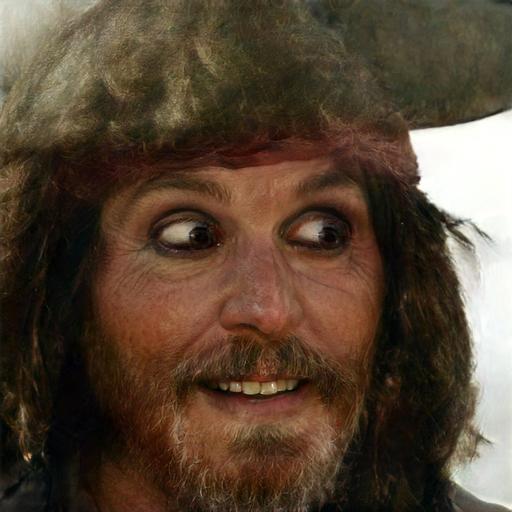} 
		\\
		& Input & SG2 & SG2$\mathcal{W}+$ & e4e & PTI & Ours
    \end{tabular}
    }
	\caption{Reconstruction and editing quality comparison using famous character images collected from the web. In each example, the editing is performed using the same editing weight.}
    \label{fig:appendix5}
\end{figure*}

\begin{figure*}
\setlength{\tabcolsep}{1pt}
\centering
{
	\renewcommand{\arraystretch}{0.5}
    \begin{tabular}{c c c c c c c}
		& Input & SG2 & SG2$\mathcal{W}+$ & e4e & PTI & Ours\\
        \raisebox{0.45in}{\rotatebox[origin=t]{90}{Inversion}}&
        \includegraphics[width=0.16\textwidth]{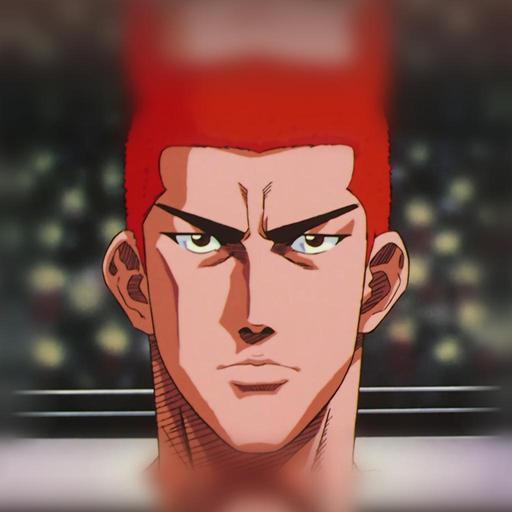}&
        \includegraphics[width=0.16\textwidth]{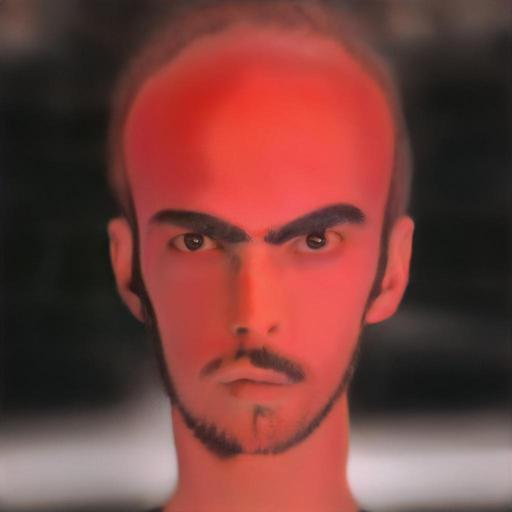}&        
        \includegraphics[width=0.16\textwidth]{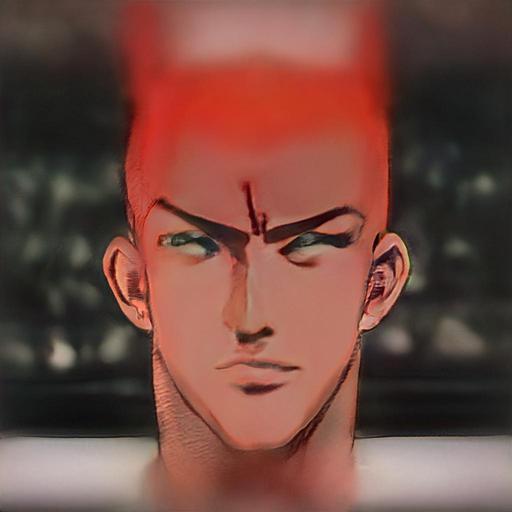}&        
        \includegraphics[width=0.16\textwidth]{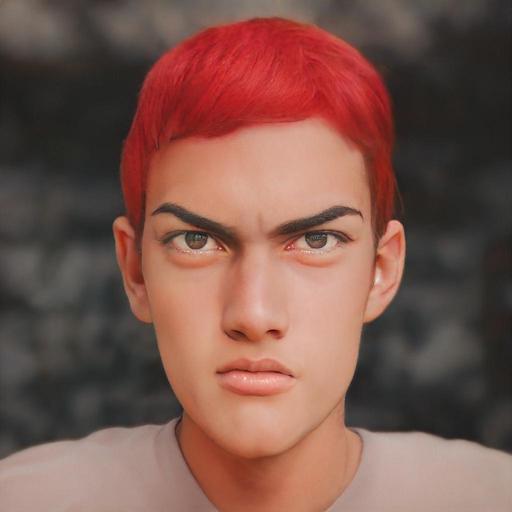}&
        \includegraphics[width=0.16\textwidth]{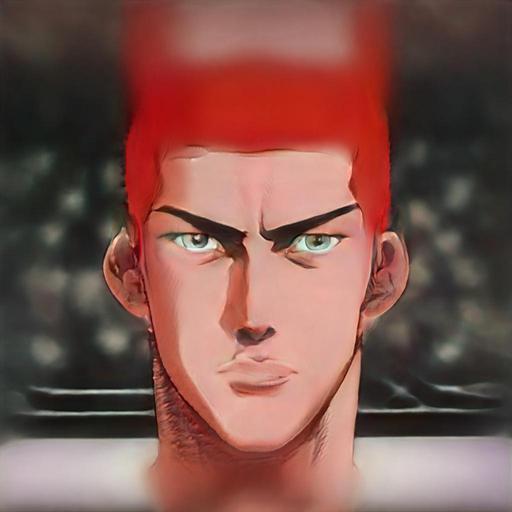}&
        \includegraphics[width=0.16\textwidth]{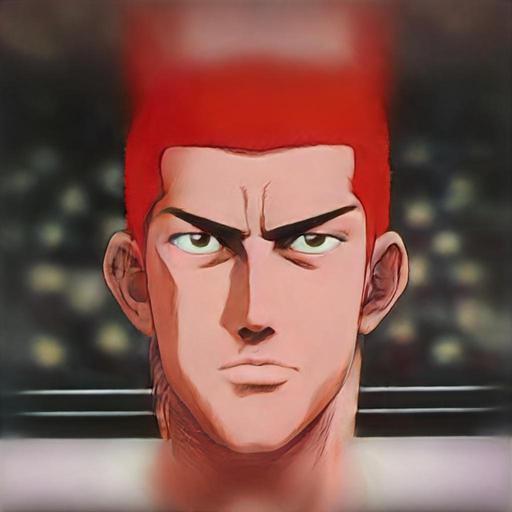} 
		\\
        \raisebox{0.45in}{\rotatebox[origin=t]{90}{Smile}}&&
        \includegraphics[width=0.16\textwidth]{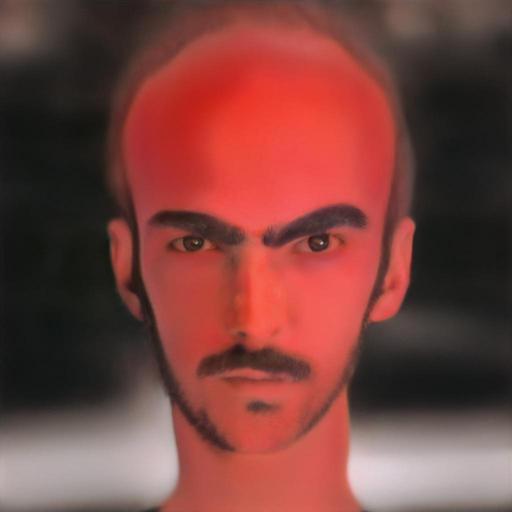}&        
        \includegraphics[width=0.16\textwidth]{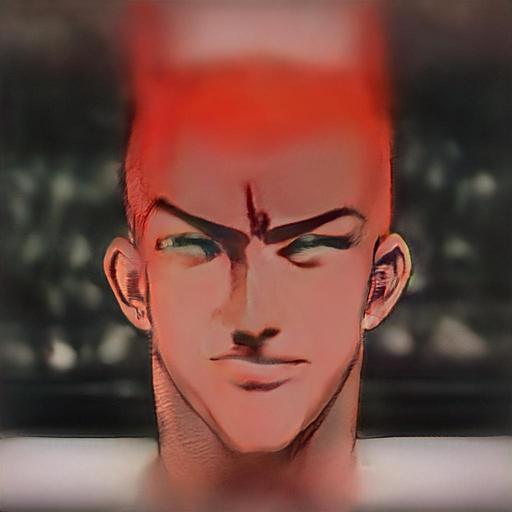}&        
        \includegraphics[width=0.16\textwidth]{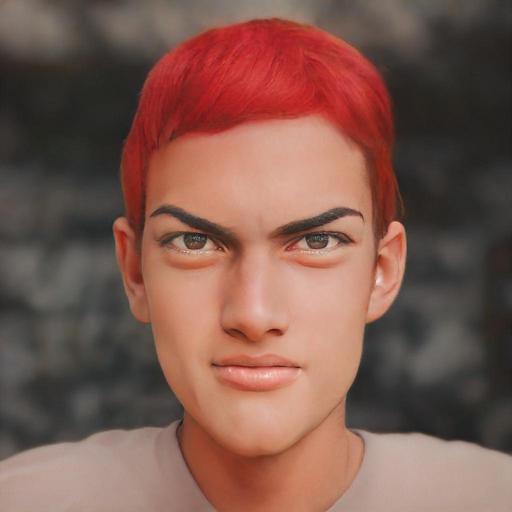}&
        \includegraphics[width=0.16\textwidth]{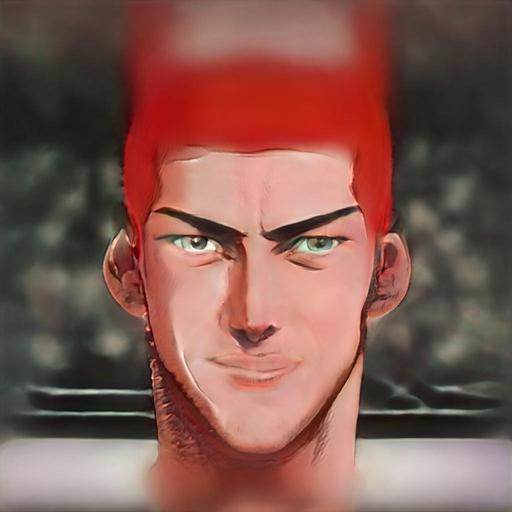}&
        \includegraphics[width=0.16\textwidth]{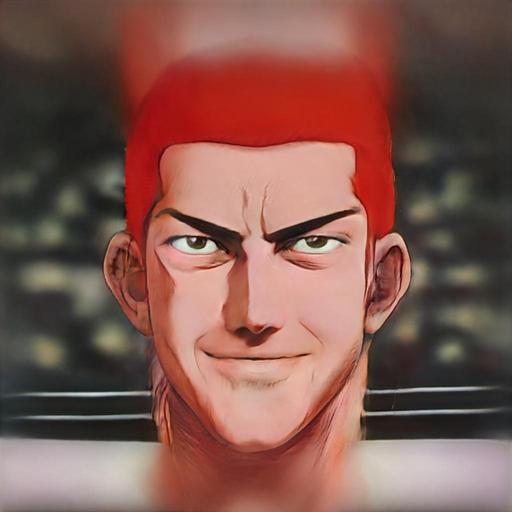} 
		\\[3pt]
        \raisebox{0.45in}{\rotatebox[origin=t]{90}{Inversion}}&
        \includegraphics[width=0.16\textwidth]{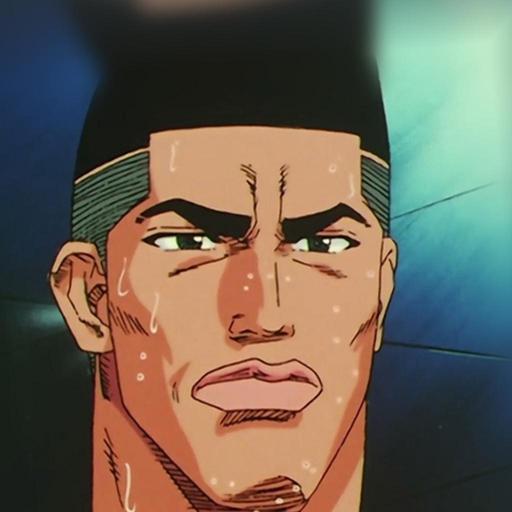}&
        \includegraphics[width=0.16\textwidth]{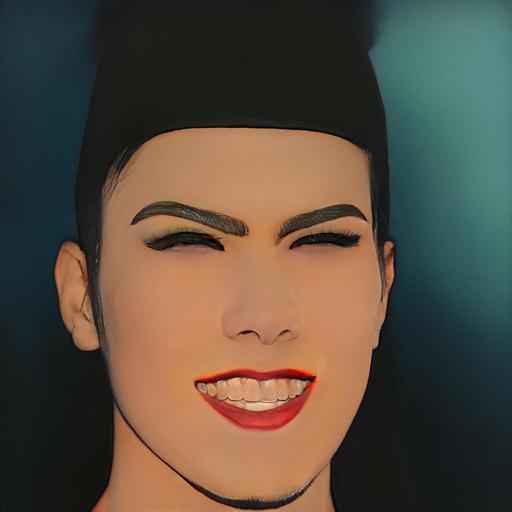}&        
        \includegraphics[width=0.16\textwidth]{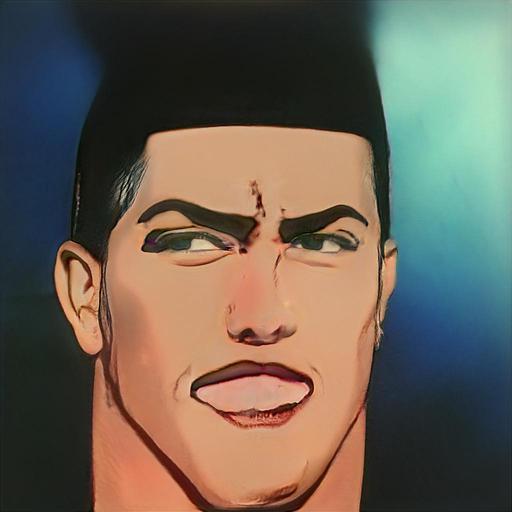}&        
        \includegraphics[width=0.16\textwidth]{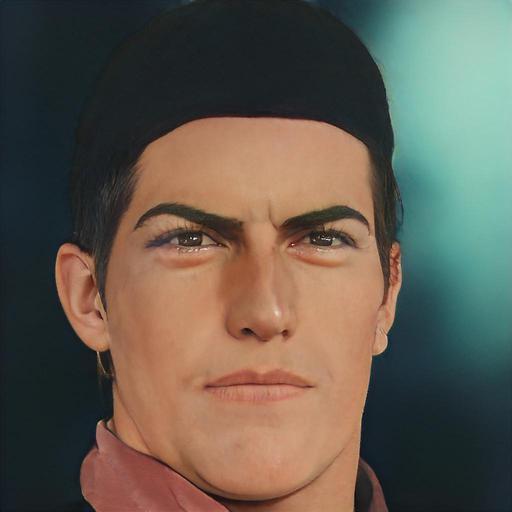}&
        \includegraphics[width=0.16\textwidth]{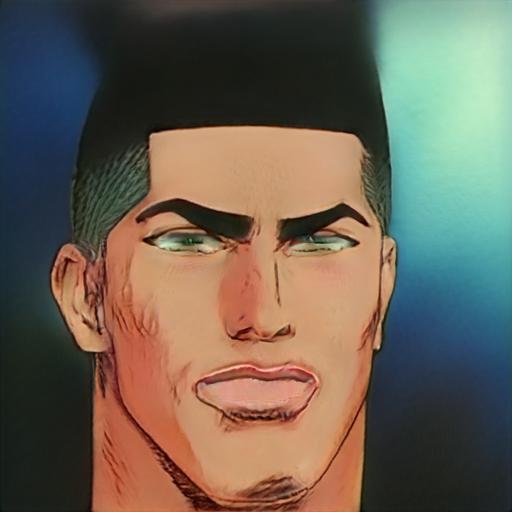}&
        \includegraphics[width=0.16\textwidth]{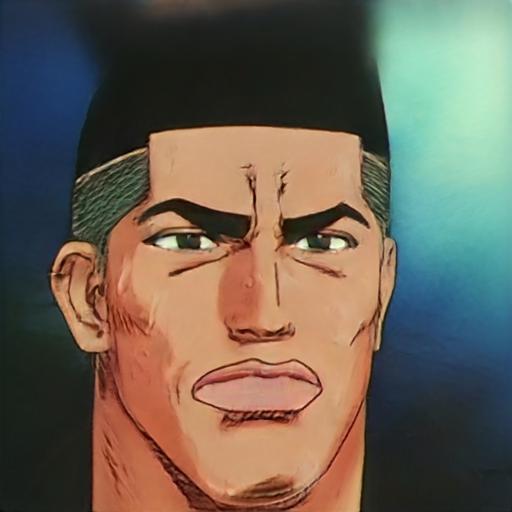} 
		\\
        \raisebox{0.45in}{\rotatebox[origin=t]{90}{Pose}}&&
        \includegraphics[width=0.16\textwidth]{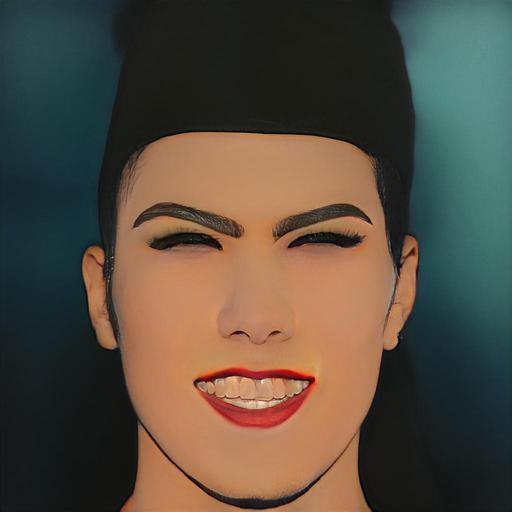}&        
        \includegraphics[width=0.16\textwidth]{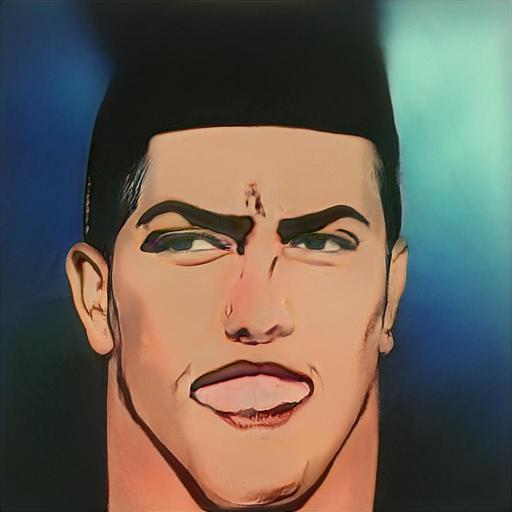}&        
        \includegraphics[width=0.16\textwidth]{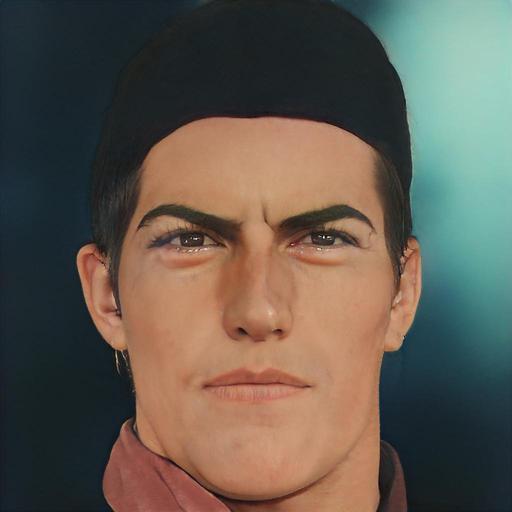}&
        \includegraphics[width=0.16\textwidth]{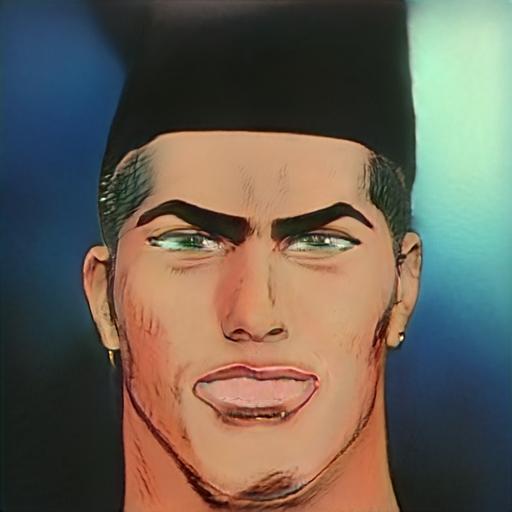}&
        \includegraphics[width=0.16\textwidth]{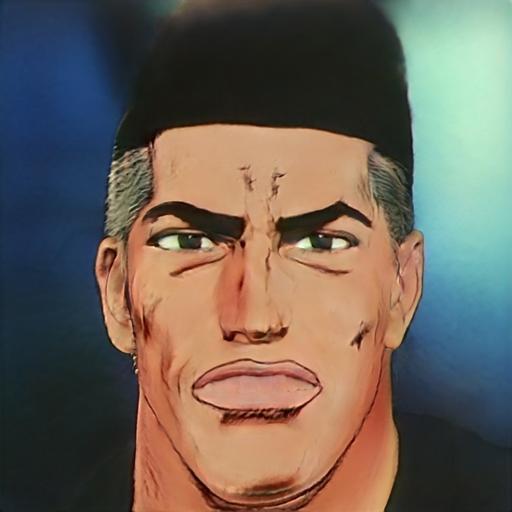} 
		\\[3pt]
        \raisebox{0.45in}{\rotatebox[origin=t]{90}{Inversion}}&
        \includegraphics[width=0.16\textwidth]{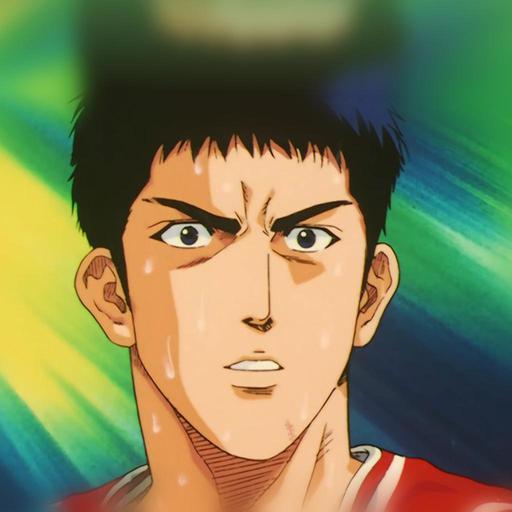}&
        \includegraphics[width=0.16\textwidth]{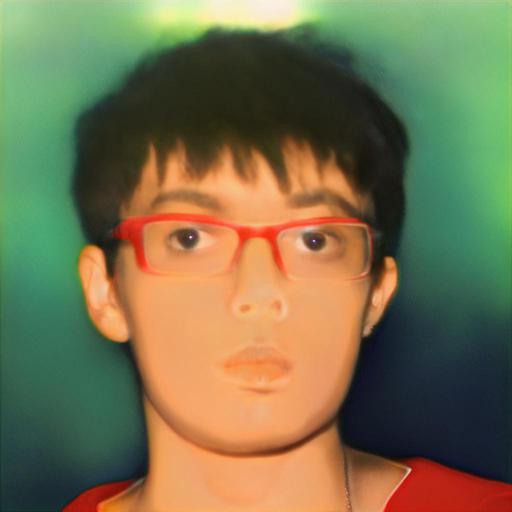}&        
        \includegraphics[width=0.16\textwidth]{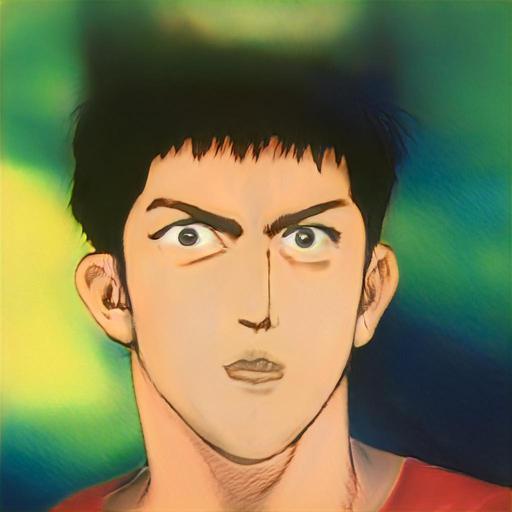}&        
        \includegraphics[width=0.16\textwidth]{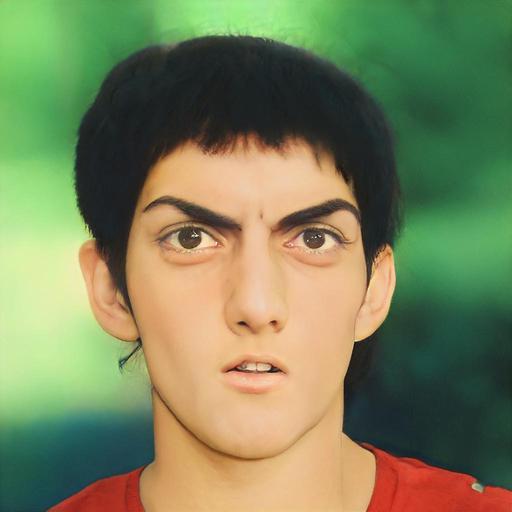}&
        \includegraphics[width=0.16\textwidth]{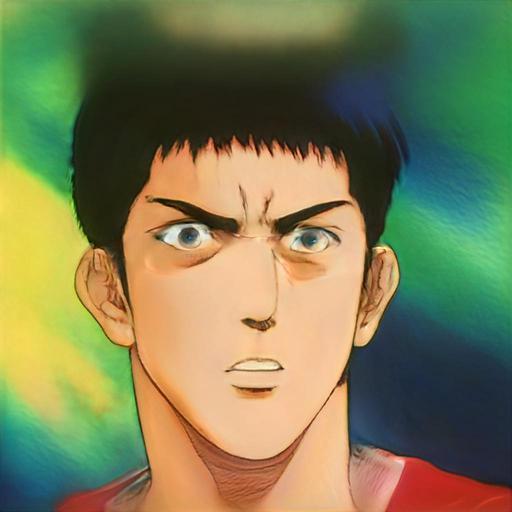}&
        \includegraphics[width=0.16\textwidth]{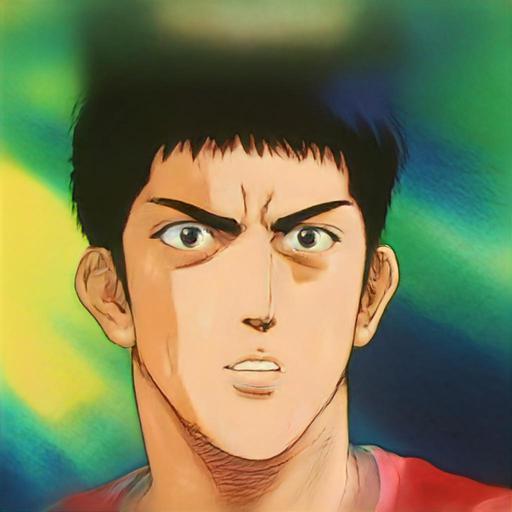} 
		\\
        \raisebox{0.45in}{\rotatebox[origin=t]{90}{Age}}&&
        \includegraphics[width=0.16\textwidth]{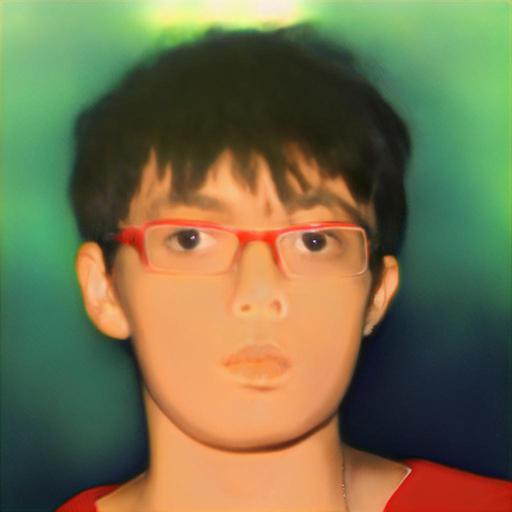}&        
        \includegraphics[width=0.16\textwidth]{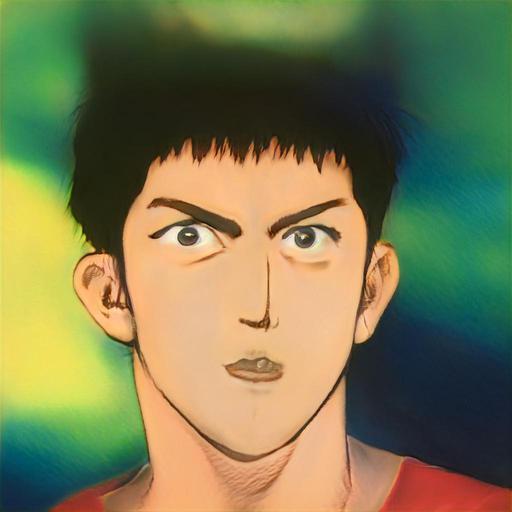}&        
        \includegraphics[width=0.16\textwidth]{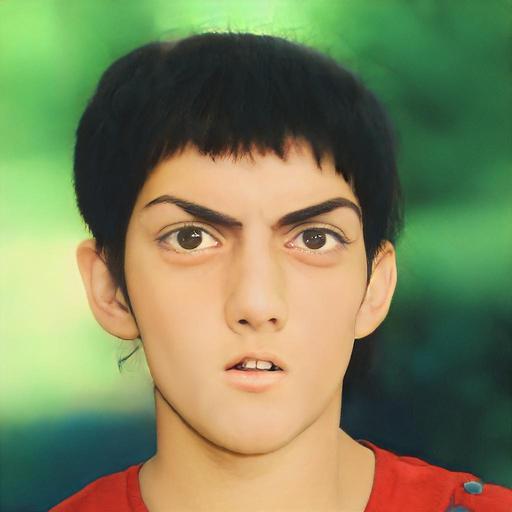}&
        \includegraphics[width=0.16\textwidth]{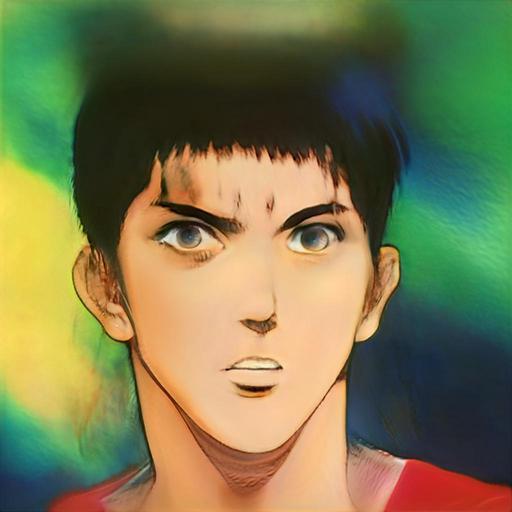}&
        \includegraphics[width=0.16\textwidth]{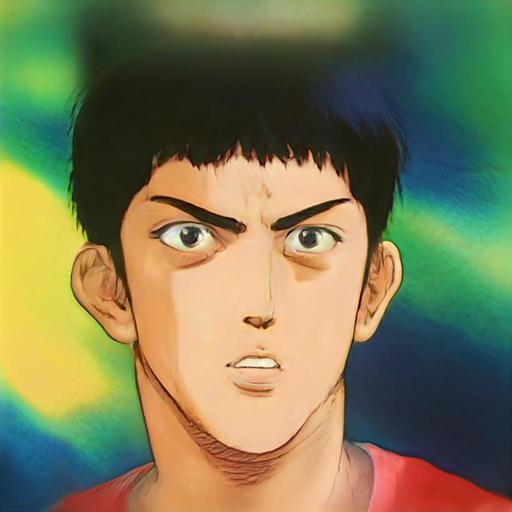} 
		\\
		& Input & SG2 & SG2$\mathcal{W}+$ & e4e & PTI & Ours
    \end{tabular}
    }
	\caption{Reconstruction and editing quality comparison using cartoon images collected from the web. In each example, the editing is performed using the same editing weight.}
    \label{fig:appendix6}
\end{figure*}

\begin{figure*}
\setlength{\tabcolsep}{1pt}
\centering
{
	\renewcommand{\arraystretch}{0.5}
    \begin{tabular}{c c c c c c c}
		& Input & SG2 & SG2$\mathcal{W}+$ & e4e & PTI & Ours\\
        \raisebox{0.45in}{\rotatebox[origin=t]{90}{Inversion}}&
        \includegraphics[width=0.16\textwidth]{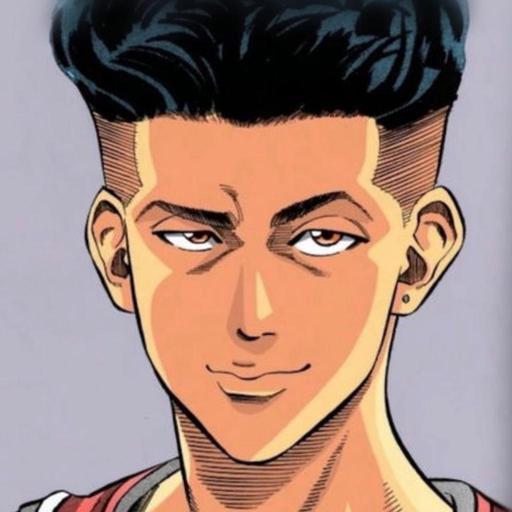}&
        \includegraphics[width=0.16\textwidth]{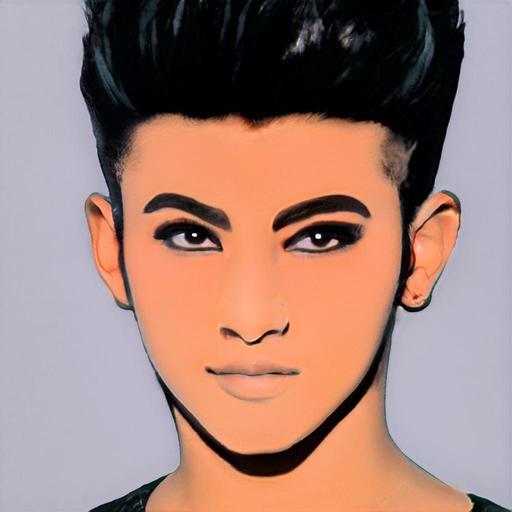}&        
        \includegraphics[width=0.16\textwidth]{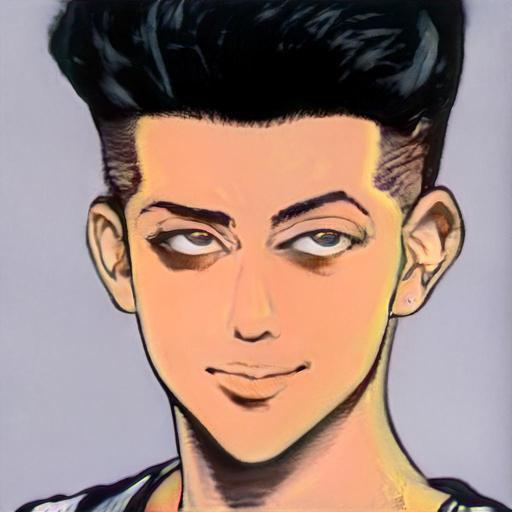}&        
        \includegraphics[width=0.16\textwidth]{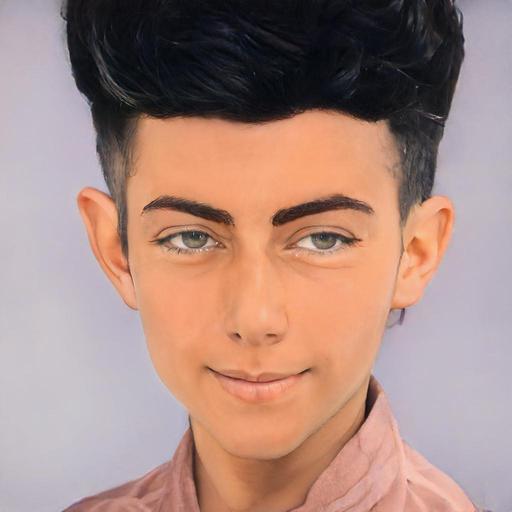}&
        \includegraphics[width=0.16\textwidth]{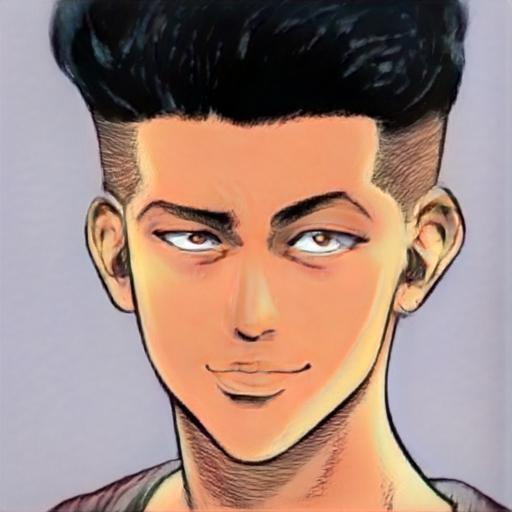}&
        \includegraphics[width=0.16\textwidth]{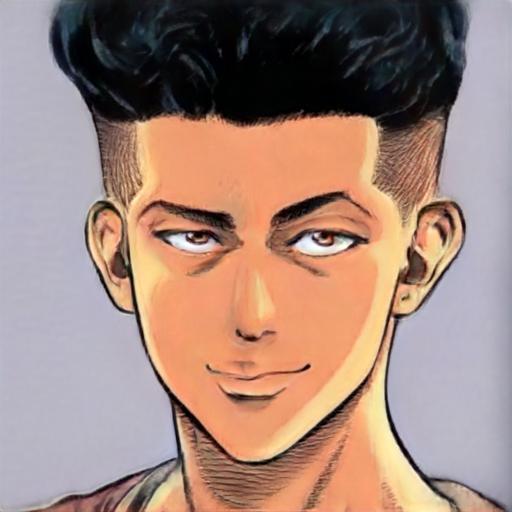} 
		\\
        \raisebox{0.45in}{\rotatebox[origin=t]{90}{Smile}}&&
        \includegraphics[width=0.16\textwidth]{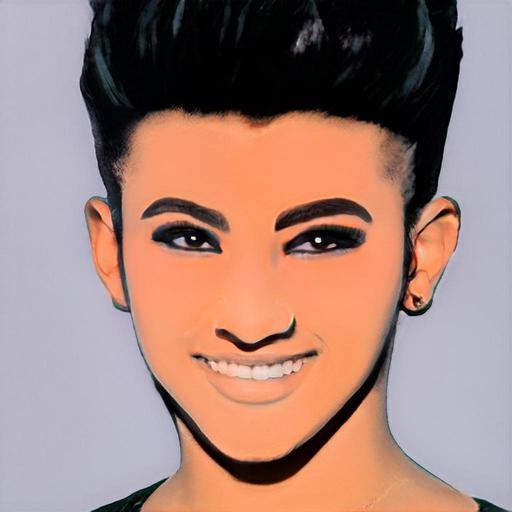}&        
        \includegraphics[width=0.16\textwidth]{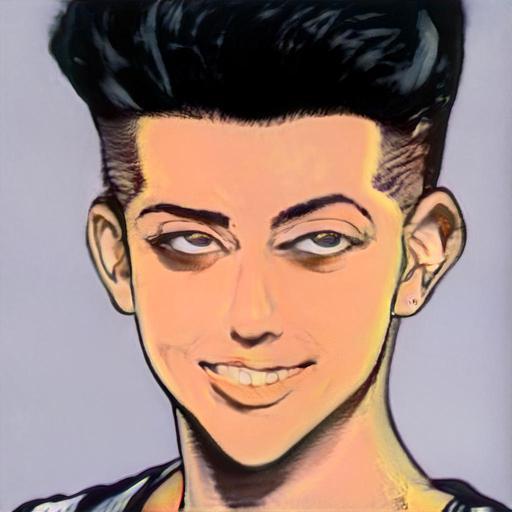}&        
        \includegraphics[width=0.16\textwidth]{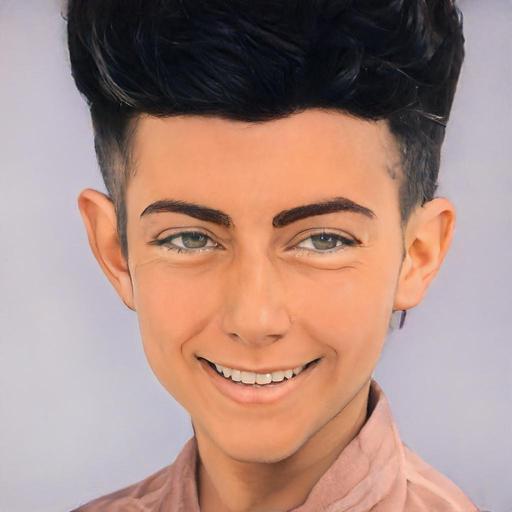}&
        \includegraphics[width=0.16\textwidth]{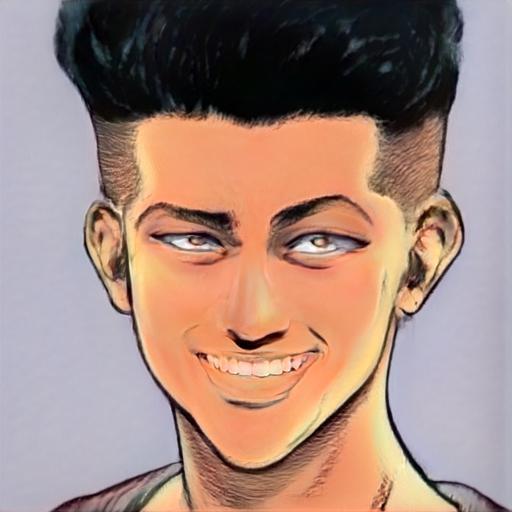}&
        \includegraphics[width=0.16\textwidth]{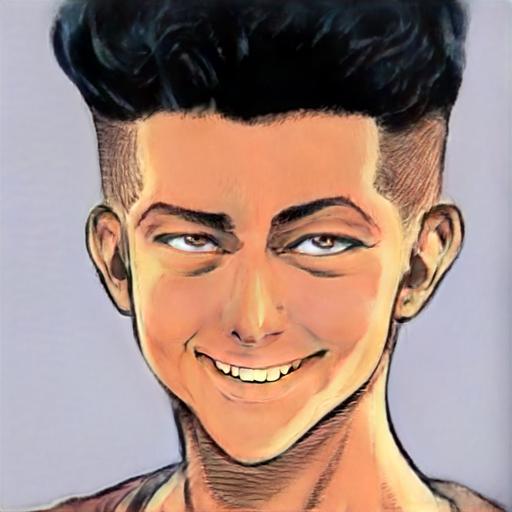} 
		\\[3pt]
        \raisebox{0.45in}{\rotatebox[origin=t]{90}{Inversion}}&
        \includegraphics[width=0.16\textwidth]{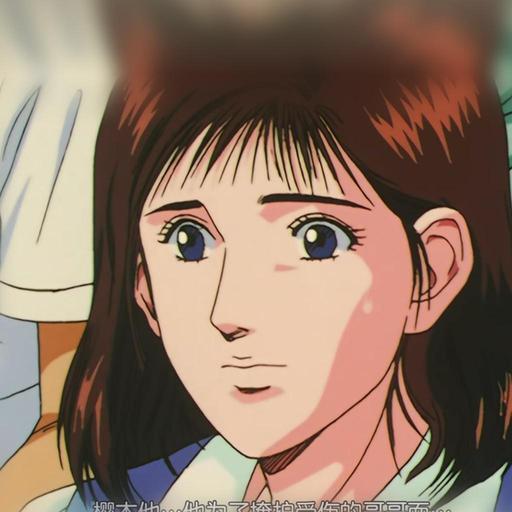}&
        \includegraphics[width=0.16\textwidth]{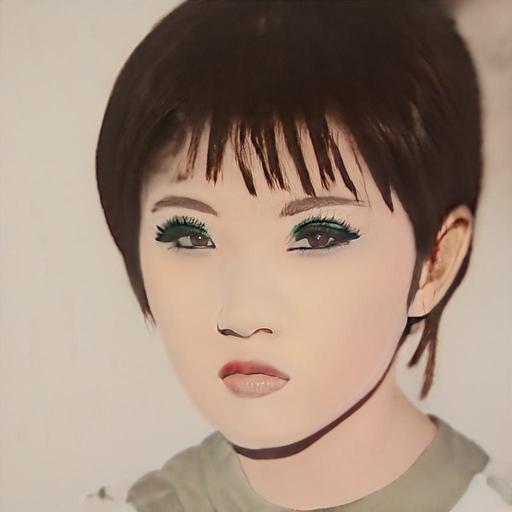}&        
        \includegraphics[width=0.16\textwidth]{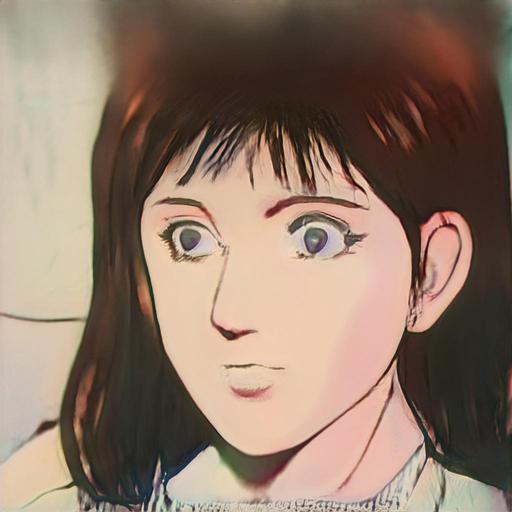}&        
        \includegraphics[width=0.16\textwidth]{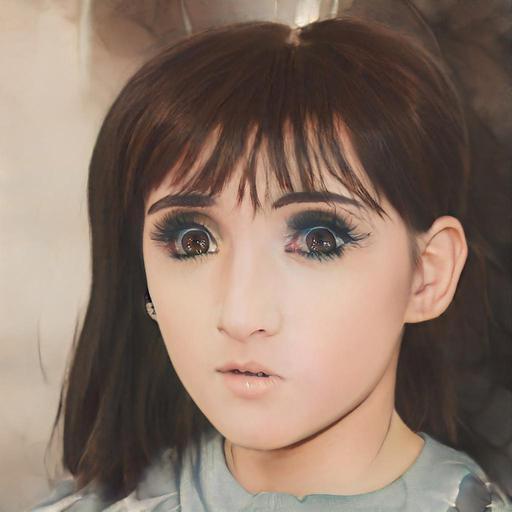}&
        \includegraphics[width=0.16\textwidth]{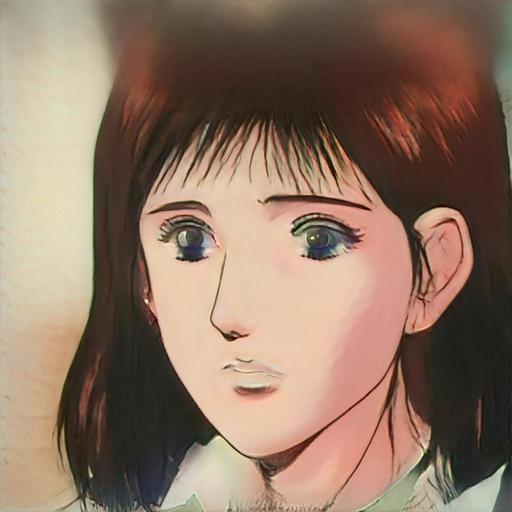}&
        \includegraphics[width=0.16\textwidth]{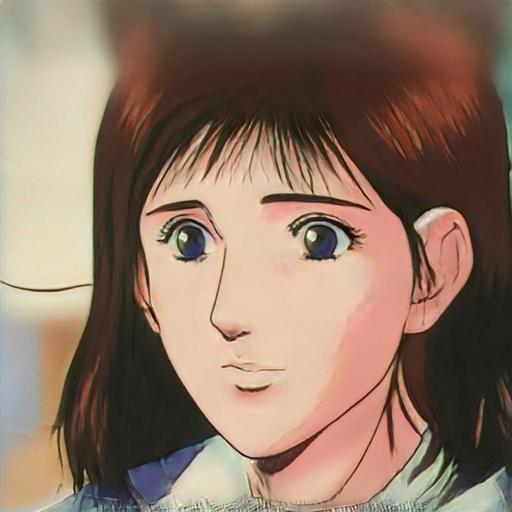} 
		\\
        \raisebox{0.45in}{\rotatebox[origin=t]{90}{Pose}}&&
        \includegraphics[width=0.16\textwidth]{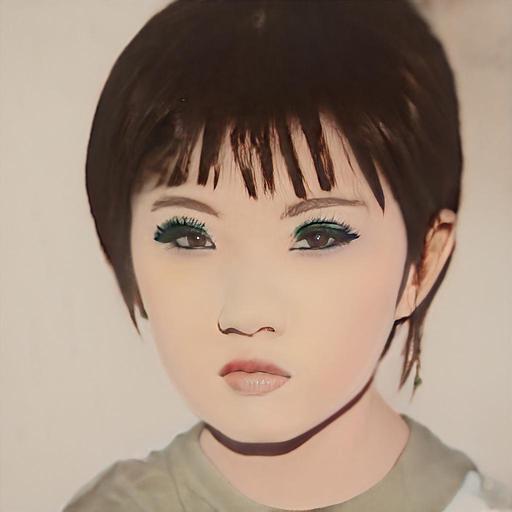}&        
        \includegraphics[width=0.16\textwidth]{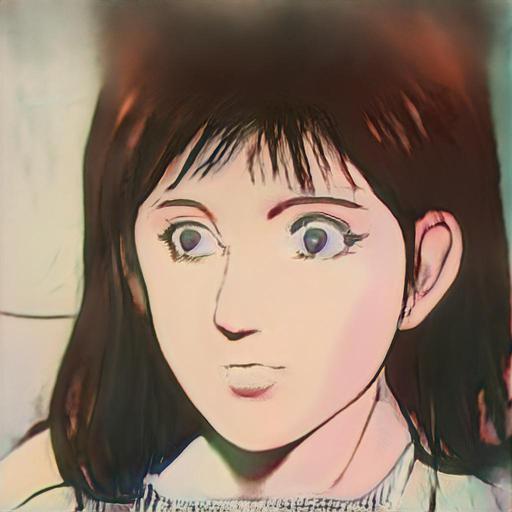}&        
        \includegraphics[width=0.16\textwidth]{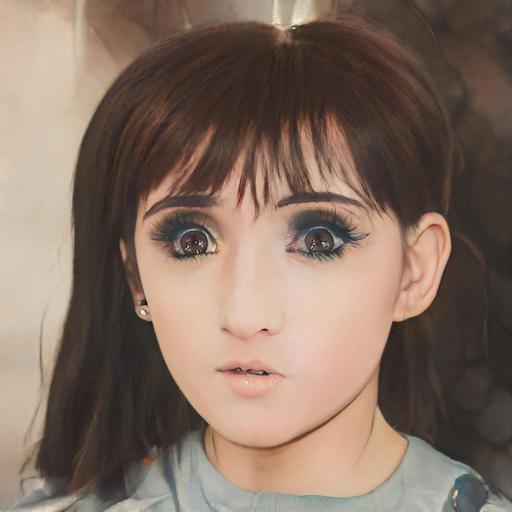}&
        \includegraphics[width=0.16\textwidth]{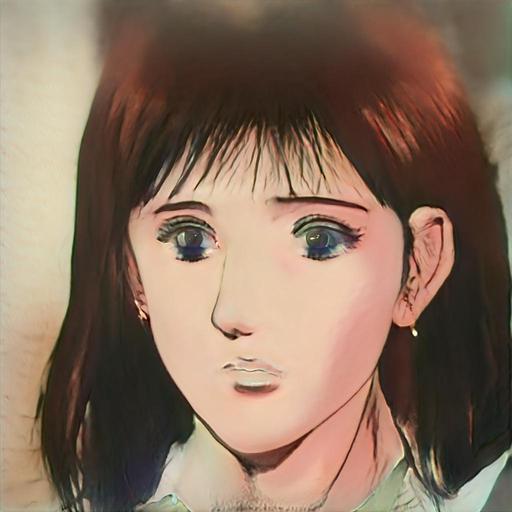}&
        \includegraphics[width=0.16\textwidth]{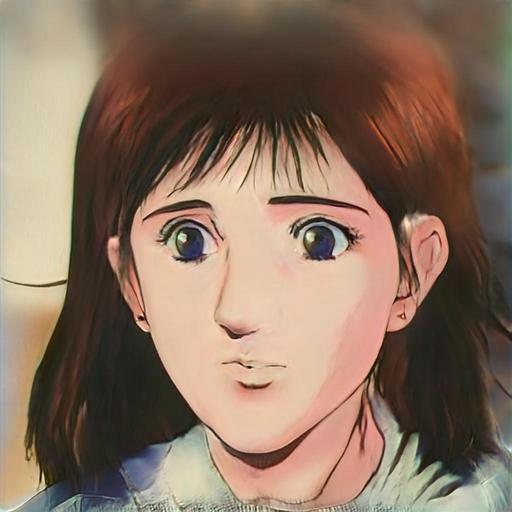} 
		\\[3pt]
        \raisebox{0.45in}{\rotatebox[origin=t]{90}{Inversion}}&
        \includegraphics[width=0.16\textwidth]{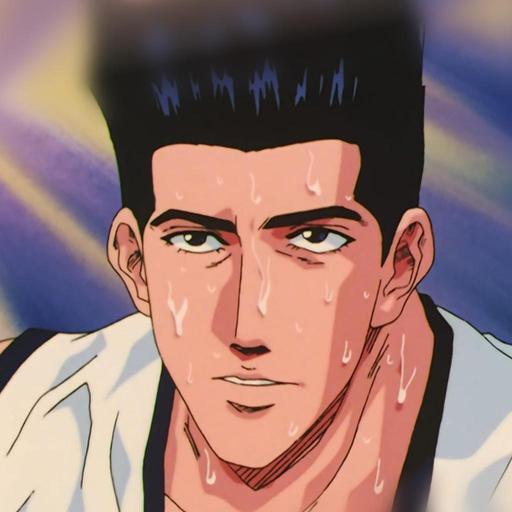}&
        \includegraphics[width=0.16\textwidth]{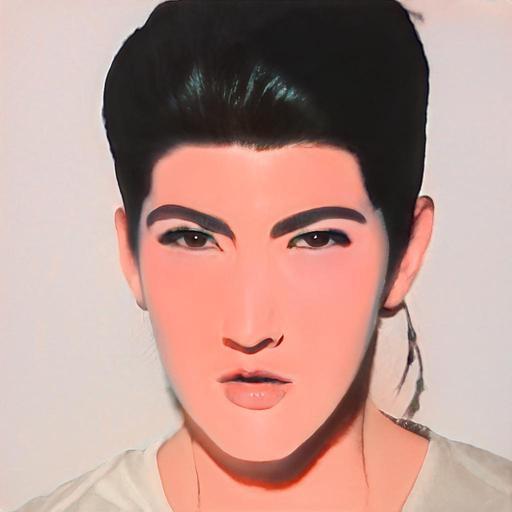}&        
        \includegraphics[width=0.16\textwidth]{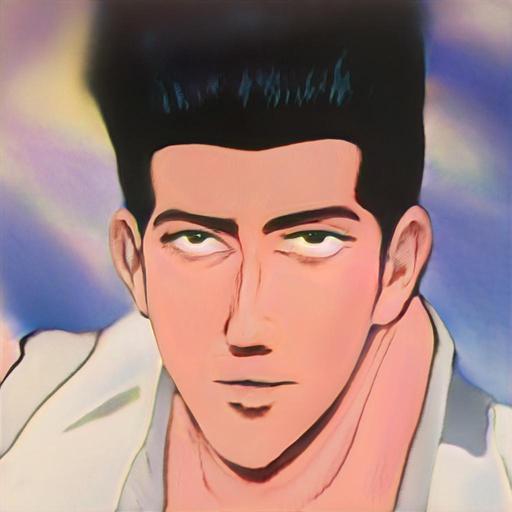}&        
        \includegraphics[width=0.16\textwidth]{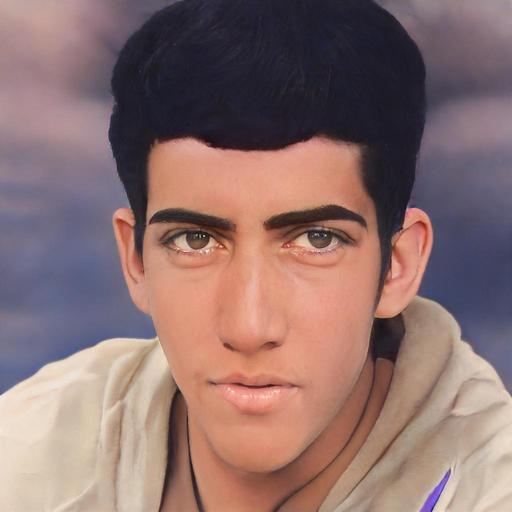}&
        \includegraphics[width=0.16\textwidth]{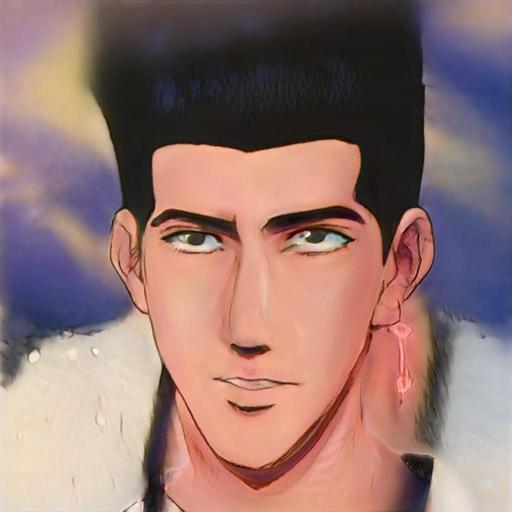}&
        \includegraphics[width=0.16\textwidth]{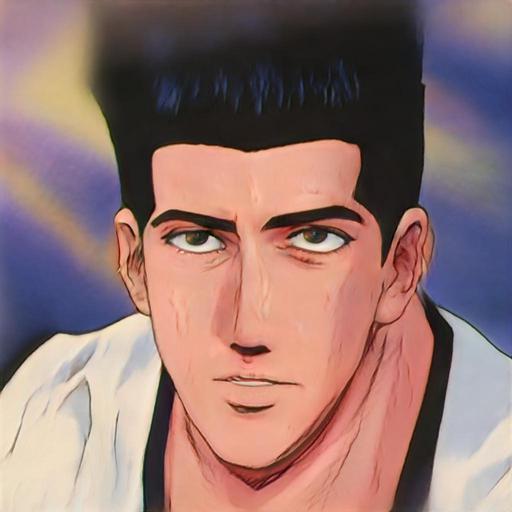} 
		\\
        \raisebox{0.45in}{\rotatebox[origin=t]{90}{Age}}&&
        \includegraphics[width=0.16\textwidth]{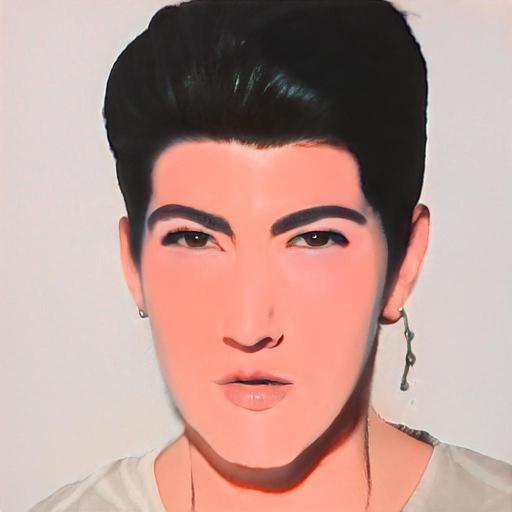}&        
        \includegraphics[width=0.16\textwidth]{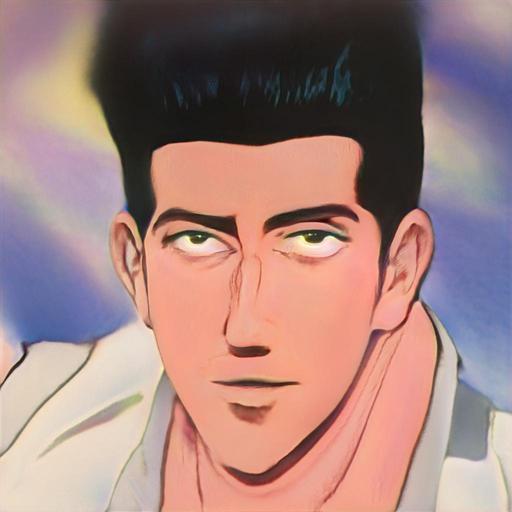}&        
        \includegraphics[width=0.16\textwidth]{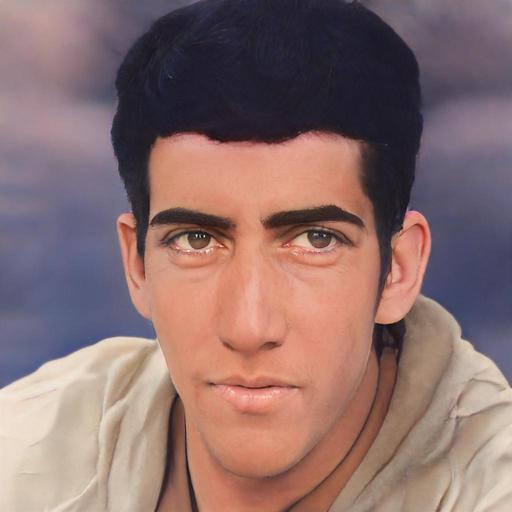}&
        \includegraphics[width=0.16\textwidth]{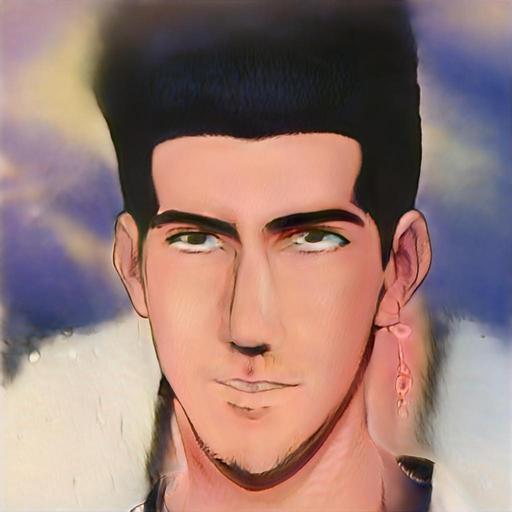}&
        \includegraphics[width=0.16\textwidth]{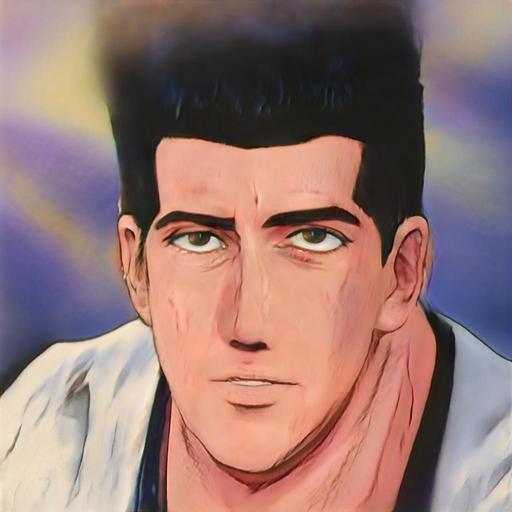} 
		\\
		& Input & SG2 & SG2$\mathcal{W}+$ & e4e & PTI & Ours
    \end{tabular}
    }
	\caption{Reconstruction and editing quality comparison using cartoon images collected from the web. In each example, the editing is performed using the same editing weight.}
    \label{fig:appendix7}
\end{figure*}

\begin{figure*}
\setlength{\tabcolsep}{1pt}
\centering
{
	\renewcommand{\arraystretch}{0.5}
    \begin{tabular}{c c c c c c c}
		& Input & SG2 & SG2$\mathcal{W}+$ & e4e & PTI & Ours\\
        \raisebox{0.45in}{\rotatebox[origin=t]{90}{Inversion}}&
        \includegraphics[width=0.16\textwidth]{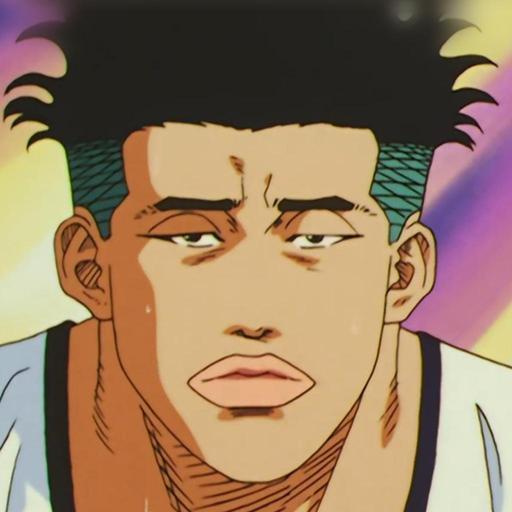}&
        \includegraphics[width=0.16\textwidth]{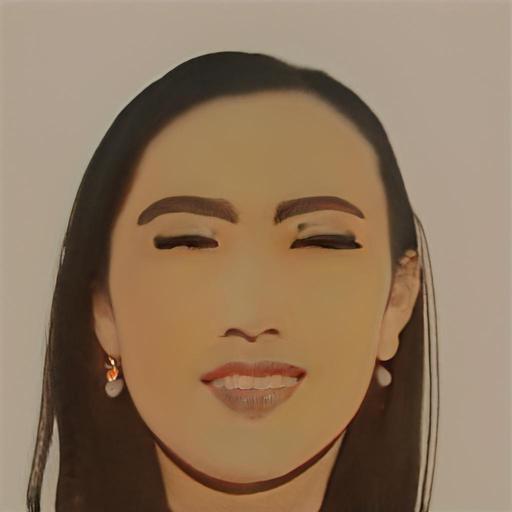}&        
        \includegraphics[width=0.16\textwidth]{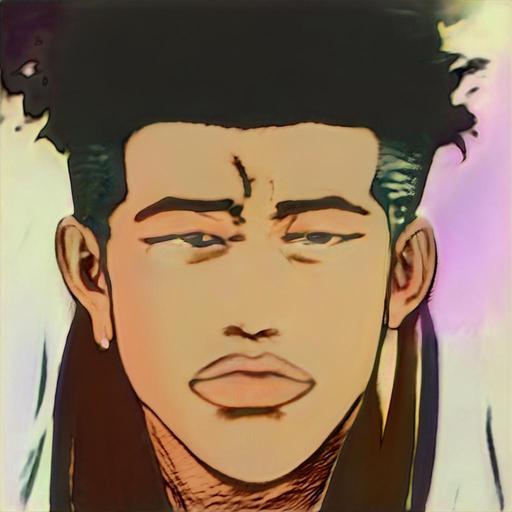}&        
        \includegraphics[width=0.16\textwidth]{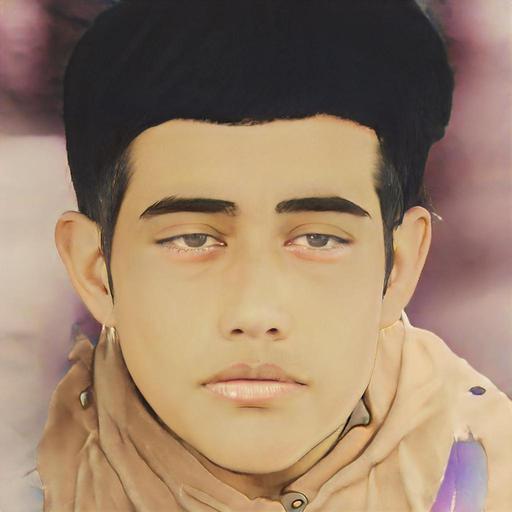}&
        \includegraphics[width=0.16\textwidth]{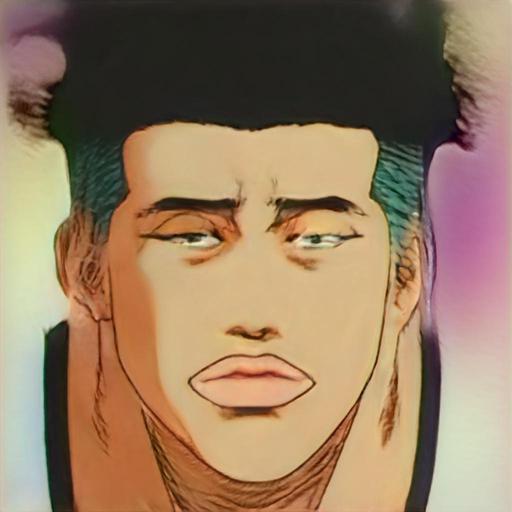}&
        \includegraphics[width=0.16\textwidth]{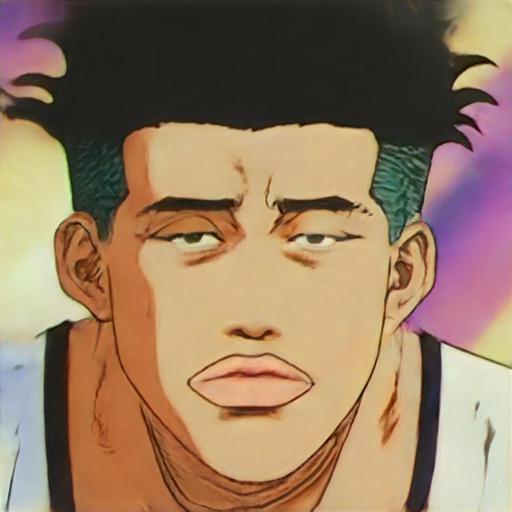} 
		\\
        \raisebox{0.45in}{\rotatebox[origin=t]{90}{Smile}}&&
        \includegraphics[width=0.16\textwidth]{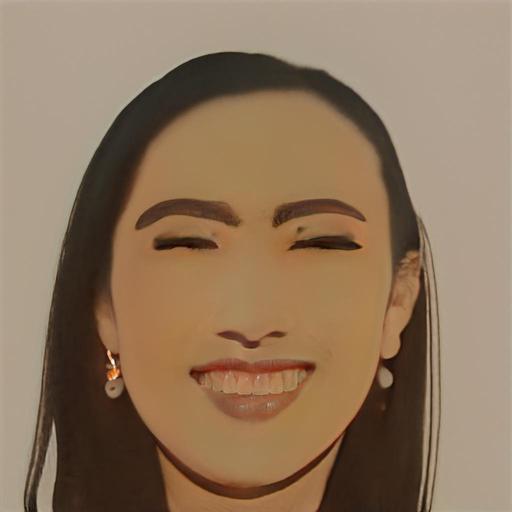}&        
        \includegraphics[width=0.16\textwidth]{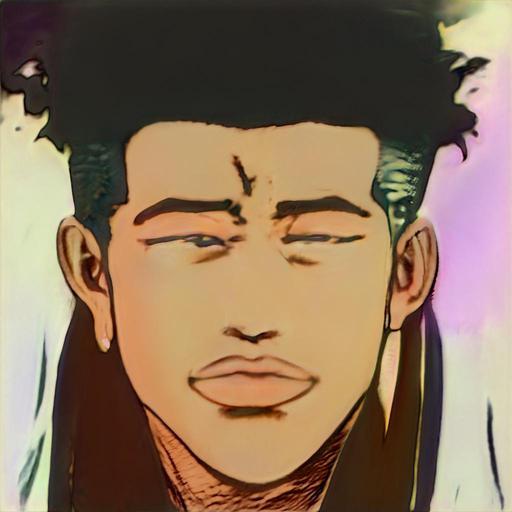}&        
        \includegraphics[width=0.16\textwidth]{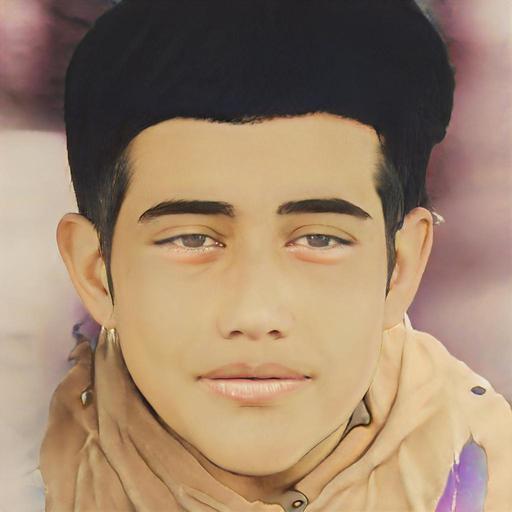}&
        \includegraphics[width=0.16\textwidth]{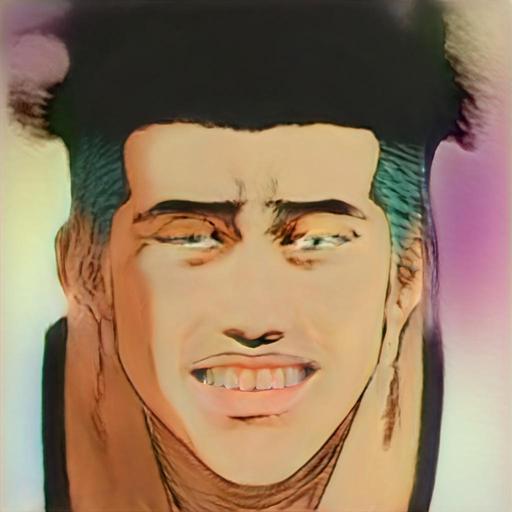}&
        \includegraphics[width=0.16\textwidth]{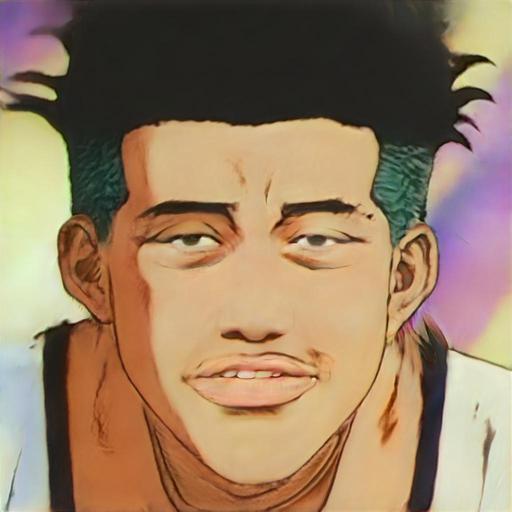} 
		\\[3pt]
        \raisebox{0.45in}{\rotatebox[origin=t]{90}{Inversion}}&
        \includegraphics[width=0.16\textwidth]{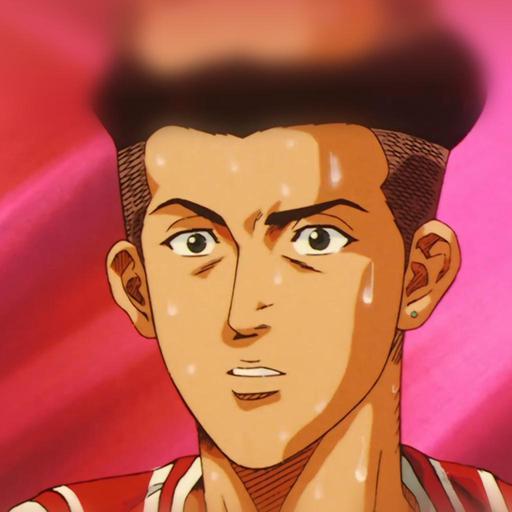}&
        \includegraphics[width=0.16\textwidth]{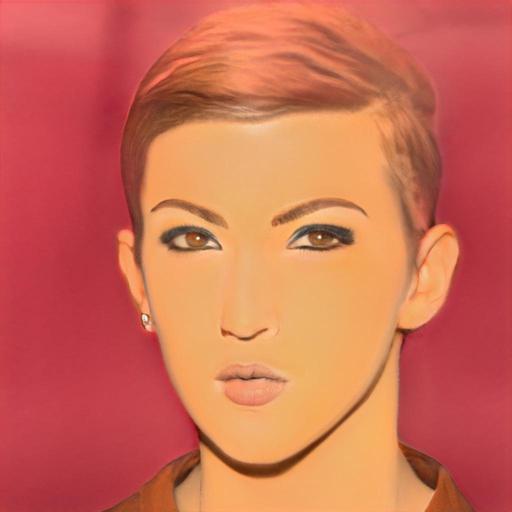}&        
        \includegraphics[width=0.16\textwidth]{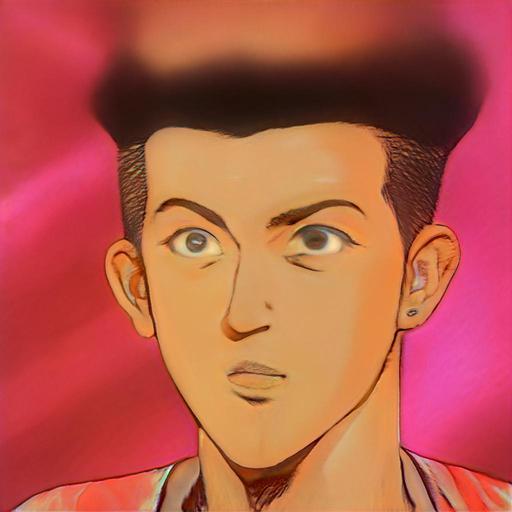}&        
        \includegraphics[width=0.16\textwidth]{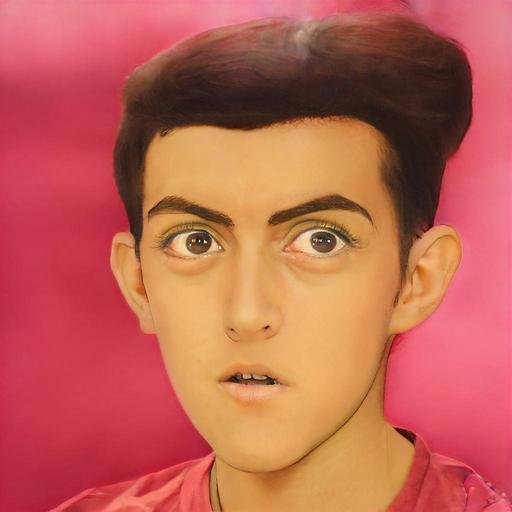}&
        \includegraphics[width=0.16\textwidth]{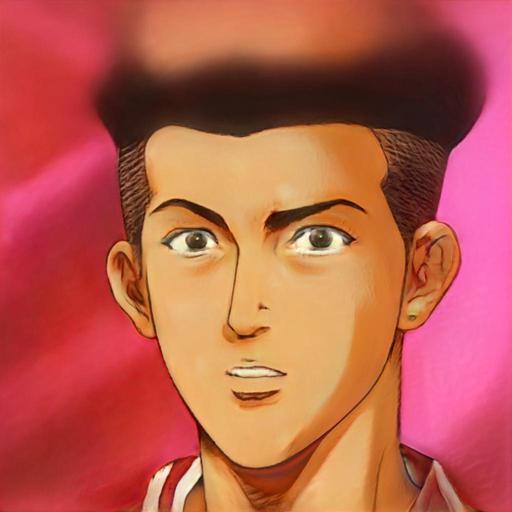}&
        \includegraphics[width=0.16\textwidth]{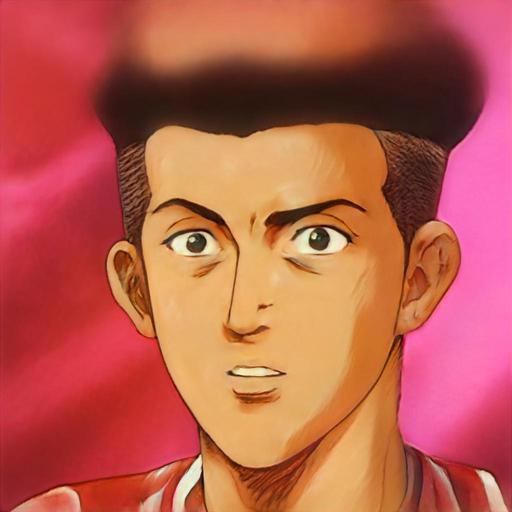} 
		\\
        \raisebox{0.45in}{\rotatebox[origin=t]{90}{Pose}}&&
        \includegraphics[width=0.16\textwidth]{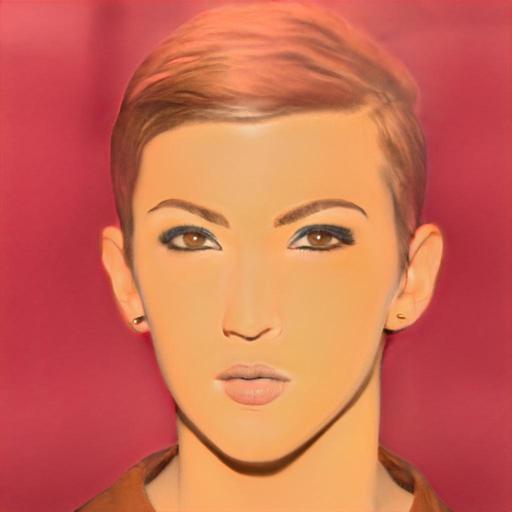}&        
        \includegraphics[width=0.16\textwidth]{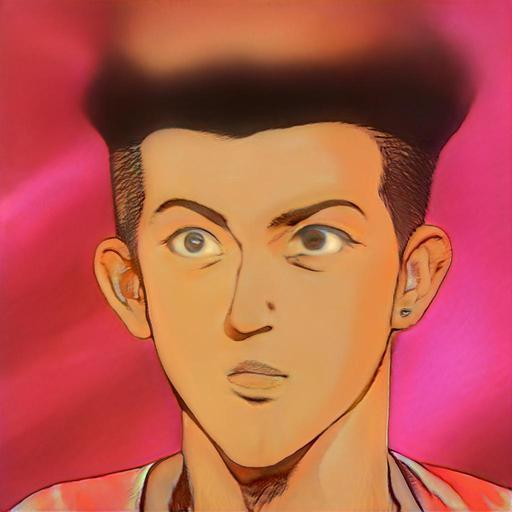}&        
        \includegraphics[width=0.16\textwidth]{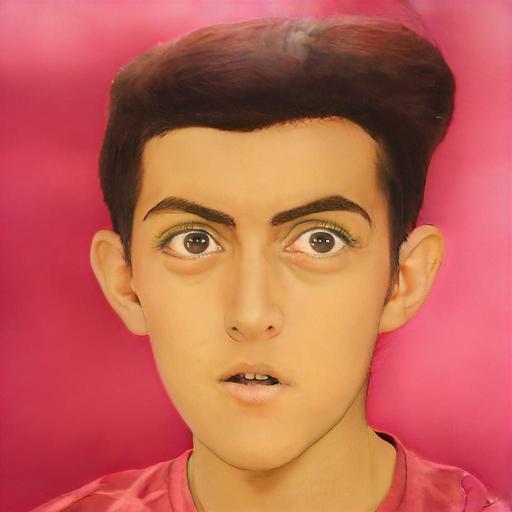}&
        \includegraphics[width=0.16\textwidth]{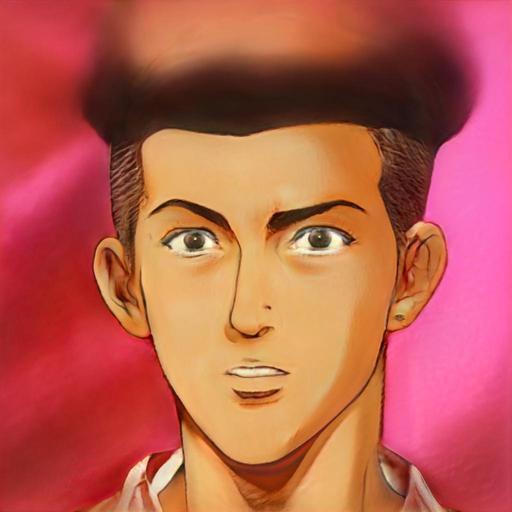}&
        \includegraphics[width=0.16\textwidth]{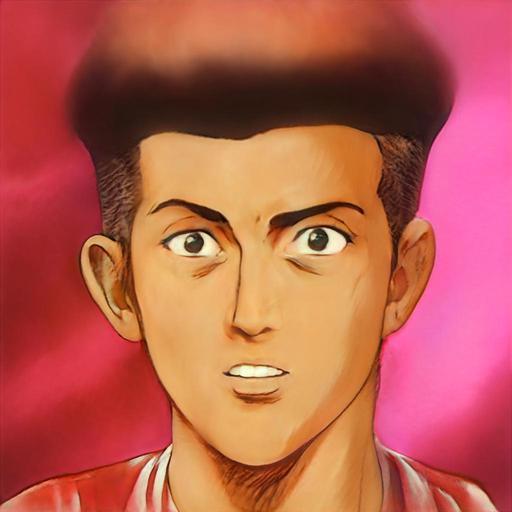} 
		\\[3pt]
        \raisebox{0.45in}{\rotatebox[origin=t]{90}{Inversion}}&
        \includegraphics[width=0.16\textwidth]{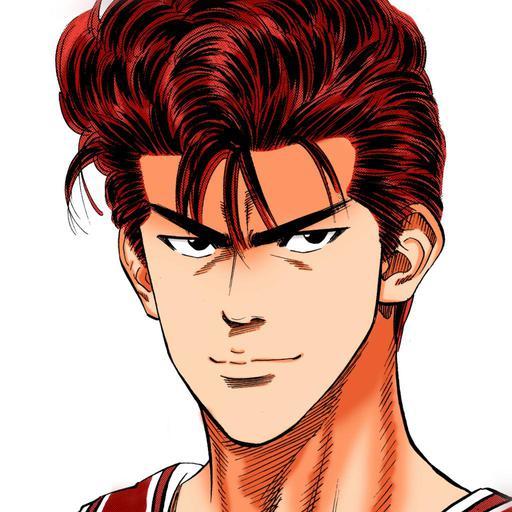}&
        \includegraphics[width=0.16\textwidth]{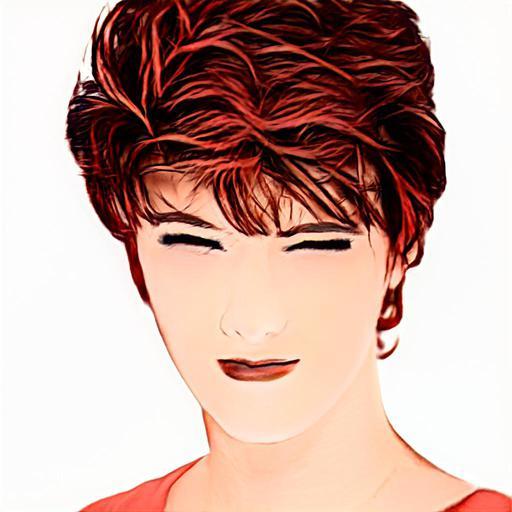}&        
        \includegraphics[width=0.16\textwidth]{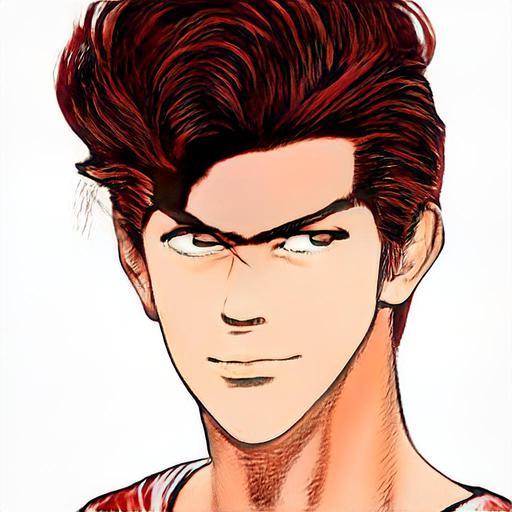}&        
        \includegraphics[width=0.16\textwidth]{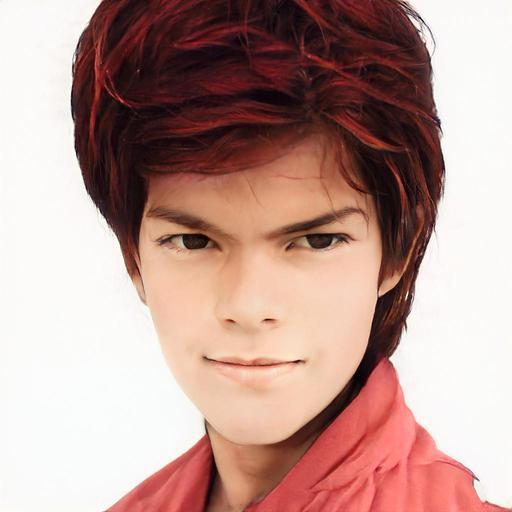}&
        \includegraphics[width=0.16\textwidth]{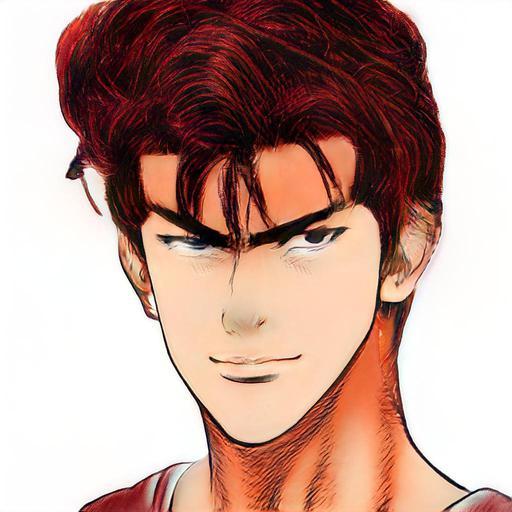}&
        \includegraphics[width=0.16\textwidth]{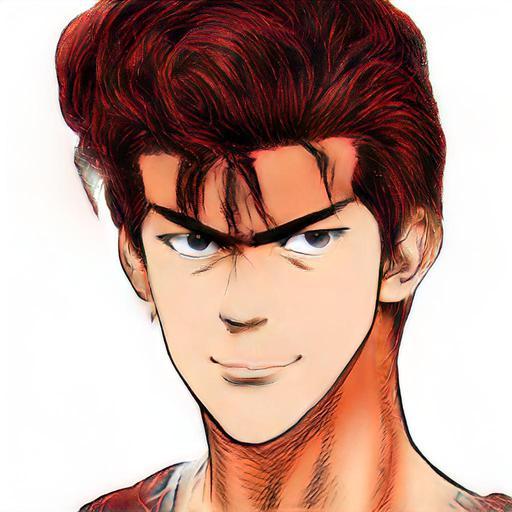} 
		\\
        \raisebox{0.45in}{\rotatebox[origin=t]{90}{Age}}&&
        \includegraphics[width=0.16\textwidth]{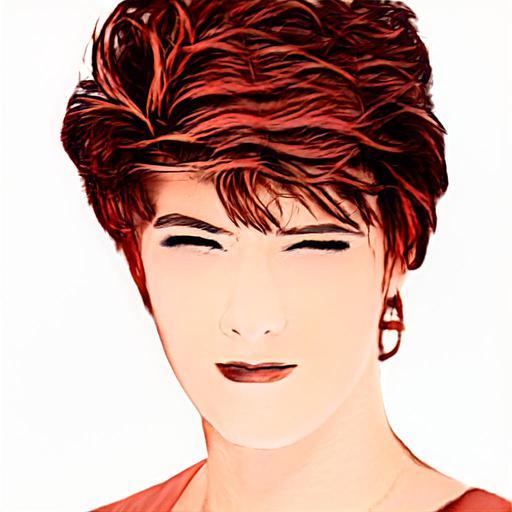}&        
        \includegraphics[width=0.16\textwidth]{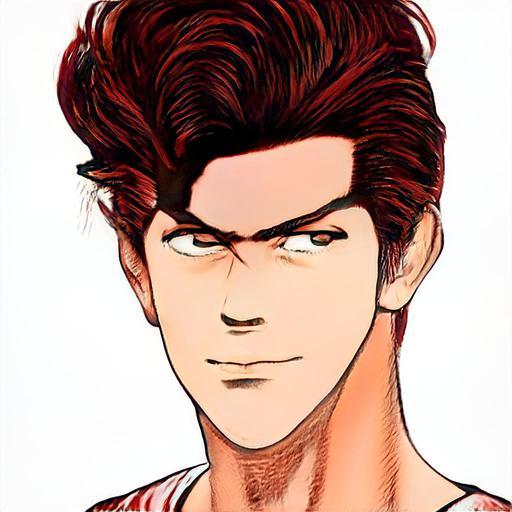}&        
        \includegraphics[width=0.16\textwidth]{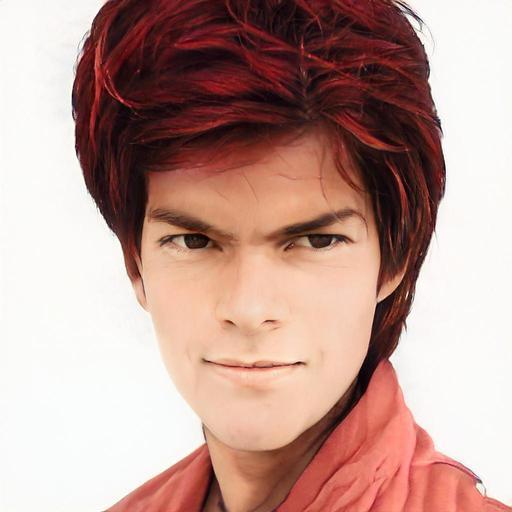}&
        \includegraphics[width=0.16\textwidth]{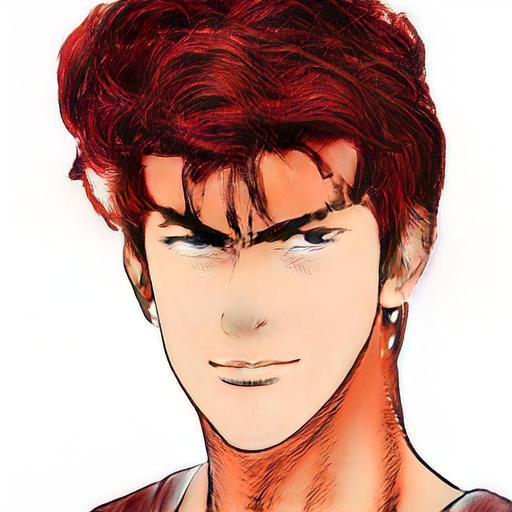}&
        \includegraphics[width=0.16\textwidth]{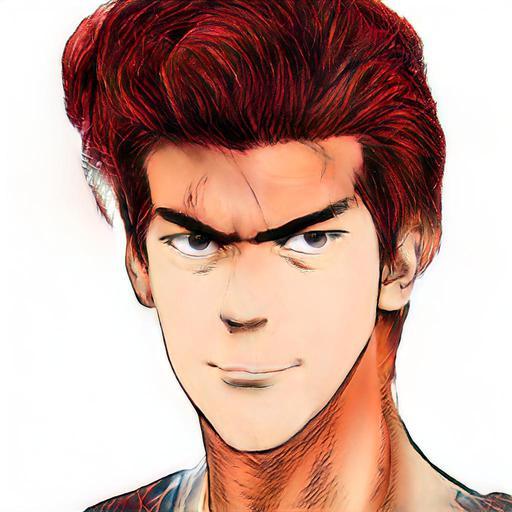} 
		\\
		& Input & SG2 & SG2$\mathcal{W}+$ & e4e & PTI & Ours
    \end{tabular}
    }
	\caption{Reconstruction and editing quality comparison using cartoon images collected from the web. In each example, the editing is performed using the same editing weight.}
    \label{fig:appendix8}
\end{figure*}

Figures \ref{fig:appendix0} to \ref{fig:appendix8} present additional reconstruction and editing results using images from CelebA-HQ test set, famous character images, and cartoon images. 

\section{More Editing Directions}
\begin{figure*}
\setlength{\tabcolsep}{1pt}
\centering
{
    \begin{tabular}{c c c c c c}
        \raisebox{0.5in}{\rotatebox[origin=t]{90}{Input}}&
        \includegraphics[width=0.185\textwidth]{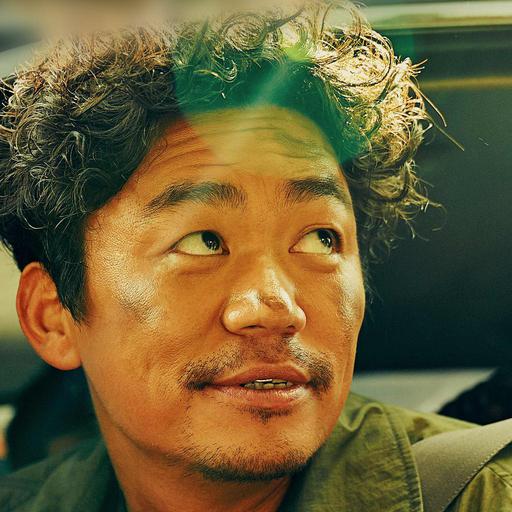}&
        \includegraphics[width=0.185\textwidth]{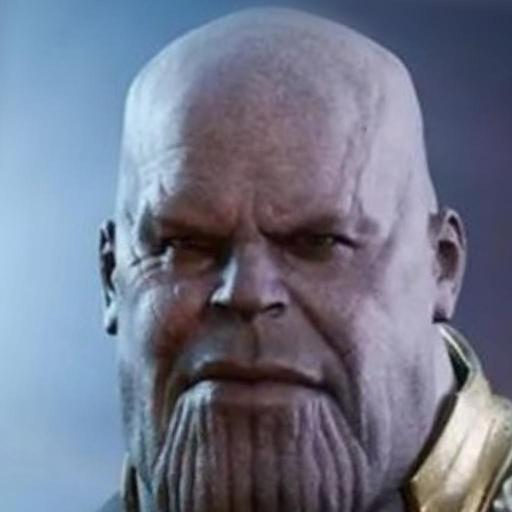}&        
        \includegraphics[width=0.185\textwidth]{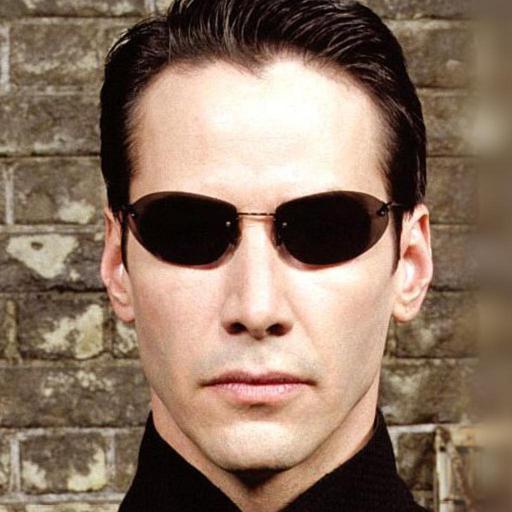}&
        \includegraphics[width=0.185\textwidth]{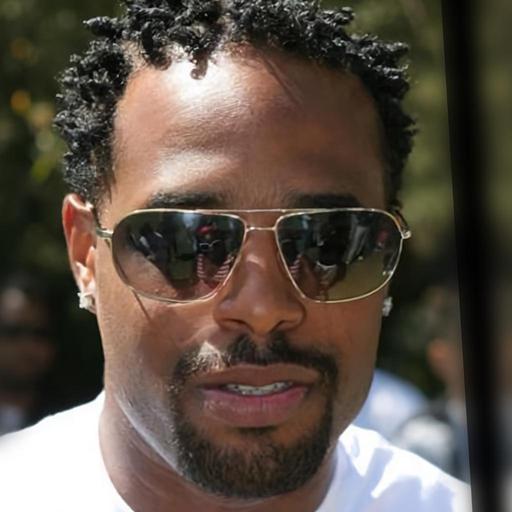}&
        \includegraphics[width=0.185\textwidth]{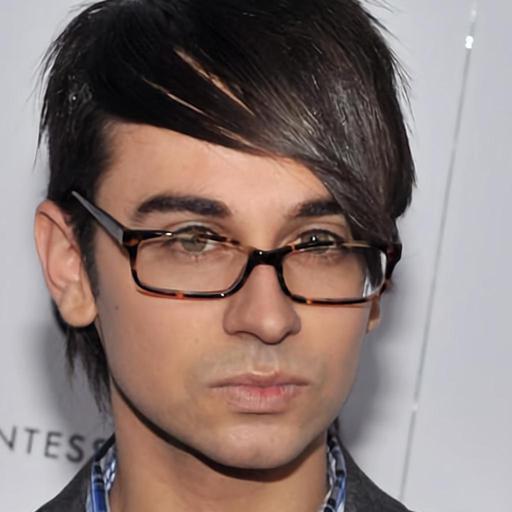} 
		\tabularnewline
        \raisebox{0.5in}{\rotatebox[origin=t]{90}{PTI}}&
        \includegraphics[width=0.185\textwidth]{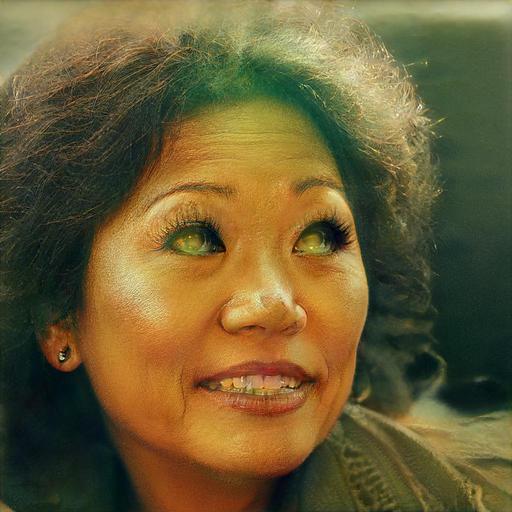}&
        \includegraphics[width=0.185\textwidth]{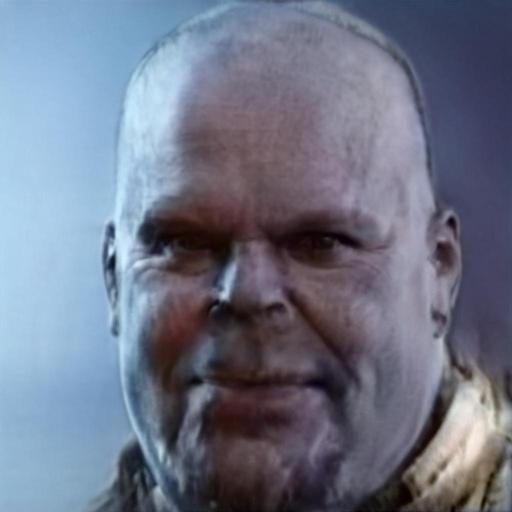}&        
        \includegraphics[width=0.185\textwidth]{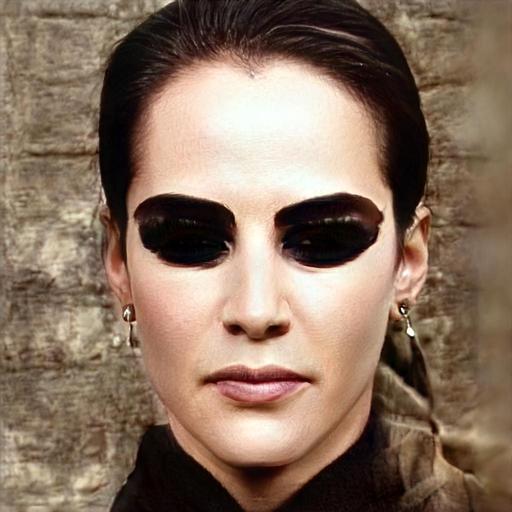}&
        \includegraphics[width=0.185\textwidth]{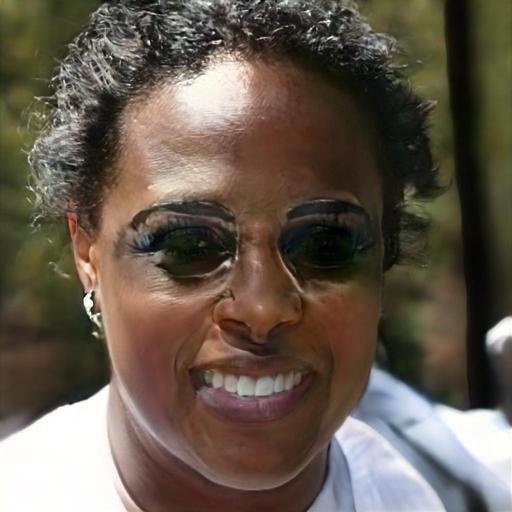}&
        \includegraphics[width=0.185\textwidth]{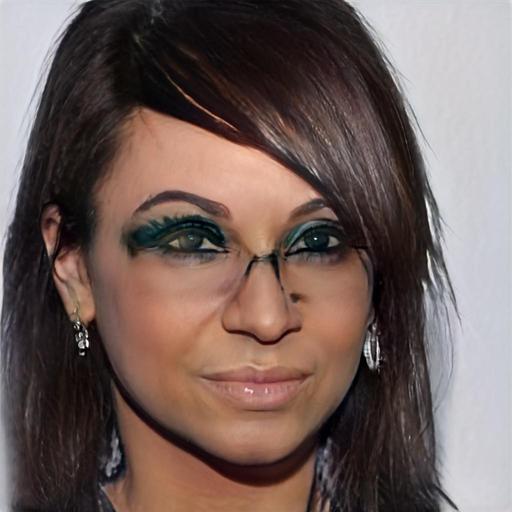} 
		\tabularnewline
        \raisebox{0.5in}{\rotatebox[origin=t]{90}{Ours}}&
        \includegraphics[width=0.185\textwidth]{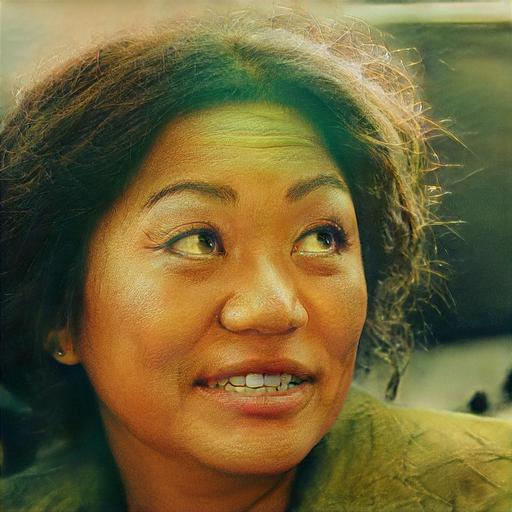}&
        \includegraphics[width=0.185\textwidth]{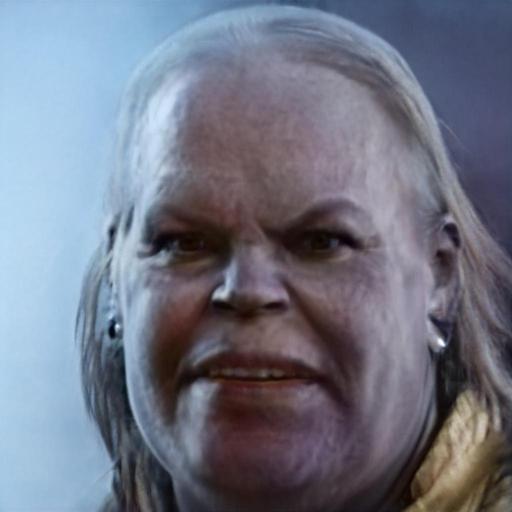}&        
        \includegraphics[width=0.185\textwidth]{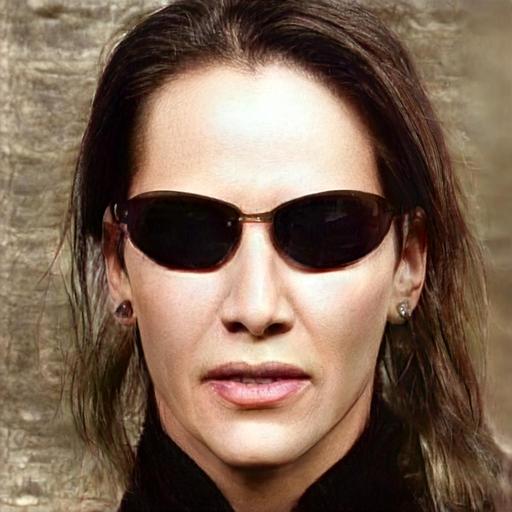}&
        \includegraphics[width=0.185\textwidth]{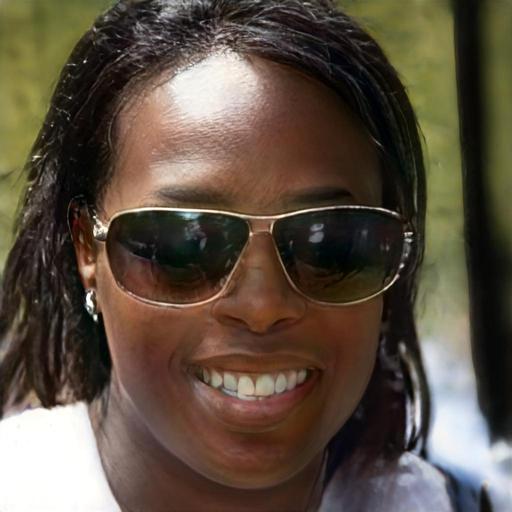}&
        \includegraphics[width=0.185\textwidth]{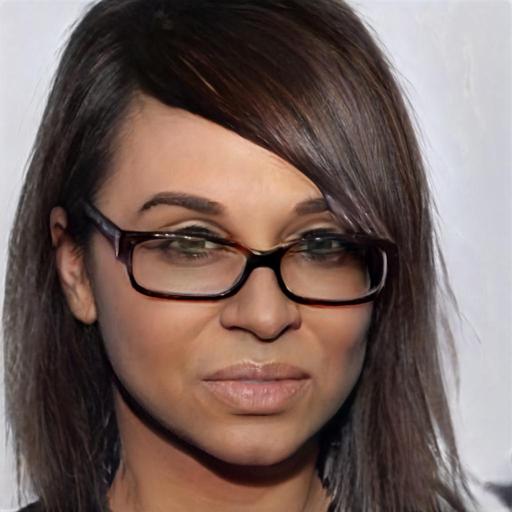} 
		\tabularnewline
		& Male$\rightarrow$Female & Male$\rightarrow$Female & Male$\rightarrow$Female & Male$\rightarrow$Female & Male$\rightarrow$Female 
		\tabularnewline
        \raisebox{0.5in}{\rotatebox[origin=t]{90}{Input}}&
        \includegraphics[width=0.185\textwidth]{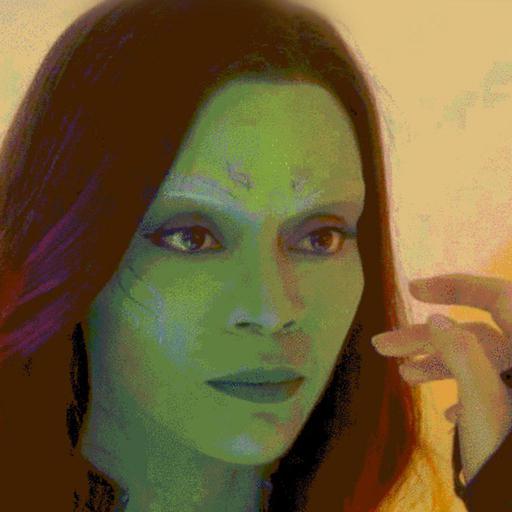}&
        \includegraphics[width=0.185\textwidth]{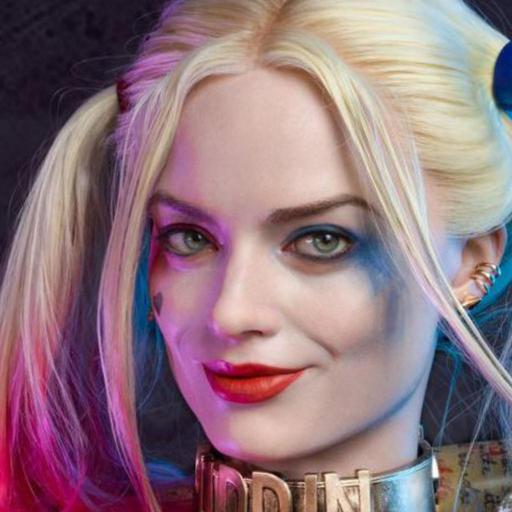}&        
        \includegraphics[width=0.185\textwidth]{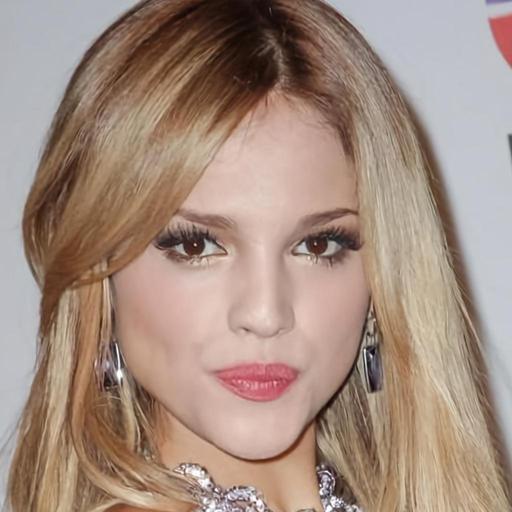}&
        \includegraphics[width=0.185\textwidth]{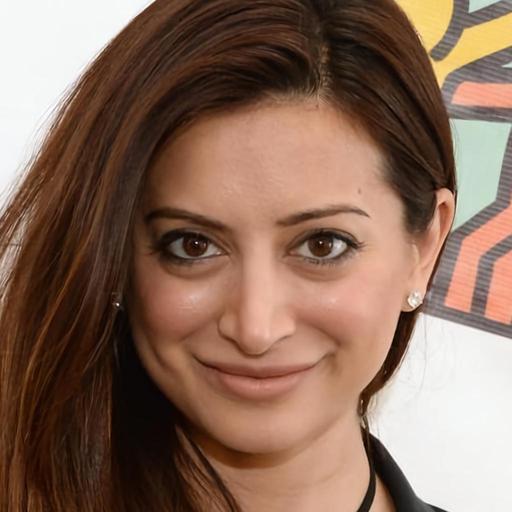}&
        \includegraphics[width=0.185\textwidth]{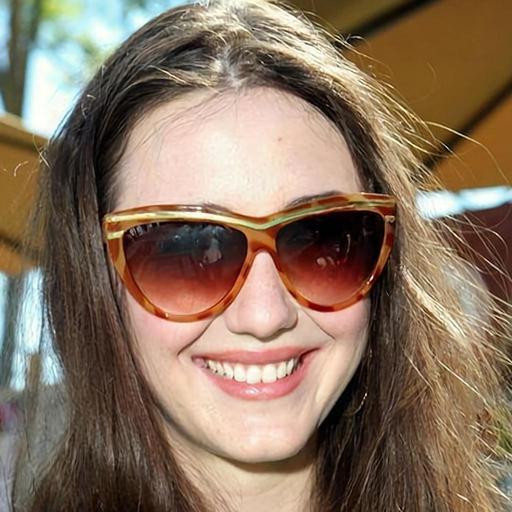} 
		\tabularnewline
        \raisebox{0.5in}{\rotatebox[origin=t]{90}{PTI}}&
        \includegraphics[width=0.185\textwidth]{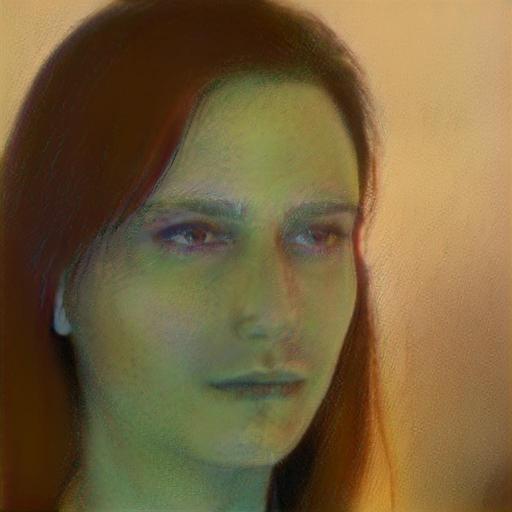}&
        \includegraphics[width=0.185\textwidth]{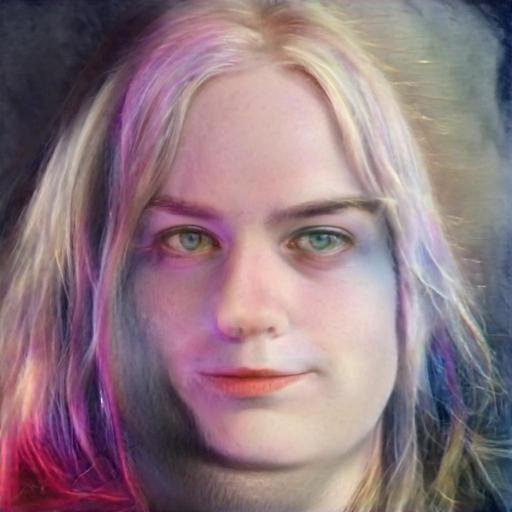}&        
        \includegraphics[width=0.185\textwidth]{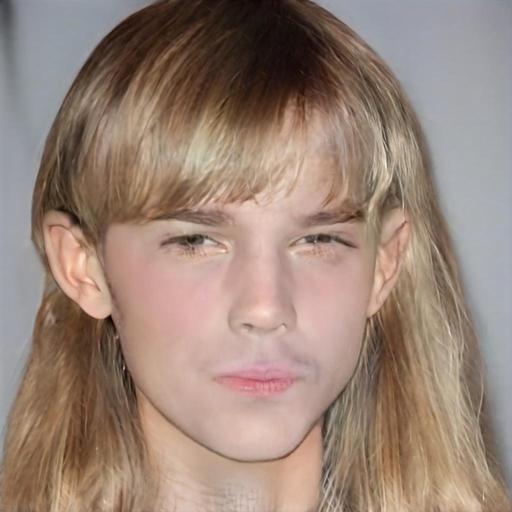}&
        \includegraphics[width=0.185\textwidth]{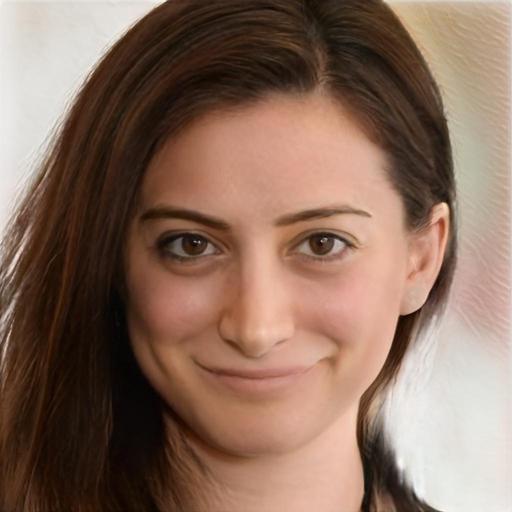}&
        \includegraphics[width=0.185\textwidth]{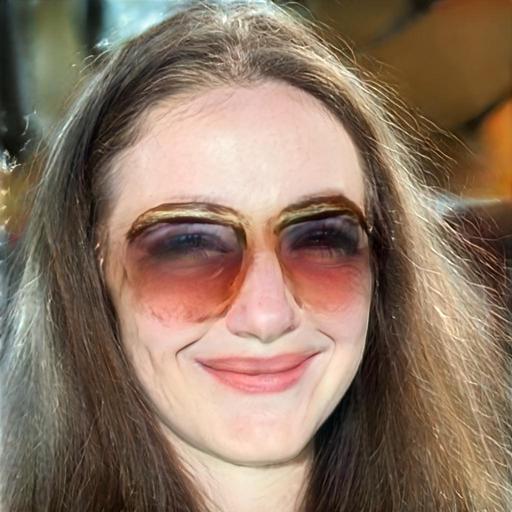} 
		\tabularnewline
        \raisebox{0.5in}{\rotatebox[origin=t]{90}{Ours}}&
        \includegraphics[width=0.185\textwidth]{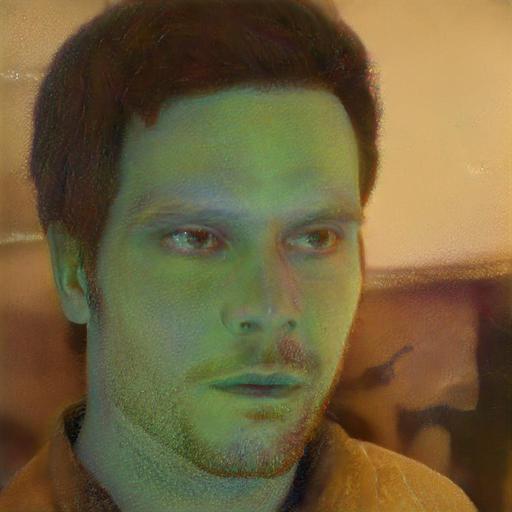}&
        \includegraphics[width=0.185\textwidth]{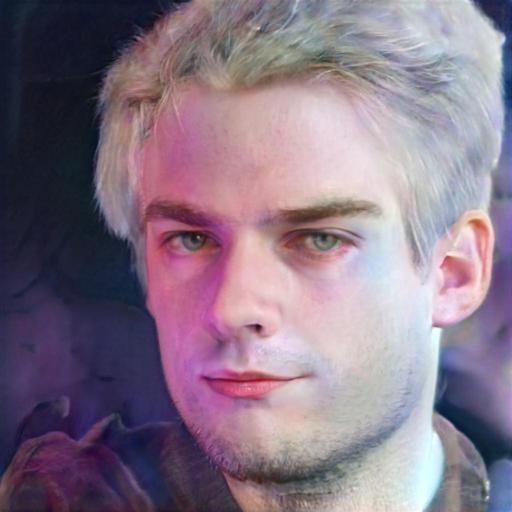}&        
        \includegraphics[width=0.185\textwidth]{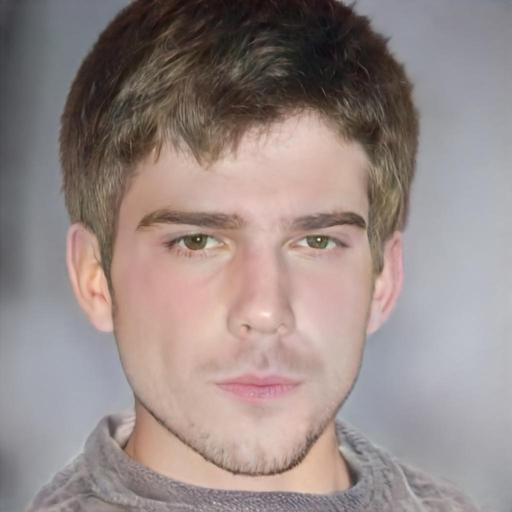}&
        \includegraphics[width=0.185\textwidth]{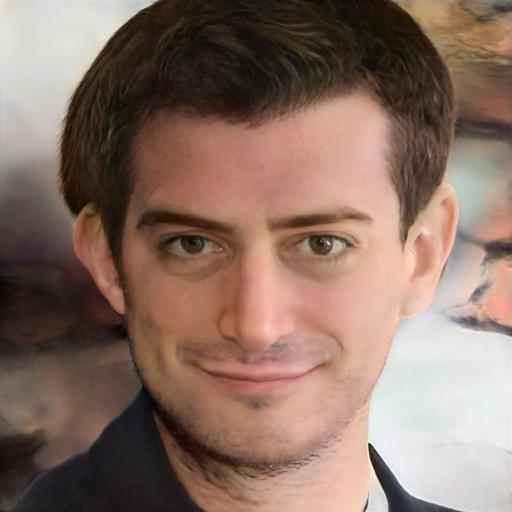}&
        \includegraphics[width=0.185\textwidth]{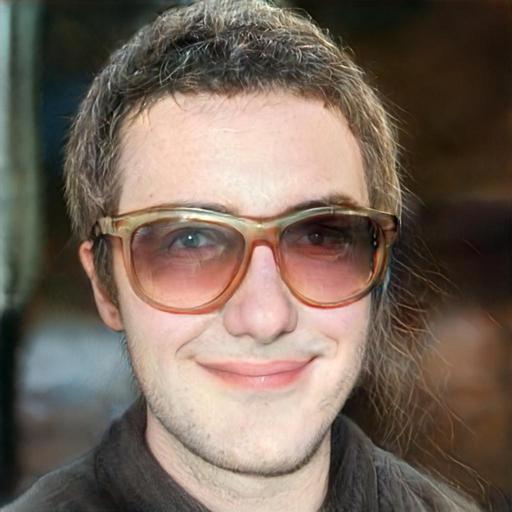} 
		\tabularnewline
		& Female$\rightarrow$Male & Female$\rightarrow$Male & Female$\rightarrow$Male & Female$\rightarrow$Male & Female$\rightarrow$Male 
    \end{tabular}
    }
	\caption{Editing quality comparison using more editing directions. In each example, the editing is performed using the same editing weight.}
    \label{fig:appendix9}
\end{figure*}

\begin{figure*}
\setlength{\tabcolsep}{1pt}
\centering
{
    \begin{tabular}{c c c c c c}
        \raisebox{0.5in}{\rotatebox[origin=t]{90}{Input}}&
        \includegraphics[width=0.185\textwidth]{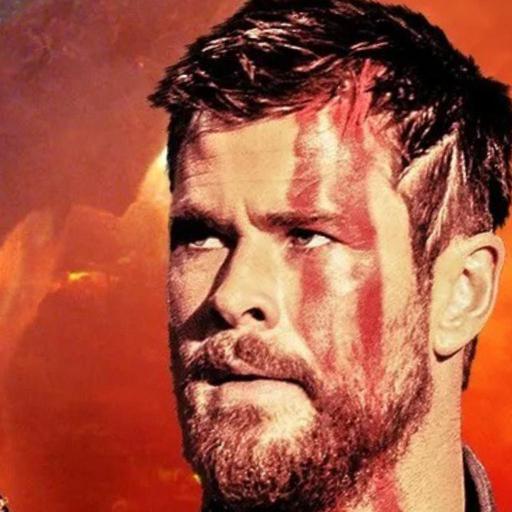}&        
        \includegraphics[width=0.185\textwidth]{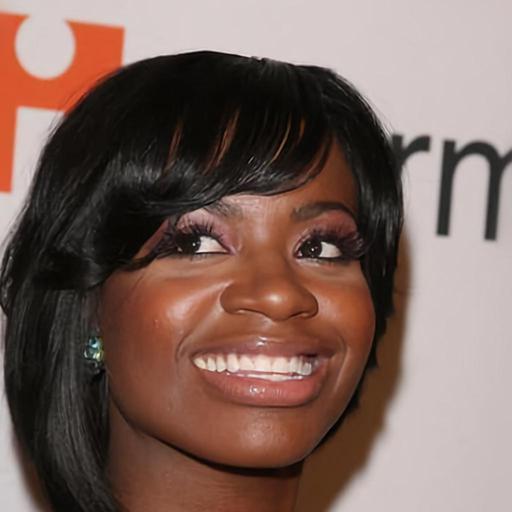}&
        \includegraphics[width=0.185\textwidth]{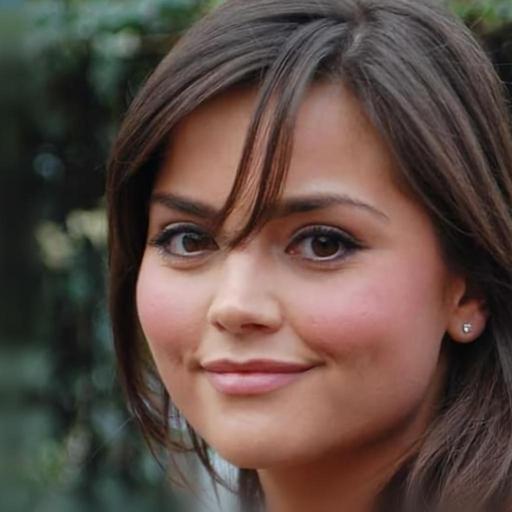}&
        \includegraphics[width=0.185\textwidth]{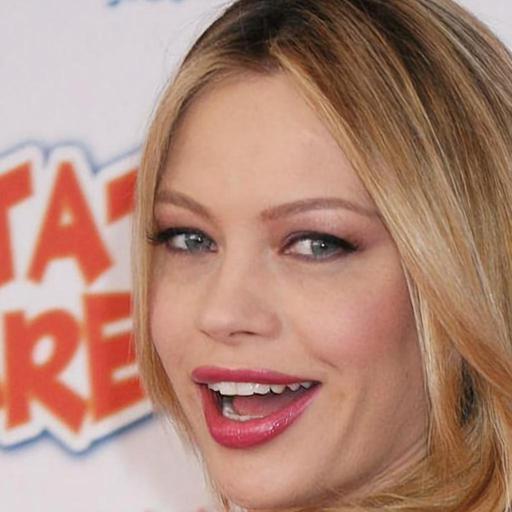}& 
        \includegraphics[width=0.185\textwidth]{images/more_directions/input/28004.jpg}
		\tabularnewline
        \raisebox{0.5in}{\rotatebox[origin=t]{90}{PTI}}&
        \includegraphics[width=0.185\textwidth]{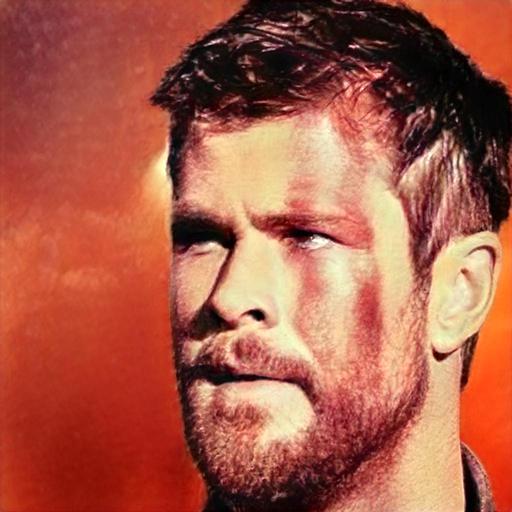}&        
        \includegraphics[width=0.185\textwidth]{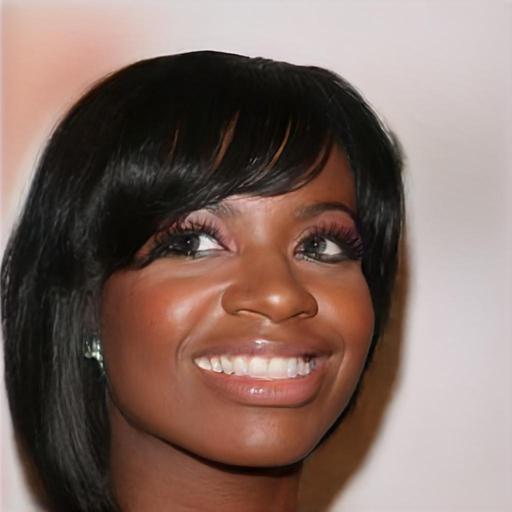}&
        \includegraphics[width=0.185\textwidth]{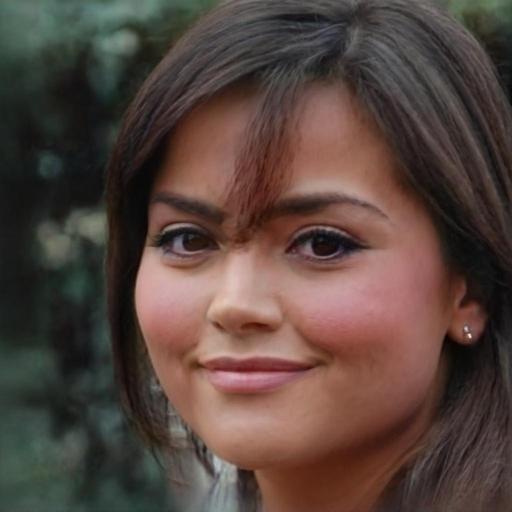}&
        \includegraphics[width=0.185\textwidth]{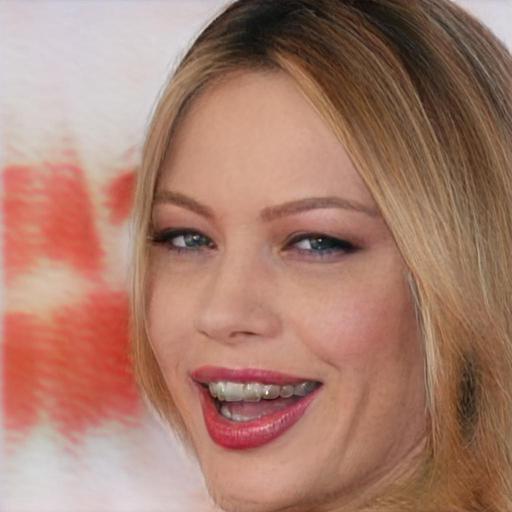} &
        \includegraphics[width=0.185\textwidth]{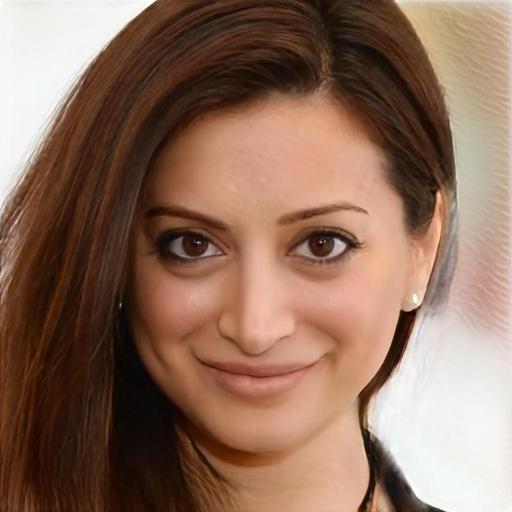}
		\tabularnewline
        \raisebox{0.5in}{\rotatebox[origin=t]{90}{Ours}}&
        \includegraphics[width=0.185\textwidth]{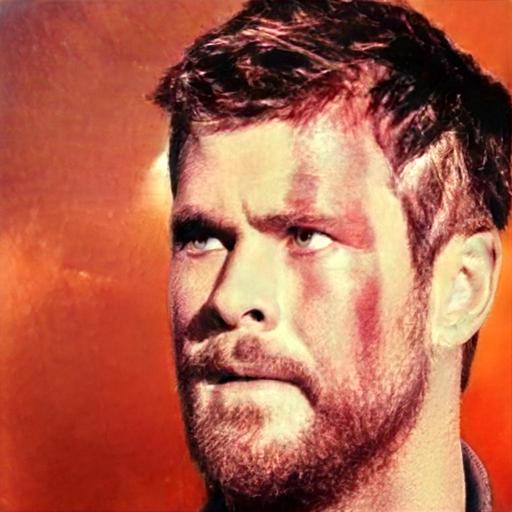}&        
        \includegraphics[width=0.185\textwidth]{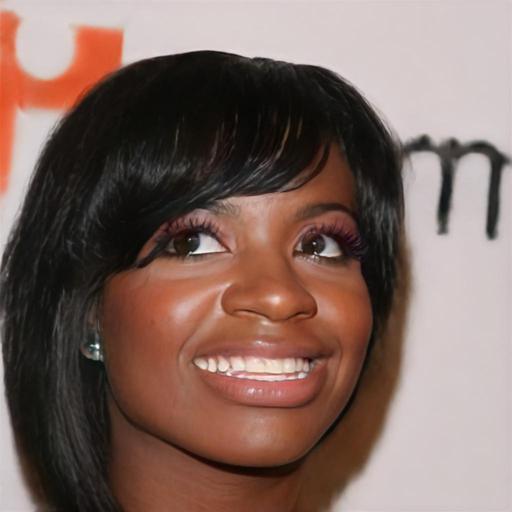}&
        \includegraphics[width=0.185\textwidth]{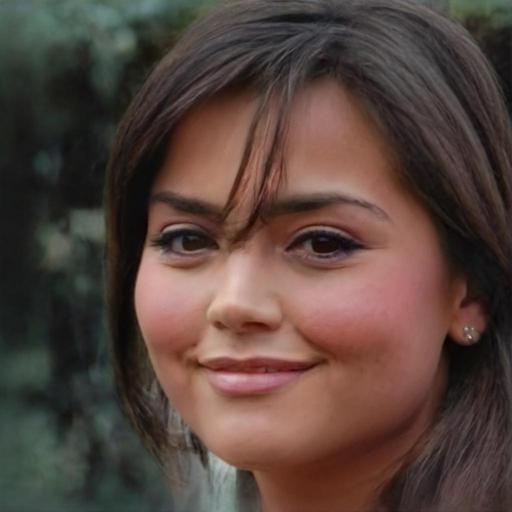}&
        \includegraphics[width=0.185\textwidth]{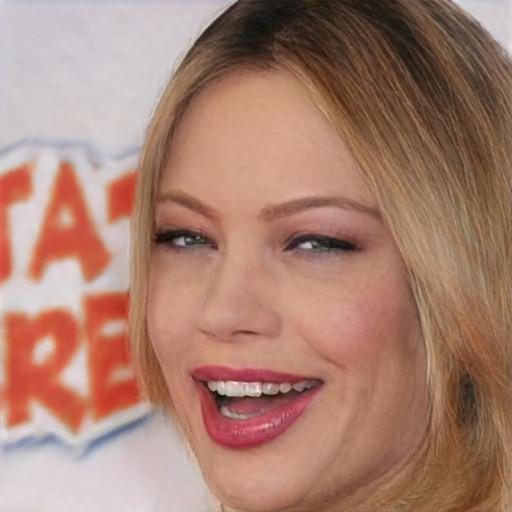}&
        \includegraphics[width=0.185\textwidth]{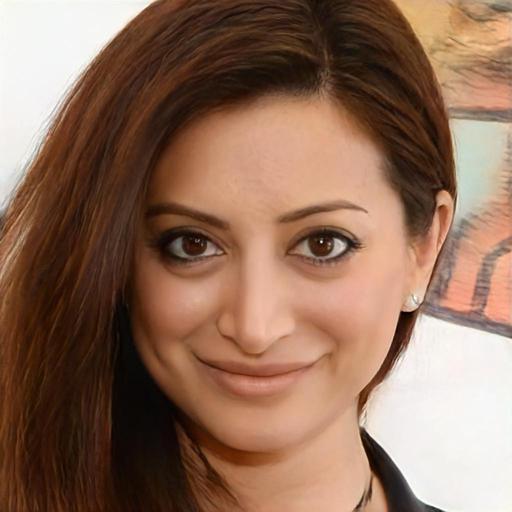}
		\tabularnewline
		&Eyes open &Eyes open& Eyes close & Eyes close & Eye and eyebrow distance
		\tabularnewline
		\raisebox{0.5in}{\rotatebox[origin=t]{90}{Input}}&
        \includegraphics[width=0.185\textwidth]{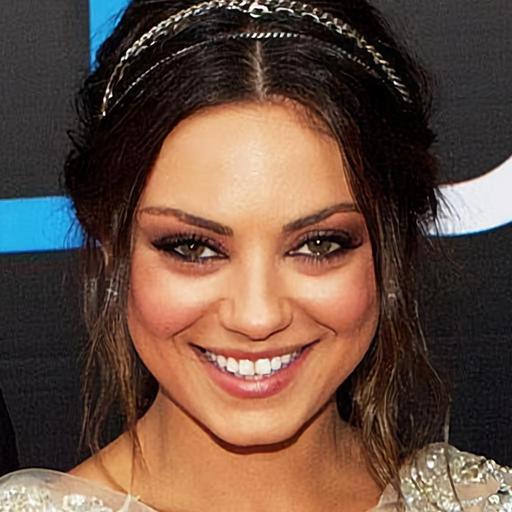}&
        \includegraphics[width=0.185\textwidth]{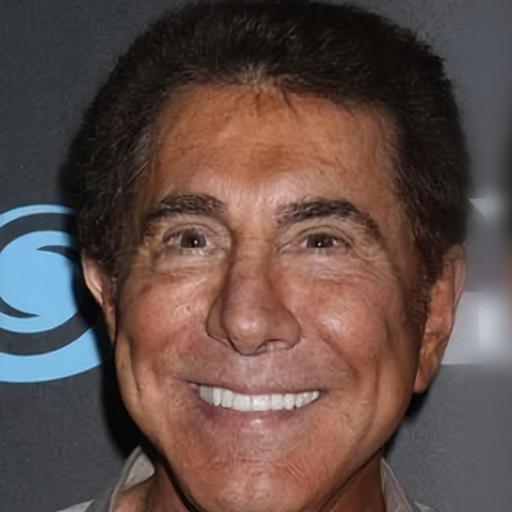}&        
        \includegraphics[width=0.185\textwidth]{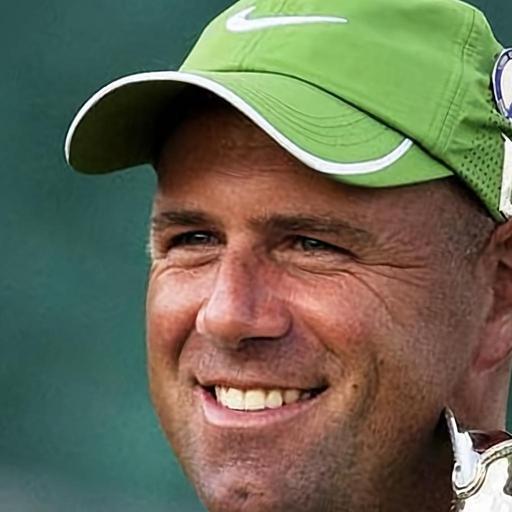}&
        \includegraphics[width=0.185\textwidth]{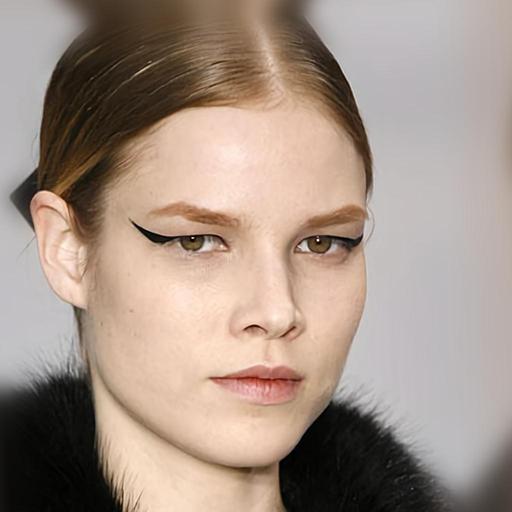}&
        \includegraphics[width=0.185\textwidth]{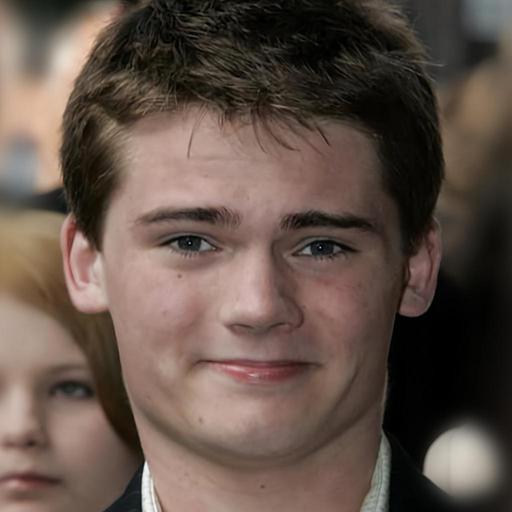} 
		\tabularnewline
        \raisebox{0.5in}{\rotatebox[origin=t]{90}{PTI}}&
        \includegraphics[width=0.185\textwidth]{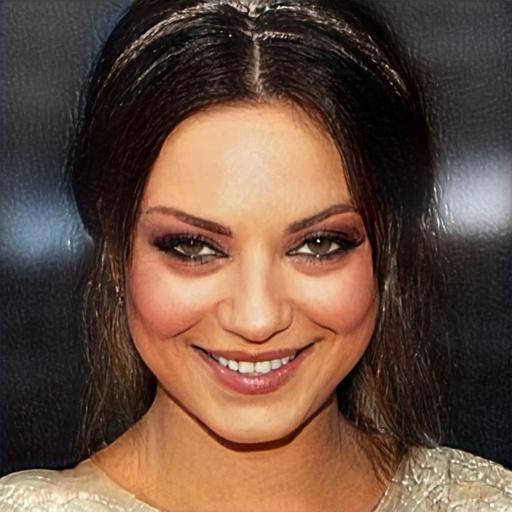}&
        \includegraphics[width=0.185\textwidth]{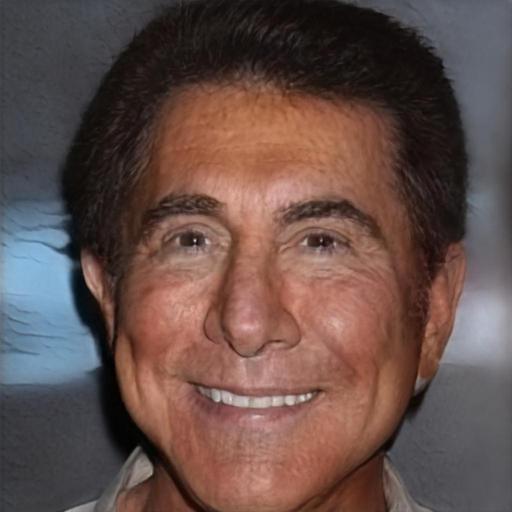}&        
        \includegraphics[width=0.185\textwidth]{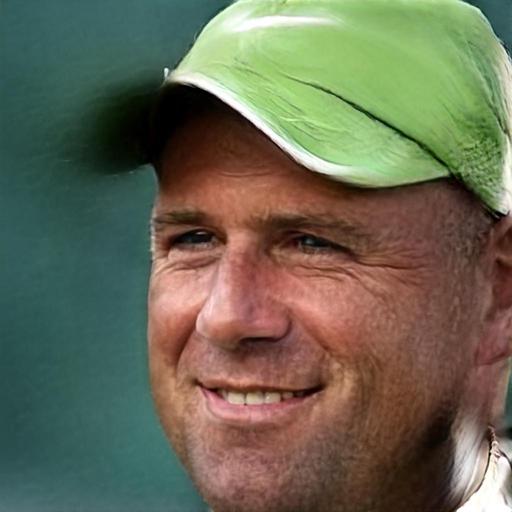}&
        \includegraphics[width=0.185\textwidth]{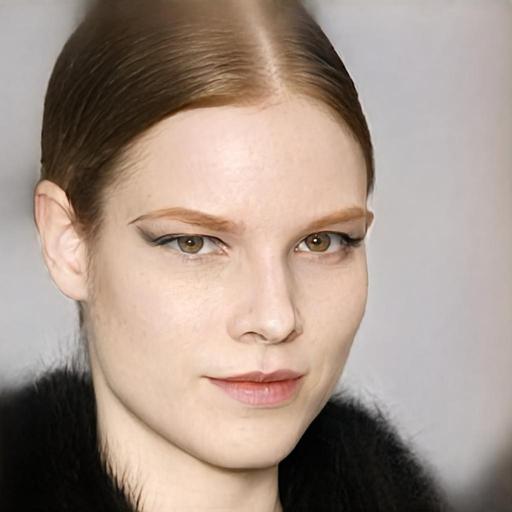}&
        \includegraphics[width=0.185\textwidth]{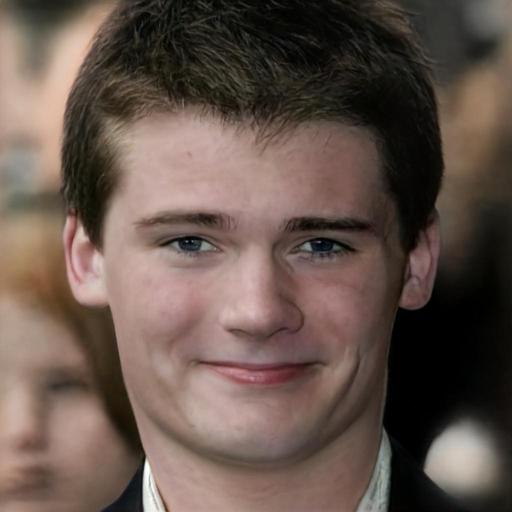} 
		\tabularnewline
        \raisebox{0.5in}{\rotatebox[origin=t]{90}{Ours}}&
        \includegraphics[width=0.185\textwidth]{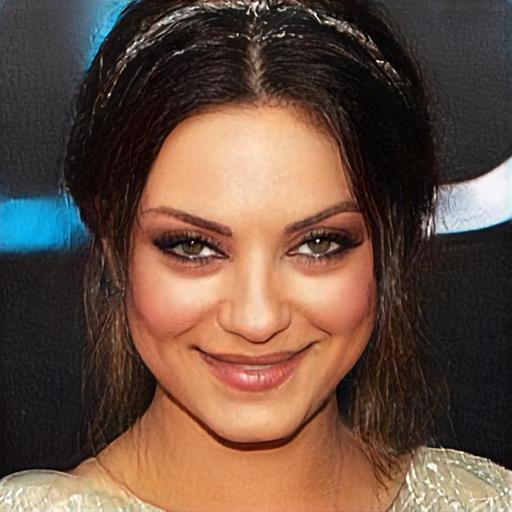}&
        \includegraphics[width=0.185\textwidth]{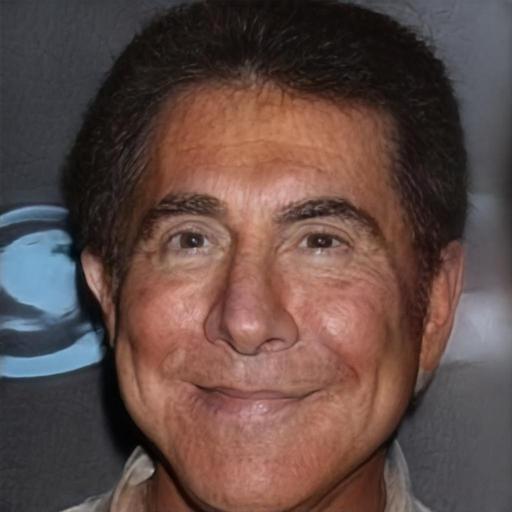}&        
        \includegraphics[width=0.185\textwidth]{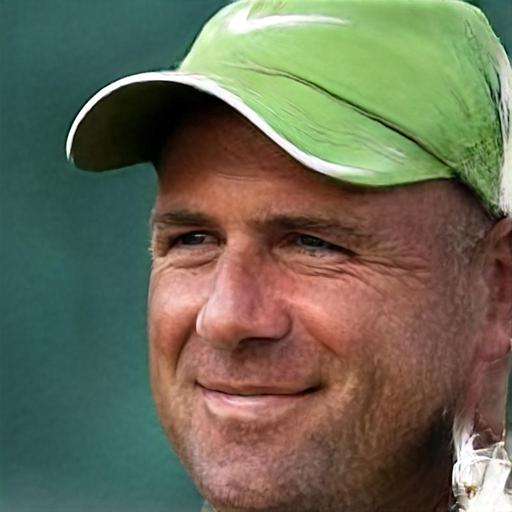}&
        \includegraphics[width=0.185\textwidth]{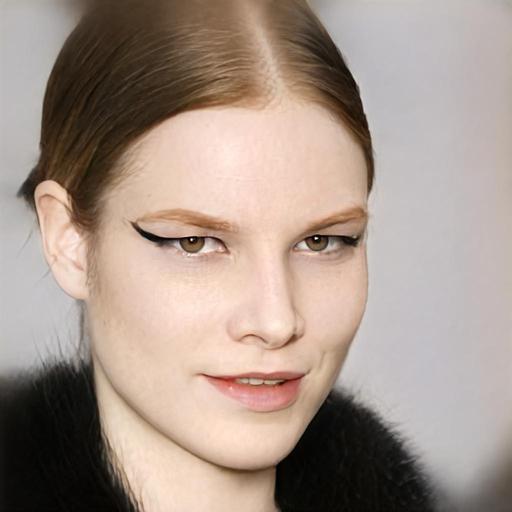}&
        \includegraphics[width=0.185\textwidth]{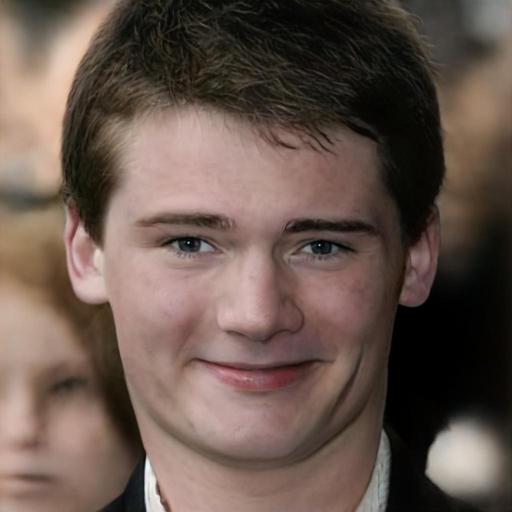} 
		\tabularnewline
		&Mouth close & Mouth close & Mouth close & Lip ratio & Lip ratio
    \end{tabular}
    }
	\caption{Editing quality comparison using more editing directions. In each example, the editing is performed using the same editing weight.}
    \label{fig:appendix10}
\end{figure*}

Figures \ref{fig:appendix9} and \ref{fig:appendix10} present the editing results using more directions\footnote{From https://twitter.com/robertluxemburg/status/1207087801344372736}, including eyes open/close, eye and eyebrow distance, and lip ratio.

\section{Editing with StyleClip}
\begin{figure*}
\setlength{\tabcolsep}{1pt}
\centering
{
    \begin{tabular}{c c c c c c}
        \raisebox{0.5in}{\rotatebox[origin=t]{90}{Input}}&
        \includegraphics[width=0.185\textwidth]{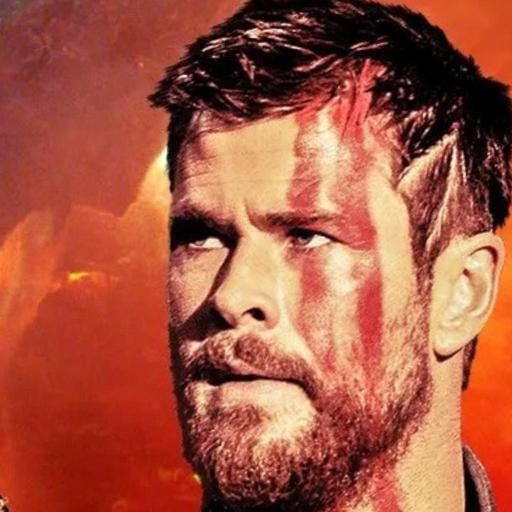}&        
        \includegraphics[width=0.185\textwidth]{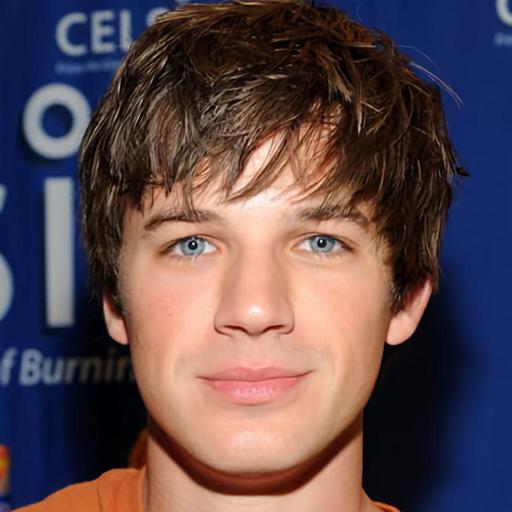}&
        \includegraphics[width=0.185\textwidth]{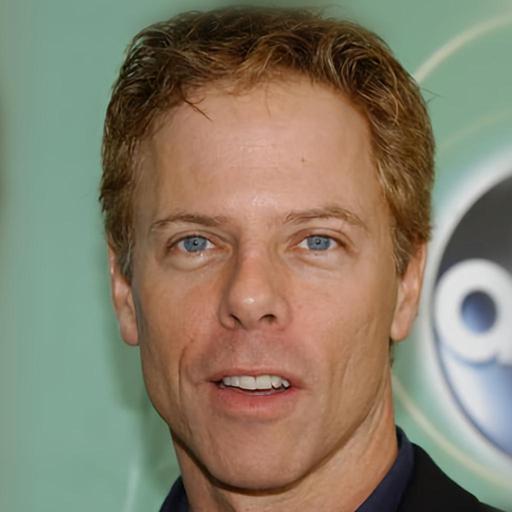}&
        \includegraphics[width=0.185\textwidth]{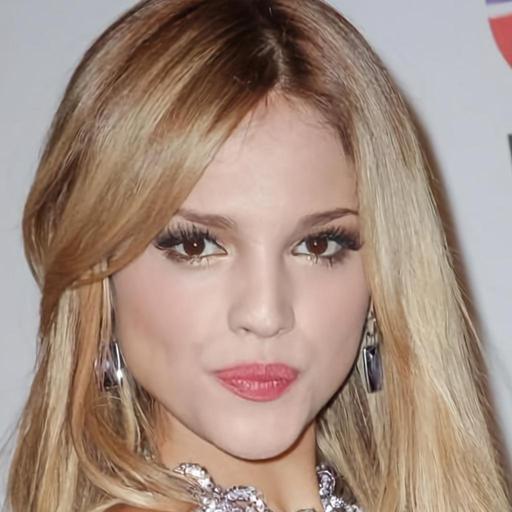}& 
        \includegraphics[width=0.185\textwidth]{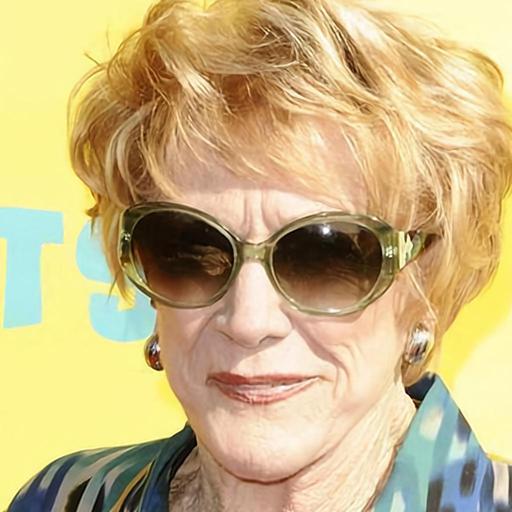}
		\tabularnewline
        \raisebox{0.5in}{\rotatebox[origin=t]{90}{StyleClip}}&
        \includegraphics[width=0.185\textwidth]{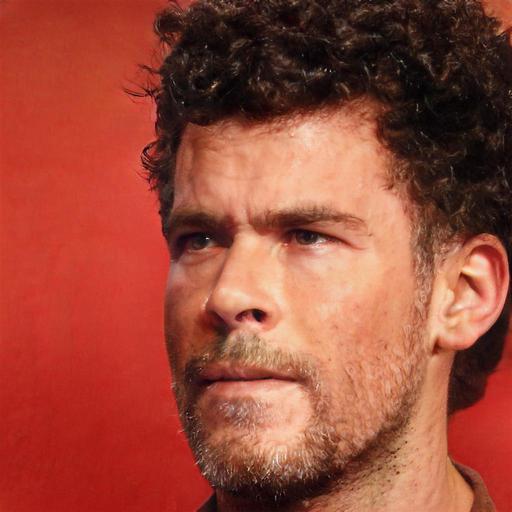}&        
        \includegraphics[width=0.185\textwidth]{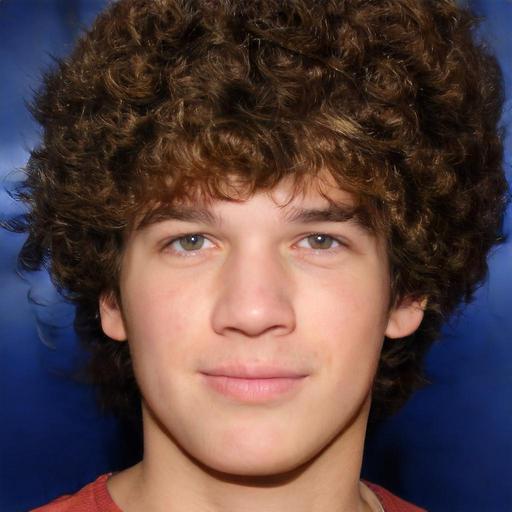}&
        \includegraphics[width=0.185\textwidth]{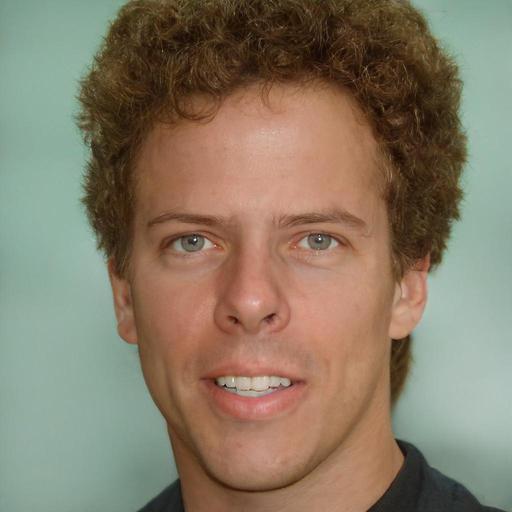}&
        \includegraphics[width=0.185\textwidth]{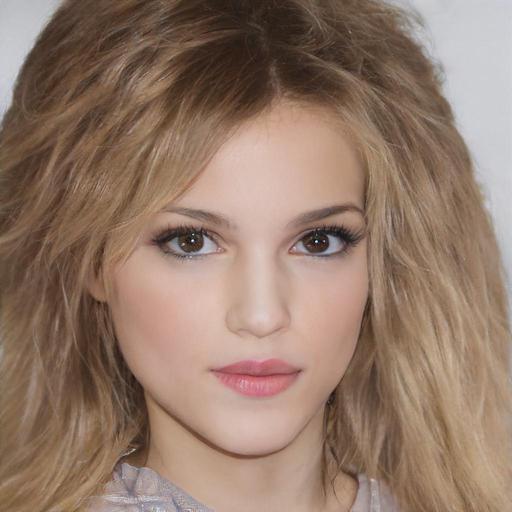} &
        \includegraphics[width=0.185\textwidth]{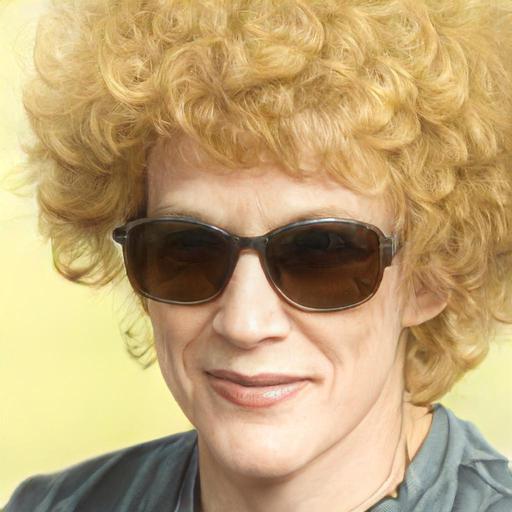}
		\tabularnewline
        \raisebox{0.5in}{\rotatebox[origin=t]{90}{PTI}}&
        \includegraphics[width=0.185\textwidth]{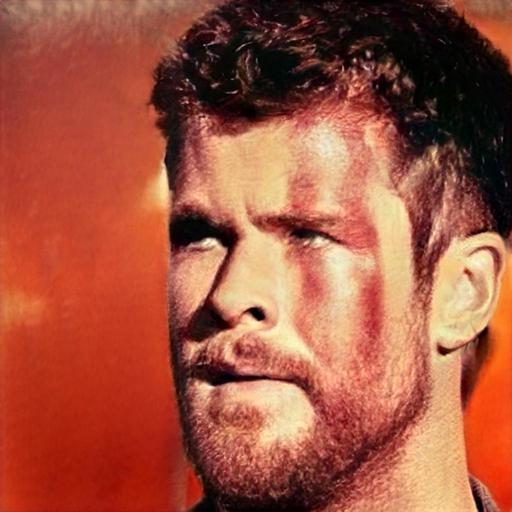}&        
        \includegraphics[width=0.185\textwidth]{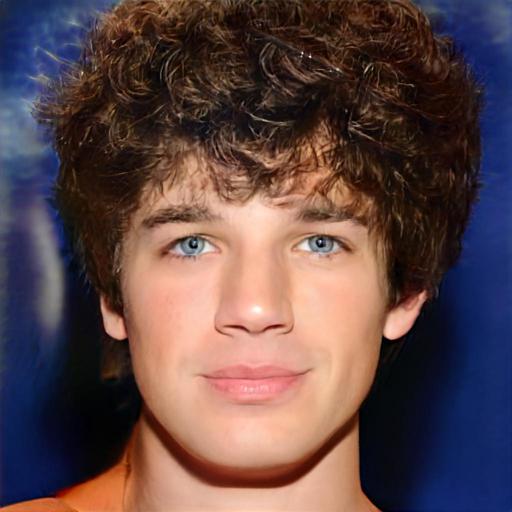}&
        \includegraphics[width=0.185\textwidth]{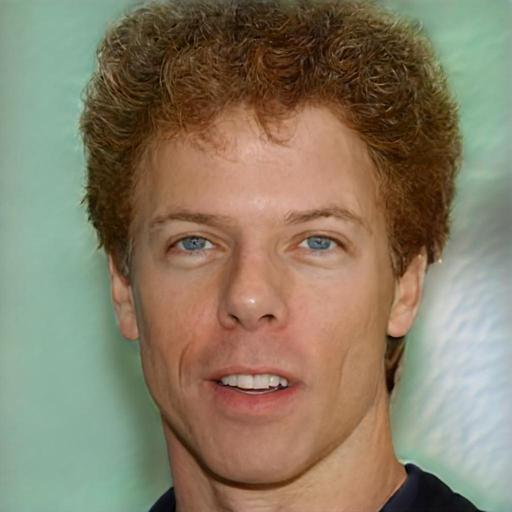}&
        \includegraphics[width=0.185\textwidth]{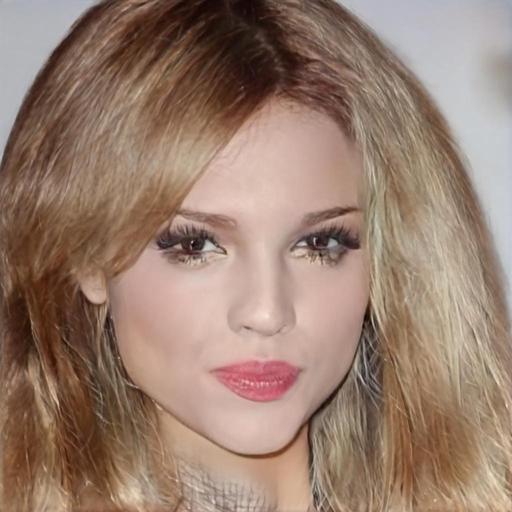} &
        \includegraphics[width=0.185\textwidth]{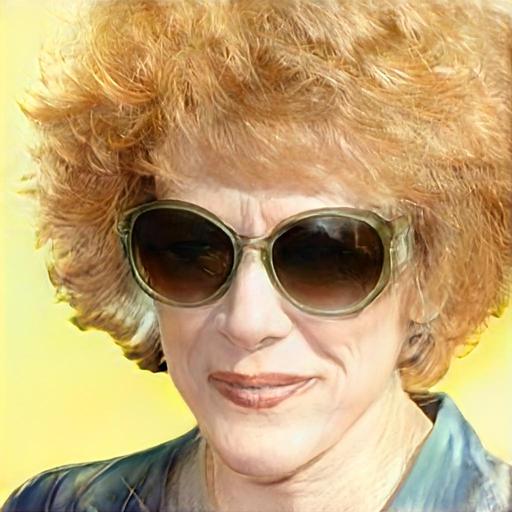}
		\tabularnewline
        \raisebox{0.5in}{\rotatebox[origin=t]{90}{PTI with e4e}}&
        \includegraphics[width=0.185\textwidth]{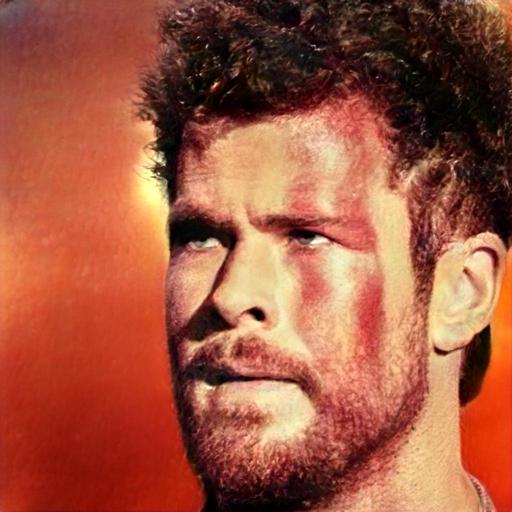}&        
        \includegraphics[width=0.185\textwidth]{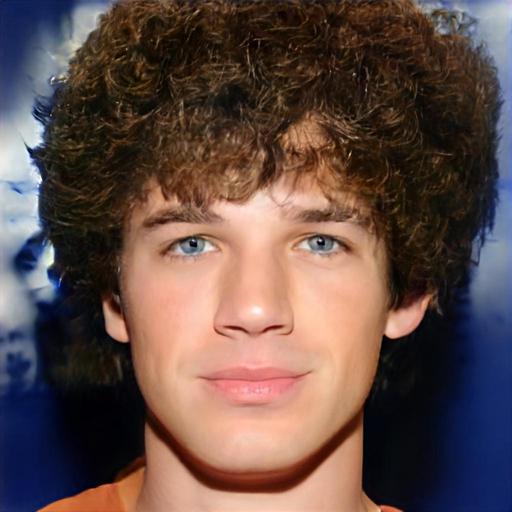}&
        \includegraphics[width=0.185\textwidth]{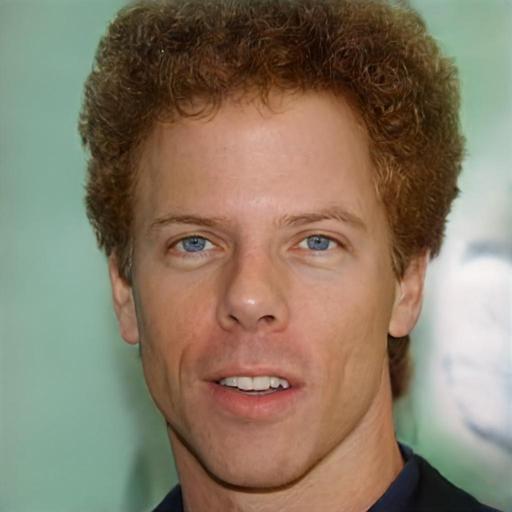}&
        \includegraphics[width=0.185\textwidth]{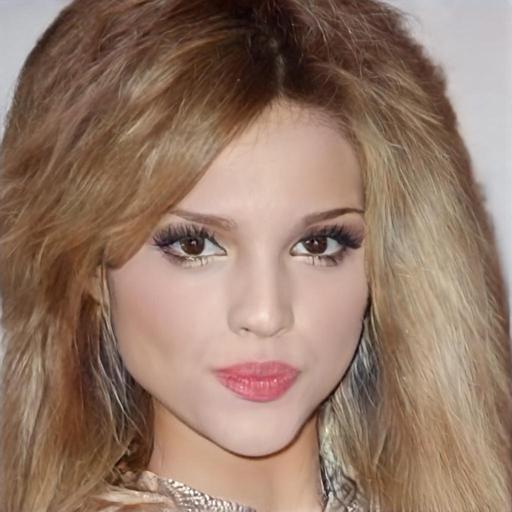} &
        \includegraphics[width=0.185\textwidth]{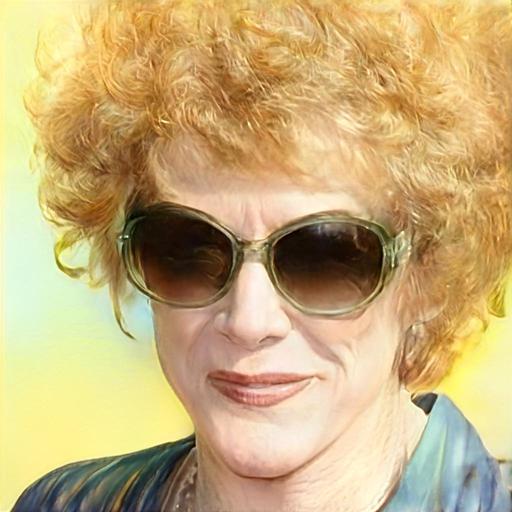}
		\tabularnewline
        \raisebox{0.5in}{\rotatebox[origin=t]{90}{Ours}}&
        \includegraphics[width=0.185\textwidth]{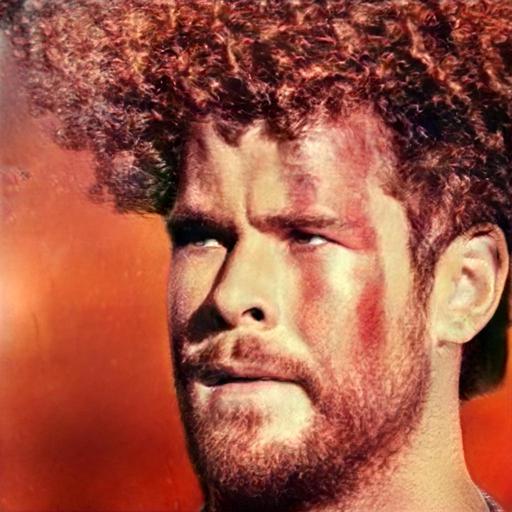}&        
        \includegraphics[width=0.185\textwidth]{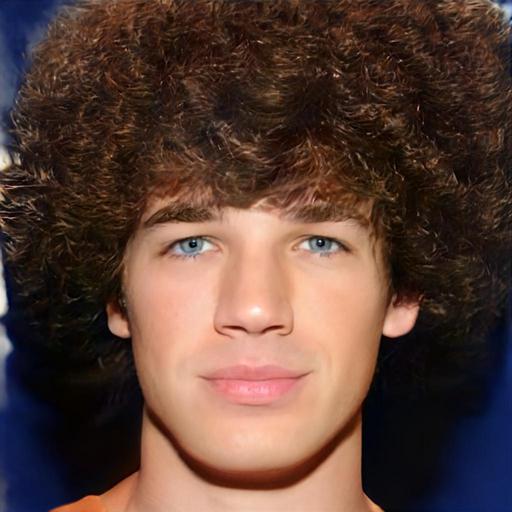}&
        \includegraphics[width=0.185\textwidth]{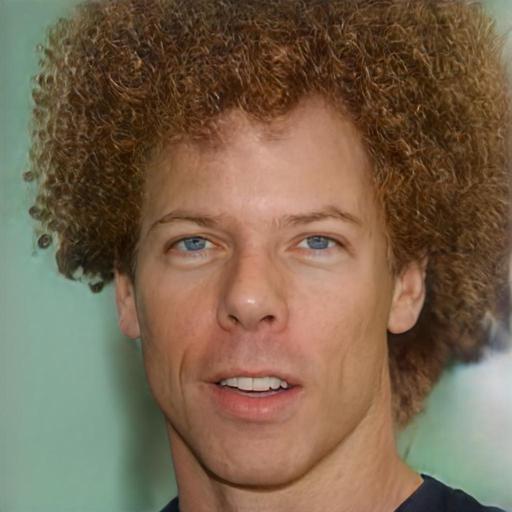}&
        \includegraphics[width=0.185\textwidth]{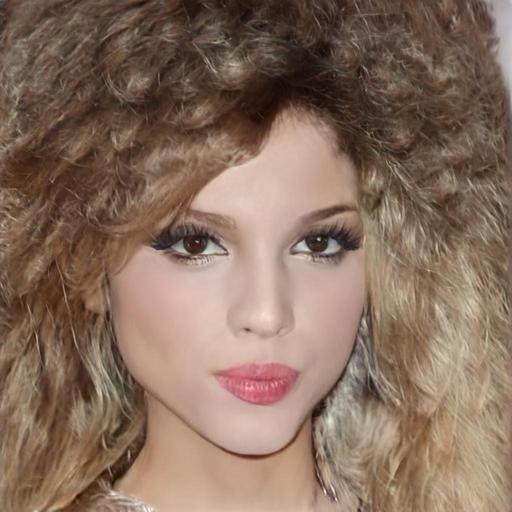}&
        \includegraphics[width=0.185\textwidth]{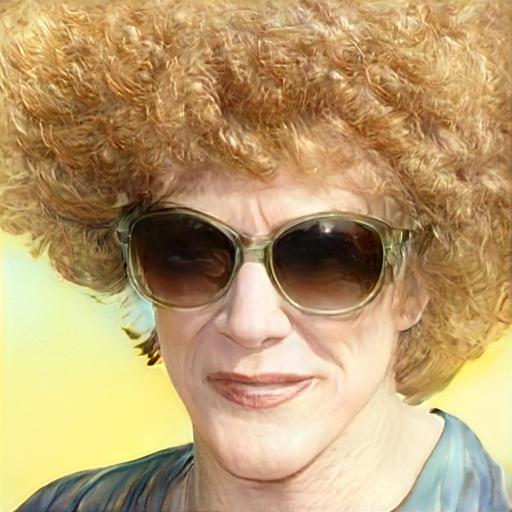}
		\tabularnewline
    \end{tabular}
    }
	\caption{Afro hairstyle edits using StyleClip \cite{Patashnik2021}. Second row: the original StyleClip model. Third row: PTI uses the optimization-based method for the pivot code. Forth row: PTI uses the e4e model for the pivot code.}
    \label{fig:appendix11}
\end{figure*}

\begin{figure*}
\setlength{\tabcolsep}{1pt}
\centering
{
    \begin{tabular}{c c c c c c}
        \raisebox{0.5in}{\rotatebox[origin=t]{90}{Input}}&
        \includegraphics[width=0.185\textwidth]{images/styleclip/input/19.jpg}&        
        \includegraphics[width=0.185\textwidth]{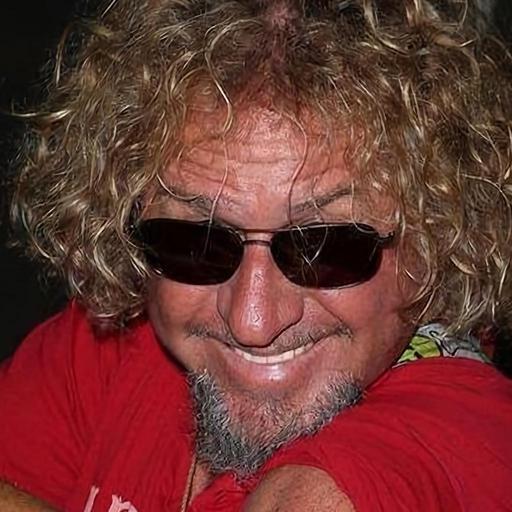}&
        \includegraphics[width=0.185\textwidth]{images/styleclip/input/28233.jpg}&
        \includegraphics[width=0.185\textwidth]{images/styleclip/input/28109.jpg}& 
        \includegraphics[width=0.185\textwidth]{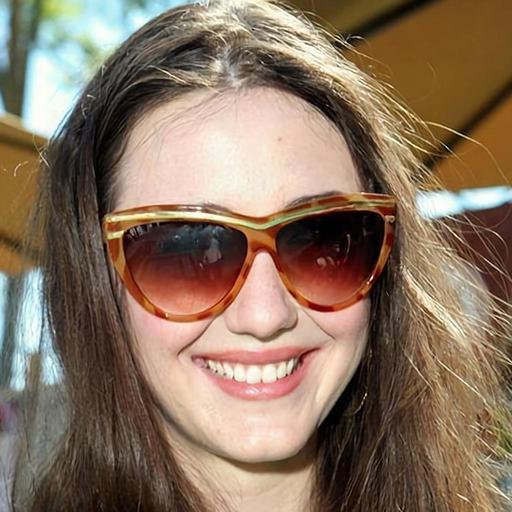}
		\tabularnewline
        \raisebox{0.5in}{\rotatebox[origin=t]{90}{StyleClip}}&
        \includegraphics[width=0.185\textwidth]{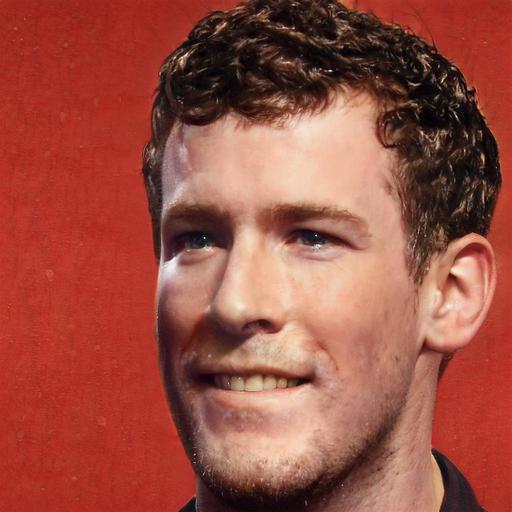}&        
        \includegraphics[width=0.185\textwidth]{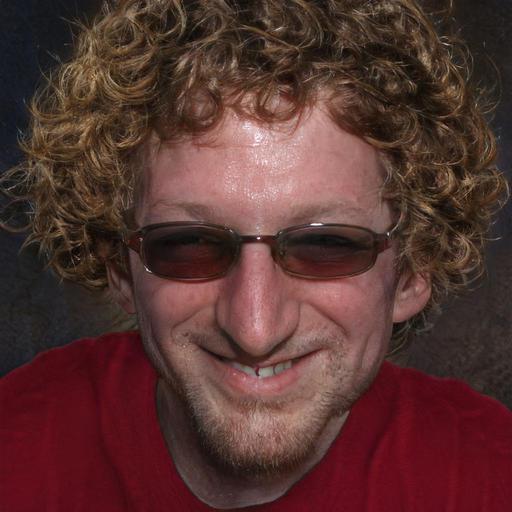}&
        \includegraphics[width=0.185\textwidth]{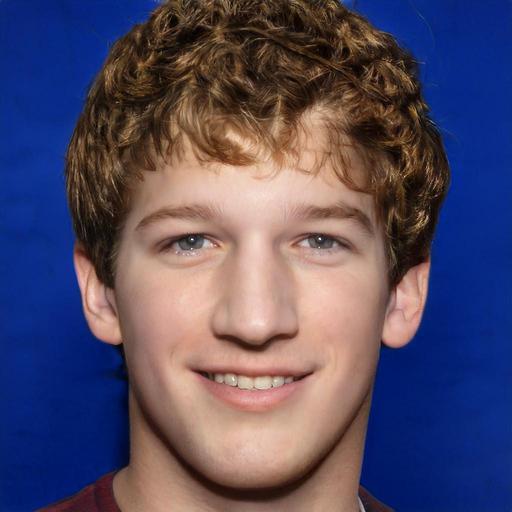}&
        \includegraphics[width=0.185\textwidth]{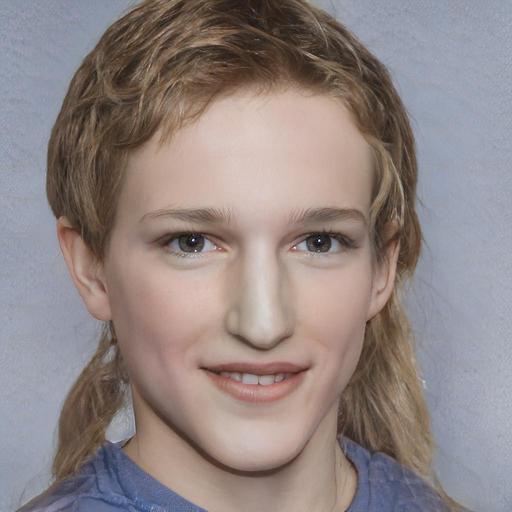} &
        \includegraphics[width=0.185\textwidth]{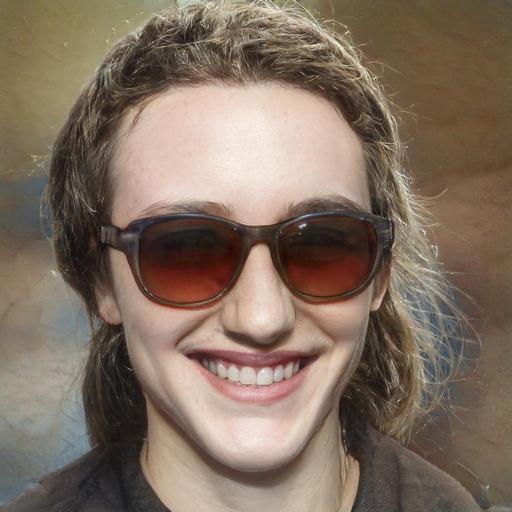}
		\tabularnewline
        \raisebox{0.5in}{\rotatebox[origin=t]{90}{PTI}}&
        \includegraphics[width=0.185\textwidth]{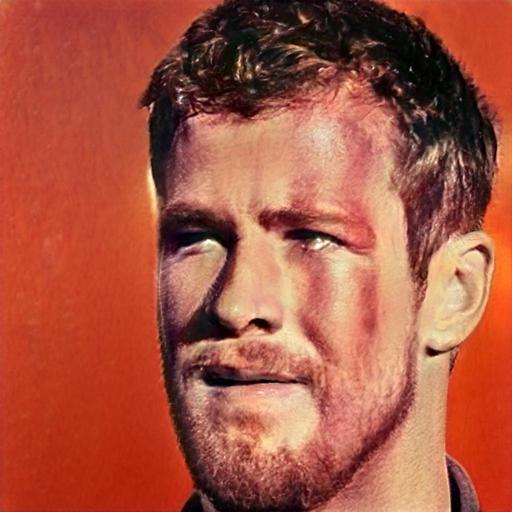}&        
        \includegraphics[width=0.185\textwidth]{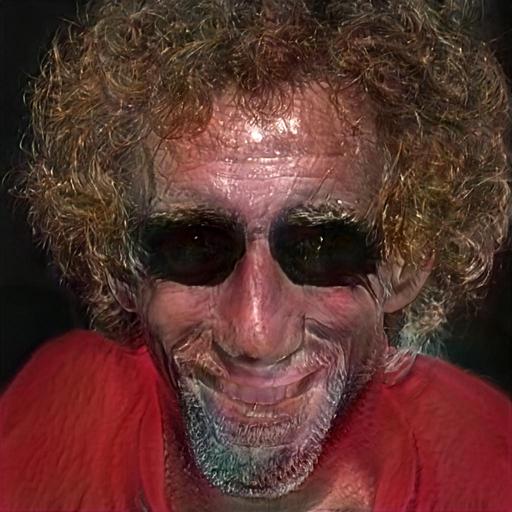}&
        \includegraphics[width=0.185\textwidth]{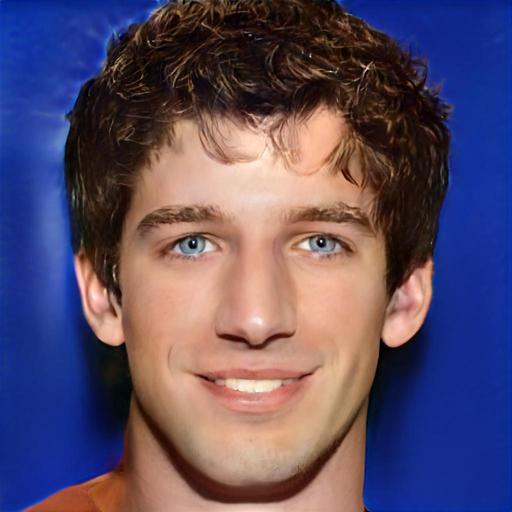}&
        \includegraphics[width=0.185\textwidth]{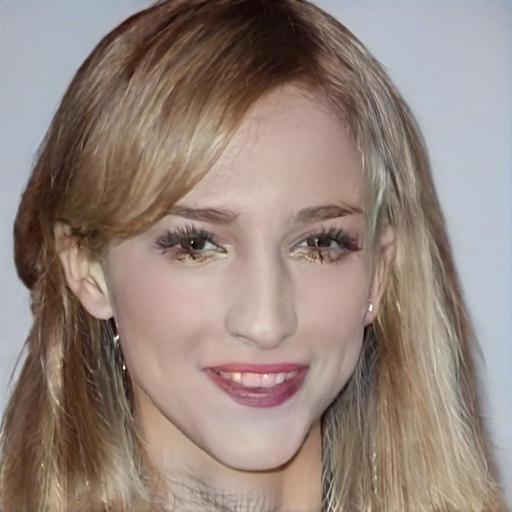} &
        \includegraphics[width=0.185\textwidth]{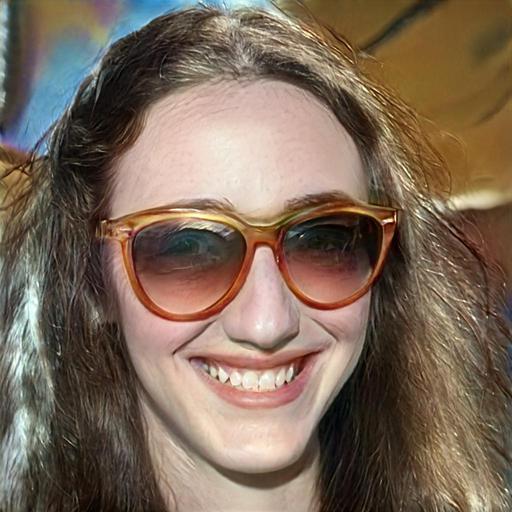}
		\tabularnewline
        \raisebox{0.5in}{\rotatebox[origin=t]{90}{PTI with e4e}}&
        \includegraphics[width=0.185\textwidth]{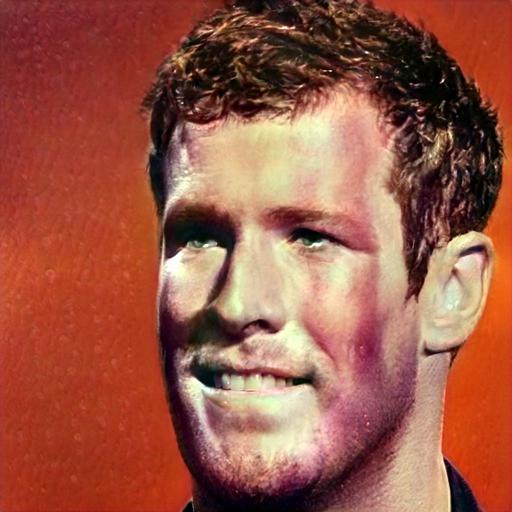}&        
        \includegraphics[width=0.185\textwidth]{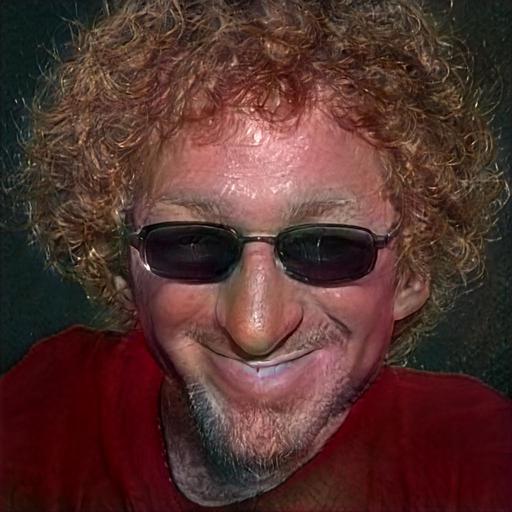}&
        \includegraphics[width=0.185\textwidth]{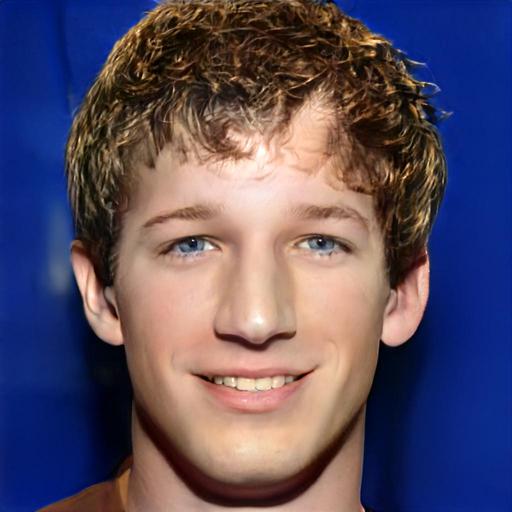}&
        \includegraphics[width=0.185\textwidth]{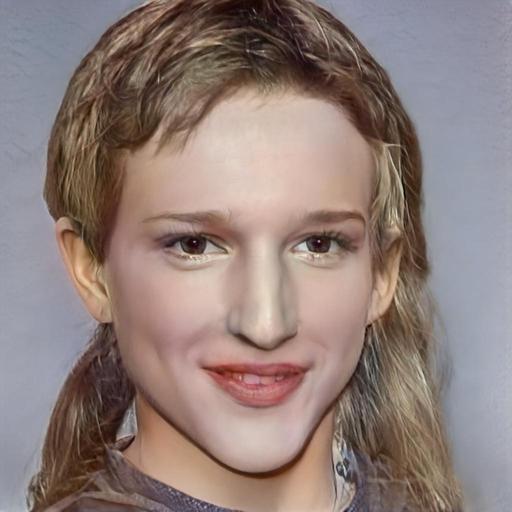} &
        \includegraphics[width=0.185\textwidth]{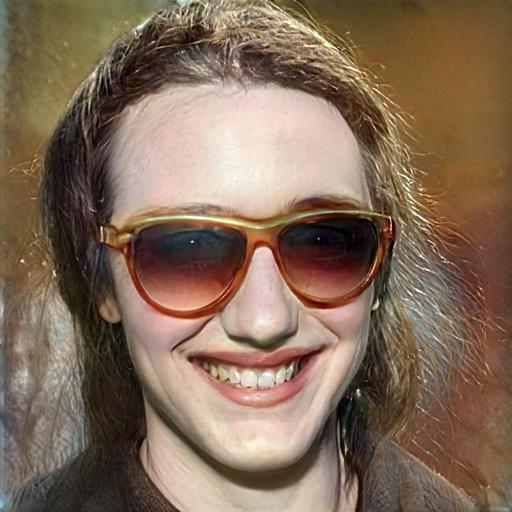}
		\tabularnewline
        \raisebox{0.5in}{\rotatebox[origin=t]{90}{Ours}}&
        \includegraphics[width=0.185\textwidth]{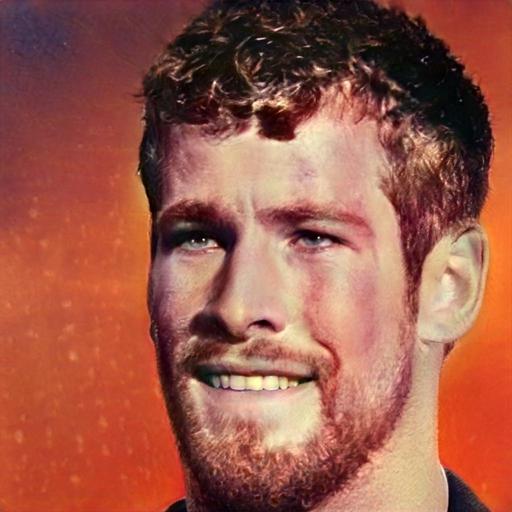}&        
        \includegraphics[width=0.185\textwidth]{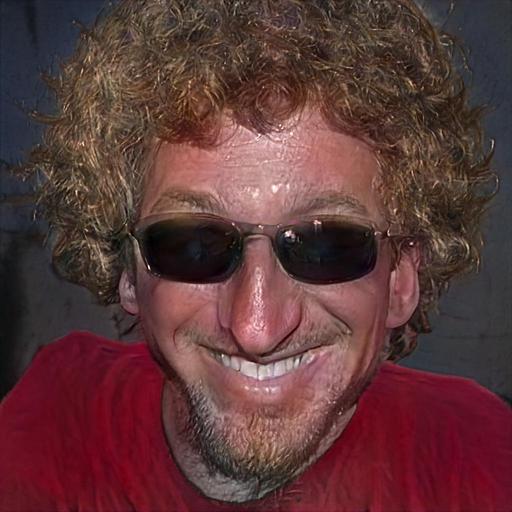}&
        \includegraphics[width=0.185\textwidth]{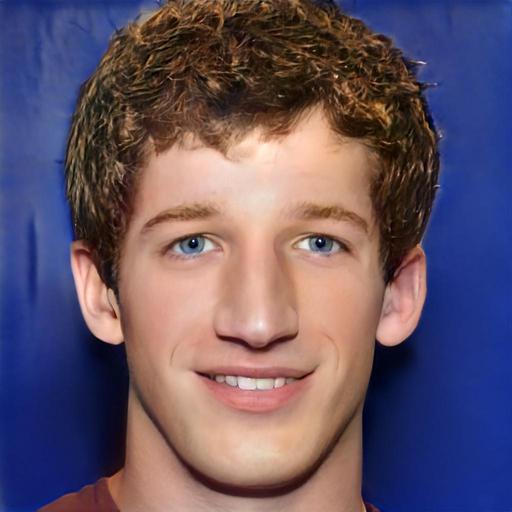}&
        \includegraphics[width=0.185\textwidth]{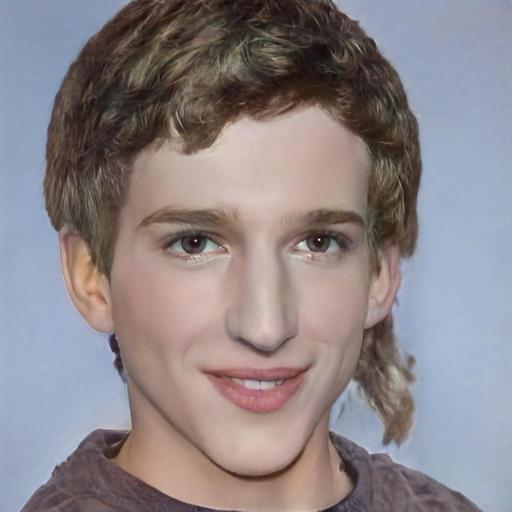} &
        \includegraphics[width=0.185\textwidth]{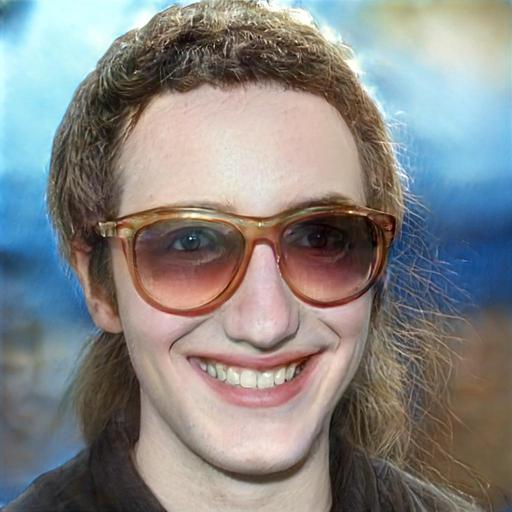}
		\tabularnewline
    \end{tabular}
    }
	\caption{Celebrity edits (Mark Zuckerberg) using StyleClip \cite{Patashnik2021}. Second row: the original StyleClip model. Third row: PTI uses the optimization-based method for the pivot code. Forth row: PTI uses the e4e model for the pivot code.}
    \label{fig:appendix12}
\end{figure*}

Figures \ref{fig:appendix11} and \ref{fig:appendix12} present the editing results using StyleClip \cite{Patashnik2021}, including hairstyle edits and celebrity edits.

\begin{figure}[b]
\setlength{\tabcolsep}{1pt}
\centering
{\small
    \begin{tabular}{c c c c c c}
        \raisebox{0.2in}{\rotatebox[origin=t]{90}{Input}}&
        \includegraphics[width=0.085\textwidth]{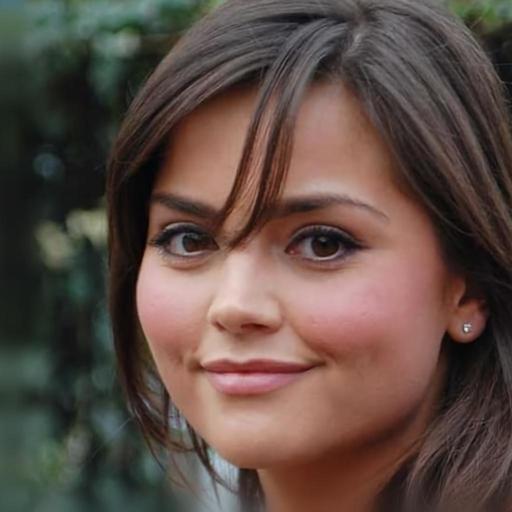}&
        \includegraphics[width=0.085\textwidth]{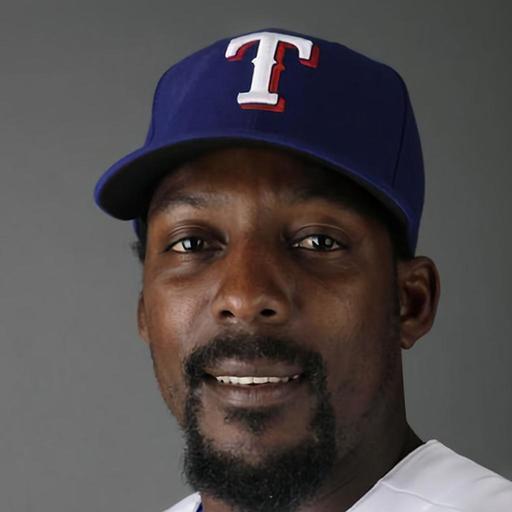}&        
        \includegraphics[width=0.085\textwidth]{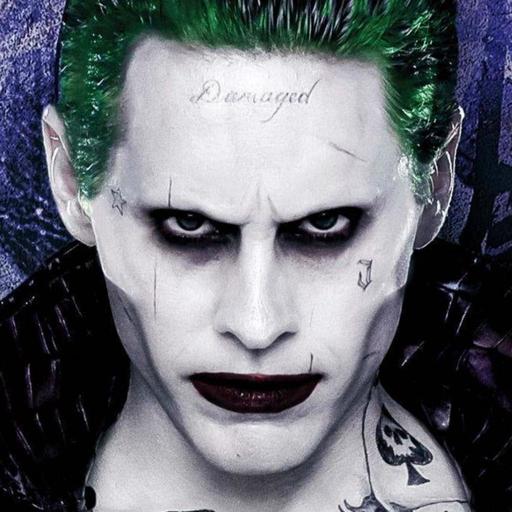}&
        \includegraphics[width=0.085\textwidth]{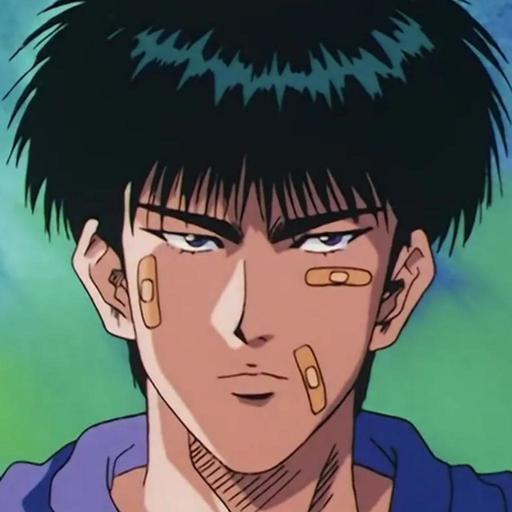}&
        \includegraphics[width=0.085\textwidth]{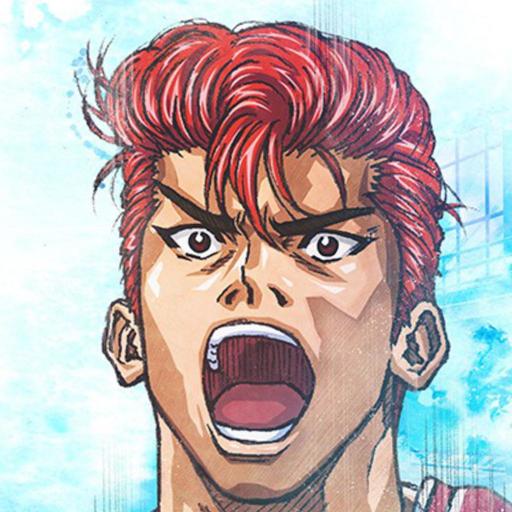} 
		\tabularnewline
        \raisebox{0.25in}{\rotatebox[origin=t]{90}{Optim. in \W}}&
        \includegraphics[width=0.085\textwidth]{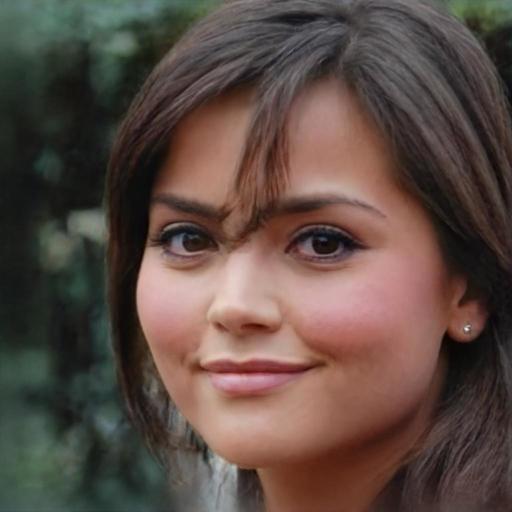}&
        \includegraphics[width=0.085\textwidth]{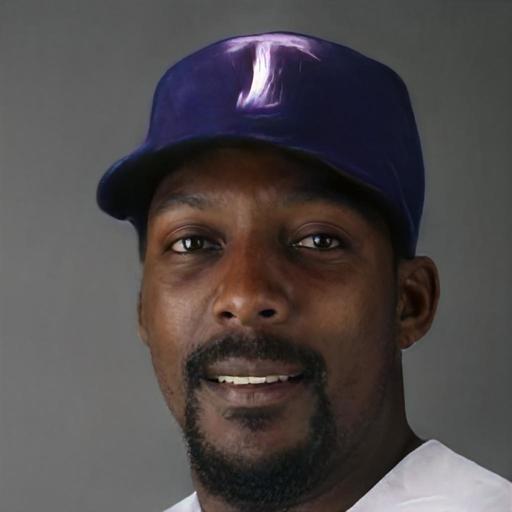}&        
        \includegraphics[width=0.085\textwidth]{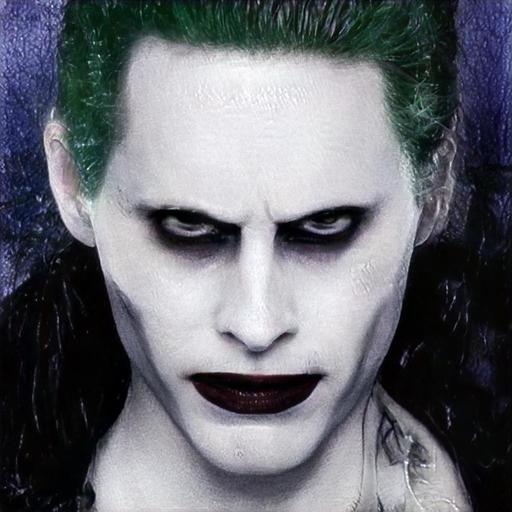}&
        \includegraphics[width=0.085\textwidth]{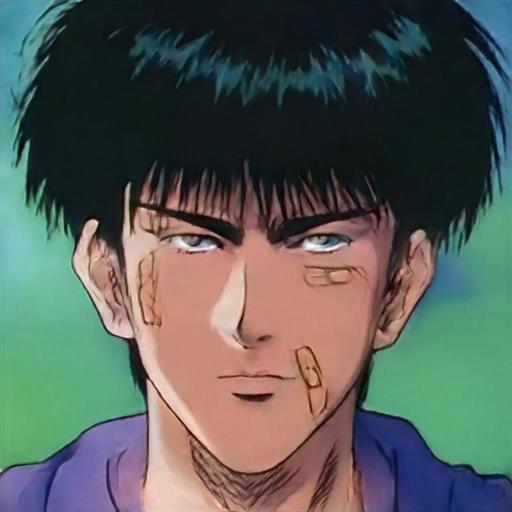}&
        \includegraphics[width=0.085\textwidth]{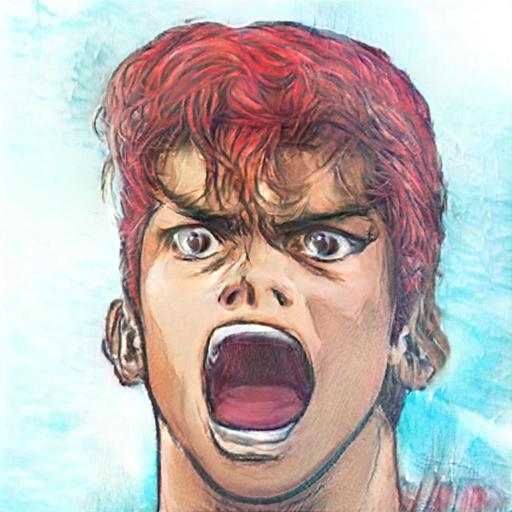} 
		\tabularnewline
        \raisebox{0.23in}{\rotatebox[origin=t]{90}{Optim. in \Wp}}&
        \includegraphics[width=0.085\textwidth]{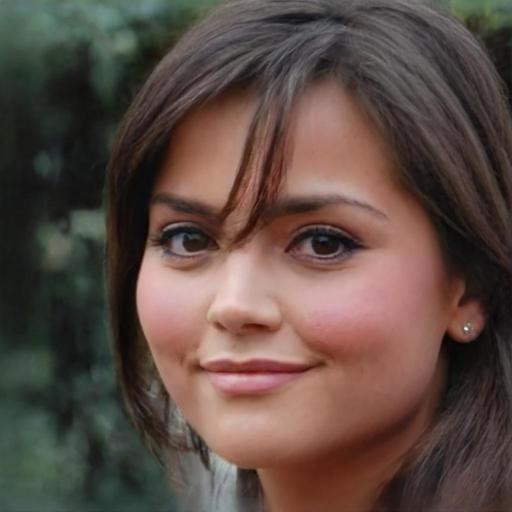}&
        \includegraphics[width=0.085\textwidth]{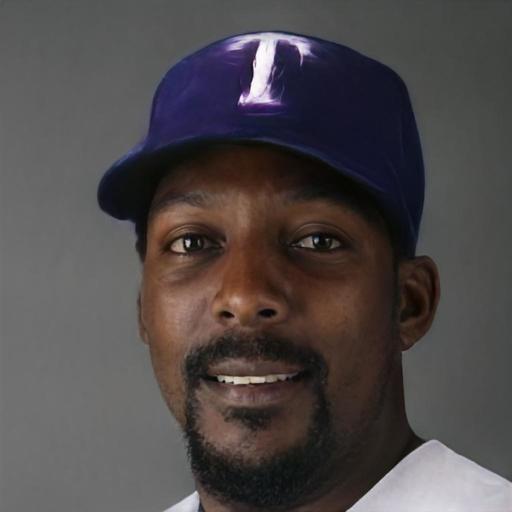}&        
        \includegraphics[width=0.085\textwidth]{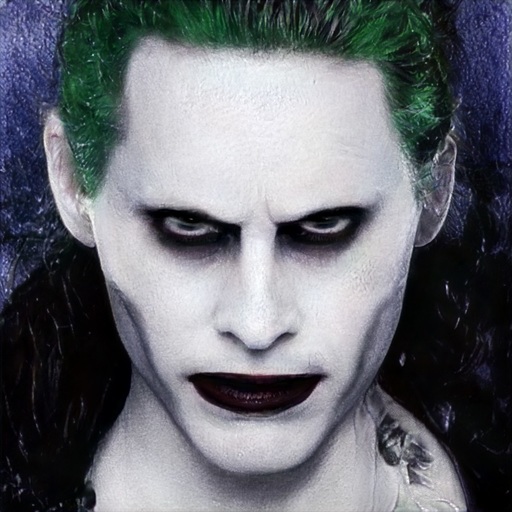}&
        \includegraphics[width=0.085\textwidth]{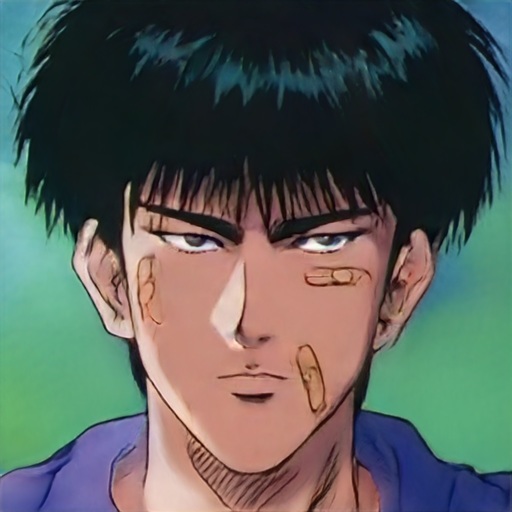}&
        \includegraphics[width=0.085\textwidth]{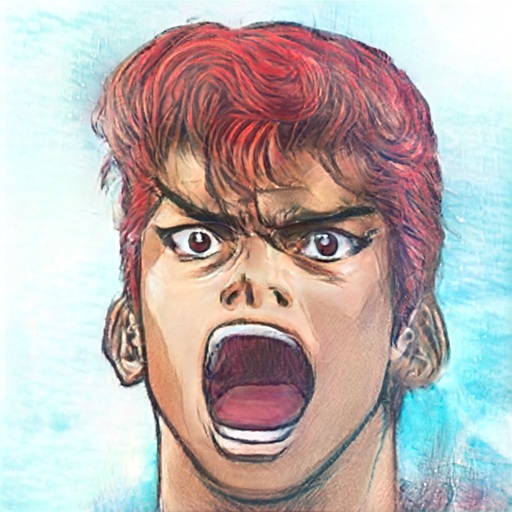} 
    \end{tabular}
    }
	\caption{Reconstruction quality comparison. Optim. in \Wp means that applying the optimization step in the \Wp space.}
    \label{fig:on_w}
\end{figure}

\begin{table}[b]
	\centering
	\caption{Optim. in \Wp means that applying the optimization step in the \Wp space. First row: reconstruction evaluation via identity similarity. Second row: editability evaluation via rotation angle.}
	\begin{tabular}{L{1.3cm}P{1.5cm}P{1.9cm}P{1.9cm}}
	\toprule
	 & PTI \cite{Roich2021} & Optim. in \W  & Optim. in \Wp \\
	\midrule
	Identity$\uparrow$ &
	\multicolumn{1}{c}{$0.845$} &
	\multicolumn{1}{c}{$0.857$} &
	\multicolumn{1}{c}{$0.866$} \\
	Angle$\uparrow$  &
	\multicolumn{1}{c}{$16.64$} &
	\multicolumn{1}{c}{$22.40$} &
	\multicolumn{1}{c}{$22.20$} \\
	\bottomrule
	\end{tabular}
	\label{table:on_w}
\end{table}

\section{Optimization in \W or \Wp space}
In section \ref{sec:optimization}, we apply the optimization in the \Wp space (denoted as Optim. in \Wp). Here, we compare this design choice with applying the optimization in the \W space (denoted as Optim. in \W). The quantitative results in Table \ref{table:on_w} show that Optim. in \Wp achieves lower distortion with a subtle drop in editability. The qualitative results in Figure \ref{fig:on_w} also show that Optim. in \Wp achieves more accurate reconstruction. Thus, we believe that Optim. in \Wp is cost-effective compared to the \W space. The reason is that slight and local changes (15 iterations in our experiments) to the pivot code can be applied without damaging its editing capability.

\section{Training Methodology for $\mathcal{W}+$$\rightarrow$$\mathcal{W}$}
\begin{figure}[b]
\setlength{\tabcolsep}{2pt}
\centering
{\small
    \begin{tabular}{c c c c c c}
        \raisebox{0.18in}{\rotatebox[origin=t]{90}{Hard}}&
        \includegraphics[width=0.08\textwidth]{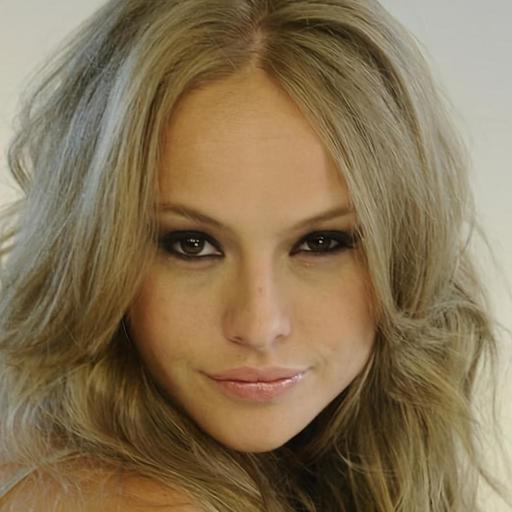}&
        \includegraphics[width=0.08\textwidth]{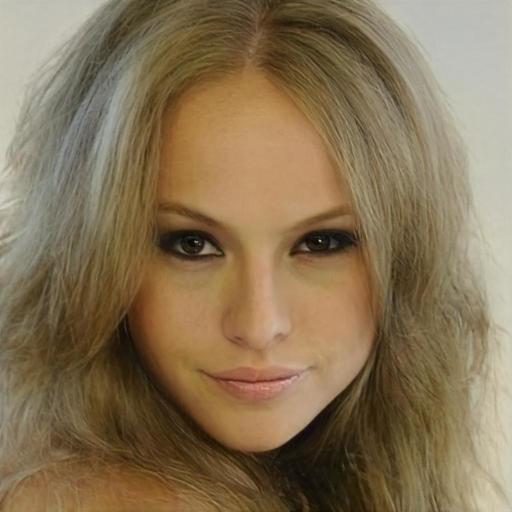}&        
        \includegraphics[width=0.08\textwidth]{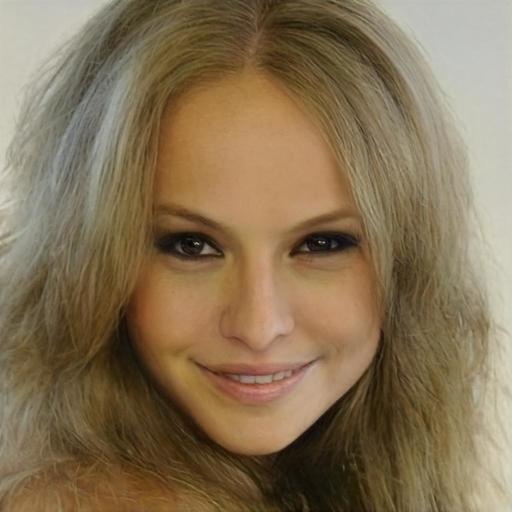}&
        \includegraphics[width=0.08\textwidth]{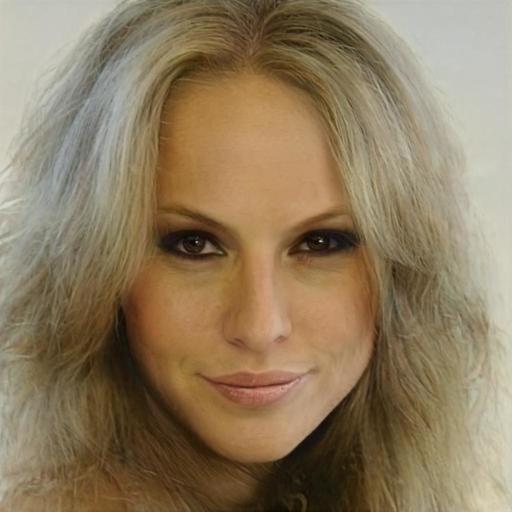}&
        \includegraphics[width=0.08\textwidth]{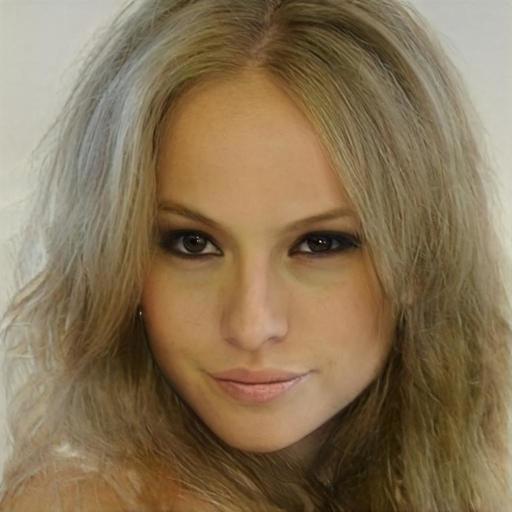} 
		\\
        \raisebox{0.2in}{\rotatebox[origin=t]{90}{Soft$\rightarrow$Hard}}&
		\includegraphics[width=0.08\textwidth]{images/appendix/hard/28349_real.jpg}&
        \includegraphics[width=0.08\textwidth]{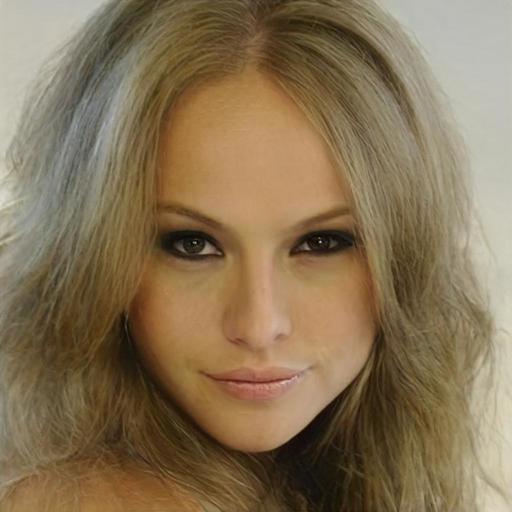}&        
        \includegraphics[width=0.08\textwidth]{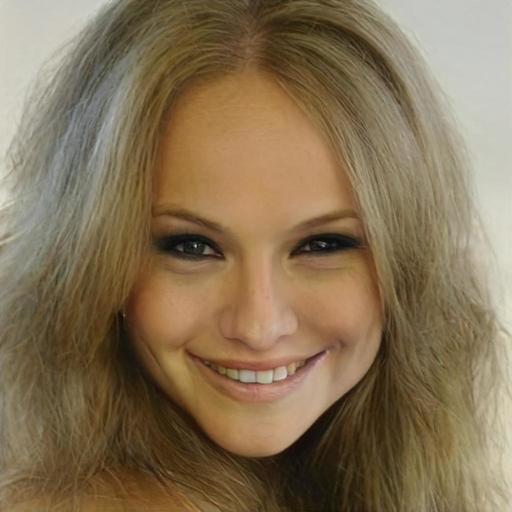}&
        \includegraphics[width=0.08\textwidth]{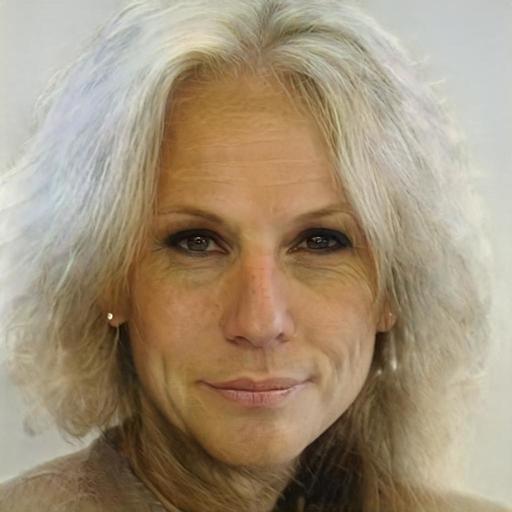}&
        \includegraphics[width=0.08\textwidth]{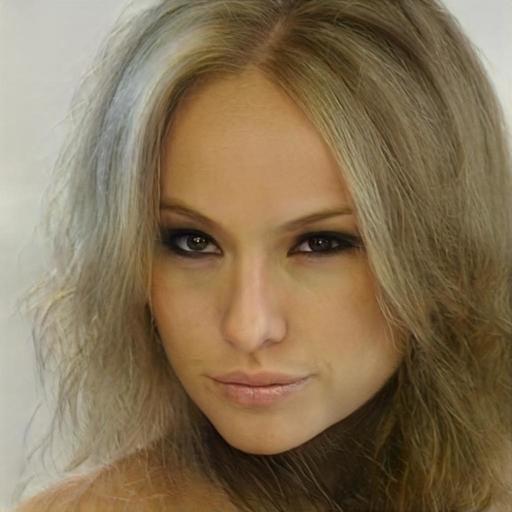} 
		\\
		& Input  & Inversion & Smile & Age & Pose
    \end{tabular}
    }
	\caption{Training methodology for $\mathcal{W}+$$\rightarrow$$\mathcal{W}$. Soft$\rightarrow$Hard denotes the training methodology described in section \ref{sec:wp2w}. Hard dentoes the training methodology described in section \ref{sec:w2wp}. The editing is performed using the same editing weight. Soft$\rightarrow$Hard ahieves superior editing quality in terms of editing magnitude.}
    \label{fig:hard}
\end{figure}

In section \ref{sec:cycle_encoding}, we propose an training methodology for $\mathcal{W}+$$\rightarrow$$\mathcal{W}$ that first gradually increases the weight of the delta regularization loss and then sequentially sets from $\Delta_{N-1}=0$ to $\Delta_1=0$. Increasing the weight of the delta regularization loss can be viewed as a ``soft'' operation, and setting $\Delta_i=0$ can be viewed as a ``hard'' operation. Here, we compare our design choice (denoted as Soft$\rightarrow$Hard) with sequentially setting from $\Delta_{N-1}=0$ to $\Delta_1=0$ (denoted as Hard). The quantitative results in Table \ref{table:hard} show that our method outperforms the hard operation in both reconstruction and editability. Figure \ref{fig:hard} also shows that our method achieves superior editability compared to the hard operation.

\begin{table}[b]
	\centering
	\caption{Training methodology for $\mathcal{W}+$$\rightarrow$$\mathcal{W}$. First row: reconstruction evaluation via identity similarity. Second row: editability evaluation via rotation angle.}
	\begin{tabular}{L{1.3cm}P{1.5cm}P{1.9cm}P{1.9cm}}
	\toprule
	 & PTI \cite{Roich2021} & Hard & Soft$\rightarrow$Hard \\
	\midrule
	Identity$\uparrow$ &
	\multicolumn{1}{c}{$0.845$} &
	\multicolumn{1}{c}{$0.863$} &
	\multicolumn{1}{c}{$0.866$} \\
	Angle$\uparrow$  &
	\multicolumn{1}{c}{$16.64$} &
	\multicolumn{1}{c}{$11.84$} &
	\multicolumn{1}{c}{$22.20$} \\
	\bottomrule
	\end{tabular}
	\label{table:hard}
\end{table}

\section{Details of the Models in Ablation Study}
In section \ref{sec:ablation}, we evaluate the proposed cycle encoding by comparing it with three variants: training the encoder in \W, training the encoder in $\mathcal{W}$$\rightarrow$$\mathcal{W}+$, and training the encoder in $\mathcal{W}+$$\rightarrow$$\mathcal{W}$. Here, we provide the details of the above model configurations. For the first configuration, we train the encoder only in the \W space for 750K iterations using the same hyperparameters as those used in the full approach. For the second configuration, we first train the encoder in the \Wp space for 500K iterations and then apply the training methodology described in section \ref{sec:wp2w} for 250K iterations. For the third configuration, we train the encoder using the training methodology described in section \ref{sec:w2wp} for 750K iterations.

\end{document}